\documentclass{article}

\PassOptionsToPackage{numbers, compress}{natbib}

\usepackage{hyperref}       %
\usepackage[final]{neurips_2020}

\usepackage[utf8]{inputenc} %
\usepackage[T1]{fontenc}    %
\usepackage{url}            %
\usepackage{booktabs}       %
\usepackage{amsfonts}       %
\usepackage{nicefrac}       %
\usepackage{microtype}      %

\usepackage{multirow}
\usepackage{amsmath}
\usepackage{amsthm}
\usepackage{dsfont}
\usepackage{amssymb}
\usepackage{amscd}
\usepackage{mathtools}
\mathtoolsset{showonlyrefs=true}
\usepackage{enumerate}
\usepackage{algorithmic}
\usepackage{algorithm}
\usepackage{bbold}
\usepackage{lmodern}
\usepackage{framed}
\usepackage{natbib}
\usepackage{geometry}
\usepackage{tabularx}
\usepackage{xcolor}

\hypersetup{ %
    colorlinks,%
    citecolor=black,%
    filecolor=black,%
    linkcolor=black,%
    urlcolor=black
}

\newtheorem{theorem}{Theorem}
\newtheorem{prop}{Proposition}

\newtheorem{remark}{Remark}
\newtheorem{lemma}{Lemma}

\newcommand{\argmax}{\operatorname*{argmax}} %
\newcommand{\argmin}{\operatorname*{argmin}}

\newcommand{\kl}{\operatorname*{KL}}

\newcommand{\cst}{\operatorname*{cst}}
\newcommand{\mm}{\operatorname{mm}}

\newcommand{\E}{\mathbb{E}}

\newcommand{\R}{\mathbb{R}}

\newcommand{\sep}{|}

\newcommand{\s}{\mathcal{S}}
\newcommand{\A}{\mathcal{A}}

\newcommand{\gc}{\mathcal{G}}
\newcommand{\hc}{\mathcal{H}}

\usepackage{mleftright}

\DeclareMathOperator{\FC}{FC}
\DeclareMathOperator{\Conv}{Conv}
\newcommand{\wor}{\textit{w/o}}
\newcommand{\wir}{\textit{w/}}

\newcommand{\un}{\mathbf{1}}

\makeatletter
\newcommand\footnoteref[1]{\protected@xdef\@thefnmark{\ref{#1}}\@footnotemark}
\makeatother

\title{Leverage the Average: an Analysis of KL Regularization in Reinforcement Learning}

\author{%
  Nino Vieillard \\
  Google Research, Brain Team\\
  Université de Lorraine, CNRS,  Inria \\IECL,  F-54000 Nancy, France \\
  \texttt{vieillard@google.com} \\
   \And
   Tadashi Kozuno\thanks{Work done while at DeepMind.} \\
   Okinawa Institute of Science and Technology \\
   \texttt{tadashi.kozuno@oist.jp} \\
  \AND
  Bruno Scherrer \\
  Université de Lorraine, CNRS,  Inria \\IECL,  F-54000 Nancy, France \\
   \texttt{bruno.scherrer@inria.fr} \\
   \And
   Olivier Pietquin \\
  Google Research, Brain Team \\
   \texttt{pietquin@google.com} \\
  \AND
  Rémi Munos \\
  DeepMind \\
   \texttt{munos@google.com} \\
  \And
  Matthieu Geist\\
  Google Research, Brain Team\\
  \texttt{mfgeist@google.com}
}

\begin{document}

\maketitle

\begin{abstract}
Recent Reinforcement Learning (RL) algorithms making use of  Kullback-Leibler (KL) regularization as a core component have shown outstanding performance. Yet, only little is understood theoretically about why KL regularization helps, so far. We study KL regularization within an approximate value iteration scheme and show that it implicitly averages $q$-values. Leveraging this insight, we provide a very strong performance bound, the very first to combine two desirable aspects: a linear dependency to the horizon (instead of quadratic) and an error propagation term involving an averaging effect of the estimation errors (instead of an accumulation effect). We also study the more general case of an additional entropy regularizer. The resulting abstract scheme encompasses many existing RL algorithms. Some of our assumptions do not hold with neural networks, so we complement this theoretical analysis with an extensive empirical study.
\end{abstract}

\section{Introduction}

In Reinforcement Learning (RL), Kullback-Leibler (KL) regularization consists in penalizing a new policy from being too far from the previous one, as measured by the KL divergence. It is at the core of efficient deep RL algorithms, such as Trust Region Policy Optimization (TRPO)~\citep{schulman2015trust} (motivated by trust region constraints) or Maximum a Posteriori Policy Optimization (MPO)~\citep{abdolmaleki2018maximum} (arising from the view of control as probabilistic inference
~\cite{levine2018reinforcement,fellows2019virel}), but without much theoretical guarantees. Recently, \citet{geist2019theory} have analyzed algorithms operating in the larger scope of regularization by Bregman divergences. They concluded that regularization doesn't harm in terms of  convergence, rate of convergence, and propagation of errors, but these results are not better than the corresponding ones in unregularized approximate dynamic programming (ADP). 

Building upon their formalism, we show that using a KL regularization implicitly averages the successive estimates of the $q$-function in the ADP scheme. Leveraging this insight, we provide a strong performance bound, the very first to combine two desirable aspects: 1) it has a linear dependency to the time horizon $(1-\gamma)^{-1}$, 2) it exhibits an error averaging property of the KL regularization. The linear dependency in the time horizon contrasts with the standard quadratic dependency of usual ADP, which is tight~\citep{scherrer2012use}. The only approaches achieving a linear dependency we are aware of make use of non-stationary policies~\citep{bagnell2004policy,scherrer2012use} and never led to practical deep RL algorithms. More importantly, the bound involves the norm of the average of the errors, instead of a discounted sum of the norms of the errors for classic ADP.
This means that, while standard ADP is not guaranteed to converge for the ideal case of independent and centered errors, 
KL regularization allows convergence to the optimal policy in that case.
The sole algorithms that also enjoy this compensation of errors are Dynamic Policy Programming (DPP)~\citep{azar2012dynamic} and Speedy Q-learning (SQL)~\citep{azar2011speedy}, that also build (implicitly) on KL regularization, as we will show for SQL. However, their dependency to the horizon is quadratic, and they are not well amenable to a deep learning setting~\citep{vieillard2019momentum}.

We also study the case of an additional entropy regularization, usual in practical algorithms, and specifically the interplay between both regularizations. The resulting abstract framework encompasses a wide variety of existing RL algorithms, the connections between some of them being known~\citep{geist2019theory}, but many other being new, thanks to the implicit average of $q$-values. We highlight that, even though our analysis covers the case where only the entropy regularization is considered, it does not explain why it helps without an additional KL term. Some argue that having a higher entropy helps exploration~\citep{schulman2017equivalence}, other that it has beneficial effects on the optimization landscape~\citep{ahmed2019understanding}, but it also biases the solution of the MDP~\citep{geist2019theory}. %

Our analysis requires some assumptions, notably that the regularized greedy step is done without approximation. If this is reasonable with discrete actions and a linear parameterization, it does not hold when neural networks are considered. Given their prevalence today, we complement our thorough analysis with an extensive empirical study, that aims at observing the core effect of regularization in a realistic deep RL setting.

\section{Background and Notations}
\label{sec:background}

Let $\Delta_X$ be the set of probability distributions over a finite set $X$ and $Y^X$ the set of applications from $X$ to the set $Y$. An MDP is a tuple $\{\s,\A,P,r,\gamma\}$ with $\s$ the finite
state space, $\A$ the finite set of actions, $P\in\Delta_\s^{\s\times\A}$ the Markovian transition kernel, $r\in\mathbb{R}^{\s\times\A}$ the reward function bounded by $r_\text{max}$, and $\gamma\in(0,1)$ the discount factor. For $\tau\geq 0$, we write $v^\tau_\text{max} = \frac{r_\text{max} + \tau\ln|\A|}{1-\gamma}$ and simply $v_\text{max} = v^0_\text{max}$. We  write $\un\in\R^{\s\times\A}$ the vector whose components are all equal to 1. 
A policy $\pi\in\Delta_\A^\s$ associates a distribution over actions to each state. Its (state-action) value function is defined as
$%
    q_\pi(s,a) = \E_\pi\left[\sum_{t=0}^\infty \gamma^t r(S_t,A_t)\middle \vert S_0=s, A_0=a\right]
$, %
$\E_\pi$ being the expectation over trajectories induced by $\pi$. Any optimal policy satisfies $\pi_* \in\argmax_{\pi\in\Delta^\s_\A} q_\pi$ (all scalar operators applied on vectors should be understood point-wise), and $q_*=q_{\pi_*}$.
The following notations will be useful. For $f_1,f_2\in\R^{\s\times\A}$,%
$%
    \langle f_1,f_2 \rangle = \left(\sum_{a} f_1(s,a) f_2(s,a)\right)_{s}\in\R^\s
$. %
This will be used  with $q$-values and (log) policies.
We write $P_\pi$ the stochastic kernel induced by $\pi$, and for $q\in\mathbb{R}^{\s\times \A}$ we have
$%
    P_\pi q = \left(\sum_{s'}P(s'|s,a) \sum_{a'} \pi(a'|s')  q(s',a')\right)_{s,a}\in\R^{\s\times\A}
$. %
For $v\in\R^\s$, we also define
$%
    P v = \left(\sum_{s'} P(s'|s,a) v(s')\right)_{s,a}\in\R^{\s\times\A}
$, %
hence $P_\pi q = P\langle \pi, q\rangle$.

The Bellman evaluation operator is  $T_\pi q = r + \gamma P_\pi q$, its unique fixed point being $q_\pi$. The set of greedy policies w.r.t. $q\in\R^{\s\times\A}$ is $\gc(q) = \argmax_{\pi\in\Delta_\A^\s} \langle q,\pi\rangle$. A classical approach to estimate an optimal policy is Approximate Modified Policy Iteration (AMPI)~\citep{puterman1978modified,scherrer2015approximate},
\begin{equation}
    \begin{cases}
        \pi_{k+1}\in\gc(q_k)\\ q_{k+1} = (T_{\pi_{k+1}})^m q_k + \epsilon_{k+1}
    \end{cases},
\end{equation}
which reduces to Approximate Value Iteration (AVI, $m=1$) and Approximate Policy Iteration (API,  $m=\infty$) as special cases. The term $\epsilon_{k+1}$ accounts for errors made when applying the Bellman operator. For example, the classic DQN~\citep{mnih2015human} is encompassed by this abstract ADP scheme, with $m=1$ and the error arising from fitting the neural network (regression step of DQN).
The typical
use of $m$-step rollouts in (deep) RL actually corresponds to an AMPI scheme with $m>1$. Next, we add regularization to this scheme.

\section{Regularized MPI}
\label{sec:regMPI}

In this work, we consider the entropy $\hc(\pi) = -\langle \pi,\ln \pi\rangle \in\R^\s$ and the KL divergence $\kl(\pi_1||\pi_2) = \langle \pi_1, \ln \pi_1 - \ln \pi_2 \rangle \in\R^\s$.
First, we introduce a slight variation of the Mirror Descent MPI scheme~\citep{geist2019theory} (handling both KL and entropy penalties, based on $q$-values).

\paragraph{Mirror Descent MPI.}

For $q\in\R^{\s\times\A}$ and an associated policy $\mu \in\Delta^\s_\A$, we define the regularized greedy policy as
$%
    \gc_\mu^{\lambda,\tau}(q) = \argmax_{\pi\in\Delta^\s_\A} \left(\langle \pi, q\rangle - \lambda \kl(\pi||\mu) + \tau \hc(\pi) \right)
$. %
Observe that with $\lambda=\tau=0$, we get the usual greediness. Notice also that with $\lambda=0$, the KL term disappears, so does the dependency to $\mu$. In this case we write $\gc^{0,\tau}$. 
We also account for the regularization in the Bellman evaluation operator. Recall that the standard operator is $T_\pi q = r + \gamma P \langle\pi, q\rangle$. Given the form of the regularized greediness, it is natural to replace the term $\langle \pi, q\rangle$ by the regularized one, giving $T_{\pi\sep\mu}^{\lambda,\tau} q = r + \gamma P \left(\langle \pi, q\rangle - \lambda \kl(\pi||\mu) + \tau \hc(\pi)\right)$.
These lead to the following MD-MPI($\lambda$,$\tau$) scheme. It is initialized with $q_0\in\R^{\s\times\A}$ such that $\|q_0\|_\infty\leq v_\text{max}$ and with $\pi_0$ the uniform policy, without much loss of generality (notice that the greedy policy is unique whenever $\lambda>0$ or $\tau>0$):
\begin{equation}
    \begin{cases}
        \pi_{k+1} = \gc_{\pi_k}^{\lambda,\tau}(q_k)
        \\
        q_{k+1} = (T^{\lambda,\tau}_{\pi_{k+1}\sep\pi_k})^m q_k + \epsilon_{k+1}
    \end{cases}.\label{eq:md-mpi}
\end{equation}

\paragraph{Dual Averaging MPI.}

We provide an equivalent formulation of scheme~\eqref{eq:md-mpi}. This will be the basis of our analysis, and it also allows drawing connections to other algorithms, originally not introduced as using a KL regularization. All the technical details are provided in the Appendix, but we give an intuition here, for the case $\tau=0$ (no entropy).
Let $\pi_{k+1} = \gc_{\pi_k}^{\lambda,0} (q_k)$. This optimization problem can be solved analytically, yielding $\pi_{k+1}\propto \pi_k \exp\frac{q_k}{\lambda}$. By direct induction, $\pi_0$ being uniform, we have %
$ %
    \pi_{k+1}\propto \pi_k \exp\frac{q_k}{\lambda} \propto \dots \propto \exp \frac{1}{\lambda}\sum_{j=0}^k q_j
$. %
This means that penalizing the greedy step with a KL divergence provides a policy being a softmax over the scaled sum of all past $q$-functions (no matter how they are obtained). This is reminiscent of dual averaging in convex optimization, hence the name.

We now introduce the Dual Averaging MPI (DA-MPI) scheme. 
Contrary to MD-MPI, we have to distinguish the cases $\tau=0$ and $\tau\neq 0$.
DA-MPI($\lambda$,0) and DA-MPI($\lambda$,$\tau>0$) are %
\begin{equation}
    \begin{cases}
        \pi_{k+1} = \gc^{0,\frac{\lambda}{k+1}}(h_k)
        \\
        q_{k+1} = (T_{\pi_{k+1}\sep\pi_k}^{\lambda,0})^m q_k + \epsilon_{k+1}
        \\
        h_{k+1} = \frac{k+1}{k+2} h_k + \frac{1}{k+2} q_{k+1}
    \end{cases} 
    \text{and }
    \begin{cases}
        \pi_{k+1} = \gc^{0,\tau}(h_k)
        \\
        q_{k+1} = (T_{\pi_{k+1}\sep\pi_k}^{\lambda,\tau})^m q_k + \epsilon_{k+1}
        \\
        h_{k+1} = \beta h_{k} + (1-\beta) q_{k+1} \text{ with } \beta = \frac{\lambda}{\lambda+\tau}
    \end{cases},\label{eq:da-mpi}
\end{equation}
with $h_0=q_0$ for $\tau=0$ and $h_{-1}=0$ for $\tau>0$.
The following result is proven in Appx.~\ref{subappx:proof_equivalence}.%
\begin{prop}
    \label{thm:equivalence}
    For any $\lambda>0$, MD-MPI($\lambda$,0) and DA-MPI($\lambda$,0) are equivalent (but not in the limit $\lambda \rightarrow 0$). Moreover, for any $\tau>0$,  MD-MPI($\lambda$,$\tau$) and DA-MPI($\lambda$,$\tau$) are equivalent.
\end{prop}

\begin{table}[tbh]
    \centering
    \caption{Algorithms encompassed by MD/DA-MPI (in italic if new compared to~\citep{geist2019theory}).}
    \label{tab:covered_algs}
    \begin{tabular}{l|c|c|c}
         & only entropy & only KL & both \\
         \hline
        reg. & Soft Q-learning~\citep{fox2015taming,haarnoja2017reinforcement},  & DPP~\citep{azar2012dynamic},   & \textit{CVI}~\citep{kozuno2019theoretical},    
        \\ 
        eval. & SAC~\citep{haarnoja2018soft}, \textit{Mellowmax}~\citep{asadi2016alternative} & \textit{SQL}~\citep{azar2011speedy} &  \textit{AL}~\citep{baird1999reinforcement,bellemare2016increasing} 
        \\ \hline
        unreg. & \textit{softmax DQN}~\citep{song2019revisiting}  & TRPO~\citep{schulman2015trust}, MPO~\citep{abdolmaleki2018maximum},   & \textit{softened LSPI}~\citep{perolat2016softened},
        \\
        eval. & & \textit{Politex}~\citep{abbasi2019politex}, \textit{MoVI}~\citep{vieillard2019momentum} & \textit{MoDQN}~\citep{vieillard2019momentum}
    \end{tabular}
\end{table}

\paragraph{Links to existing algorithms.}

Equivalent schemes~\eqref{eq:md-mpi} and~\eqref{eq:da-mpi} encompass (possibly variations of) many existing RL algorithms (see Tab.~\ref{tab:covered_algs} and details below). Yet, we think important to highlight that many of them don't consider regularization in the evaluation step (they use $T_{\pi_{k+1}}$ instead of $T^{\lambda,\tau}_{\pi_{k+1}\sep\pi_k}$), something we abbreviate as ``\wor''. If it does not preclude convergence in the case $\tau=0$~\citep[Thm.~4]{geist2019theory}, it is known for the case $\tau>0$ and $\lambda=0$ that the resulting Bellman operator may have multiple fixed points~\citep{asadi2016alternative}, which is not desirable. Therefore, we only consider a regularized evaluation for the analysis, but we will compare both approaches empirically.
Now, we present the approaches encompassed by scheme~\eqref{eq:md-mpi} (see also  Appx.~\ref{subappx:conection_mdmpi}).
Soft Actor Critic (SAC)~\citep{haarnoja2018soft} and soft Q-learning~\citep{haarnoja2017reinforcement} are variations of MD-MPI($0$,$\tau$), as is softmax DQN~\citep{song2019revisiting} but \wor. The Mellowmax policy~\citep{asadi2016alternative} is equivalent to MD-MPI($0$,$\tau$). TRPO and MPO are variations of MD-MPI($\lambda$,$0$), \wor. DPP~\citep{azar2012dynamic} is almost a reparametrization of MD-MPI($\lambda$,$0$), and Conservative Value Iteration (CVI)~\citep{kozuno2019theoretical} is a reparametrization of MD-MPI$_1$($\lambda$,$\tau$), which consequently also generalizes Advantage Learning (AL)~\citep{baird1999reinforcement,bellemare2016increasing}.
Next, we present the approaches encompassed by schemes~\eqref{eq:da-mpi} (see also Appx.~\ref{subappx:connection_dampi}).
Politex~\citep{abbasi2019politex} is a PI scheme for the average reward case, building upon prediction with expert advice. In the discounted case, it is  DA-MPI($\lambda$,0), \wor. Momentum Value Iteration (MoVI)~\citep{vieillard2019momentum} is a limit case of DA-MPI($\lambda$,0), \wor{}, as $\lambda\rightarrow 0$, and its practical extension to deep RL momentum DQN (MoDQN) is a limit case of DA-MPI($\lambda$,$\tau$), \wor. SQL~\citep{azar2011speedy} is a limit case of DA-MPI($\lambda$, 0) as $\lambda\rightarrow 0$. Softened LSPI~\citep{perolat2015approximate} deals with zero-sum Markov games, but specialized to single agent RL it is a limit case of DA-MPI($\lambda$,$\tau$), \wor.

\section{Theoretical Analysis}
\label{sec:analysis}

Here, we analyze the propagation of errors of MD-MPI, through the equivalent DA-MPI, for the case $m=1$ (that is regularized VI, the  extension to $m>1$ remaining an open question). We   provide component-wise bounds that assess the quality of the learned policy, depending on $\tau=0$ or not. From these, $\ell_p$-norm bounds could be derived, using~\citep[Lemma~5]{scherrer2015approximate}.

\paragraph{Analysis of DA-VI($\lambda$,0).}

This corresponds to scheme~\eqref{eq:da-mpi}, left, with $m=1$.
The following Thm. is proved in Appx.~\ref{subsec:proof:davi_type1_noEntropy}.
\begin{theorem}
    \label{thm:davi_type1_noEntropy}
    Define $E_k = -\sum_{j=1}^k \epsilon_j$, $A_k^1 = (I-\gamma P_{\pi_*})^{-1} - (I-\gamma P_{\pi_{k}})^{-1}$ and $g^1(k) = \frac{4}{1-\gamma} \frac{v_\text{max}^\lambda}{k}$. Assume that $\|q_k\|_\infty \leq v_\text{max}$. We have
    $%
        0 \leq q_* - q_{\pi_{k}} \leq \left|A^1_k \frac{E_k}{k}\right| + g^1(k)\un
    $. %
\end{theorem}
\begin{remark}
\label{rk:rk}
    The assumption $\|q_k\|_\infty\leq v_\text{max}$ is not strong. It can be enforced by simply clipping the result of the evaluation step in $[-v_\text{max},v_\text{max}]$.
    See also Appx.~\ref{appx:rk}.
\end{remark}
To ease the discussion, we express an $\ell_\infty$-bound as a direct corollary of Thm.~\ref{thm:davi_type1_noEntropy}:
\begin{equation}
    \|q_* - q_{\pi_k}\|_{\infty}\leq 
    \frac{2}{1-\gamma} \left\|\frac{1}{k}\sum_{j=1}^k \epsilon_j\right\|_\infty
    + \frac{4}{1-\gamma} \frac{v^\lambda_\text{max}}{k}
    .
\end{equation}
We also recall the typical propagation of errors of AVI without regularization (\textit{e.g.}~\citep{scherrer2015approximate}, we scale the sum by $1-\gamma$ to make explicit the normalizing factor of a discounted sum):
\begin{equation}
    \|q_* - q_{\pi_k}\|_\infty \leq 
    \frac{2\gamma}{(1-\gamma)^2}\bigg( (1-\gamma) \sum_{j=1}^k \gamma^{k-j} \|\epsilon_j\|_\infty\bigg)
    + \frac{2}{1-\gamma} \gamma^k v_\text{max}
    .
\end{equation}
For each bound, the first term can be decomposed as a factor, the \emph{horizon term} ($(1-\gamma)^{-1}$ is the average horizon of the MDP), scaling the \emph{error term}, that expresses how the errors made at each iteration reflect in the final performance. The second term reflects the influence of the initialization over iterations, without errors it give the \emph{rate of convergence} of the algorithms. We discuss these three terms.

\textbf{Rate of convergence.} It is slower for DA-VI($\lambda$,0) than for AVI, $\gamma^k = o(\frac{1}{k})$. This was to be expected, as the KL term slows down the policy updates. It is not where the benefits of KL regularization arise. However, notice that for $k$ small enough and $\gamma$ close to 1, we may have $\frac{1}{k} \leq \gamma^k$. This term has also a linear dependency to $\lambda$ (through $v^\lambda_\text{max}$), suggesting that a lower $\lambda$ is better. This is intuitive, a larger $\lambda$ leads to smaller changes of the policy, and thus to a slower convergence.

\textbf{Horizon term.} We have a linear dependency to the horizon, instead of a quadratic one, which is very strong. Indeed, it is known that the square dependency to the horizon is tight for API and AVI~\citep{scherrer2012use}. The only algorithms based on ADP having a linear dependency we are aware of make use of non-stationary policies~\citep{scherrer2012use,bagnell2004policy}, and have never led to practical (deep) RL algorithms. Minimizing directly the Bellman residual would also lead to a linear dependency (\textit{e.g.},~\citep[Thm.~1]{piot2014difference}), but it comes with its own drawbacks~\citep{geist2017bellman} (\textit{e.g.}, bias problem with stochastic dynamics, and it is not used in deep RL, as far as we know).

\textbf{Error term.} For AVI, the error term is a discounted \emph{sum of the norms} of the successive estimation errors, while in our case it is the \emph{norm of the average} of these estimation errors. The difference is fundamental, it means that the KL regularization allows for a compensation of the errors made at each iteration. Assume that the sequence of errors is a martingale difference. AVI would not converge in this case, while DA-VI($\lambda$, 0) converges to the optimal policy ($\|\frac{1}{k}\sum_{j=1}^k \epsilon_j\|_\infty$ converges to 0 by the law of large numbers). As far as we know, only SQL and DPP have such an error term, but they have a worse dependency to the horizon.

Thm.~\ref{thm:davi_type1_noEntropy} is the first result showing that an RL algorithm can benefit from both a linear dependency to the horizon and from an averaging of the errors, and we argue that this explains, at least partially, the beneficial effect of using a KL regularization. Notice that Thm.~4 of~\citet{geist2019theory} applies to DA-VI($\lambda$, 0), as they study more generally MPI regularized by a Bregman divergence. Although they bound a regret rather than $q_* - q_{\pi_k}$, their result is comparable to AVI, with a quadratic dependency to the horizon and a discounted sum of the norms of the errors. Therefore, our result significantly improves previous analyses.

\begin{figure}[tbh]
    \centering
    \begin{minipage}[c]{0.33\linewidth}
        \includegraphics[width=\linewidth]{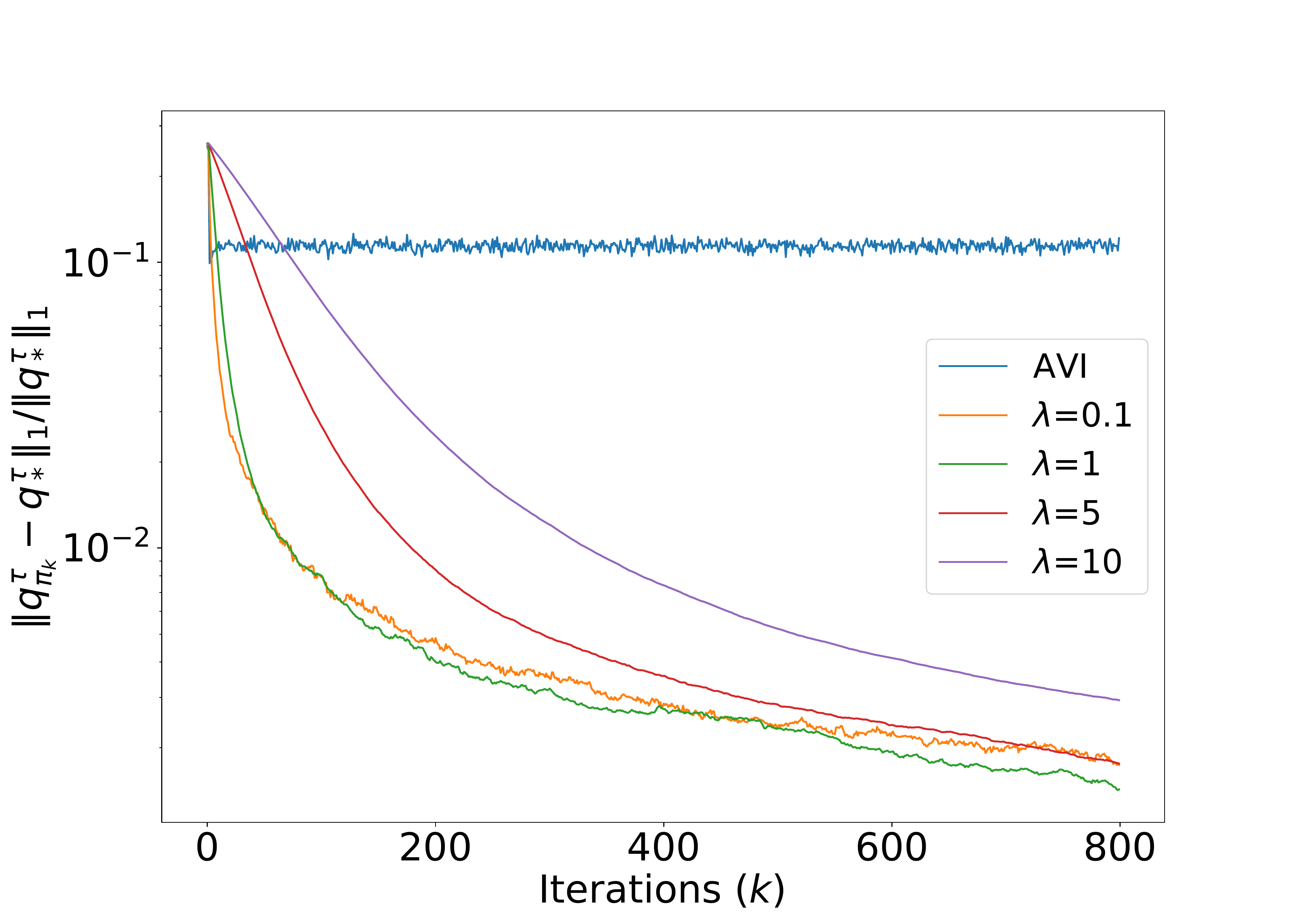}
    \end{minipage} \hfill
    \begin{minipage}[c]{0.33\linewidth}
        \includegraphics[width=\linewidth]{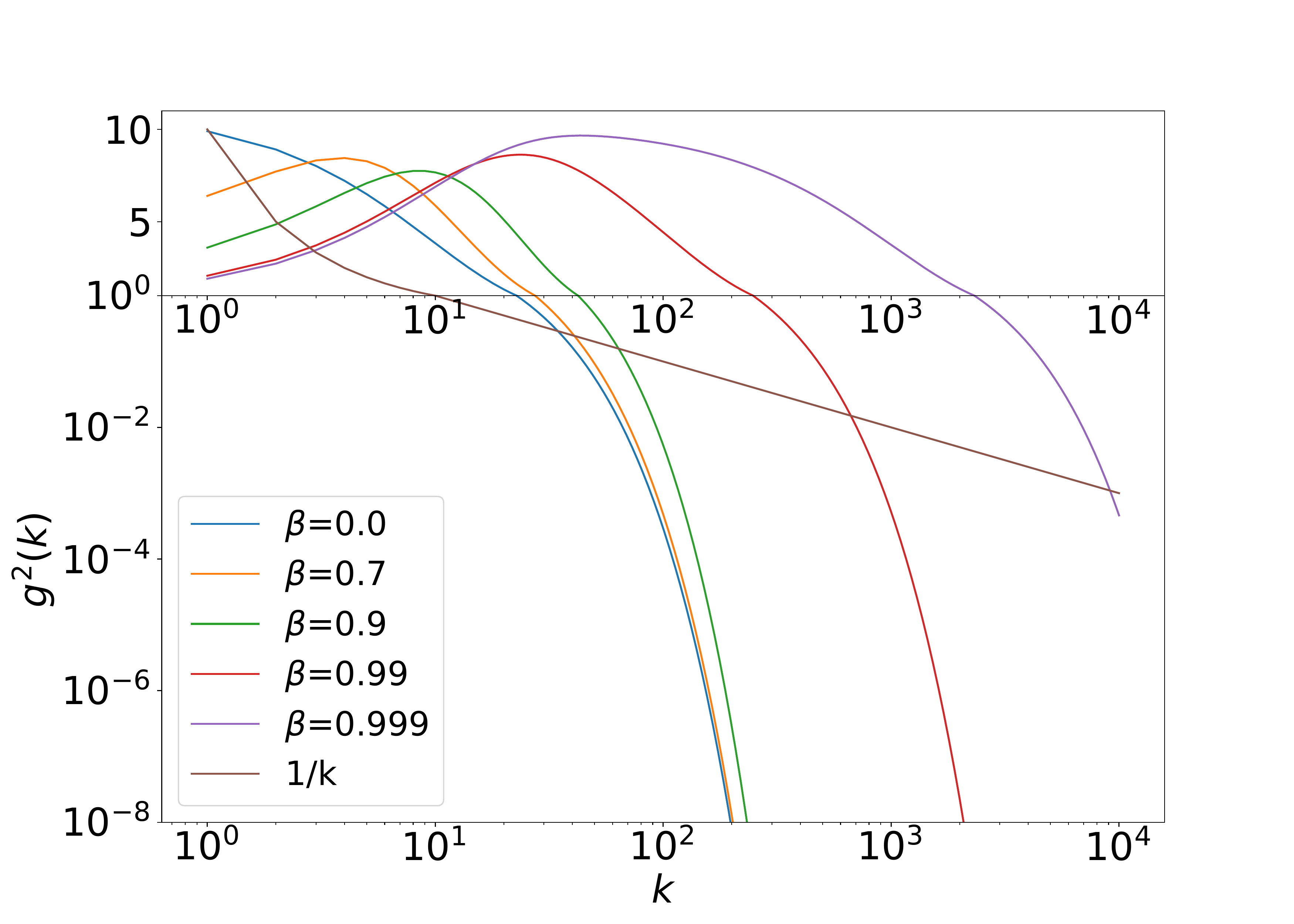}
    \end{minipage}\hfill
    \begin{minipage}[c]{0.33\linewidth}
        \includegraphics[width=\linewidth]{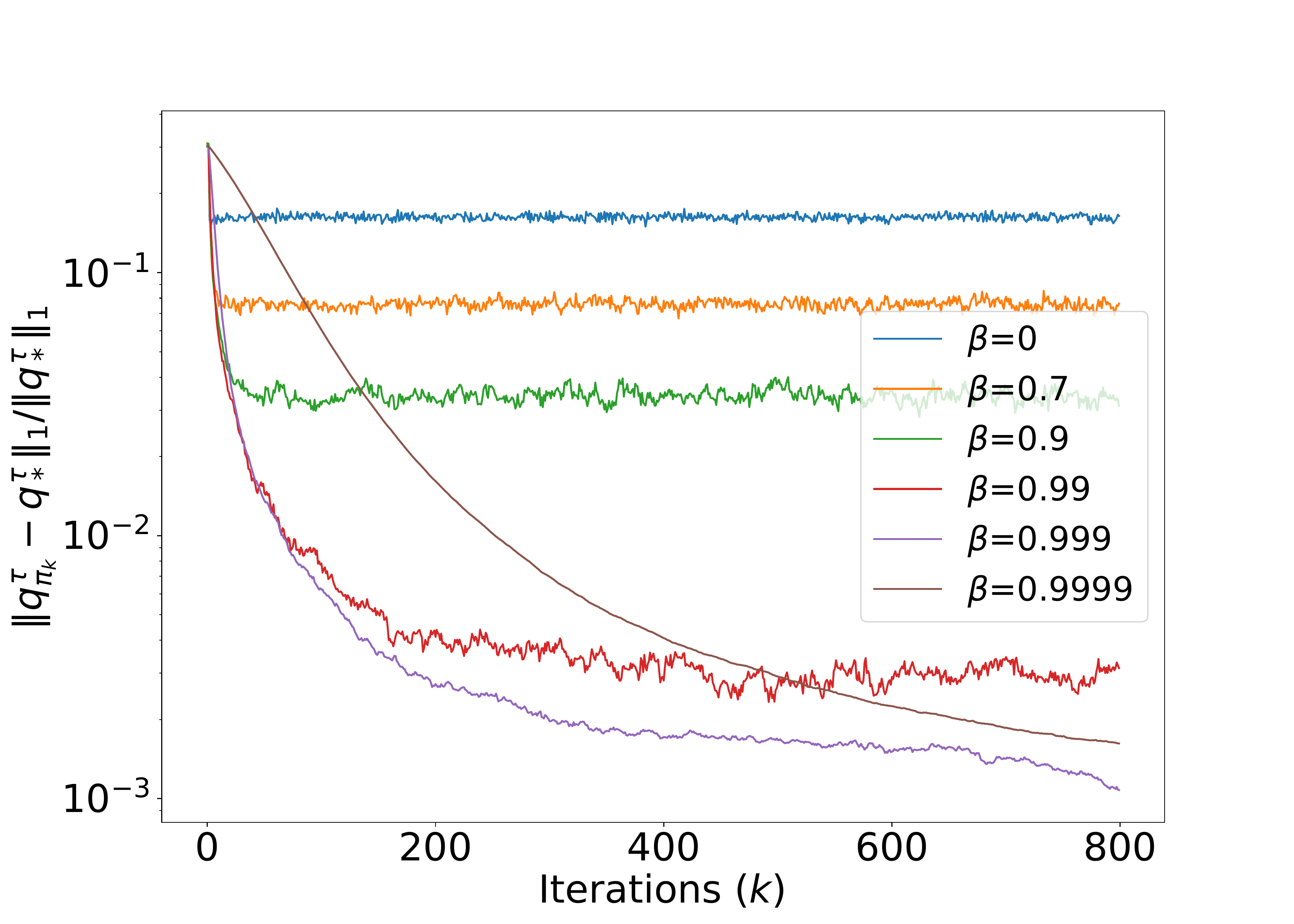}
    \end{minipage}
    \caption{\textbf{Left}: behavior for Thm~\ref{thm:davi_type1_noEntropy}. \textbf{Middle}: function $g^2(k)$. \textbf{Right}: behavior for Thm~\ref{thm:davi_type1_withEntropy}.}
    \label{fig:illustration_thm1}
\end{figure}

We illustrate the bound with a simple experiment\footnote{
\label{foo}
We illustrate the bounds in a simple tabular setting with access to a generative model. Considering random MDPs (called Garnets), at each iteration of DA-VI we sample a single transition for each state-action couple and apply the resulting sampled Bellman operator. The error $\epsilon_k$ is the difference between the sampled and the exact operators. The sequence of these estimation errors is thus a martingale difference w.r.t. its natural filtration~\citep{azar2011speedy} (one can think about bounded, centered and roughly independent errors). More details about this practical setting are provided in Appx.~\ref{appx:garnet}.
},
see Fig.~\ref{fig:illustration_thm1}, left. We observe that AVI doesn't converge, while DA-VI($\lambda$,0) does, and that higher values of $\lambda$ slow down the convergence. Yet, they are also a bit more stable. This is not explained by our bound but is quite intuitive (policies changing less between iterations).
\paragraph{Analysis of DA-VI($\lambda$,$\tau$).}
This is scheme~\eqref{eq:da-mpi}, right, with $m=1$.
Due to the non-vanishing entropy term in the greedy step, it cannot converge to the unregularized optimal $q$-function. Yet, without errors and with $\lambda=0$, it would converge to the solution of the MDP regularized by the scaled entropy (that is, considering the reward augmented by the scaled entropy). Our bound will show that adding a KL penalty does not change this. To do so, we introduce a few notations.
The proofs of the following claims can be found in~\citep{geist2019theory}, for example.
We already have defined the operator $T_\pi^{0,\tau}$. It has a unique fixed point, that we write $q^\tau_\pi$. The unique optimal $q$-function is $q^\tau_* = \max_{\pi} q^\tau_\pi$. We write $\pi_*^\tau = \gc^{0,\tau}(q_*^\tau)$ the associated unique optimal policy, and $q_{\pi_*^\tau}^\tau = q_*^\tau$.
The next result is proven in Appx.~\ref{subsec:proof:davi_type1_withEntropy}.
\begin{theorem}
    \label{thm:davi_type1_withEntropy}
    For a sequence of policies $\pi_0,\dots,\pi_k$, we define $P_{k:j} = P_{\pi_k} P_{\pi_{k-1}} \dots P_{\pi_j}$ if $j\leq k$, $P_{k:j} = I$ else. We define $A^2_{k:j} = P_{\pi_*^\tau}^{k-j} + (I-\gamma P_{\pi_{k+1}})^{-1} P_{k:j+1}(I-\gamma P_{\pi_j})$. We define $g^2(k)=\gamma^k (1 + \frac{1-\beta}{1-\gamma}) \sum_{j=0}^k (\frac{\beta}{\gamma})^j v_\text{max}^\tau$, with $\beta$ as defined in Eq.
   ~\eqref{eq:da-mpi}. Finally, we define $E^\beta_k = (1-\beta)\sum_{j=1}^k \beta^{k-j} \epsilon_j$. With these notations:
    $%
        0 \leq q_*^\tau - q_{\pi_{k+1}}^\tau \leq \sum_{j=1}^k \gamma^{k-j}\left|A^2_{k:j} E^\beta_j\right| + g^2(k)\un
    $. %
\end{theorem}
Again, to ease the discussion, we express an $\ell_\infty$-bound as a direct corollary of Thm.~\ref{thm:davi_type1_withEntropy}:
\begin{equation}
    \|q_*^\tau - q_{\pi_{k+1}}^\tau\|_\infty \leq \frac{2}{(1-\gamma)^2} \bigg((1-\gamma)\sum_{j=1}^k \gamma^{k-j} \|E^\beta_j\|_\infty\bigg) %
    + \gamma^k (1 + \frac{1-\beta}{1-\gamma}) \sum_{j=0}^k \left(\frac{\beta}{\gamma} \right)^j v_\text{max}^\tau.
\end{equation}
There is a square dependency to the horizon, as for AVI. We discuss the other terms.

\textbf{Rate of convergence.} It is given by the function $g^2$, defined in Thm.~\ref{thm:davi_type1_withEntropy}. If $\beta = \gamma$, we have $g^2(k) = 2 (k+1) \gamma^k v_\text{max}^\tau$. If $\beta \neq \gamma$, we have $g^2(k) = (1+\frac{1-\beta}{1-\gamma}) \frac{\beta^{k+1} - \gamma^{k+1}}{\beta - \gamma}$. In all cases, $g^2(k) = o(\frac{1}{k})$, so it is asymptotically faster than in Thm.~\ref{thm:davi_type1_noEntropy}, but the larger the $\beta$, the slower the initial convergence. This is illustrated in Fig.~\ref{fig:illustration_thm1}, middle (notice that it's a logarithmic plot, except for the upper part of the $y$-axis).

\textbf{Error rate.} As with AVI, the error term is a discounted sum of the norms of errors. However, contrary to AVI, each error term is not an iteration error, but a moving average of past iteration errors, $E_k^\beta = \beta E_{k-1}^\beta + (1-\beta)\epsilon_k$. In the ideal case where the sequence of these errors is a martingale difference with respect to the natural filtration, this term no longer vanishes, contrary to $\frac{1}{k} E_k$. However, it can reduce the variance. For simplicity, assume that the $\epsilon_j$'s are i.i.d. of variance 1. In this case, it is easy to see that the variance of $E^\beta_k$ is bounded by $1-\beta<1$, that tends toward 0 for $\beta$ close to 1. Therefore, we advocate that DA-VI$_1$($\lambda$,$\tau$) allows for a better control of the error term than AVI (retrieved for $\beta=0$). Notice that if asymptotically this error term predominates, the non-asymptotic behavior is also driven by the convergence rate $g^2$, which will be faster for $\beta$ closer to 0. Therefore, there is a trade-off, illustrated in Fig.~\ref{fig:illustration_thm1}, right (for the same simple experiment\footnoteref{foo}). Higher values of $\beta$ lead to better asymptotic performance, but at the cost of slower initial convergence rate.

\textbf{Interplay between the KL and the entropy terms.} The l.h.s. of the bound of Thm.~\ref{thm:davi_type1_withEntropy} solely depends on the entropy scale $\tau$, while the r.h.s. solely depends on the term $\beta=\frac{\lambda}{\lambda + \tau}$. DA-VI($\lambda$,$\tau$) approximates the optimal policy of the regularized MDP, while we are usually interested in the solution of the original one. We have that $\|q_* - q_{\pi_*^\tau}\|_\infty \leq \frac{\tau\ln|\A|}{1-\gamma}$~\citep{geist2019theory}, this bias can be controlled by setting an (arbitrarily) small $\tau$. This does not affect the r.h.s. of the bound, as long as the scale of the KL term follows (such that $\frac{\lambda}{\lambda + \tau}$ remains fixed to the chosen value). So, Thm.~\ref{thm:davi_type1_withEntropy} suggests to set $\tau$ to a very small value and to choose $\lambda$ such that we have a given value of $\beta$. However, adding an entropy term has been proven efficient empirically, be it with arguments of exploration and robustness~\citep{haarnoja2018soft} or regarding the optimization landscape~\citep{ahmed2019understanding}. Our analysis does not cover this aspect. Indeed, it applies to $\lambda=\beta=0$ (that is, solely entropy regularization), giving the propagation of errors of SAC, as a special case of~\citep[Thm.~3]{geist2019theory}. In this case, we retrieve the bound of AVI ($E_j^0 = \epsilon_j$, $g^2(k)\propto \gamma^k$), up to the bounded quantity. Thus, it does not show an advantage of using solely an entropy regularization, but it shows the advantage for considering an additional KL regularization, if the entropy is of interest for other reasons.

We end this discussion with some related works. The bound of Thm.~\ref{thm:davi_type1_withEntropy} is similar to the one of CVI, despite a quite different proof technique. Notably, both involve a moving average of the errors. This is not surprising, CVI being a reparameterization of DA-VI. The core difference is that by bounding the distance to the regularized optimal $q$-function (instead of the unregularized one), we indeed show to what the algorithm converges without error.
\citet{shani2019adaptive} study a variation of TRPO, for which they show a convergence rate of $\mathcal{O}(\frac{1}{\sqrt{k}})$, improved to $\mathcal{O}(\frac{1}{k})$ when an additional entropy regularizer is considered. This is to be compared to the convergence rate of our variation of TRPO, $\mathcal{O}(\frac{1}{k}) = o(\frac{1}{\sqrt{k}})$ (Thm.~\ref{thm:davi_type1_noEntropy}) improved to $g^2(k) = o(\frac{1}{k})$ with an additional entropy term (Thm.~\ref{thm:davi_type1_withEntropy}). Our rates are much better. However, this is only part of the story. We additionally show a compensation of errors in both cases, something not covered by their analysis. They also provide a sample complexity, but it is much worse than the one of SQL, that we would improve (thanks to the improved horizon term). Therefore, our results are stronger and more complete.

\paragraph{Limitations of our analysis.}

Our analysis provides strong theoretical arguments in favor of considering KL regularization in RL. Yet, it has also some limitations.
First, it does not provide arguments for using only entropy regularization, as already extensively discussed (even though it provides arguments for combining it with a KL regularization).
Second, we study how the errors propagate over iterations, and show that KL allows for a compensation of these errors, but we say nothing about how to control these errors. This depends heavily on how the $q$-functions are approximated and on the data used to approximate them. We could easily adapt the analysis of~\citet{azar2011speedy} to provide sample complexity bounds for MD-VI in the case of a tabular representation and with access to a generative model, but providing a more general answer is difficult, and beyond the scope of this paper.
Third, we assumed that the greedy step was performed exactly. This assumption would be reasonable with a linear parameterization and discrete actions, but not if the policy and the $q$-function are approximated with neural networks. In this case, the equivalence between MD-VI and DA-VI no longer holds, suggesting various ways of including the KL regularizer (explicitly, MD-VI, or implicitly, DA-VI).
Therefore, we complement our thorough theoretical analysis with an extensive empirical study, to analyse the core effect of regularization in deep RL.

\section{Empirical study}
\label{sec:experiments}

Before all, we would like to highlight that if regularization is a core component of successful deep RL algorithms (be it with entropy, KL, or both), it is never the sole component. For example, SAC uses a twin critic~\citep{fujimoto2018addressing}, TRPO uses a KL hard constraint rather than a KL penalty~\citep{schulman2017proximal}, or MPO uses retrace~\citep{munos2016safe} for value function evaluation. All these further refinements play a role in the final performance. On the converse, our goal is to study the core effect of regularization, especially of KL regularization, in a deep RL context. To achieve this, we notice that DA-VI and MD-VI are extensions of AVI. One of the most prevalent VI-based deep RL algorithm being DQN~\citep{mnih2016asynchronous}, our approach is to start from a reasonably tuned version of it~\citep{castro2018dopamine} and to provide the minimal modifications to obtain deep versions of MD-VI or DA-VI. Notably, we fixed the meta-parameters to the best values for DQN.

\paragraph{Practical algorithms.}

We describe briefly the variations we consider, a complementary high-level view is provided in Appx.~\ref{subappx:high_level} and all practical details in Appx.~\ref{subappx:algorithms}. We modify DQN by adding an actor. For the \textbf{evaluation step}, we keep the DQN loss, modified to account for regularization (that we'll call ``\wir'', and that simply consists in adding the regularization term to the target $q$-network). Given that many approaches ignore the regularization there, we'll also consider the DQN loss (denoted ``\wor'' before, not covered by our analysis). For the \textbf{greedy step}, MD-VI and DA-VI are no longer equivalent. For MD-VI, there are two ways of approximating the regularized policy. The first one, denoted ``\emph{MD direct}'', consists in directly solving the optimization problem corresponding to the regularized greediness, the policy being a neural network. This is reminiscent of TRPO (with a penalty rather than a constraint). The second one, denoted ``\emph{MD indirect}'', consists in computing the analytical solution to the greedy step ($\pi_{k+1} \propto \pi_k^\beta \exp(\frac{1}{\lambda}\beta q_k)$) and to approximate it with a neural network. This is reminiscent of MPO. For DA-VI, we have to distinguish $\tau>0$ from $\tau=0$. In the first case, the regularized greedy policy can be computed analytically from an $h$-network, that can be computed by fitting a moving average of the online $q$-network and of a target $h$-network. This is reminiscent of MoDQN. If $\tau=0$, DA-VI($\lambda$,0) is not practical in a deep learning setting, as it requires averaging over iterations. Updates of target networks are too fast to consider them as new iterations, and a moving average is more convenient. So, we only consider the limit case $\lambda,\tau\rightarrow 0$ with $\beta=\frac{\lambda}{\lambda + \tau}$ kept constant. This is MoDQN with fixed $\beta$, and the evaluation step is necessarily unregularized ($\lambda=\tau=0$). To sum up, we have six variations (three kinds of greediness, evaluation regularized or not), restricted to five variations for $\tau=0$.

\paragraph{Research questions.}

Before describing the empirical study, we state the research questions we would like to address. The first is to know if regularization, without further refinements, helps, compared to the baseline DQN. The second one is to know if adding regularization in the evaluation step, something required by our analysis, provides improved empirical results. The third one is to compare the different kinds of regularized greediness, which are no longer equivalent with approximation. The last one is to study the effect of entropy, not covered by our analysis, and its interplay with the KL term.

\paragraph{Environments.}

We consider two environments here (more are provided in Appx.~\ref{appx:exp}). The light Cartpole from Gym~\citep{brockman2016openai} allows for a large sweep over the parameters, and to average each result over 10 seeds. We also consider the Asterix Atari game~\citep{bellemare2013arcade}, with sticky actions, to assess the effect of regularization on a large-scale problem. The sweep over parameters is smaller, and each result is averaged over 3 seeds. 

\paragraph{Visualisation.} For each environment, we present results as a table, the rows corresponding to the type of evaluation (\wir{} or \wor), the columns to the kind of greedy step. Each element of this table is a grid, varying $\beta$ for the rows and $\tau$ for the columns. One element of this grid is the average undiscounted return per episode obtained during training, averaged over the number of seeds. On the bottom of this table, we show the limit cases with the same principle, varying with $\lambda$ for MD-VI and with $\beta$ for DA-VI (ony \wor, as explained before). The scale of colors is common to all these subplots, and the performance of DQN is indicated on this scale for comparison. Additional visualisations are provided in Appx.~\ref{appx:exp}.

\begin{figure}%
    \centering
    \includegraphics[width=.75\linewidth,trim={2cm 4cm 0.5cm 6cm},clip]{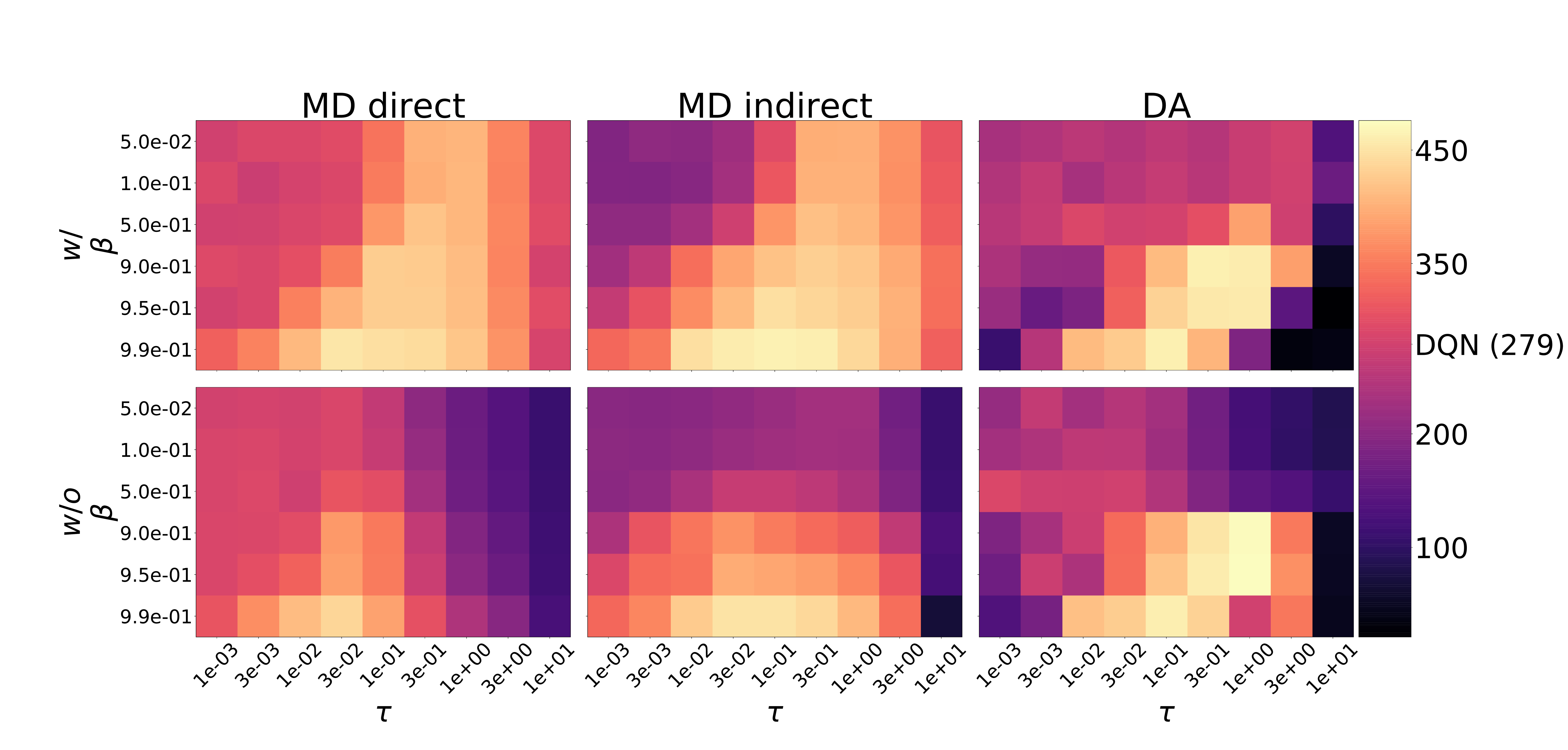}  \\
    \includegraphics[width=.75\linewidth,trim={2cm 1cm 4cm 6cm},clip]{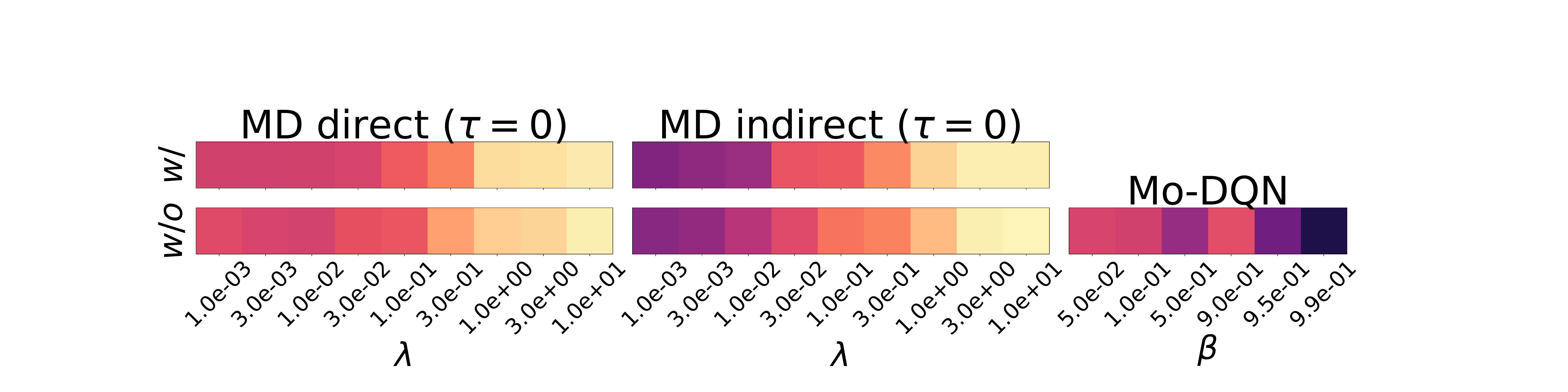}
    \caption{Cartpole.%
    }
    \label{fig:exp_core}
\end{figure}

\begin{figure}%
    \centering
    \includegraphics[width=.75\linewidth,trim={2cm 4cm 0.5cm 6cm},clip]{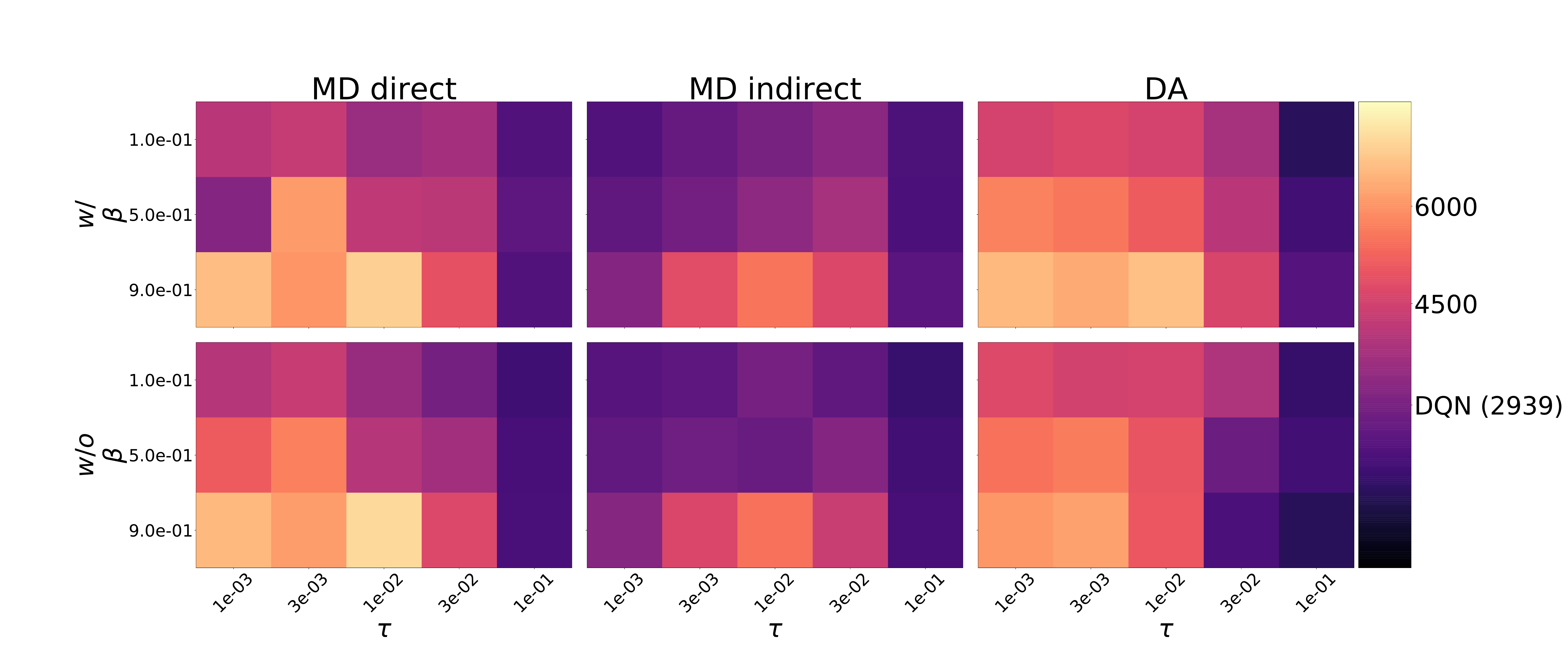}  \\
    \includegraphics[width=.75\linewidth,trim={2cm 0.5cm 4cm 4cm},clip]{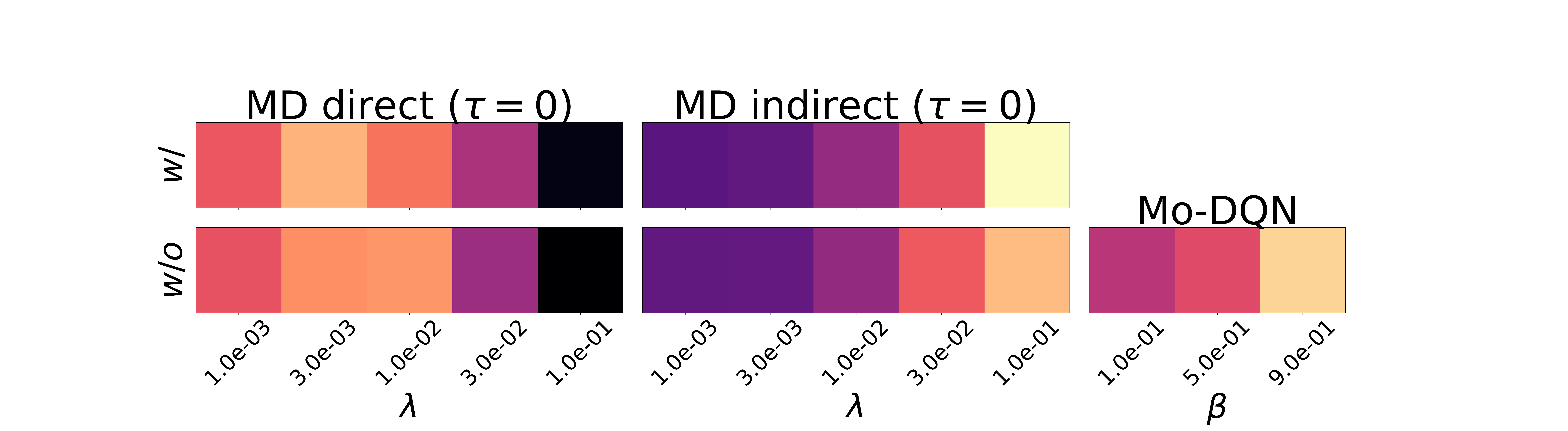}
    \caption{Asterix. %
    }
    \label{fig:exp_core2}
\end{figure}

\paragraph{Discussion.}

Results are provided in Fig.~\ref{fig:exp_core} and~\ref{fig:exp_core2}. First, we observe that regularization helps. Indeed, the results obtained by all these variations are better than the one of DQN, the baseline, for a large range of the parameters, sometime to a large extent. We also observe that, for a given value of $\tau$, the results are usually better for medium to large values of $\beta$ (or $\lambda$), suggesting that \textbf{KL regularization is beneficial} (even though too large KL regularization can be harmful in some case, for example for MD direct, $\tau=0$, on Asterix). 

Then, we study the effect of regularizing the evaluation step, something suggested by our analysis. The effect of this can be observed by comparing the first row to the second row of each table. One can observe that the range of good parameters is larger in the first row (especially for large entropy), suggesting that \textbf{regularizing the evaluation step helps}. Yet, we can also observe that when $\tau=0$ (no entropy), there is much less difference between the two rows. This suggests that adding the entropy regularization to the evaluation step might be more helpful (but adding the KL term too is costless and never harmful).

Next, we study the effect of the type of greediness. MD-direct shows globally better results than MD-indirect, but MD-indirect provides the best result on both environments (by a small margin), despite being more sensitive to the parameters. DA is more sensitive to parameters than MD for Cartpole, but less for Asterix, its best results being comparable to those of MD. This let us think that \textbf{the best choice of greediness is problem dependent}, something that goes beyond our theoretical analysis.

Last, we discuss the effect of entropy. As already noticed, for a given level of entropy, medium to large values of the KL parameter improve performance, suggesting that entropy works better in conjunction with KL, something appearing in our bound. Now, observing the table corresponding to $\tau=0$ (no entropy), we observe that we can obtain comparable best performance with solely a KL regularization, especially for MD. This suggests that \textbf{entropy is better with KL, and KL alone might be sufficient}. We already explained that some beneficial aspects of entropy, like exploration or better optimization landscape, are not explained by our analysis. However, we hypothesize that KL might have similar benefits. For examples, entropy enforces stochastic policies, which helps for exploration. KL has the same effect (if the initial policy is uniform), but in an adaptive manner (exploration decreases with training time).

\section{Conclusion}
\label{sec:conclusion}

We provided an explanation of the effect of KL regularization in RL, through the implicit averaging of $q$-values. We provided a very strong performance bound for KL regularization, the very first RL bound showing both a linear dependency to the horizon and an averaging the estimation errors. We also analyzed the effect of KL regularization with an additional entropy term. The introduced abstract framework encompasses a number of existing approaches, but some assumptions we made do not hold when neural networks are used. Therefore, we complemented our thorough theoretical analysis with an extensive empirical study. It confirms that KL regularization is helpful, and that regularizing the evaluation step is never detrimental. It also suggests that KL regularization alone, without entropy, might be sufficient (and better than entropy alone).

The core issue of our analysis is that it relies heavily on the absence of errors in the greedy step, something we deemed impossible with neural networks. However, \citet{vieillard2020munchausen} proposed subsequently a reperameterization of our regularized approximate dynamic scheme. The resulting approach, called ``Munchausen Reinforcement Learning'', is simple and general, and provides agents outperforming the state of the art. Crucially, thanks to this reparameterization, there's no error in their greedy step and our bounds apply readily. More details can be found in~\cite{vieillard2020munchausen}.

\clearpage 

\paragraph{Broader impact.}
Our core contribution is theoretical. We unify a large body of the literature under KL-regularized reinforcement learning, and provide strong performance bounds, among them the first one ever to combine a linear dependency to the horizon and an averaging of the errors. We complement these results with an empirical study. It shows that the insights provided by the theory can still be used in a deep learning context, when some of the assumptions are not satisfied. As such, we think the broader impact of our contribution to be the same as the one of reinforcement learning. 

\paragraph{Funding transparency statement.} Nothing to disclose.

\bibliography{bib}
\bibliographystyle{plainnat}

\clearpage
\onecolumn

\appendix

\paragraph{Content.} These appendices complement the core paper with the following:
\begin{itemize}
    \item Appx.~\ref{appx:LF} is a warm-up that states a few facts about the Legendre-Fenchel transform, useful all along the derivations.
    \item Appx.~\ref{appx:connections-algorithms} justifies the connections drawn in Sec.~\ref{sec:regMPI} between MD-MPI or DA-MPI and the literature.
    \item Appx.~\ref{appx:proofs} provides the proofs of all stated theoretical results, as well as some necessary lemmata.
    \item Appx.~\ref{appx:garnet} provides details about the experiment used to illustrate the bounds in Sec.~\ref{sec:analysis}.
    \item Appx.~\ref{appx:exp} provides additional details regarding the practical algorithms and the experiments, as well as additional experiments and visualisations.
\end{itemize}

\section{Convex Conjugacy for KL and Entropy Regularization}
\label{appx:LF}

Let $q\in\R^{\s\times\A}$ and $\mu\in\Delta_\A^\s$, and consider the general greedy step $\pi'\in\gc^{\lambda,\tau}_\mu$, the optimization being understood here state-wise.
\begin{equation}
    \pi' \in \argmax_{\pi\in\Delta_\A^\s} \left(\langle \pi, q \rangle - \lambda \kl(\pi||\mu) + \tau \hc(\pi)\right).\label{eq:appx:greedy_general}
\end{equation}
The function $\lambda\kl(\pi||\mu) - \tau\hc(\pi)$ being convex in $\pi$, this optimization problem is related to the Legendre-Fenchel transform (\textit{e.g.}, \citet[Ch.~E]{hiriart2012fundamentals}), or convex conjugate (which is the maximum rather than the maximizer). First, we consider a simple case, $\lambda=0$ and $\tau=1$. It is well known in this case that the maximum (the convex conjugate) is the log-sum-exp function and the maximizer (the gradient of the convex conjugate) is the softmax (\textit{e.g.}, \citet[Ex.~3.25]{boyd2004convex}):
\begin{align}
    \max_{\pi\in\Delta_\A^\s}  \left(\langle \pi, q \rangle + \hc(\pi)\right) &= \ln \langle \un, \exp q\rangle \in\R^\s,
    \\
    \argmax_{\pi\in\Delta_\A^\s}  \left(\langle \pi, q \rangle + \hc(\pi)\right) &= \frac{\exp q}{\langle \un, \exp q\rangle} \in\R^{\s\times\A},
\end{align}
with $\un\in\R^{\s\times\A}$ the vector of which all components are equal to 1. We made use of the notations introduced in Sec.~\ref{sec:background}, and overload $v\in\R^\s$ to $v\in\R^{\s\times\A}$ as $v(s,a) = v(s)$. To make things clear, it gives
\begin{align}
    \left[\ln \langle \un, \exp q\rangle\right](s) &=  \ln \sum_{a\in\A} \exp q(s,a)
    \\
    \text{and }
    \left[\frac{\exp q}{\langle \un, \exp q\rangle}\right](s,a) &= \frac{\exp q(s,a)}{\sum_{a'\in\A} q(s,a')}.
\end{align}
Notice also that a direct consequence of this is that
\begin{equation}
    \ln \langle \un, \exp q\rangle = \langle \pi', q \rangle + \hc(\pi') \text{ with } \pi' = \frac{\exp q}{\langle \un, \exp q\rangle}.
\end{equation}
From this simple case, we can easily handle the general case. We have
\begin{align}
    \langle \pi, q \rangle - \lambda \kl(\pi||\mu) + \tau \hc(\pi)
    &= \langle \pi, q \rangle - \lambda\langle \pi, \ln\pi - \ln \mu \rangle - \tau \langle \pi, \ln\pi\rangle
    \\
    &= \langle \pi, q  + \lambda\ln\mu \rangle - (\lambda+\tau)\langle\pi,\ln\pi\rangle
    \\
    &= (\lambda+\tau)\left( \left\langle \pi, \frac{q + \lambda \ln\mu}{\lambda+\tau} \right\rangle + \hc(\pi) \right).
\end{align}
From this, we can deduce directly that the maximum of~\eqref{eq:appx:greedy_general} is
\begin{align}
    \max_{\pi\in\Delta_\A^\s} \left(\langle \pi, q \rangle - \lambda \kl(\pi||\mu) + \tau \hc(\pi)\right)
    &= (\lambda+\tau)\ln\left\langle1, \exp \frac{q +\lambda\ln\mu}{\lambda+\tau}\right\rangle
    \\
    &= (\lambda+\tau)\ln \left\langle\mu^{\frac{\lambda}{\lambda+\tau}}, \exp \frac{q}{\lambda+\tau}\right\rangle
    \label{eq:appx:maximum}
    \\
    &= (\lambda+\tau) \left(\ln \sum_{a\in\A} \mu(a|s)^{\frac{\lambda}{\lambda+\tau}} \exp\frac{q(s,a)}{\lambda+\tau}\right)_{s\in\s}, 
\end{align}
and that the maximizer of~\eqref{eq:appx:greedy_general} is
\begin{align}
    \argmax_{\pi\in\Delta_\A^\s} \left(\langle \pi, q \rangle - \lambda \kl(\pi||\mu) + \tau \hc(\pi)\right)
    &= \frac{\exp \frac{q +\lambda\ln\mu}{\lambda+\tau}}{\langle \un, \exp \frac{q +\lambda\ln\mu}{\lambda+\tau}\rangle}
    = \frac{\mu^{\frac{\lambda}{\lambda+\tau}} \exp \frac{q}{\lambda+\tau}}{\langle \un, \mu^{\frac{\lambda}{\lambda+\tau}} \exp \frac{q}{\lambda+\tau}\rangle}
    \label{eq:appx:maximizer}
    \\
    &= \left(\frac{ \mu(a|s)^{\frac{\lambda}{\lambda+\tau}} \exp\frac{q(s,a)}{\lambda+\tau}}{\sum_{a'\in\A}  \mu(a'|s)^{\frac{\lambda}{\lambda+\tau}} \exp\frac{q(s,a')}{\lambda+\tau}}\right)_{(s,a)\in\s\times\A}
\end{align}
Again, the relationship between the maximum and the maximizer gives
\begin{align}
    (\lambda+\tau)\ln \left\langle\mu^{\frac{\lambda}{\lambda+\tau}}, \exp \frac{q}{\lambda+\tau}\right\rangle
    &= \langle \pi', q \rangle - \lambda \kl(\pi'||\mu) + \tau \hc(\pi')
    \label{eq:appx:relationship}
    \\
    \text{with }
    \pi' &= \frac{\mu^{\frac{\lambda}{\lambda+\tau}} \exp \frac{q}{\lambda+\tau}}{\langle \un, \mu^{\frac{\lambda}{\lambda+\tau}} \exp \frac{q}{\lambda+\tau}\rangle}.
\end{align}

\section{Connections to existing algorithms}
\label{appx:connections-algorithms}

In this section, we justify the connections stated in Sec.~\ref{sec:regMPI} between the considered regularized ADP schemes and the literature.

\subsection{Connection of MD-MPI($\lambda$,$\tau$) to other algorithms}
\label{subappx:conection_mdmpi}

\paragraph{Connection to SAC.} We stated that SAC~\citep{haarnoja2018soft} is a variation of MD-MPI(0,$\tau$). SAC was introduced as PI scheme ($m=\infty$), while it is practically implemented as VI scheme ($m=1$). We keep the VI viewpoint for this discussion. The MD-VI(0,$\tau$) scheme is given by
\begin{equation}
    \begin{cases}
        \pi_{k+1} = \gc^{0,\tau}(q_k)
        \\
        q_{k+1} = T_{\pi_{k+1}}^{0,\tau} q_k + \epsilon_{k+1}
    \end{cases}.\label{eq:appx:mdvi-sac}
\end{equation}
The regularized Bellman operator can be rewritten as follows:
\begin{align}
    T_{\pi_{k+1}}^{0,\tau} q_k &= T_{\pi_{k+1}} q_k + \gamma P \tau \hc(\pi_{k+1})
    = r + \gamma P \left( \langle \pi_{k+1}, q_k \rangle - \tau \langle \pi_{k+1}, \ln \pi_{k+1}\rangle \right)
    \\
    &= r + \gamma P \langle \pi_{k+1}, q_k - \tau \ln\pi_{k+1}\rangle.
\end{align}
This is exactly the Bellman operator considered in SAC. For the greedy step, we have directly from Eq.~\eqref{eq:appx:maximizer} that $\pi_{k+1} \propto \exp \frac{q_k}{\tau}$. In SAC, continuous actions are considered, so the policy cannot be computed (due to the partition function). Therefore, it is approximated with a neural network by minimizing a reverse KL divergence (that allows getting rid of the partition function) between the neural policy and the target policy (the solution of the original greedy step):
\begin{equation}
    \pi_{k+1} = \argmin_{\pi_\theta} \E_s[\kl(\pi_\theta||\pi^*_{k+1})]
    = \argmin_{\pi_\theta} \E_s[\kl(\pi_\theta||\exp \frac{q}{\tau})] \text{ with } \pi^*_{k+1}
    = \frac{\exp\frac{q_k}{\tau}}{\langle \un, \exp\frac{q_k}{\tau}\rangle}.
\end{equation}

\paragraph{Connection to Soft Q-learning.} We stated that Soft Q-learning~\citep{fox2015taming,haarnoja2017reinforcement} is also a variation of MD-MPI(0,$\tau$). It is indeed a VI scheme, so a variation of MD-VI(0,$\tau$) depicted in Eq.~\eqref{eq:appx:mdvi-sac}. As a direct consequence of Eq.~\eqref{eq:appx:relationship}, $\pi_{k+1}\propto \exp \frac{q_k}{\tau}$ being the maximizer, we have
\begin{equation}
    \langle \pi_{k+1}, q_k\rangle + \tau \hc(\pi_{k+1}) = \tau \ln\langle \un, \exp \frac{q_k}{\tau}\rangle. 
\end{equation}
This allows rewriting the evaluation step as follows:
\begin{align}
    q_{k+1} &= T_{\pi_{k+1}}^{0,\tau} q_k + \epsilon_{k+1}
    \\
    &= r + \gamma P \left(\langle \pi_{k+1}, q_k\rangle + \tau \hc(\pi_{k+1})\right) + \epsilon_{k+1}
    \\
    \Leftrightarrow
    q_{k+1} &= r+ \gamma P \left(\tau \ln\langle \un, \exp \frac{q_k}{\tau}\rangle\right) + \epsilon_{k+1}.\label{eq:appx:softQL}
\end{align}
Eq.~\eqref{eq:appx:softQL} is equivalent to Eq.~\eqref{eq:appx:mdvi-sac}, and it is the Bellman operator upon which Soft Q-learning is built (replacing the hard maximum by the log-sum-exp). \citet{haarnoja2017reinforcement} additionally handle continuous actions, which requires some refinements. 

\paragraph{Connection to Softmax DQN.} We stated that Softmax DQN~\citep{song2019revisiting} is a variation of MD-MPI(0,$\tau$), but \wor{} (without regularization in the evaluation step). Therefore, it is scheme~\eqref{eq:appx:mdvi-sac}, but replacing $T_{\pi_{k+1}}^{0,\tau}$ by $T_{\pi_{k+1}}$:
\begin{equation}
    \begin{cases}
        \pi_{k+1} = \gc^{0,\tau}(q_k)
        \\
        q_{k+1} = T_{\pi_{k+1}} q_k + \epsilon_{k+1}
    \end{cases}.
\end{equation}
Given that $\pi_{k+1}\propto\exp\frac{q_k}{\tau}$, this amounts to iterating the following so called softmax operator
\begin{align}
    q_{k+1} &=  T_{\pi_{k+1}} q_k + \epsilon_{k+1}
    \\
    &= r + \gamma P \left\langle \frac{\exp\frac{q_k}{\tau}}{\langle \un,\exp\frac{q_k}{\tau}\rangle}, q_k \right\rangle + \epsilon_{k+1},
\end{align}
which is the core update rule of softmax DQN. Notice that this operator might not be a contraction (depending on the value of $\tau$), and that it can have multiple fixed points~\citep{asadi2016alternative}.

\paragraph{Connection to the mellowmax policy.} \citet{asadi2016alternative} introduced a so-called mellowmax policy as a convergent alternative to the softmax operator. This can be indeed seen as an alternative way of regularizing the evaluation step. We explain here why. To do so, we reframe the mellowmax idea with our notations. \citet{asadi2016alternative} introduced the mellowmax operator as
\begin{equation}
    \mm_\tau(q) = \tau \ln\left\langle \un, \frac{1}{|\A|} \exp\frac{q}{\tau}\right\rangle. 
\end{equation}
One can easily see that it is indeed the convex conjugate of the KL with respect to the uniform policy (that behaves like the entropy). Indeed, from Eq.~\eqref{eq:appx:maximum}, we have directly that
\begin{equation}
    \mm_\tau(q) = \max_{\pi\in\Delta^\s_\A} \left(\langle \pi, q\rangle - \tau \kl(\pi||\pi_U)\right),
\end{equation}
with $\pi_U$ the uniform policy. From~\citet{geist2019theory}, we know that the following equivalent schemes,
\begin{equation}
    \begin{cases}
        \pi_{k+1} =  \argmax_{\pi\in\Delta^\s_\A} \left(\langle \pi, q\rangle - \tau \kl(\pi||\pi_U)\right)
        \\
        q_{k+1} = T_{\pi_{k+1}} q_k - \gamma P \tau \kl(\pi_{k+1}||\pi_U)
    \end{cases}
    \Leftrightarrow
    q_{k+1} = r + \gamma P \mm_\tau(q_k),
\end{equation}
are convergent (MDP regularized with $\tau\kl(\cdot||\pi_U)$, the equivalence being from Eq.~\eqref{eq:appx:relationship}). This is not the viewpoint of \citet{asadi2016alternative}. They try to find a policy $\pi'_{k+1}$ such that $q_{k+1} = r + \gamma P \mm_\tau(q_k) = r + \gamma P\langle \pi'_{k+1}, q_k\rangle$. To account for the possible existence of multiple policies, they look for the one with maximal entropy and solve (numerically) for
\begin{equation}
    \pi'_{k+1} = \max_{\pi\in\Delta^\s_\A: \langle \pi, q_k\rangle = \mm_\tau(q_k)} \hc(\pi).
\end{equation}
Then, they apply $q_{k+1} = r + \gamma P\langle \pi'_{k+1}, q_k\rangle$. If there is no error when computing $\pi'_{k+1}$, this is equivalent to adding the regularization to the evaluation step.

\paragraph{Connection to TRPO.} We stated that TRPO~\citep{schulman2015trust} is a variation of MD-MPI($\lambda$, 0), \wor{}. More precisely, it is a variation of MD-PI($\lambda$, 0):
\begin{equation}
    \begin{cases}
        \pi_{k+1} =\gc^{\lambda, 0}(q_k)
        \\
        q_{k+1} = T_{\pi_{k+1}}^\infty q_k + \epsilon_k = q_{\pi_{k+1}} + \epsilon_k
    \end{cases}.
    \label{eq:trpo-mdmpi}
\end{equation}
In TRPO, the $q$-function is evaluated using Monte Carlo rollouts. The greedy policy is approximated with a neural network by directly solving the expected greedy step:
\begin{equation}
    \pi_{k+1} = \argmin_{\pi_\theta} \E_s[\langle \pi_\theta, q_k\rangle - \lambda\kl(\pi_\theta||\pi_k)].
\end{equation}
TRPO is indeed a bit different, as it uses importance sampling to sample actions according to $\pi_k$ (which is especially useful for continuous actions, but does not change the objective function), it uses a constraint based on the KL rather than a regularization, and it considers the KL in the other direction:
\begin{align}
    \pi_{k+1} = \argmin_{\pi_\theta:\E_s[\kl(\pi_k||\pi_\theta)]\leq \epsilon} \E_s[\E_{a\sim \pi_k(.|s)}[\langle \frac{\pi_\theta}{\pi_k}, q_k\rangle]].
\end{align}
However, from an abstract viewpoint, TRPO is close to scheme~\eqref{eq:trpo-mdmpi}.

\paragraph{Connection to MPO.} We stated that MPO~\citep{abdolmaleki2018maximum} is also a variation of MD-MPI($\lambda$, 0), \wor{}:
\begin{equation}
    \begin{cases}
        \pi_{k+1} =\gc^{\lambda, 0}(q_k)
        \\
        q_{k+1} = T_{\pi_{k+1}}^m q_k + \epsilon_k 
    \end{cases}.
    \label{eq:mpo-mdmpi}
\end{equation}
The evaluation step is done by combining a TD approach with eligibility traces (a geometric average of $m$-step returns), rather than using $m$-step returns (that amounts to using the $T_\pi^m$ operator). For the greedy step, the analytic solution can be computed for any state-action couple, and generalized to the whole state-action space by minimizing a KL between this analytical solution and a neural network:
\begin{align}
    \pi_{k+1} &= \argmin_{\pi_\theta} \E_s[\kl(\pi^*_{k+1}||\pi_\theta)] = \argmax_{\pi_\theta} \E_s[\E_{a\sim \pi^*_{k+1}(.|s)}[\ln \pi_\theta(a|s)]] 
    \\
    \text{with } \pi^*_{k+1} &= \frac{\pi_k \exp\frac{q_k}{\lambda}}{\langle \un,  \pi_k \exp\frac{q_k}{\lambda} \rangle}.
\end{align}
The greedy step of MPO is indeed a bit different, the algorithm being derived from an expectation-maximization principle based on a probabilistic inference view of RL. The term $\lambda$ is not fixed but learnt by the minimization of a convex dual function (coming from viewing the KL term as a constraint rather than a regularization), and an additional KL penalty is added (not necessarily redundant with the initial one, as the KL there is in the other direction):
\begin{equation}
    \pi_{k+1} = \argmax_{\pi_\theta:\E_s[\kl(\pi_k||\pi_\theta)]\leq \epsilon} \E_s[\E_{a\sim \pi^*_{k+1}(.|s)}[\ln \pi_\theta(a|s)]].
\end{equation}
However, from an abstract viewpoint, MPO is close to scheme~\eqref{eq:mpo-mdmpi}.

\paragraph{Connection to DPP.} We stated that DPP~\citep{azar2012dynamic} is a variation of MD-MPI($\lambda$, 0). More precisely, it is close to be a reparameterization of MD-VI($\lambda$, 0), the difference being mainly the error term:
\begin{equation}
    \begin{cases}
        \pi_{k+1} =\gc^{\lambda, 0}(q_k)
        \\
        q_{k+1} = T_{\pi_{k+1}}^{\lambda,0} q_k + \epsilon_k
    \end{cases}.
    \label{eq:dpp-mdmpi}
\end{equation}
To derive the DPP update rule from Eq.~\eqref{eq:dpp-mdmpi}, we consider $\epsilon_k=0$. The greedy policy is, according to~\eqref{eq:appx:maximizer},
\begin{equation}
    \pi_{k+1} = \frac{\pi_k\exp\frac{q_k}{\lambda}}{\langle \un, \pi_k\exp\frac{q_k}{\lambda} \rangle}.
\end{equation}
Define $v_{k+1}$ as (the second equality coming from Eq.~\eqref{eq:appx:relationship})
\begin{equation}
    v_{k+1} = \langle \pi_{k+1}, q_k\rangle -\lambda \kl(\pi_{k+1}||\pi_k) = \lambda \ln\langle \pi_k, \exp\frac{q_k}{\lambda}\rangle.
\end{equation}
With this, we have
\begin{equation}
    q_{k+1} = T_{\pi_{k+1}}^{\lambda,0} q_k
    = r + \gamma P (\langle \pi_{k+1}, q_k\rangle -\lambda \kl(\pi_{k+1}||\pi_k))
    = r + \gamma P v_{k+1}
\end{equation}
Let us define $\psi_{k+1}\in\mathbb{R}^{\s\times\A}$ as
\begin{equation}
    \psi_{k+1} = \lambda \ln\left(\pi_k \exp\frac{q_k}{\lambda}\right) = r + \gamma P v_k + \lambda\ln\pi_k.
    \label{eq:dpp:psi}
\end{equation}
Thus, we have
\begin{align}
    \pi_k &= \frac{\exp\frac{\psi_k}{\lambda}}{\langle \un, \exp\frac{\psi_k}{\lambda} \rangle}
    \label{eq:dpp:pi}
    \\ \text{and }
    v_k &= \lambda \ln \langle \un, \frac{\psi_k}{\lambda}\rangle. 
    \label{eq:dpp:v}
\end{align}
Injecting Eqs.~\eqref{eq:dpp:pi} and~\eqref{eq:dpp:v} into~\eqref{eq:dpp:psi}, we get
\begin{equation}
    \psi_{k+1} = r + \gamma P \lambda \ln\langle \un,\frac{\psi_k}{\lambda}\rangle + \psi_k - \lambda\ln\langle \un,\frac{\psi_k}{\lambda}\rangle.
\end{equation}
This is how DPP is justified from a DP viewpoint~\citep[Appx.~A]{azar2012dynamic}. It is a bit different from the DPP algorithm analyzed by~\citet{azar2012dynamic}, for which $\ln\langle \un,\frac{\psi_k}{\lambda}\rangle$ is replaced by $\langle \pi_{k}, \psi_k\rangle$ (both terms being equal in the limit $\lambda\rightarrow 0$), and that consider an estimation error $\epsilon'_{k+1}$:
\begin{equation}
    \psi_{k+1} = r + \gamma P \langle \pi_{k}, \psi_k\rangle + \psi_k - \langle \pi_{k}, \psi_k\rangle + \epsilon'_{k+1}.
\end{equation}
We advocate that the error $\epsilon'_k$ is usually harder to control than $\epsilon_k$ (or equivalently that $q_k$ is easier to estimate than $\psi_k$), because the function $\psi_*$ (the optimal $\psi$-function for the MDP) is equal to $-\infty$ for any suboptimal action~\citep[Cor.~ 4]{azar2012dynamic}.

\paragraph{Connection to CVI.} We stated that CVI is a reparametrization of MD-VI($\lambda$,$\tau$), that we recall (without the error term, to do the reparameterization):
\begin{equation}
    \begin{cases}
        \pi_{k+1} =\gc^{\lambda, \tau}(q_k)
        \\
        q_{k+1} = T_{\pi_{k+1}}^{\lambda,\tau} q_k 
    \end{cases}.
    \label{eq:cvi-mdmpi}
\end{equation}
We now show how to derive the CVI update rule from this. The regularized greedy policy is, thanks to Eq.~\eqref{eq:appx:maximizer}, and writing $\beta = \frac{\lambda}{\lambda + \tau}$:
\begin{equation}
    \pi_{k+1} = \frac{\pi_k^\beta \exp\frac{\beta q_k}{\lambda}}{\langle \un,  \pi_k^\beta \exp\frac{\beta q_k}{\lambda} \rangle}.
\end{equation}
Similarly to DPP, we can define $v_{k+1}$ as (still using Eq.~\eqref{eq:appx:relationship} for the second equality):
\begin{equation}
    v_{k+1} = \langle \pi_{k+1}, q_k\rangle -\lambda \kl(\pi_{k+1}||\pi_k) + \tau\hc(\pi_{k+1}) = \frac{\lambda}{\beta} \ln\langle \pi_k^\beta, \exp\frac{\beta q_k}{\lambda}\rangle.
\end{equation}
With this, we have
\begin{equation}
    q_{k+1} = T_{\pi_{k+1}}^{\lambda,0} q_k
    = r + \gamma P (\langle \pi_{k+1}, q_k\rangle -\lambda \kl(\pi_{k+1}||\pi_k) + \tau\hc(\pi_{k+1}))
    = r + \gamma P v_{k+1}.
\end{equation}
Let us define $\psi_{k+1}\in\mathbb{R}^{\s\times\A}$ as
\begin{equation}
    \psi_{k+1} = \frac{\lambda}{\beta} \ln\left(\pi_k^\beta \exp\frac{\beta q_k}{\lambda}\right) = r + \gamma P v_k + \lambda\ln\pi_k.
    \label{eq:cvi:psi}
\end{equation}
Thus, we have
\begin{align}
    \pi_k &= \frac{\exp\frac{\beta\psi_k}{\lambda}}{\langle \un, \exp\frac{\beta\psi_k}{\lambda} \rangle}
    \label{eq:cvi:pi}
    \\ \text{and }
    v_k &= \frac{\lambda}{\beta} \ln \langle \un, \frac{\beta \psi_k}{\lambda}\rangle. 
    \label{eq:cvi:v}
\end{align}
Injecting Eqs.~\eqref{eq:cvi:pi} and~\eqref{eq:cvi:v} into~\eqref{eq:cvi:psi}, we get
\begin{equation}
    \psi_{k+1} = r + \gamma P \frac{\lambda}{\beta} \ln\langle \un,\frac{\beta \psi_k}{\lambda}\rangle + \beta (\psi_k - \frac{\lambda}{\beta}\ln\langle \un,\frac{\beta\psi_k}{\lambda}\rangle).
\end{equation}
This is exactly the CVI update rule. Notice that setting $\beta=1$, i.e., $\tau=0$ (no entropy term), we retrieve DPP (which was to be expected). As we obtain CVI, by considering $\lambda + \tau \rightarrow 0$ while keeping $\beta = \frac{\lambda}{\lambda + \tau}$ constant, we retrieve advantage learning in the limit~\citep{baird1999reinforcement,bellemare2016increasing}, that DA-VI($\lambda$,$\tau$) thus generalizes.

\subsection{Connection of DA-MPI($\lambda$,$\tau$) to other algorithms}
\label{subappx:connection_dampi}

\paragraph{Connection to Politex.} Politex~\citep{abbasi2019politex} addresses the average reward criterion. It is a PI scheme, up to the fact that the policy, instead of being greedy according to the last $q$-function, is softmax according to the sum of all past $q$-function. In the discounted reward case considered here, this is exactly DA-PI($\lambda$,0), \wor{} (without regularization in the evaluation step):
\begin{equation}
    \begin{cases}
        \pi_{k+1} = \gc^{0,\frac{\lambda}{k+1}}(h_k)
        \\
        q_{k+1} = T_{\pi_{k+1}}^\infty q_k + \epsilon_{k+1} = q_{\pi_{k+1}} + \epsilon_{k+1}
        \\
        h_{k+1} = \frac{k+1}{k+2} h_k + \frac{1}{k+2} q_{k+1}
    \end{cases}
\end{equation}
Indeed, by definition $h_k = \frac{1}{k+1}\sum_{j=0}^k q_j$ and the greedy policy is
\begin{equation}
    \pi_{k+1} =\gc^{0,\frac{\lambda}{k+1}}(h_k) = \frac{\exp\frac{(k+1) h_k}{\lambda}}{\langle \un, \exp\frac{(k+1) h_k}{\lambda}\rangle} = \frac{\exp\frac{\sum_{j=0}^k q_j}{\lambda}}{\langle \un, \exp\frac{\sum_{j=0}^k q_j}{\lambda}\rangle}.
\end{equation}
This is exactly the Politex algorithm, but for the discounted reward case (that changes how the $q$-function is defined, and thus estimated).

\paragraph{Connection to MoVI.} MoVI~\citep{vieillard2019momentum} is a VI scheme, up to the fact that the policy, instead of being greedy according to the last $q$-function, is greedy according to the average of past $q$-functions. It is indeed is a limiting case of DA-VI($\lambda$, 0), $\wor$: 
\begin{equation}
    \begin{cases}
        \pi_{k+1} = \gc^{0,\frac{\lambda}{k+1}}(h_k)
        \\
        q_{k+1} = T_{\pi_{k+1}} q_k + \epsilon_{k+1}
        \\
        h_{k+1} = \frac{k+1}{k+2} h_k + \frac{1}{k+2} q_{k+1}
    \end{cases}.
\end{equation}
It is well known that the limit of a softmax, when the temperatures goes to zero, is the greedy policy: $\gc^{0,\frac{\lambda}{k+1}}(h_k) \rightarrow \gc(h_k)$ as $\lambda \rightarrow 0$. So, DA-VI($\lambda\rightarrow 0$, 0), \wor{}, is the following scheme,
\begin{equation}
    \begin{cases}
        \pi_{k+1} \in \gc(h_k)
        \\
        q_{k+1} = T_{\pi_{k+1}} q_k + \epsilon_{k+1}
        \\
        h_{k+1} = \frac{k+1}{k+2} h_k + \frac{1}{k+2} q_{k+1}
    \end{cases},
\end{equation}
that is exactly MoVI. Notice that it is different from MD-VI($\lambda\rightarrow 0$, 0), \wor, which is AVI (see also Prop.~\ref{thm:equivalence}).

\paragraph{Connection to momentum DQN.} Momentum DQN~\citep{vieillard2019momentum} was introduced as a practical heuristic to MoVI, changing the exact average by a moving average (more amenable to optimization with deep networks). We show below that it is indeed a limiting case of DA-VI($\lambda$,$\tau$), \wor{} (without regularized greedy step), that is:
\begin{equation}
    \begin{cases}
        \pi_{k+1} = \gc^{0,\tau}(h_k)
        \\
        q_{k+1} = T_{\pi_{k+1}} q_k + \epsilon_{k+1}
        \\
        h_{k+1} = \beta h_{k} + (1-\beta) q_{k+1} \text{ with } \beta = \frac{\lambda}{\lambda+\tau}
    \end{cases}.
\end{equation}
Fix $\beta\in(0,1)$, we can consider $\lambda,\tau\rightarrow 0$ with $\beta = \frac{\lambda}{\lambda+\tau}$ kept constant. In this case, the regularized greedy operator tends to the usual greedy one: $\gc^{0,\tau}(h_k) \rightarrow \gc(h_k)$ as $\tau\rightarrow 0$. In the limit, we obtain the following scheme,
\begin{equation}
    \begin{cases}
        \pi_{k+1} = \gc(h_k)
        \\
        q_{k+1} = T_{\pi_{k+1}} q_k + \epsilon_{k+1}
        \\
        h_{k+1} = \beta h_{k} + (1-\beta) q_{k+1}
    \end{cases}
\end{equation}
for a chosen $\beta$, which is exactly momentum DQN with fixed $\beta$.

\paragraph{Connection to Speedy Q-learning.} We stated that Speedy Q-learning~\citep{azar2011speedy} is a limiting case of DA-VI($\lambda$,0), which we recall (without the error term here):
\begin{equation}
    \begin{cases}
        \pi_{k+1} = \gc^{0,\frac{\lambda}{k+1}}(h_k)
        \\
        q_{k+1} = T_{\pi_{k+1}}^{\lambda,0} q_k 
        \\
        h_{k+1} = \frac{k+1}{k+2} h_k + \frac{1}{k+2} q_{k+1}
    \end{cases}.
\end{equation}
As shown in Lemma~\ref{lemma:q_to_h} in Appx.~\ref{subsec:proof:davi_type1_noEntropy}, we have
\begin{equation}
    T_{\pi_{k+1}\sep\pi_k}^{\lambda,0} q_k = (k+1) T_{\pi_{k+1}}^{0,\frac{\lambda}{k+1}} h_k - k T_{\pi_k}^{0,\frac{\lambda}{k}} h_{k-1}.
\end{equation}
With this, DA-VI$_1$($\lambda$,0) can be expressed solely in terms of $h_k$ and $\pi_k$:
\begin{equation}
    \begin{cases}
        \pi_{k+1} = \gc^{0,\frac{\lambda}{k+1}}(h_k)
        \\
        h_{k+1} = \frac{k+1}{k+2} h_k + \frac{1}{k+2} \left((k+1) T_{\pi_{k+1}}^{0,\frac{\lambda}{k+1}} h_k - k T_{\pi_k}^{0,\frac{\lambda}{k}} h_{k-1} \right).
    \end{cases}
    \label{eq:davi_only_with_h}
\end{equation}
As before, as $\lambda\rightarrow 0$, the regularized greedy step tends to the greedy step, $\gc^{0,\frac{\lambda}{k+1}}(h_k) \rightarrow \gc(h_k)$. Regarding the evaluation step, we can write, by definition of the regularized Bellman operator and using Eq.~\eqref{eq:appx:relationship},
\begin{align}
    T_{\pi_{k+1}}^{0,\frac{\lambda}{k+1}} h_k &= 
    r + \gamma P \left(\langle\pi_{k+1}, h_k \rangle + \frac{\lambda}{k+1} \hc(\pi_{k+1})\right)
    \\
    &= r + \gamma P \left(\frac{\lambda}{k+1}\ln\langle \un, \exp\frac{(k+1)h_k}{\lambda}\rangle\right).
\end{align}
It is a classical result that the convex conjugate of the entropy tends to the hard maximum as the associated temperature goes to zero. For any $s\in\s$,
\begin{equation}
    \lim_{\lambda\rightarrow 0} \frac{\lambda}{k+1} \ln\sum_{a\in\A} \exp\frac{(k+1)h_k(s,a)}{\lambda} = \frac{1}{k+1} \max_{a\in\A}\left( (k+1) h_k(s,a)\right) = \max_{a\in\A} h_k(s,a).
\end{equation}
Writing $T_*$ the Bellman optimality operator, defined as $T_*q = \max_\pi T_\pi q$, we thus have
\begin{equation}
    \lim_{\lambda\rightarrow 0} T_{\pi_{k+1}}^{0,\frac{\lambda}{k+1}} h_k = T_* h_k.
\end{equation}
Thus, writing the limit of scheme~\eqref{eq:davi_only_with_h} as $\lambda\rightarrow 0$, we obtain
\begin{equation}
    h_{k+1} = (1-\frac{1}{k+2}) h_k + \frac{1}{k+2} \left((k+1) T_* h_k - k T_* h_{k-1} \right),
\end{equation}
which is exactly the Speedy Q-learning update rule.

\paragraph{Connection to softened LSPI.}
\citep{perolat2016softened} address the problem of learning a Nash equilibria in zero-sum Markov games. They show that state of the art algorithms can be derived by minimizing the norm of the (projected) Bellman residual using a Newton descent, and propose more stable algorithms by using instead a quasi-Newton descent. Single agent reinforcement learning is a special case of zero-sum Markov games, and in this case the algorithm they propose can be written as follows, in an abstract way\footnote{Specialized to single agent RL, their algorithm adopts a linear parameterization of the $q$-function and estimate $q_{\pi_{k+1}}$ either with LSTD~\citep{bradtke1996linear} or by minimizing the norm of the Bellman residual.}:
\begin{equation}
    \begin{cases}
        \pi_{k+1} \in \gc(h_k)
        \\
        q_{k+1} = T_{\pi_{k+1}}^\infty q_k + \epsilon_{k+1} = q_{\pi_{k+1}} + \epsilon_{k+1}
        \\
        h_{k+1} = \beta h_k + (1-\beta) q_{k+1}
    \end{cases}.
\end{equation}
Using the same arguments as for the connection to momentum DQN, this is a limit case of DA-PI($\lambda$,$\tau$), \wor, as $\lambda,\tau\rightarrow 0$ with $\beta = \frac{\lambda}{\lambda+\tau}$ kept constant. It is also closely related to Politex (the policy is greedy instead of being softmax, moving average of the $q$-values instead of an average).

\section{Proofs of Theoretical Results}
\label{appx:proofs}

In this section, we prove the results stated in the paper.

\subsection{Proof of Proposition~\ref{thm:equivalence}}
\label{subappx:proof_equivalence}

\paragraph{Sketch of proof.}

As explained in the paper, the optimization problem $\pi_{k+1} = \gc_{\pi_k}^{\lambda,0} (q_k)$ can be solved analytically, yielding $\pi_{k+1}\propto \pi_k \exp\frac{q_k}{\lambda}$. By direct induction, $\pi_0$ being uniform, we have %
$ %
    \pi_{k+1}\propto \pi_k \exp\frac{q_k}{\lambda} \propto \dots \propto \exp \frac{1}{\lambda}\sum_{j=0}^k q_j
$. %
Thus, the policy is indeed softmax according to the sum of $q$-values. Defining $h_k$ as the average of past $q$-values basically provides the stated DA-VI($\lambda$,0). The case with an additionnal entropy term is a bit more involved, but the principle is the same.

\paragraph{Proof.}

We start by proving the equivalence for the case $\tau=0$.
Recall that we assumed, with little loss of generality, that $\pi_0$ is the uniform policy. We recall MD-MPI($\lambda$,0):
\begin{equation}
    \begin{cases}
        \pi_{k+1} = \gc_{\pi_k}^{\lambda,0}(q_k)
        \\
        q_{k+1} = (T_{\pi_{k+1}\sep\pi_k}^{\lambda,0})^m q_k + \epsilon_{k+1}
    \end{cases}.
    \label{eq:appx:proof_equiv_md0}
\end{equation}
Let us define $h_0=q_0$ and $h_k$ for $k\geq 1$ as the average of past $q$-functions.
\begin{align}
    h_k = \frac{1}{k+1}\sum_{j=0}^k q_j = \frac{k}{k+1} h_{k-1} + \frac{1}{k+1} q_k.
\end{align}
As a direct consequence of Eq.~\eqref{eq:appx:maximizer}, we have that $\pi_{k+1}\propto \pi_k\exp\frac{q_k}{\lambda}$. By direct induction, 
\begin{equation}
    \pi_{k+1} \propto \pi_k \exp \frac{q_k}{\lambda} \propto \pi_{k-1} \exp\frac{q_k+q_{k-1}}{\lambda}
    \propto \dots \propto \exp\frac{\sum_{j=0}^k q_j}{\lambda} = \exp\frac{(k+1)h_k}{\lambda}.
\end{equation}
Still thanks to Eq.~\eqref{eq:appx:maximizer}, this means that $\pi_{k+1}$ satisfies
\begin{equation}
    \pi_{k+1} = \argmax_{\pi\in\Delta_\A^\s}\left(\langle\pi, h_k\rangle + \frac{\lambda}{k+1} \hc(\pi)\right) = \gc^{0,\frac{\lambda}{k+1}}(h_k).
\end{equation}
This shows that Eq.~\eqref{eq:appx:proof_equiv_md0} is equivalent to
\begin{equation}
    \begin{cases}
        \pi_{k+1} = \gc^{0,\frac{\lambda}{k+1}}(h_k)
        \\
        q_{k+1} = (T_{\pi_{k+1}\sep\pi_k}^{\lambda,0})^m q_k + \epsilon_{k+1}
        \\
        h_{k+1} = \frac{k+1}{k+2} h_k + \frac{1}{k+2} q_{k+1}
    \end{cases},
    \label{eq:appx:proof_equiv_da0}
\end{equation}
which is DA-MPI($\lambda$,0), and this shows the first part of the result. In the limit $\lambda\rightarrow 0$, the regularized greediness becomes the usual greediness (hard maximum over $q$-values) and the (regularized) evaluation operator becomes the standard one. However, notice that schemes are not equivalent in the limit: scheme~\eqref{eq:appx:proof_equiv_md0} tends to classic VI, while scheme~\eqref{eq:appx:proof_equiv_da0} tends to Speedy Q-learning~\citep{azar2011speedy} (see the justification of the connection to SQL in Appx.~\ref{subappx:connection_dampi}).

Next, we prove the equivalence for the case $\tau>0$. We recall MD-MPI($\lambda$,$\tau$):
\begin{equation}
    \begin{cases}
        \pi_{k+1} = \gc_{\pi_k}^{\lambda,\tau}(q_k)
        \\
        q_{k+1} = (T_{\pi_{k+1}\sep\pi_k})^m q_k + \epsilon_{k+1}
    \end{cases}.\label{eq:appx:proof_equiv_md1}
\end{equation}
Thanks to Eq.~\eqref{eq:appx:maximizer}, we have that $\pi_{k+1}\propto\exp\frac{q_k+\lambda\ln \pi_k}{\lambda+\tau}$. We define $\beta=\frac{\lambda}{\lambda + \tau}$ (and thus $1-\beta = \frac{\tau}{\lambda+\tau}$ and $\frac{\beta}{\lambda} = \frac{1}{\lambda+\tau}$). By induction, we have (writing ``$\cst$'' any function depending solely on states, not necessarily the same for different lines):
\begin{align}
    \ln \pi_{k+1} &= \frac{\beta}{\lambda} q_k + \beta \ln \pi_k + \cst
    \\
    &= \frac{\beta}{\lambda}\left(q_k + \beta q_{k-1} + \beta^2 q_{k-2} + \dots\right) + \cst
    \\
    &= \frac{\beta}{\lambda(1-\beta)}\left((1-\beta)(q_k + \beta q_{k-1} + \beta^2 q_{k-2} + \dots)\right) + \cst.
\end{align}
We now define $h_k$ as the moving average of past $q$-values, with $h_{-1} = 0$:
\begin{equation}
    h_k = \beta h_{k-1} + (1-\beta) q_k = (1-\beta)\sum_{j=0}^k \beta^{k-j} q_j.
    \label{eq:def:h}
\end{equation}
Noticing also that $\frac{\beta}{\lambda(1-\beta)} = \frac{1}{\tau}$, this shows that
\begin{equation}
    \pi_{k+1} \propto \exp\frac{h_k}{\tau}.
\end{equation}
As before, this means that $\pi_{k+1}$ is the solution of an entropy regularized greedy step with respect to $h_k$:
\begin{equation}
    \pi_{k+1} = \argmax_{\pi\in\Delta_\A^\s}\left(\langle\pi, h_k\rangle + \tau \hc(\pi)\right) = \gc^{0,\tau}(h_k).
\end{equation}
This means that Eq.~\eqref{eq:appx:proof_equiv_md1} is equivalent to
\begin{equation}
    \begin{cases}
        \pi_{k+1} = \gc^{0,\tau}(h_k)
        \\
        q_{k+1} = (T_{\pi_{k+1}\sep\pi_k}^{\lambda,\tau})^m q_k + \epsilon_{k+1}
        \\
        h_{k+1} = \beta h_{k} + (1-\beta) q_{k+1} \text{ with } \beta = \frac{\lambda}{\lambda+\tau}
    \end{cases},
\end{equation}
which is the DA-MPI($\lambda$,$\tau$) scheme. This concludes the proof.

\subsection{Proof of Theorem~\ref{thm:davi_type1_noEntropy}}
\label{subsec:proof:davi_type1_noEntropy}

Here, we provide the bound for DA-VI($\lambda$,0), which we recall:
\begin{equation}
    \begin{cases}
        \pi_{k+1} = \gc^{0,\frac{\lambda}{k+1}}(h_k)
        \\
        q_{k+1} = T_{\pi_{k+1}\sep\pi_k}^{\lambda, 0} q_k + \epsilon_{k+1}
        \\
        h_{k+1} = \frac{k+1}{k+2} h_k + \frac{1}{k+2} q_{k+1}
    \end{cases}.\label{eq:appx:davi10}
\end{equation}

\paragraph{Sketch of proof.}

The quantity of interest is $q_* - q_{\pi_{k+1}}$, it can be decomposed as $q_* - q_{\pi_{k+1}} = q_* - h_k + h_k - q_{\pi_{k+1}}$. Lemma~\ref{lemma:value_residual} allows expressing the quantity of interest essentially as a function of the Bellman residual $T_{\pi_{k+1}} h_k - h_k$. Controlling this residual is the key to state our bound. To achieve this, we first derive Lemma~\ref{lemma:q_to_h} that expresses the evaluation step (the update of the $q$-function) as a difference of Bellman operators applied to successive h-functions (the averages of $q$-values). Thanks to this, we're able to derive a Bellman-like recursion for $h_k$ in Lemma~\ref{lemma:h_bellman_tau_zero}, using notably Lemma~\ref{lemma:q_to_h} and a telescoping argument. The rest of the proof consists in exploiting this Bellman-like recursion to control the residual and eventually boud the quantity of interest.

\paragraph{Proof.}

We start by stating a useful lemma.
\begin{lemma}
    \label{lemma:value_residual}
    For any $q\in\R^{\s\times\A}$ and $\pi\in\Delta_\A^\s$, we have
    \begin{equation}
        q_\pi - q = (I-\gamma P_\pi)^{-1}(T_\pi q - q).
    \end{equation}
\end{lemma}
\begin{proof}
    This result is classic, and appears many times in the literature (\textit{e.g.}, \citet{kakade2002approximately}). We provide a one line proof for completeness, relying on basic properties of the Bellman operator:
    \begin{equation}
        q_\pi - q = T_\pi q_\pi - T_\pi q + T_\pi q - q = \gamma P_\pi (q_\pi - q) + T_\pi q - q \Leftrightarrow q_\pi-q = (I-\gamma P_\pi)^{-1}(T_\pi q - q).
    \end{equation}
\end{proof}
The aim is to bound the quantity $q_* - q_{\pi_{k+1}}$, the difference between the optimal value function and the value function computed by DA-VI($\lambda$,0). Thanks to Lemma~\ref{lemma:value_residual}, we can decompose this term as
\begin{align}
    q_* - q_{\pi_{k+1}} &= q_* - h_k + h_k - q_{\pi_{k+1}}
    \\
    &= (I-\gamma P_{\pi_*})^{-1}(T_{\pi_*} h_k - h_k) - (I-\gamma P_{\pi_{k+1}})^{-1}(T_{\pi_{k+1}}h_k - h_k). \label{eq:appx:davi10:decomposition}
\end{align}
Notice that $q_* = q_{\pi_*}$ for any optimal policy $\pi_*$. There exists an optimal deterministic policy~\citep{puterman2014markov}, so we will consider a deterministic $\pi_*$. As for any deterministic policy, $\hc(\pi_*)=0$. Using the definition of $\pi_{k+1}$, we have
\begin{align}
    \pi_{k+1} = \gc^{0,\frac{\lambda}{k+1}}(h_k)
    &\Rightarrow
    \langle \pi_{k+1}, h_k \rangle + \frac{\lambda}{k+1} \hc(\pi_{k+1}) \geq \langle \pi_*, h_k\rangle + \frac{\lambda}{k+1}\underbrace{\hc(\pi_*)}_{=0}
    \\
    &\Rightarrow
    r + \gamma P \left(\langle \pi_{k+1}, h_k \rangle + \frac{\lambda}{k+1} \hc(\pi_{k+1})\right) \geq r + \gamma P \langle \pi_*, h_k\rangle 
    \\
    &\Rightarrow
    T_{\pi_{k+1}}^{0,\frac{\lambda}{k+1}} h_k = T_{\pi_{k+1}} h_k + \gamma \frac{\lambda}{k+1} P\hc(\pi_{k+1}) \geq T_{\pi_*} h_k.
\end{align}
Injecting this into Eq.~\eqref{eq:appx:davi10:decomposition}, we obtain, using the fact that for any $\pi$ the matrix $(I-\gamma P_\pi)^{-1} = \sum_{t\geq 0} \gamma^t P_\pi^t$ is positive,
\begin{equation}
    q_* - q_{\pi_{k+1}} \leq (I-\gamma P_{\pi_*})^{-1}(T_{\pi_{k+1}}^{0,\frac{\lambda}{k+1}} h_k - h_k)
    - (I-\gamma P_{\pi_{k+1}})^{-1}(T_{\pi_{k+1}}^{0,\frac{\lambda}{k+1}}h_k - h_k - \gamma \frac{\lambda}{k+1} P\hc(\pi_{k+1})).
    \label{eq:appx:davi10:decomposition2}
\end{equation}
So, what we have to do is to control the residual $T_{\pi_{k+1}}^{0,\frac{\lambda}{k+1}} h_k - h_k$.

To do so, the following lemma will be useful.
\begin{lemma}
    \label{lemma:q_to_h}
    For any $k\geq 1$, we have that
    \begin{equation}
        T_{\pi_{k+1}\sep\pi_k}^{\lambda,0} q_k = (k+1) T_{\pi_{k+1}}^{0,\frac{\lambda}{k+1}} h_k - k T_{\pi_k}^{0,\frac{\lambda}{k}} h_{k-1}.
    \end{equation}
    For $k=0$, we have
    \begin{equation}
        T_{\pi_1\sep\pi_0}^{\lambda,0}q_0 = T_{\pi_1}^{0,\lambda} h_0 - \gamma \lambda P \hc(\pi_0).
    \end{equation}
\end{lemma}
\begin{proof}
    To prove this result, we will start by working on the optimization problem related to the regularized greedy step $\gc_{\pi_k}^{\lambda,0} q_k$:
    \begin{equation}
        \langle \pi, q_k\rangle - \lambda\kl(\pi||\pi_k) = \langle \pi, q_k\rangle - \lambda \langle \pi, \ln \pi - \ln \pi_k\rangle = \langle \pi, q_k + \lambda \ln\pi_k\rangle - \lambda \langle \pi,\ln\pi\rangle.
    \end{equation}
    For DA-VI$_1$($\lambda$,0), $\pi_{k+1}\in\gc^{0,\frac{\lambda}{k+1}}(h_k)$ (see Eq.~\eqref{eq:appx:davi10}), so according to Eq.~\eqref{eq:appx:maximizer}, $\pi_{k+1}\propto \exp\frac{(k+1)h_k}{\lambda}$. Therefore, we have, using also the definition of $h_k$
    \begin{align}
        q_k + \lambda \ln\pi_k &= q_k + \lambda(\frac{k}{\lambda} h_{k-1} - \ln\langle 1, \exp\frac{k h_{k-1}}{\lambda}\rangle)
        \\
        &= (k+1) h_k - \lambda \ln\langle 1, \exp\frac{k h_{k-1}}{\lambda}\rangle.
    \end{align}
    Therefore, we have
    \begin{equation}
        \langle \pi, q_k\rangle - \lambda \kl(\pi||\pi_k) = \langle \pi, (k+1) h_k\rangle - \lambda \langle \pi, \ln\pi\rangle - \lambda \ln\langle \un, \exp\frac{k h_{k-1}}{\lambda}\rangle.
    \end{equation}
    The maximizer is $\pi_{k+1}$, obviously. It is also the maximizer of $\langle \pi, (k+1) h_k\rangle - \lambda \langle \pi, \ln\pi\rangle$ (the third term not depending on $\pi$), and the associated maximum is, according to Eq.~\eqref{eq:appx:maximum}, $\lambda \ln \langle \un, \exp\frac{(k+1)h_k}{\lambda}\rangle$. This gives
    \begin{align}
        \langle \pi_{k+1}, q_k\rangle - \lambda \kl(\pi_{k+1}||\pi_k) &=
        \lambda \ln \langle \un, \exp\frac{(k+1)h_k}{\lambda}\rangle - \lambda \ln\langle \un, \exp\frac{k h_{k-1}}{\lambda}\rangle
        \\
        &= (k+1)\frac{\lambda}{k+1} \ln \langle \un, \exp\frac{(k+1)h_k}{\lambda}\rangle - k\frac{\lambda}{k} \ln\langle \un, \exp\frac{k h_{k-1}}{\lambda}\rangle.
    \end{align}
    Still from Eq.~\eqref{eq:appx:maximum}, we know that $\frac{\lambda}{k+1} \ln \langle 1, \exp\frac{(k+1)h_k}{\lambda}\rangle$ is the maximum of $\langle\pi, h_k\rangle + \frac{\lambda}{k+1} \hc(\pi)$, the associated maximizer being again $\pi_{k+1}$, so using Eq.~\eqref{eq:appx:relationship}, we can conclude that
    \begin{equation}
        \langle \pi_{k+1}, q_k\rangle - \lambda \kl(\pi_{k+1}||\pi_k)  = (k+1)\left(\langle\pi_{k+1}, h_k\rangle + \frac{\lambda}{k+1}\hc(\pi_{k+1})\right) - k\left(\langle\pi_{k}, h_{k-1}\rangle + \frac{\lambda}{k}\hc(\pi_{k})\right). 
    \end{equation}
    Noticing that $r = (k+1) r - k r$, we have the first part of the result:
    \begin{equation}
        T_{\pi_{k+1}\sep\pi_k}^{\lambda,0} q_k = (k+1) T_{\pi_{k+1}}^{0,\frac{\lambda}{k+1}} h_k - k T_{\pi_k}^{0,\frac{\lambda}{k}} h_{k-1}.
    \end{equation}
    This only holds for $k\geq 1$. For $k=0$, using the fact that $h_0 = q_0$, 
    \begin{align}
        T_{\pi_1\sep\pi_0}^{\lambda,0} q_0 &= r + \gamma P (\langle \pi_1, q_0\rangle - \lambda \kl(\pi_1||\pi_0))
        \\
        &= r + \gamma P (\langle \pi_1, h_0\rangle - \lambda \langle\pi_1,\ln \pi_1 - \ln\pi_0\rangle)
        \\
        &= r + \gamma P (\langle \pi_1, h_0\rangle + \lambda \hc(\pi_1) + \lambda \langle\pi_1, \ln\pi_0\rangle)
        \\
        &= T_{\pi_1}^{0,\lambda}h_0 - \gamma \lambda P \hc(\pi_0),
    \end{align}
    where we used in the last line the fact that, $\pi_0$ being uniform,
    \begin{equation}
        \langle\pi_1, \ln\pi_0\rangle = \langle\pi_1,\ln\frac{1}{|\A|}\rangle
        = - \ln |\A| \langle\pi_1, 1\rangle = - \ln |\A| = -\hc(\pi_0).
    \end{equation}
    This concludes the proof.
\end{proof}

Using this lemma, we can provide a Bellman-like induction on $h_k$.
\begin{lemma}
    \label{lemma:h_bellman_tau_zero}
    Define $E_k = -\sum_{j=1}^k \epsilon_j$.
    For any $k\geq 1$, we have that 
    \begin{equation}
        h_{k+1} = \frac{k+1}{k+2} T_{\pi_{k+1}}^{0,\frac{\lambda}{k+1}} h_k
        + \frac{1}{k+2} \left(q_0 - E_{k+1} -\gamma\lambda P\hc(\pi_0)\right).
    \end{equation}
\end{lemma}
\begin{proof}
    Using the definition of $h_k$, Lemma~\ref{lemma:q_to_h}, the fact that $q_{k+1} = T_{\pi_{k+1}\sep\pi_k}^{\lambda,0} q_k + \epsilon_{k+1}$, and the definition $E_{k} = -\sum_{j=1}^k \epsilon_j$, we have
    \begin{align}
        (k+2) h_{k+1} &= \sum_{j=0}^{k+1} q_j
        \\
        &= q_0 + q_1 + \sum_{j=1}^k q_{j+1}
        \\
        &= q_0 + T_{\pi_1\sep\pi_0}^{\lambda,0} q_0 + \epsilon_1 + \sum_{j=1}^k \left(T_{\pi_{j+1}}^{\lambda,0} q_j + \epsilon_{j+1}\right)
        \\
        &= q_0 + T_{\pi_1}^{0,\lambda}h_0 - \gamma \lambda P \hc(\pi_0) + \sum_{j=1}^k\left((j+1) T_{\pi_{j+1}}^{0,\frac{\lambda}{j+1}} h_j - j T_{\pi_{j}}^{0,\frac{\lambda}{j}} h_{j-1}\right) - E_{k+1}
        \\
        &= q_0 - E_{k+1} -\gamma\lambda P\hc(\pi_0) + (k+1) T_{\pi_{k+1}}^{0,\frac{\lambda}{k+1}} h_k 
        \\
        \Leftrightarrow
        h_{k+1}  &= \frac{k+1}{k+2} T_{\pi_{k+1}}^{0,\frac{\lambda}{k+1}} h_k
        + \frac{1}{k+2} \left(q_0 - E_{k+1} -\gamma\lambda P\hc(\pi_0)\right).
    \end{align}
\end{proof}

We now have the tools to work on the residual of interest. Starting from Lemma~\ref{lemma:h_bellman_tau_zero}, and using the fact that $(k+2)h_{k+1} = (k+1)h_k + q_{k+1}$,
\begin{align}
    h_{k+1}  &= \frac{k+1}{k+2} T_{\pi_{k+1}}^{0,\frac{\lambda}{k+1}} h_k
        + \frac{1}{k+2} \left(q_0 - E_{k+1} -\gamma\lambda P\hc(\pi_0)\right)
    \\
    \Leftrightarrow
    (k+1) h_k + q_{k+1} &= q_0 - E_{k+1} -\gamma\lambda P\hc(\pi_0) + (k+1) T_{\pi_{k+1}}^{0,\frac{\lambda}{k+1}} h_k \label{eq:tmp:cmt}
    \\
    \Leftrightarrow
    T_{\pi_{k+1}}^{0,\frac{\lambda}{k+1}} h_k - h_k &= \frac{1}{k+1}\left(q_{k+1} - q_0 + E_{k+1} + \gamma \lambda P \hc(\pi_0)\right). 
\end{align}
Injecting this last result into decomposition~\eqref{eq:appx:davi10:decomposition2}, we get
\begin{align}
    q_* - q_{\pi_{k+1}} &\leq (I-\gamma P_{\pi_*})^{-1}(T_{\pi_{k+1}}^{0,\frac{\lambda}{k+1}} h_k - h_k)
    - (I-\gamma P_{\pi_{k+1}})^{-1}(T_{\pi_{k+1}}^{0,\frac{\lambda}{k+1}}h_k - h_k - \gamma \lambda P\hc(\pi_{k+1}))
    \\
    &= (I-\gamma P_{\pi_*})^{-1}\left(\frac{1}{k+1}\left(q_{k+1} - q_0 + E_{k+1} + \gamma \lambda P \hc(\pi_0)\right)\right)
    \\
    &~~~~ - (I-\gamma P_{\pi_{k+1}})^{-1}\left(\frac{1}{k+1}\left(q_{k+1} - q_0 + E_{k+1} + \gamma \lambda P \hc(\pi_0)\right) - \gamma \frac{\lambda}{k+1} P\hc(\pi_{k+1})\right)
    \\
    &\leq (I-\gamma P_{\pi_*})^{-1}\left(\frac{1}{k+1}\left(q_{k+1} - q_0 + E_{k+1} + \gamma \lambda P \hc(\pi_0)\right)\right)
    \\
    &~~~~ - (I-\gamma P_{\pi_{k+1}})^{-1}\left(\frac{1}{k+1}\left(q_{k+1} - q_0 + E_{k+1}  - \gamma \lambda P\hc(\pi_{k+1})\right)\right),
\end{align}
where we used for the last inequality the fact that $-(I-\gamma P_{\pi_{k+1}})^{-1} P \hc(\pi_0) \leq 0$.
Next, using the fact that $q_* - q_{\pi_{k+1}}\geq 0$ and rearranging terms, we have
\begin{align}
    q_* - q_{\pi_{k+1}} \leq &\left|\left((I-\gamma P_{\pi_*})^{-1}- (I-\gamma P_{\pi_{k+1}})^{-1}\right)\frac{E_{k+1}}{k+1}\right|
    \\
    &+ (I-\gamma P_{\pi_*})^{-1}\left|\frac{q_{k+1} - q_0 + \gamma \lambda P \hc(\pi_0)}{k+1}\right| \\
    &+ (I-\gamma P_{\pi_{k+1}})^{-1}\left|\frac{q_{k+1} - q_0 + \gamma \lambda P \hc(\pi_{k+1})}{k+1}\right|.
\end{align}
We assumed that $\|q_{k+1}\|_\infty\leq v_\text{max}\leq v_\text{max}^\lambda$ (see also Rk.~\ref{rk:rk}). When introducing the algorithm, we assumed that $\|q_0\|_\infty\leq v_\text{max}$. Therefore, $\|q_0 - \gamma \lambda P \hc(\pi_0)\|_\infty \leq v_\text{max}^\lambda$. Writing $\un$ the vector whose components are all 1, we get $|q_{k+1} - q_0 + \gamma \lambda P \hc(\pi_0)|\leq 2  v_\text{max}^\lambda \un$. Notice that for any policy $\pi$, we have that $P_\pi \un = \un$. Therefore, we have
\begin{equation}
    (I-\gamma P_{\pi_*})^{-1}\left|\frac{q_{k+1} - q_0 + \gamma \lambda P \hc(\pi_0)}{k+1}\right| \leq \frac{2}{1-\gamma}\frac{v_\text{max}^\lambda }{k+1} \un.
\end{equation}
With the same arguments, we have that
\begin{equation}
    (I-\gamma P_{\pi_{k+1}})^{-1}\left|\frac{q_{k+1} - q_0 + \gamma \lambda P \hc(\pi_{k+1})}{k+1}\right| \leq \frac{2}{1-\gamma}\frac{v_\text{max}^\lambda }{k+1} \un.
\end{equation}
We finally have
\begin{equation}
    q_* - q_{\pi_{k+1}} \leq \left|\left((I-\gamma P_{\pi_*})^{-1}- (I-\gamma P_{\pi_{k+1}})^{-1}\right)\frac{E_{k+1}}{k+1}\right| + \frac{4}{1-\gamma} \frac{v^\lambda_\text{max}}{k+1} \un,
\end{equation}
which is the stated result.

\subsection{About Remark~\ref{rk:rk}}
\label{appx:rk}

We stated in Rk.~\ref{rk:rk}, in the context of DA-VI($\lambda$,0), that the assumption $\|q_k\|_\infty\leq v_\text{max}$ is not strong with approximation, as this just requires clipping the $q$-values. Indeed, without approximation, it's not even necessary to clip the $q$-values.

\paragraph{No approximation.}
We will proceed by induction. Assume that $\|q_k\|_\infty \leq v_\text{max}$. We assumed generally that $\|q_0\|_\infty \leq v_\text{max}$.
Without error, the considered scheme is
\begin{equation}
    \begin{cases}
        \pi_{k+1} = \gc^{0,\frac{\lambda}{k+1}}(h_k)
        \\
        q_{k+1} = T_{\pi_{k+1}\sep\pi_k}^{\lambda, 0} q_k 
        \\
        h_{k+1} = \frac{k+1}{k+2} h_k + \frac{1}{k+2} q_{k+1}
    \end{cases} \Leftrightarrow
    \begin{cases}
        \pi_{k+1} = \gc^{\lambda,0}(q_k)
        \\
        q_{k+1} = T_{\pi_{k+1}\sep\pi_k}^{\lambda, 0} q_k 
    \end{cases}.
\end{equation}
As $\pi_{k+1} = \gc^{\lambda,0}(q_k)$, we have that
\begin{equation}
    q_{k+1} = T_{\pi_{k+1}\sep\pi_k}^{\lambda, 0} q_k  \geq T_{\pi_{k}\sep\pi_k}^{\lambda, 0} q_k = T_{\pi_k} q_k \geq -v_\text{max} \un,
\end{equation}
The inequality making use of the induction argument. On the other hand, making use of the positiveness of the KL divergence, we have that
\begin{equation}
    q_{k+1} = T_{\pi_{k+1}\sep\pi_k}^{\lambda, 0} q_k  \leq T_{\pi_{k+1}} q_k \leq v_\text{max}\un,
\end{equation}
where again the inequality comes from the induction argument. This allows concluding, $\|q_{k+1}\|_\infty\leq v_\text{max}$.

\paragraph{With approximation.} Knowing a bound of the $q$-values without approximation, we can clip $q_k$ such that it satisfies the bound, the effect of the clipping being part of the error. For example, assume that the evaluation step is approximated with a least-squares problems, a parameterized $q$-function, the target being a sampling of $T_{\pi_{k+1}\sep\pi_k}^{\lambda, 0} q_k$, $q_k$ being the previous approximation (for example the target network). We can clip the result of the least-squares in $[-v_\text{max},+v_\text{max}]$ and call the resulting function $q_{k+1}$. The resulting error is defined as $\epsilon_{k+1} = q_{k+1} - T_{\pi_{k+1}\sep\pi_k}^{\lambda, 0} q_k$.

\subsection{Proof of Theorem~\ref{thm:davi_type1_withEntropy}}
\label{subsec:proof:davi_type1_withEntropy}

In this section, we provide a bound for DA-VI($\lambda$,$\tau$). First, we recall the scheme:
\begin{equation}
    \begin{cases}
        \pi_{k+1} = \gc^{0,\tau}(h_k)
        \\
        q_{k+1} = T_{\pi_{k+1}\sep\pi_k}^{\lambda, \tau} q_k + \epsilon_{k+1}
        \\
        h_{k+1} = \beta h_{k} + (1-\beta) q_{k+1} \text{ with } \beta = \frac{\lambda}{\lambda+\tau}
    \end{cases}.
\end{equation}
We recall that due to the entropy term, this scheme cannot converge to the unregularized optimal $q_*$ function. Yet, without errors and with $\lambda=0$, it would converge to the solution of the MDP regularized by the scaled entropy~\citep{geist2019theory} (optimizing for the reward augmented by the scaled entropy). Our bound will show that adding a KL penalty does not change this. We recall the notations introduced in the main paper.
We already have defined the operator $T_\pi^{0,\tau}$. It has a unique fixed point, which we write $q^\tau_\pi$. The unique optimal $q$-function is $q^\tau_* = \max_{\pi} q^\tau_\pi$. We write $\pi_*^\tau = \gc^{0,\tau}(q_*^\tau)$ the associated unique optimal policy, and $q_{\pi_*^\tau}^\tau = q_*^\tau$.

\paragraph{Sketch of proof.}

The proof is similar to the one of Thm.~\ref{thm:davi_type1_noEntropy}, albeit a bit more technical. Thanks to Lemma~\ref{lemma:value_residual_reg} (that generalizes Lemma~\ref{lemma:value_residual}), we decompose the quantity of interest $q_*^\tau - q_{\pi_{k+1}}^\tau$ as a function of $q_*^\tau - h_k$ and of $T_{\pi_{k+1}}^{0,\tau} h_k - h_k$, to be respectively upper-bounded and lower-bounded. To achieve this, we first derive Lemma~\ref{lemma:q_to_h_entropy} that expresses the evaluation step as a difference of Bellman operators applied to successive h-functions (similarly to Lemma~\ref{lemma:q_to_h}). Thanks to this, we're able to derive a Bellman-like recursion for $h_k$ in Lemma~\ref{lemma:h_equal_Th}, using notably Lemma~\ref{lemma:q_to_h_entropy} and a telescoping argument (similarly to Lemma~\ref{lemma:h_bellman_tau_zero}). The end of the proof is then close to the classic propagation of errors of AVI, involving moving averages of the errors instead of the errors, as well as some additional terms.

\paragraph{Proof.}

The following lemma, generalizing Lemma~\ref{lemma:value_residual} to the regularized Bellman operator, will be useful:
\begin{lemma}
    \label{lemma:value_residual_reg}
    Let $\tau \geq 0$. For any $q\in\mathbb{R}^{\s\times\A}$ and $\pi \in \Delta^\s_\A$, we have
    \begin{equation}
        q_\pi^\tau - q = (I-\gamma P_\pi)^{-1}(T_{\pi}^{0,\tau} q - q).
    \end{equation}
\end{lemma}
\begin{proof}
    The proof is the same as the one of Lemma~\ref{lemma:value_residual}, relying on the fact that the regularized Bellman operator has the same properies as the Bellman operator~\citep{geist2019theory}:
    \begin{equation}
        q_\pi^\tau - q = T_\pi^{0,\tau} q^\tau_\pi - T_\pi^{0,\tau} q + T_\pi^{0,\tau} q - q = \gamma P_\pi (q^\tau_\pi - q) + T_\pi^{0,\tau} q - q \Leftrightarrow q^\tau_\pi-q = (I-\gamma P_\pi)^{-1}(T^{0,\tau}_\pi q - q).
    \end{equation}
\end{proof}

We will bound the quantity $q_*^\tau - q_{\pi_{k+1}}^\tau$, using the following decomposition, based on Lemma~\ref{lemma:value_residual_reg}:
\begin{align}
    q_*^\tau - q_{\pi_{k+1}}^\tau &= q_*^\tau - h_k + h_k - q_{\pi_{k+1}}^\tau
    \\
    &= (q_*^\tau - h_k) - (I-\gamma P_{\pi_{k+1}})^{-1}(T_{\pi_{k+1}}^{0,\tau} h_k - h_k).
    \label{eq:Xerror}
\end{align}
To do so, we will upper-bound $q_*^\tau - h_k$ and lower-bound $T_{\pi_{k+1}}^{0,\tau} h_k - h_k$ (we recall that the matrix $(I-\gamma P_{\pi_{k+1}})^{-1}$ is non-negative). This requires a Bellman-like induction on $h_k$. For this,
the following intermediate lemma, similar to Lemma~\ref{lemma:q_to_h}, will be useful.
\begin{lemma}
    \label{lemma:q_to_h_entropy}
    For any $k\geq 0$, we have that
    \begin{equation}
        T_{\pi_{k+1}\sep \pi_k}^{\lambda,\tau} q_k = \frac{1}{1-\beta}\left(T_{\pi_{k+1}}^{0,\tau} h_k - \beta T_{\pi_{k}}^{0,\tau} h_{k-1}\right).
    \end{equation}
\end{lemma}
\begin{proof}
    We have that, for any $\pi$,
    \begin{align}
        \langle\pi,q_k\rangle - \lambda \kl(\pi||\pi_k) + \tau \hc(\pi)
        &= \langle \pi, q_k \rangle - \lambda \langle \pi, \ln\pi - \ln \pi_k\rangle - \tau \langle \pi,\ln\pi\rangle
        \\
        &= \langle \pi, q_k + \lambda\ln\pi_k\rangle - (\lambda + \tau) \langle \pi, \ln\pi\rangle.
    \end{align}
    As $\pi_{k+1} \propto \exp\frac{h_k}{\tau}$, using also the fact that $\beta=\frac{\lambda}{\lambda+\tau}$ and $1-\beta=\frac{\tau}{\lambda+\tau}$, as well as the definition of $h_k$~\eqref{eq:def:h}, we have
    \begin{align}
        q_k + \lambda \ln\pi_k &= q_k + \lambda \left(
        \frac{h_{k-1}}{\tau} - \ln \langle \un, \exp\frac{h_{k-1}}{\tau}\rangle 
        \right)
        \\
        &= \frac{1}{1-\beta}\left((1-\beta) q_k + \beta h_{k-1} - \beta \tau \ln \langle \un, \exp\frac{h_{k-1}}{\tau}\rangle \right)
        \\
        &= \frac{1}{1-\beta}\left(h_k - \beta \tau \ln \langle \un, \exp\frac{h_{k-1}}{\tau}\rangle \right).
    \end{align}
    Hence, injecting this in the previous result, we get
    \begin{align}
        \langle \pi, q_k + \lambda\ln\pi_k\rangle - (\lambda + \tau) \langle \pi, \ln\pi\rangle
        &= \langle \pi, q_k + \lambda\ln\pi_k\rangle - \frac{\tau}{1-\beta} \langle \pi, \ln\pi\rangle
        \\
        &= \frac{1}{1-\beta}\left(\langle\pi,h_k\rangle - \tau \langle \pi, \ln\pi\rangle - \beta\tau \ln \langle \un, \exp\frac{h_{k-1}}{\tau}\rangle
        \right).
    \end{align}
    Now, as $\pi_{k+1}\propto\exp\frac{h_k}{\tau}$, we have that $\langle\pi_{k+1}, h_k \rangle + \tau\hc(\pi_{k+1}) = \tau\ln\langle \un,\exp\frac{h_k}{\tau}\rangle$ (again from Eq.~\eqref{eq:appx:relationship}), therefore
    \begin{align}
        &\langle\pi_{k+1},q_k\rangle - \lambda \kl(\pi_{k+1}||\pi_k) + \tau \hc(\pi_{k+1}) 
        \\
        = &\frac{1}{1-\beta}\left(\langle\pi_{k+1}, h_k \rangle + \tau\hc(\pi_{k+1}) - \beta (\langle\pi_{k}, h_{k-1} \rangle + \tau\hc(\pi_{k}))
        \right).
    \end{align}
    The result follows by the definition of $T_{\pi_{k+1}\sep \pi_k}^{\lambda,\tau} q_k = r + \gamma P(\langle\pi_{k+1},q_k\rangle - \lambda \kl(\pi_{k+1}||\pi_k) + \tau \hc(\pi_{k+1}) )$, and noticing that $r = \frac{1}{1-\beta}(r - \beta r)$.
\end{proof}

This result allows to build the lemma stating a Bellman-like induction for $h_k$.
\begin{lemma}
    \label{lemma:h_equal_Th}
    Define $E^\beta_{k+1}=-(1-\beta)\sum_{j=1}^{k+1} \beta^{k+1-j}\epsilon_{j} = \beta E^\beta_k + (1-\beta)\epsilon_{k+1}$ (with $E^\beta_0=0$). For any $k\geq 0$, we have that
    \begin{equation}
        h_{k+1} =  T_{\pi_{k+1}}^{0,\tau} h_k - E_{k+1} - \beta^{k+1}(T_{\pi_0}^{0,\tau} h_{-1} - h_0).
    \end{equation}
\end{lemma}
\begin{proof}
    Using the definition of $h_k$, Eq.~\eqref{eq:def:h}, the relationship between $q_{k+1}$ and $q_k$, and Lemma~\ref{lemma:q_to_h_entropy}, we have
    \begin{align}
        h_{k+1} &= (1-\beta) \sum_{j=0}^{k+1}\beta^{k+1-j}q_k
        \\
        &= (1-\beta) \beta^{k+1}q_0 + (1-\beta)\sum_{j=1}^{k+1} \beta^{k+1-j}q_j
        \\
        &= (1-\beta) \beta^{k+1}q_0 + (1-\beta)\sum_{j=0}^k \beta^{k-j}q_{j+1}
        \\
        &= (1-\beta) \beta^{k+1}q_0 + (1-\beta)\sum_{j=0}^k \beta^{k-j}\left(T_{\pi_{j+1}\sep\pi_j}^{\lambda,\tau}q_j + \epsilon_{j+1}\right)
        \\
        &= (1-\beta) \beta^{k+1}q_0 + (1-\beta)\sum_{j=0}^k \beta^{k-j}\left(\frac{1}{1-\beta}\left(T_{\pi_{j+1}}^{0,\tau} h_j - \beta T_{\pi_{j}}^{0,\tau} h_{j-1}\right)+ \epsilon_{j+1}\right).
    \end{align}
    Let define $E^\beta_{k+1}$ as
    \begin{align}
        E_{k+1} &= -(1-\beta)\sum_{j=0}^{k} \beta^{k-j}\epsilon_{j+1}
        \\
        &= -(1-\beta)\sum_{j=1}^{k+1} \beta^{k+1-j}\epsilon_{j}
        \\
        &= \beta E^\beta_k + (1-\beta)\epsilon_{k+1} \text{ with } E_0 = 0. 
    \end{align}
    We also have
    \begin{align}
        &(1-\beta)\sum_{j=0}^k \beta^{k-j}\left(\frac{1}{1-\beta}\left(T_{\pi_{j+1}}^{0,\tau} h_j - \beta T_{\pi_{j}}^{0,\tau} h_{j-1}\right)\right)
        \\
        =
        &\sum_{j=0}^k \beta^{k-j}\left(T_{\pi_{j+1}}^{0,\tau} h_j - \beta T_{\pi_{j}}^{0,\tau} h_{j-1}\right)
        \\
        =&\sum_{j=1}^{k+1} \beta^{k+1-j} T_{\pi_{j}}^{0,\tau} h_{j-1} - \sum_{j=0}^k \beta^{k+1-j} T_{\pi_{j}}^{0,\tau} h_{j-1}
        \\
        = &T_{\pi_{k+1}}^{0,\tau} h_{k} - \beta^{k+1} T_{\pi_{0}}^{0,\tau} h_{-1}.
    \end{align}
    Notice also that $h_0 = (1-\beta) q_0$.
    Putting all these parts together, we obtain
    \begin{align}
        h_{k+1} &= \beta^{k+1}h_0 - E^\beta_{k+1} + T_{\pi_{k+1}}^{0,\tau} h_{k} - \beta^{k+1} T_{\pi_{0}}^{0,\tau} h_{-1}
        \\
        &=  T_{\pi_{k+1}}^{0,\tau} h_k - E^\beta_{k+1} - \beta^{k+1}(T_{\pi_0}^{0,\tau} h_{-1} - h_0),
        \end{align}
        which is the stated result.
\end{proof}
Thanks to this result, we can now bound the terms of interest.

\paragraph{Upper-bounding $q_*^{\tau} - h_k$.} Write $e_k = E^\beta_k + \beta^{k}(T_{\pi_0}^{0,\tau} h_{-1} - h_0)$, we have from Lemma~\ref{lemma:h_equal_Th} that $h_{k+1} = T_{\pi_{k+1}}^{0,\tau} h_k - e_{k+1}$. Then, we have :
\begin{align}
    q_*^\tau - h_{k+1} &= q_*^\tau - T_{\pi_{k+1}}^{0,\tau} h_k + e_{k+1}
    \\
    &= \underbrace{T_{\pi_*^\tau}^{0,\tau} q_*^\tau - T_{\pi_*^\tau}^{0,\tau} h_k}_{=\gamma P_{\pi_*^\tau}(q_*^\tau - h_k)}
    + \underbrace{T_{\pi_*^\tau}^{0,\tau} h_k - T_{\pi_{k+1}}^{0,\tau} h_k}_{\leq 0 \text{ as } \pi_{k+1} = \gc^{0,\tau}(h_k)} + e_{k+1}
    \\
    &\leq \gamma P_{\pi_*^\tau}(q_*^\tau - h_k) + e_{k+1}.
\end{align}
By direct induction, we obtain
\begin{align}
    q_*^\tau - h_{k+1} &\leq (\gamma P_{\pi_*^\tau})^{k+1}(q_*^\tau - h_0) + \sum_{j=1}^{k+1} (\gamma P_{\pi_*^\tau})^{k+1-j} e_j
    \\
    &= (\gamma P_{\pi_*^\tau})^{k+1}(q_*^\tau - h_0) + \sum_{j=1}^{k+1} (\gamma P_{\pi_*^\tau})^{k+1-j}\left(E^\beta_j + \beta^{j}(T_{\pi_0}^{0,\tau} h_{-1} - h_0)\right).\label{eq:Xup}
\end{align}
This is the desired upper-bound.

\paragraph{Lower-bounding $T_{\pi_{k+1}}^{0,\tau} h_k - h_k$.} Using the same notation $e_k$, we have
\begin{align}
    T_{\pi_{k+1}}^{0,\tau} h_k - h_k &= \underbrace{T_{\pi_{k+1}}^{0,\tau} h_k - T_{\pi_{k}}^{0,\tau} h_k}_{\geq 0 \text{ as } \pi_{k+1} = \gc^{0,\tau}(h_k)} + T_{\pi_{k}}^{0,\tau} h_k - h_k 
    \\
    &\geq T_{\pi_{k}}^{0,\tau} h_k - h_k
    \\
    &= T_{\pi_{k}}^{0,\tau} \left(T_{\pi_{k}}^{0,\tau} h_{k-1} - e_k\right) - \left(T_{\pi_{k}}^{0,\tau} h_{k-1} - e_k\right)
    \text{ by Lemma~\ref{lemma:h_equal_Th}}
    \\
    &= \gamma P_{\pi_k} \left(T_{\pi_{k}}^{0,\tau} h_{k-1} - h_{k-1}\right) - (I-\gamma P_{\pi_k})^{-1} e_k.
\end{align}
We define $P_{k:j} = P_{\pi_k} P_{\pi_{k-1}} \dots P_{\pi_{j+1}} P_{\pi_j}$ for $j\leq k$, with the convention $P_{k:k+1}=I$. By direct induction, the preceding inequality gives
\begin{align}
    T_{\pi_{k+1}}^{0,\tau} h_k - h_k &\geq
    \gamma^k P_{k:1} (T_{\pi_1}^{0,\tau} h_0 - h_0) - \sum_{j=1}^k \gamma^{k-j} P_{k:j+1} (I-\gamma P_{\pi_j}) e_j
    \\
    &= \gamma^k P_{k:1} (T_{\pi_1}^{0,\tau} h_0 - h_0) - \sum_{j=1}^k \gamma^{k-j} P_{k:j+1} (I-\gamma P_{\pi_j})(E^\beta_j + \beta^j (T_{\pi_0}^{0,\tau} h_{-1} - h_0)).
    \label{eq:Xlow}
\end{align}

\paragraph{Putting things together.} Plugging Eqs.~\eqref{eq:Xup} and~\eqref{eq:Xlow} into Eq.~\eqref{eq:Xerror}, we obtain
\begin{align}
    q_*^\tau - q_{\pi_{k+1}}^\tau \leq
    &(\gamma P_{\pi_*^\tau})^{k}(q_*^\tau - h_0) + \sum_{j=1}^{k} (\gamma P_{\pi_*^\tau})^{k-j}\left(E^\beta_j + \beta^{j}(T_{\pi_0}^{0,\tau} h_{-1} - h_0)\right)
    \\
    &+ (I-\gamma P_{\pi_{k+1}})^{-1} \bigg(-\gamma^k P_{k:1} (T_{\pi_1}^{0,\tau} h_0 - h_0)
    \\
    &+ \sum_{j=1}^k \gamma^{k-j} P_{k:j+1} (I-\gamma P_{\pi_j})(E^\beta_j + \beta^j (T_{\pi_0}^{0,\tau} h_{-1} - h_0))\bigg).
\end{align}
Using the fact that $q_*^\tau - q_{\pi_{k+1}}^\tau \geq 0$, rearranging terms, we have
\begin{align}
    q_*^\tau - q_{\pi_{k+1}}^\tau \leq
    &\sum_{j=1}^k \left|(\gamma P_{\pi_*^\tau})^{k-j} + (I-\gamma P_{\pi_{k+1}})^{-1}\gamma^{k-j} P_{k:j+1}\left(I-\gamma P_{\pi_j}\right) E^\beta_j\right|
    \\
    &+ (\gamma P_{\pi_*^\tau})^{k}|q_*^\tau - h_0| + \sum_{j=1}^{k} (\gamma P_{\pi_*^\tau})^{k-j} \beta^{j}|T_{\pi_0}^{0,\tau} h_{-1} - h_0|
    \\
    &+ (I-\gamma P_{\pi_{k+1}})^{-1} \gamma^k P_{k:1} |T_{\pi_1}^{0,\tau} h_0 - h_0|
    \\
    &+ (I-\gamma P_{\pi_{k+1}})^{-1} \sum_{j=1}^k \gamma^{k-j} P_{k:j+1} (I+\gamma P_{\pi_j}) \beta^j |T_{\pi_0}^{0,\tau} h_{-1} - h_0|.\label{eq:presquefini}
\end{align}
The first term is related to the error, the others to the initialisation. We'll work on each of these other terms.

Recall that we assumed that $\|q_0\|_\infty \leq v_\text{max} = \frac{r_\text{max}}{1-\gamma}$. Therefore, $\|q_0\|_\infty \leq v^\tau_\text{max} = \frac{r_\text{max} + \tau\ln|\A|}{1-\gamma}$. As $h_0 = (1-\beta)q_0$, we have $\|h_0\|_\infty \leq (1-\beta) v^\tau_\text{max}$. From obvious properties of regularized MDPs~\citep{geist2019theory}, we have $\|q_*^\tau\|_\infty \leq v^\tau_\text{max}$. Therefore, writing $\un\in\R^{\s\times\A}$ the vector with all components equal to 1, we have $|q_*^\tau - h_0| \leq (2-\beta) v^\tau_\text{max} \un$. Notice that for any policy $\pi$, we have $P_\pi\un = \un$, thus
\begin{equation}
    (\gamma P_{\pi_*^\tau})^{k}|q_*^\tau - h_0| \leq \gamma^k (2-\beta) v^\tau_\text{max} \un.
\end{equation}
We also have that $\|T_{\pi_1}^{0,\tau} h_0 \|_\infty\leq r_\text{max}+ \tau \ln|\A| + \gamma (1-\beta)v^\tau_\text{max} = (1-\gamma\beta) v^\tau_\text{max}$, so
\begin{equation}
    (I-\gamma P_{\pi_{k+1}})^{-1} \gamma^k P_{k:1} |T_{\pi_1}^{0,\tau} h_0 - h_0| \leq \gamma^k \frac{2 - (1+\gamma)\beta}{1-\gamma} v_\text{max}^\tau \un.
\end{equation}
By definition $h_{-1} = 0$, so we have $\|T_{\pi_0}^{0,\tau} h_{-1}\|_\infty = \|r + \gamma P \tau \hc(\pi_0)\|_\infty \leq r_\text{max} + \tau \ln|\A| = (1-\gamma) v_\text{max}^\tau$, so $\|T_{\pi_0}^{0,\tau} h_{-1} - h_0\|_\infty \leq (2-\gamma - \beta) v_\text{max}^\tau$. Therefore, we have the following bound:
\begin{equation}
    \sum_{j=1}^{k} (\gamma P_{\pi_*^\tau})^{k-j} \beta^{j}|T_{\pi_0}^{0,\tau} h_{-1} - h_0|
    \leq \gamma^k \sum_{j=1}^k \left(\frac{\beta}{\gamma}\right)^j (2-\beta - \gamma) v_\text{max}^\tau \un.
\end{equation}
Similarly, for the last term we have
\begin{equation}
    (I-\gamma P_{\pi_{k+1}})^{-1} \sum_{j=1}^k \gamma^{k-j} P_{k:j+1} (I+\gamma P_{\pi_j}) \beta^j |T_{\pi_0}^{0,\tau} h_{-1} - h_0|
    \leq \frac{1+\gamma}{1-\gamma} \gamma^k \sum_{j=1}^k \left(\frac{\beta}{\gamma}\right)^j (2-\beta - \gamma) v_\text{max}^\tau \un.
\end{equation}
Summing these four upper bounds, we obtain
\begin{align}
    &\gamma^k (2-\beta) v^\tau_\text{max} \un + \gamma^k \frac{2 - (1+\gamma)\beta}{1-\gamma} v_\text{max}^\tau \un + \gamma^k \sum_{j=1}^k \left(\frac{\beta}{\gamma}\right)^j (2-\beta - \gamma) v_\text{max}^\tau \un 
    \\
    &+ \frac{1+\gamma}{1-\gamma} \gamma^k \sum_{j=1}^k \left(\frac{\beta}{\gamma}\right)^j (2-\beta - \gamma) v_\text{max}^\tau \un
    \\
    = &2\gamma^k \frac{2 - \beta - \gamma}{1-\gamma} \sum_{j=0}^k \left(\frac{\beta}{\gamma}\right)^j v_\text{max}^\tau \un = 2\gamma^k \left(1+\frac{1 - \beta}{1-\gamma}\right) \sum_{j=0}^k \left(\frac{\beta}{\gamma}\right)^j v_\text{max}^\tau \un. 
\end{align}
Plugging this result into Eq.~\eqref{eq:presquefini}, we obtain the stated result:
\begin{align}
    q_*^\tau - q_{\pi_{k+1}}^\tau \leq
    &\sum_{j=1}^k \left|(\gamma P_{\pi_*^\tau})^{k-j} + (I-\gamma P_{\pi_{k+1}})^{-1}\gamma^{k-j} P_{k:j+1}\left(I-\gamma P_{\pi_j}\right) E^\beta_j\right|
    \\
    &+ \gamma^k \left(1+\frac{1 - \beta}{1-\gamma}\right) \sum_{j=0}^k \left(\frac{\beta}{\gamma}\right)^j v_\text{max}^\tau \un.
\end{align}

\section{Empirical illustration of the bounds}
\label{appx:garnet}

We have illustrated the bounds of Sec.~\ref{sec:background} (Fig.~\ref{fig:illustration_thm1}) in a simple tabular setting with acces to a generative model. We provide more details about this setting here.

We consider MDPs with small state and action spaces, such that a tabular representation of the $q$-function is possible. We also assume to have access to a generative model, allowing us to sample a transition for any state-action couple. We then consider sampled MD-VI($\lambda$,$\tau$), depicted in Alg.~\ref{algo:sampled-mdvi}. At each iteration of MD-VI, we sample a single transition for each state-action couple and apply the resulting sampled Bellman operator. The error $\epsilon_k$ is the difference between the sampled and the exact operators. The sequence of these estimation errors is thus a martingale difference w.r.t. its natural filtration~\citep{azar2011speedy} (one can think about bounded, centered and roughly i.i.d. errors).

We run this algorithm on randomized MDPs called Garnets. A Garnet~\citep{archibald1995generation} is an abstract MDP, built from three parameters ($N_S$, $N_A$, $N_B$), with $N_S$ and $N_A$ respectively the number of states and actions, and $N_B$ the branching factor. The principle is to directly build the transition kernel $P$ that represents the MDP. For each $(s,a) \in \s\times\A$, $N_B$ states ($s_1, \hdots s_{N_B}$) are drawn uniformly from $\s$ without replacement. Then, $N_B - 1$ numbers are drawn uniformly in $(0,1)$ and sorted as $(p_0=0, p_1, \hdots p_{N_B-1}, p_{N_B}=1)$. The transition kernel is then defined as $P(s_k|s,a)=p_{k} - p_{k-1}$ for each $1\leq k \leq N_B$. The reward function is drawn uniformly in $(0,1)$ for $10\%$ of the states, these states being drawn uniformly without replacement.

For the experiments shown in Fig.~\ref{fig:illustration_thm1}, we set $N_S=30$, $N_A=4$, $N_B=4$ and $\gamma = 0.9$. We generate 100 Garnets and run MD-VI once for each of these Garnets, for $K=800$ iterations. The results in Fig.~\ref{fig:illustration_thm1} shows the normalized average performance, $\frac{\|q_*^\tau - q_{\pi_k}^\tau\|_1}{\|q_*^\tau\|_1}$. For sampled DA-VI($\lambda$, 0), we show the behavior for various values of $\lambda$. For DA-VI($\lambda$, $\tau$), we fix $\tau$ to a small value ($\tau = 10^{-3}$) and show the behavior for various values of $\beta = \frac{\lambda}{\lambda + \tau}$. Notice that considering a large value of $\tau$ would not be interesting. In this case, the regularized optimal policy would be close to be uniform, so close to the initial policy.

\begin{algorithm}[tbh]
\caption{Sampled MD-VI($\lambda, \tau$)}
\begin{algorithmic}
\label{algo:sampled-mdvi}
\REQUIRE $K$ number of iterations, $P$ the transition kernel.
\SET $\beta = \frac{\lambda}{\lambda + \tau}$
\SET $q_0$ to the null vector %
\SET $\pi_0$ to be the uniform policy %
\FOR{$1 \leq k \leq K$}
    \FOR{$(s,a) \in \s\times\A$}
         \STATE $\pi_k(a |s) = \frac{\pi_{k-1}(a |s) ^\beta \exp\frac{q_{k-1}(s,a)}{\lambda + \tau} }{\sum_{b \in\A} \pi_{k-1}(b |s) ^\beta \exp\frac{q_{k-1}(s,b)}{\lambda + \tau}}$
    \ENDFOR
    \FOR{$(s,a) \in \s\times\A$}
       \STATE $s' \sim P(\cdot|s,a)$
       \STATE $q_k(s,a) = r(s,a) + \gamma \sum_{b \in \A} \pi_k(b|s') \left(q_{k-1}(s',b) -\lambda\ln\frac{\pi_k(b|s')}{\pi_{k-1}(b|s')}- \tau\ln\pi_k(b|s') \right)$
    \ENDFOR
\ENDFOR
\OUTPUT $\pi_{K}$
\end{algorithmic}
\end{algorithm}

\section{Algorithms and experimental details}
\label{appx:exp}
\label{appx:experimental}

This appendix provides additional details about the algorithms and the experiments:
\begin{itemize}
    \item Appx.~\ref{subappx:high_level} provides a complementary high level view of algorithms sketched in Sec.~\ref{sec:experiments}.
    \item Appx.~\ref{subappx:algorithms} provides implementation details of these algorithms, including a pseudo-code.
    \item Appx.~\ref{subappx:hyper} provides all hyperparameters used in our experiments.
    \item Appx.~\ref{subappx:additional} provides additionnal experiments (one additional gym environment, Lunar Lander, and two additional Atari games, Breakout and Seaquest), as well as additional visualisations (including all training curves on Atari games).
\end{itemize}

\subsection{High level view of practical algorithms}
\label{subappx:high_level}

DA-VI and MD-VI are extensions of VI. One of the most prevalent VI-based deep RL algorithm is probably DQN~\citep{mnih2015human}. Thus, our approach consists in modifying the DQN algorithm to study regularization. To complement the sketch of Sec.~\ref{sec:experiments}, We present the different variations we consider with a high level viewpoint here, all practical details being just after.

DQN maintains a replay buffer and a target network $q_{k}$, and computes $q_{k+1}$ by minimizing the loss (recall that `\wor{}' stands for ``without regularization''):
\begin{equation}
    \mathcal{L}_\text{\wor{}}(q) = \hat{\E}_{s,a}\left[\left([\hat{T}_{\pi_{k+1}}q_k](s,a) - q(s,a)\right)^2\right],
    \label{eq:loss:t2}
\end{equation}
with $q$ a neural network, $\pi_{k+1}\in\gc(q_k)$ the greedy policy computed analytically from $q_k$, $[\hat{T}_{\pi_{k+1}}q_k](s,a) = r(s,a) + \gamma \langle \pi_{k+1},q_k\rangle(s')$ the sampled Bellman operator (with $s'\sim P(\cdot|s,a)$), and where the empirical expectation $\E_{s,a}$ is according to the transitions in the buffer. DQN is an optimistic AVI scheme, in the sense that only a few steps of stochastic gradient descent are performed before updating the target network. We modify DQN by adding a policy network and possibly modifying the evaluation step. For the moment, we consider $\tau>0$.

\paragraph{Greedy step.} As explained before, when the greedy step is approximated, MD-VI and DA-VI are no longer equivalent. We start with MD-VI. A natural way to learn the policy network is to optimize directly for the greedy step. Let $\pi_k$ be the target policy network and $q_k$ the target $q$-network, it corresponds to (`dir' stands for direct):
\begin{equation}
    \mathcal{L}_{\text{dir}}(\pi) = \hat{\E}_{s}\left[\langle\pi, q_k\rangle(s) - \lambda \kl(\pi||\pi_k)(s) + \tau \hc(\pi)(s)\right].
    \label{eq:loss_direct_oublie}
\end{equation}
Maximizing this loss over networks gives $\pi_{k+1}$. This is reminiscent of TRPO (see Appx.~\ref{subappx:conection_mdmpi}).

One can also compute analytically the policy $\pi_{k+1}$ (see Appx.~\ref{appx:LF}), but it would require remembering all past networks. Thus, another solution is to approximate this analytical solution by a neural network (`ind' stands for indirect):
\begin{equation}
    \mathcal{L}_{\text{ind}}(\pi) = \hat{\E}_{s}\left[\kl(\pi^*_{k+1}||\pi)(s)\right] \text{ with } \pi^*_{k+1}\propto \pi_k^\beta\exp\frac{\beta q_k}{\lambda}.
\end{equation}
Minimizing this loss over networks gives $\pi_{k+1}$. This is reminiscent of MPO (see Appx.~\ref{subappx:conection_mdmpi}), up to the fact that we consider the KL in the reverse order. Indeed, MPO (or SAC) would optimise for $\hat\E_s[\kl(\pi||\pi^*_{k+1})(s)]$. The motivation to do so is to get ride of the partition function. Yet, this is equivalent to what we call the ``direct'' approach, writing $Z_k\in\R^\s$ the partition function:
\begin{align}
    -\kl(\pi||\pi_{k+1}^*) &= \langle \pi, \ln \frac{\pi_k^\beta\exp\frac{\beta q_k}{\lambda}}{Z_k} - \ln \pi \rangle 
    \\
    &= \langle\pi, \ln(\pi_k^\beta\exp\frac{\beta q_k}{\lambda})\rangle - \langle \pi, \ln Z_k\rangle - \langle \pi, \ln\pi\rangle 
    \\
    &= \frac{\beta}{\lambda} \langle \pi, q_k\rangle + \beta \langle \pi, \ln\pi_k\rangle - \langle\pi,\ln\pi\rangle - \ln Z_k
    \\
    &= \frac{\beta}{\lambda}\left(
    \langle \pi, q_k\rangle + \lambda \langle \pi, \ln\pi_k\rangle - (\lambda+\tau) \langle \pi,\ln\pi\rangle - (\lambda+\tau)\ln Z_k
    \right)
    \\
    &= \frac{\beta}{\lambda}\left( \langle\pi,q_k\rangle - \lambda \kl(\pi||\pi_k) + \tau\hc(\pi) - \ln Z_k\right).
\end{align}
So, up to the scaling $\frac{\beta}{\lambda}=\frac{1}{\lambda+\tau}$ and to the term $\ln Z_k$, which is a constant regarding the optimized policy $\pi$ and can thus safely be ignored, we obtain the loss of Eq.~\eqref{eq:loss_direct_oublie}.

When considering DA-VI, the policy can be computed analytically, $\pi_{k+1} = \gc^{0,\tau}(h_k)$, but $h_k$ has to be approximated (and can be seen as the logits of the policy). With $h_{k-1}$ and $q_k$ the target networks:%
\begin{equation}
    \mathcal{L}_\text{da}(h) = \hat{\E}_{s,a}\left[\left([\beta h_{k-1} + (1-\beta) q_k](s,a) - h(s,a)\right)^2\right].
\end{equation}
Minimizing this loss over networks $h$ gives $h_k$. This is reminiscent of momentum-DQN (see Appx.~\ref{subappx:connection_dampi}).

\paragraph{Evaluation step.} Given one of the three ways of doing the greedy step, one can choose between regularizing the evaluation step (\wir{}, as suggested by the theory) or not (\wor{}, as often done empirically). This second case is already depicted in Eq.~\eqref{eq:loss:t2} (changing the considered policy) and the first case is given by
\begin{equation}
    \mathcal{L}_\text{\wir{}}(q) = \hat{\E}_{s,a}\left[\left([\hat{T}^{\lambda,\tau}_{\pi_{k+1}}q_k](s,a) - q(s,a)\right)^2\right].
\end{equation}
So combining one of the two evaluation steps (\wir{} or \wor{}) with one of the three greedy steps (MD-dir, MD-ind or DA), we get six variations. We discuss also the limit case without entropy.

\paragraph{When $\tau=0$.} For MD-VI, one can set $\tau=0$. However, recall that for DA-VI, the resulting algorithm is different. DA-VI($\lambda$, 0) is not practical in a deep learning setting, as it requires averaging over iterations. Indeed, updates of target networks are too fast to consider them as new iterations, and a moving average is more convenient. \citet{vieillard2019momentum} used a decay on $\beta$ to mimic this behavior, but this is a heuristic that needs to be tuned. Therefore, for DA-VI we will only consider the limit case $\lambda+\tau\rightarrow 0$ with $\beta=\frac{\lambda}{\lambda+\tau}$ kept constant (that is, momentum-DQN with fixed $\beta$). In this case, type~1 and~2 are equivalent. We offer additional visualisations in Appx.~\ref{subappx:additional}.

\subsection{More on practical algorithms}
\label{subappx:algorithms}
We now detail the losses presented in the previous section, giving equations that are closer to implementation, and providing a detailed pseudo-code in Algorithm~\ref{algo:deep}. Firts, let us introduce some notations.The $q$-value is represented by a neural network $Q_\theta$ of parameters $\theta$, and the policy is represented by a network $\Pi_\phi$ of parameters $\phi$. During training, the algorithms interact with an environment, and collect transitions $(s, a, r, s')$ that are stored in a FIFO replay buffer $\mathcal{B}$. The parameters of the networks are copied regularly into old versions of themselves, with target weights $\bar{\theta}$ and $\bar\phi$. The weights $\theta$ are optimized during the evaluation step, and $\phi$ during the greedy step. 

\subsubsection{Evaluation step}

All the actor-critics we consider have the same update rule of their critic -- the $Q$-network. We consider two regressions targets, corresponding to regularizing the evaluation step or not. If not regularized, we define a regression target as
\begin{equation}
    \hat{Q}_\text{\wor{}}(r, s') = r  + \gamma \sum_{b \in \mathcal{A}} Q_{\bar\theta}(s',b) \Pi_{\phi}(b |s'),
\end{equation}
and if regularized,
\begin{equation}
    \hat{Q}_\text{\wir{}}(r, s') = \hat{Q}_2(r, s')  - \lambda \kl\left(\Pi_\phi \Vert \Pi_{\bar\phi}\right)(s')
    + \tau \mathcal{H}\left(\Pi_\phi\right)(s').
\end{equation}
The weights $\theta$ are then updated by minimizing the following regression loss with a variant of SGD
\begin{equation}
\label{eq:deep:loss_q}
    \mathcal{L}_{\text{\wir{}-\wor{}}}(\theta) = \hat{E}_\mathcal{B}\left[\left(Q_\theta(s,a) - \hat{Q}_{\text{\wir{}-\wor{}}}(r, s')\right)^2\right].
\end{equation}
Note that if $\Pi_\phi$ was greedy with respect to $Q_{\bar\theta}$, using $\mathcal{L}_\text{\wor{}}$ would reduce to Deep $q$-networks (DQN)~\cite{mnih2015human}.

\subsubsection{Greedy step} 
Let us re-write in detail the three equations from Section~\ref{subappx:high_level} that define three ways of performing the greedy step. 

\paragraph{MD-dir.}
The Direct MD update tackles directly the optimization problem derived from the greedy step. For convenience, we define a loss (the opposite of what we would like to maximize) that we minimize with SGD
\begin{equation}
\label{eq:deep:loss_ind}
    \mathcal{L}_{\text{dir}}(\phi) = \hat{E}_\mathcal{B}\Bigg[-\sum_{b \in \mathcal{A}} Q_{\bar\theta}(s,b) \Pi_{\phi}(b |s)
    + \lambda \kl\left(\Pi_\phi \Vert \Pi_{\bar\phi}\right)(s') -
    \tau \mathcal{H}\left(\Pi_\phi\right)(s') \Bigg].
\end{equation}

\paragraph{MD-ind.}
The indirect version is based on the analytical result of the optimization problem corresponding to the greedy step. We show in Appendix~\ref{subappx:conection_mdmpi} that , at iteration $k$ of MD-VI$(\lambda, \tau)$, we have $\pi_{k+1} = \gc_{\pi_k}^{\lambda, \tau}(q_k) \propto \pi_k^\beta \exp{\frac{q_k}{\tau + \lambda}}$. Hence, we would need to fit a  target that approximates this maximizer, by defining $\hat{\Pi}(a |s)$ as 
\begin{equation}
     \hat{\Pi}(a |s) = \Pi_{\bar\phi}(a | s)^\beta \exp{\frac{Q_{\bar\theta}(s,a)}{\lambda + \tau}} \left(\sum_{b\in\mathcal{A}} \Pi_{\bar\phi}(b | s)^\beta \exp{\frac{Q_{\bar\theta}(s,b)}{\lambda + \tau}} \right)^{-1}.
\end{equation}
However, the exponential term can cause numerical problems, so what we optimize during the evaluation step is actually the logarithm of the policy. To work around this, we define a network $L_\phi$ that represents the log-probabilities of a policy, and we define a regression target
\begin{align}
    \hat{L}(s,a)  = \frac{\lambda L_{\bar\phi}(a |s) + Q_{\bar\theta}(s,a)}{\lambda + \tau} - \ln \sum_{b\in\mathcal{A}} \frac{\lambda L_{\bar\phi}(b |s) + Q_{\bar\theta}(s,b)}{\lambda + \tau},
\end{align}
and then we have $\hat{\Pi}(a|s) = \exp\left(\hat{L}(s,a)\right)$ and $\Pi_\phi(a |s) = \exp\left(L_\phi(a |s)\right)$.
We then define a loss on the parameters $\phi$,
\begin{equation}
\label{eq:deep:loss_dir}
    \mathcal{L}_{\text{ind}}(\phi) = \hat{E}_\mathcal{B}\left[\kl\left(\hat\Pi \Vert \Pi_\phi \right)(s)\right].
\end{equation}

\paragraph{DA.}
The dual averaging version is inspired by the DA-VI formulation. Instead of representing directly the policy, we estimate a moving average of the $q$-values, and then compute its softmax. The moving average is estimated via a network $H_\phi$, which fits a regression target
\begin{equation}
    \hat{H}(s,a) = \beta H_{\bar\phi}(s,a) + (1 - \beta) Q_{\bar\theta}(s,a),
\end{equation}
and the policy is defined as softmax over $H_{\phi}(s, \cdot)$,
\begin{equation}
    \Pi_{\phi}(a |s) = \exp\frac{H_\phi(s,a)}{\tau} \left(\sum_{b}\exp\frac{H_\phi(s,b)}{\tau} \right)^{-1}.
\end{equation}
The weights $\phi$ are optimized by minimizng the loss
\begin{equation}
\label{eq:deep:loss_da}
    \mathcal{L}_{\text{da}}(\phi) = \hat{E}_\mathcal{B}\left[\left(H_\phi(s,a) - \hat{H}(s,a)\right)^2\right].
\end{equation}

\subsubsection{Pseudo code}

We give a general pseudo-code of the deep RL algorithms we used in Alg.~\ref{algo:deep}. Notice that for a policy $\pi$, we define the $e$-greedy policy with respect to $\pi$ as the policy that takes a random action (uniformly on $\mathcal{A}$) with probability $e$, and follows $\pi$ with probability $1-e$.
\begin{algorithm}[tbh]
\caption{(MD-dir $|$ MD-ind $|$ DA)}
\begin{algorithmic}
\label{algo:deep}
\REQUIRE $L_q(\theta)$ and $L_\pi(\phi)$, two losses, respectively for the evaluation and the greediness. The choice of these losses determines the algorithm, see Table~\ref{tab:algos}.
\REQUIRE $K\in \mathbb{N^*}$ the number of steps, $C\in \mathbb{N^*}$ the update period, $F \in \mathbb{N^*}$ the interaction period.
\SET $\theta$, $\phi$ at random
\SET $Q_\theta$ the $q$-value network, $\Pi_\phi$ the policy network, as defined in Sec.~\ref{subappx:algorithms}.
\SET $\mathcal{B} = \{\}$
\SET $\Pi_{\phi, e_k}$ the policy $e_k$-greedy w.r.t. $\Pi_\phi$
\STATE $\bar\theta = \theta, \bar\phi = \phi$
\FOR{$1 \leq k \leq K$}
    \STATE Collect a transition $t = (s, a, r, s')$ from $\Pi_{\phi, e_k}$
    \STATE $\mathcal{B} \leftarrow \mathcal{B} \cup \{t\}$
    \IF{$k \mod F == 0$}
        \STATE On a random batch of transitions $B_{q,k} \subset \mathcal{B}$, update $\theta$ with one step of SGD on $L_q$ \label{line:lq}
        \STATE On a random batch of transitions $B_{h,k} \subset \mathcal{B}$,
            update $\phi$ with one step of SGD on  $L_\pi$ \label{line:h}
    \ENDIF
    \IF{$k \mod C == 0$}
        \STATE $\bar\theta \leftarrow \theta$, $\bar\phi \leftarrow \phi$
    \ENDIF
\ENDFOR
\OUTPUT $\Pi_{\phi}$
\end{algorithmic}
\end{algorithm}

\begin{table}[tbhh]
    \centering
    \caption{Resulting algorithms given the choice of losses in Algorithm~\ref{algo:deep}}
    \begin{tabular}{c c c c}
    \toprule
         & \multicolumn{3}{c}{$L_\pi$} \\
         \cmidrule(r){2-4}
    $L_q$     &  $\mathcal{L}_{\text{dir}}$ (Eq.\eqref{eq:deep:loss_dir}) & $\mathcal{L}_{\text{ind}}$ (Eq.~\eqref{eq:deep:loss_ind}) & $\mathcal{L}_{\text{da}}$ (Eq.~\eqref{eq:deep:loss_da})\\
    \cmidrule(r){1-1} \cmidrule(r){2-4}
    $\mathcal{L}_{\text{\wir{}}}$ (Eq.~\eqref{eq:deep:loss_q}) &  MD-dir \wir{} & MD-ind \wir{} & DA \wir{}\\
    $\mathcal{L}_{\text{\wor{}}}$ (Eq.~\eqref{eq:deep:loss_q}) &  MD-dir \wor{} & MD-ind \wor{} & DA \wor{}\\    
    \bottomrule
    \end{tabular}
    \label{tab:algos}
\end{table}

\subsection{Hyperparameters}
\label{subappx:hyper}

We provide the hyperparameters used on the Atari environments in Table~\ref{tab:atari}, and on the Gym environments in Table~\ref{tab:gym}.  We use the following notations to describe neural networks: $\FC n$ is a fully connected layer with $n$ neurons; $\Conv_{a,b}^{d}c$ is a 2d convolutional layer with $c$ filters of size $a  \times b$ and a stride of~$d$. All hyperparameters are the one found in the Dopamine code base. We only tuned the learning rate and the update period of DQN on Lunar Lander (not provided in Dopamine). 

\begin{table}[tbh]%
    \centering
    \caption{Parameters used on Atari. Both the $Q$-network and policy-network have the same structure. $n_A$ is the number of actions available in a given game.}
    \begin{tabular}{ll}
    \toprule
    Parameter     & Value \\
    \midrule
    $K$ (number of steps) & $5*10^7$ \\
    $C$ (update period)    & 8000\\
    $F$ (interaction  period)    & 4\\
    $\gamma$ (discount) & 0.99\\
    $|\mathcal{B}|$ (replay buffer size) & $10^6$\\
    $|B_{\pi,k}|$ and $|B_{q,k}|$ (batch size) & 32 \\
    $e_k$ (random actions rate) & $e_0 = 0.01$, linear decay of period $2.5\cdot10^5$ steps\\
    networks structure & $\Conv_{8,8}^{4}32-\Conv_{4,4}^{2}64-\Conv_{3,3}^{1}64-\FC512-\FC n_A$\\
    activations & Relu\\
    optimizers & RMSprop ($lr=0.00025$) \\
    \bottomrule
    \end{tabular}
    \label{tab:atari}
\end{table}

\begin{table}[tbh]%
    \centering
    \caption{Parameters used on CartPole and Lunar Lander . Both the $Q$-network and policy-network have the same structure. We  have $n_A=2$ on CartPole, and $n_A=8$ on Lunar Lander.}
    \begin{tabular}{ll}
    \toprule
    Parameter     & Value \\
    \midrule
    $K$ (number of steps) & $5*10^5$ \\
    $C$ (update period)    & 100 (Cartpole), 2500 (Lunar Lander)\\
    $F$ (interaction  period)    & 4\\
    $\gamma$ (discount) & 0.99\\
    $|\mathcal{B}|$ (replay buffer size) & $5*10^4$\\
    $|B_{\pi,k}|$ and $|B_{q,k}|$ (batch size) & $128$ \\
    $e_k$ (random actions rate) & 0.01 (constant with $k$)\\
    networks structure & $\FC512-\FC512-\FC n_A$\\
    activations & Relu\\
    optimizers & Adam ($lr=0.001$) \\
    \bottomrule
    \end{tabular}
    \label{tab:gym}
\end{table}

\subsection{Additional results}
\label{subappx:additional}

\paragraph{Additional environment.} In addition to the environments considered in Sec.~\ref{sec:experiments}, we provide three additional environments: Lunar Lander (from gym), Breakout and Seaquest (from Atari). The comments on these environments are similar to the discussion of Sec.~\ref{sec:experiments}

\begin{figure}[tbh]
    \centering
    \includegraphics[width=0.8\linewidth,trim={2cm 2cm 0cm 0cm},clip]{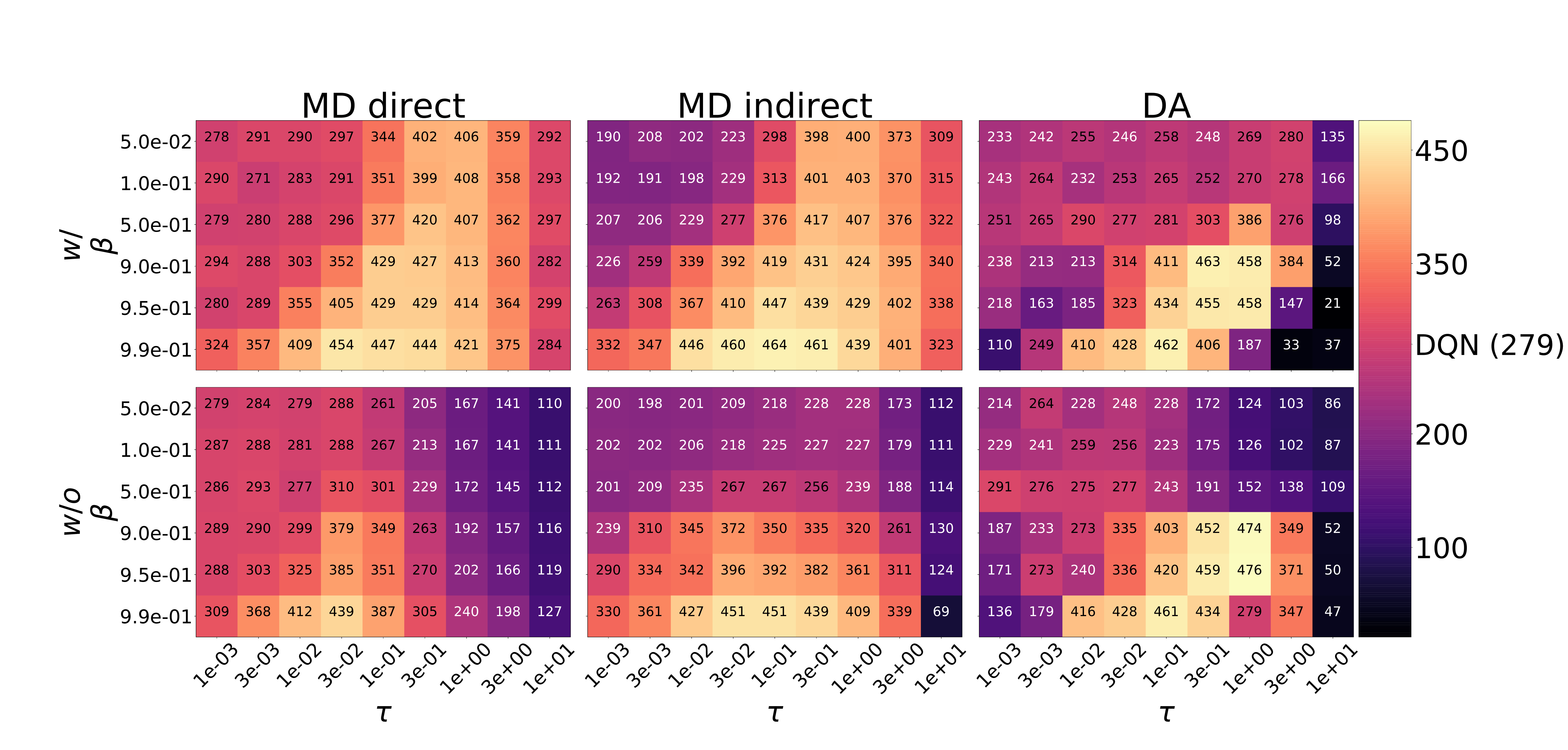}  
    \includegraphics[width=0.8\linewidth,trim={2cm 1cm 4cm 6cm},clip]{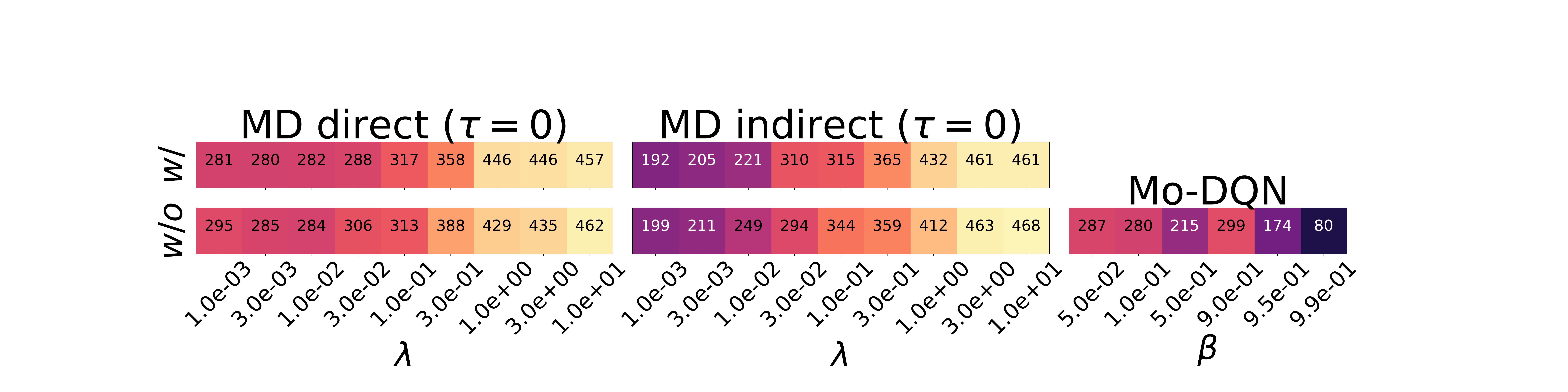}
    \caption{Cartpole with complete values.}
    
    \label{fig:cartpole_values}
\end{figure}

\begin{figure}[tbh]
    \centering
    \includegraphics[width=0.8\linewidth,trim={2cm 2cm 0cm 0cm},clip]{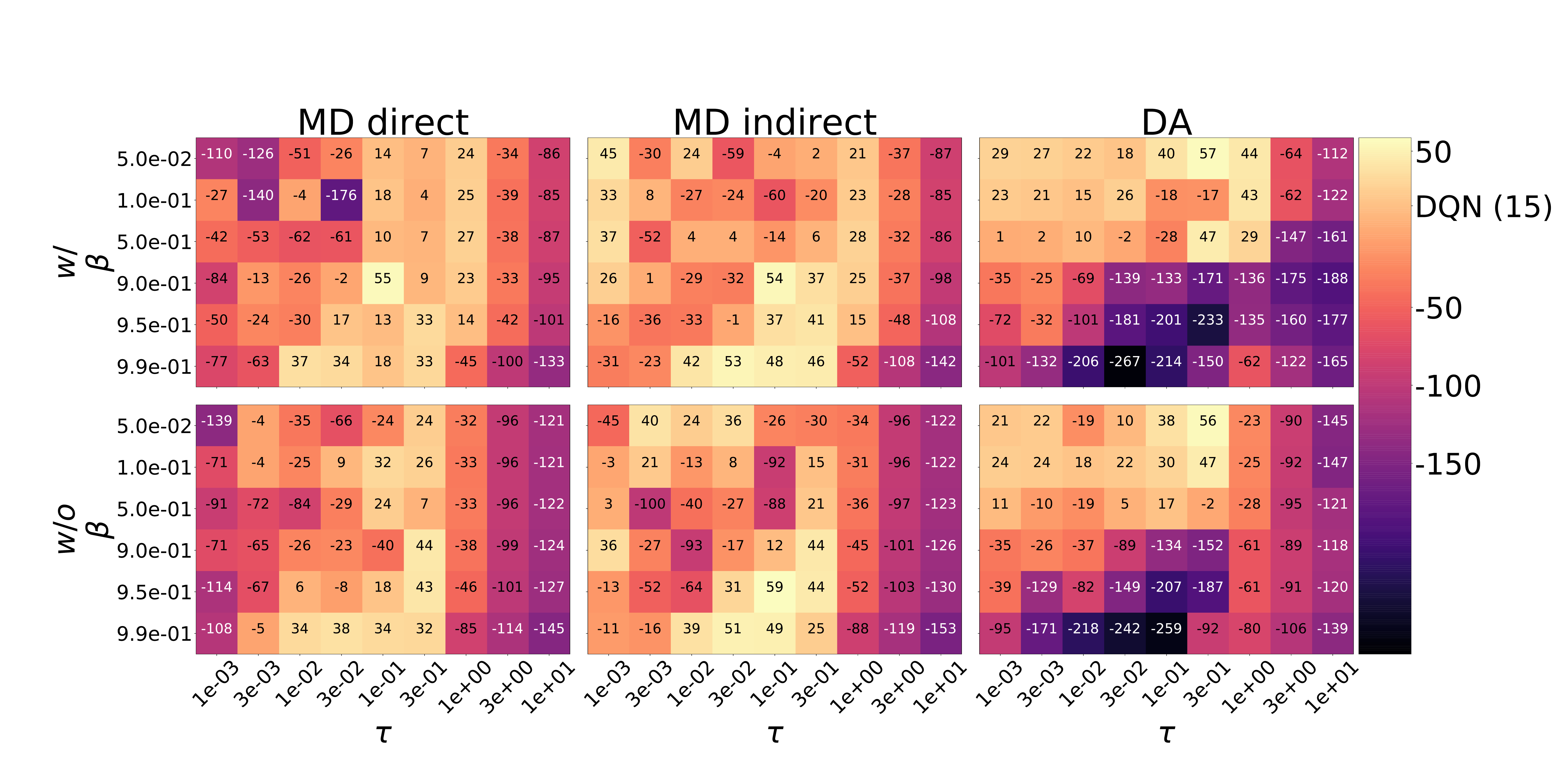}  
    \includegraphics[width=0.8\linewidth,trim={2cm 1cm 4cm 6cm},clip]{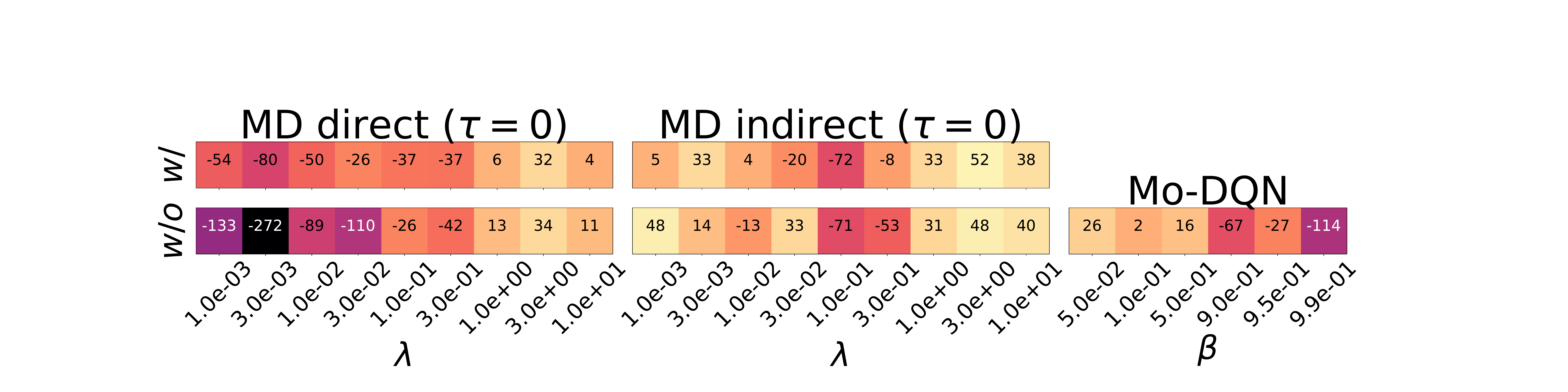}
    \caption{Lunar Lander with complete values.}
    
    \label{fig:lunar_lander_values}
\end{figure}

\begin{figure}[tbh]
    \centering
    \includegraphics[width=0.8\linewidth,trim={2cm 2cm 0cm 4cm},clip]{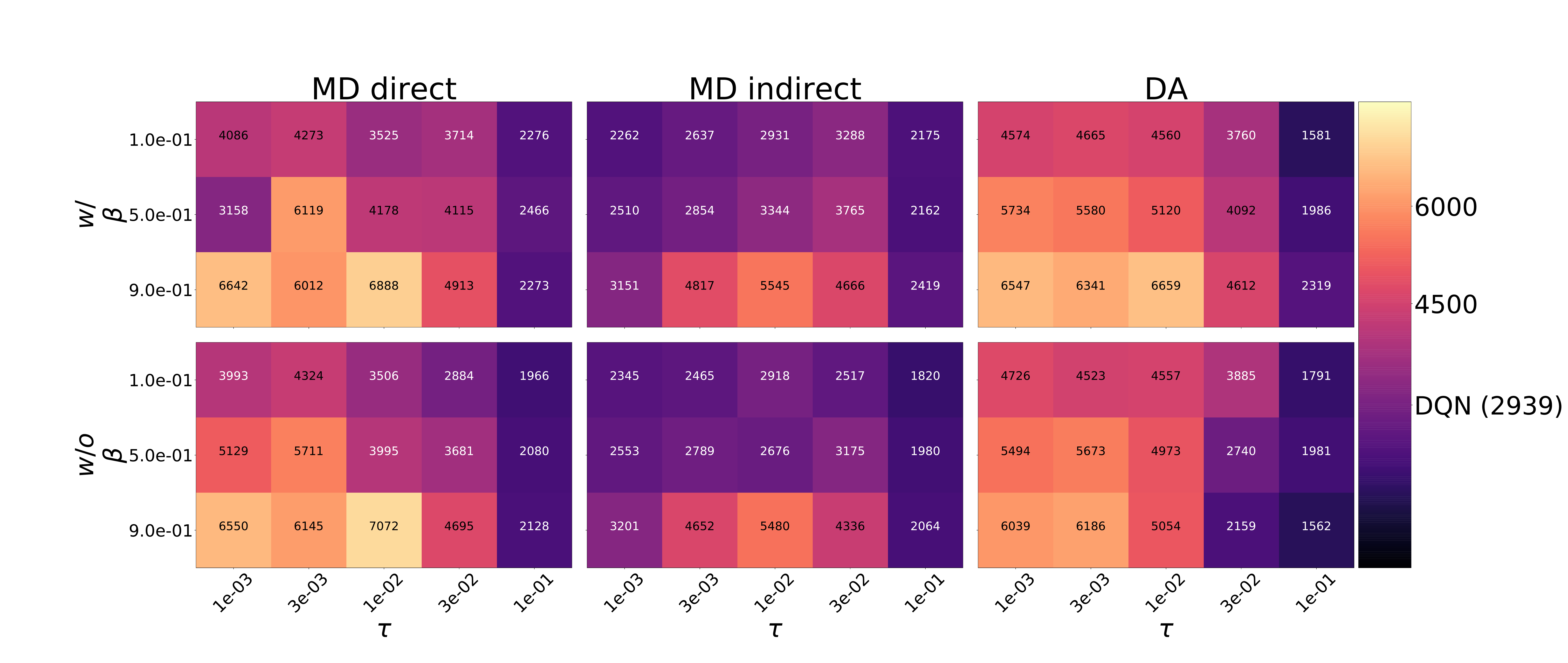}  
    \includegraphics[width=0.8\linewidth,trim={2cm .5cm 4cm 4cm},clip]{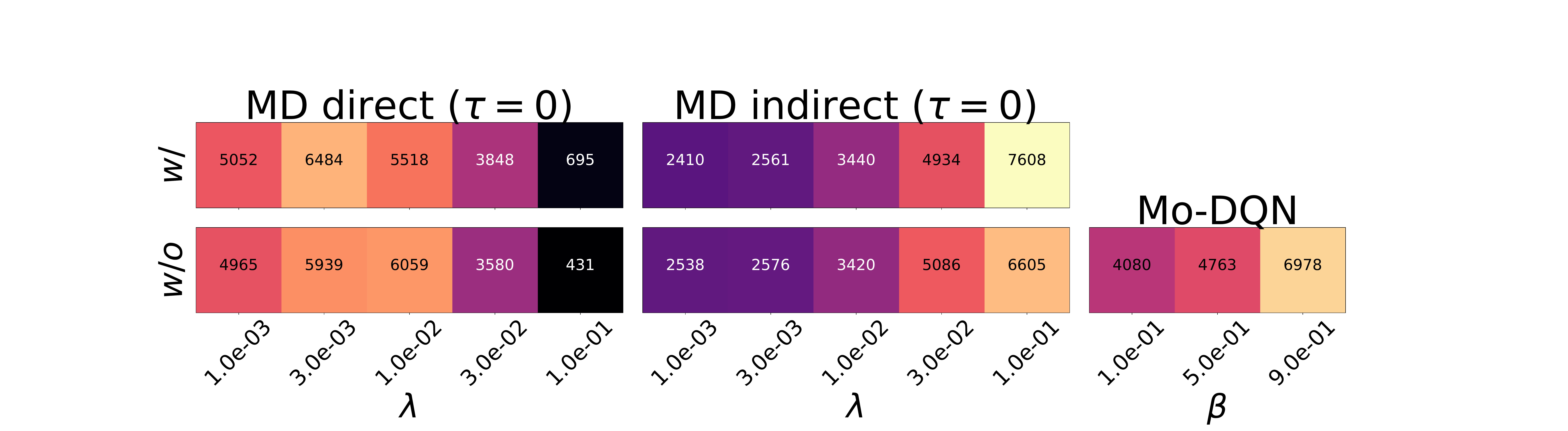}
    \vspace{-.2cm}
    \caption{Asterix with complete values.}
    
    \label{fig:asterix_values}
\end{figure}

\begin{figure}[tbh]
    \centering
    \includegraphics[width=0.8\linewidth,trim={2cm 2cm 0cm 4cm},clip]{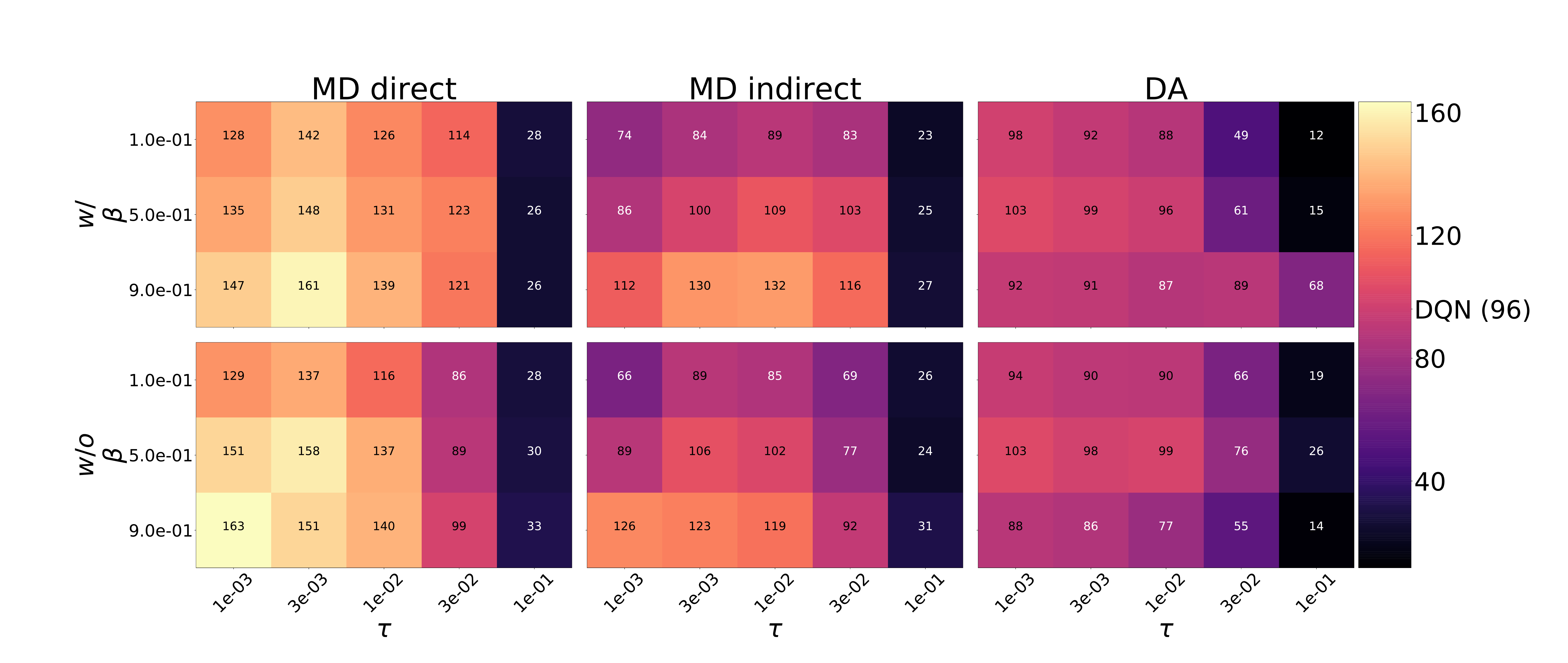}  
    \includegraphics[width=0.8\linewidth,trim={2cm .5cm 4cm 4cm},clip]{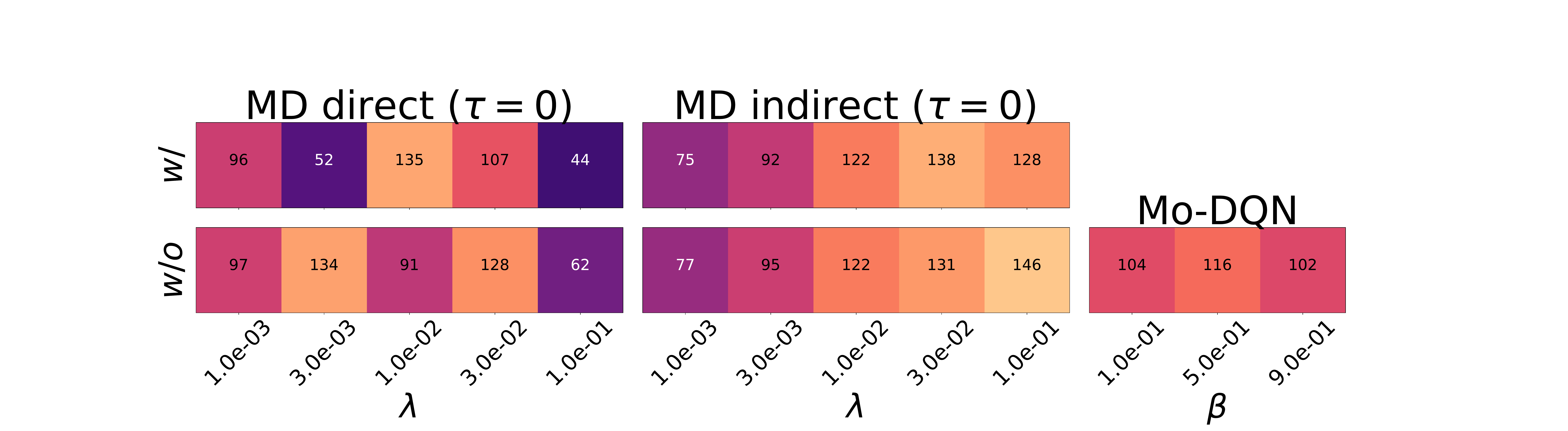}
    \vspace{-.2cm}
    \caption{Breakout with complete values.}
    
    \label{fig:breakout_values}
\end{figure}

\begin{figure}[tbh]
    \centering
    \includegraphics[width=0.8\linewidth,trim={2cm 2cm 0cm 4cm},clip]{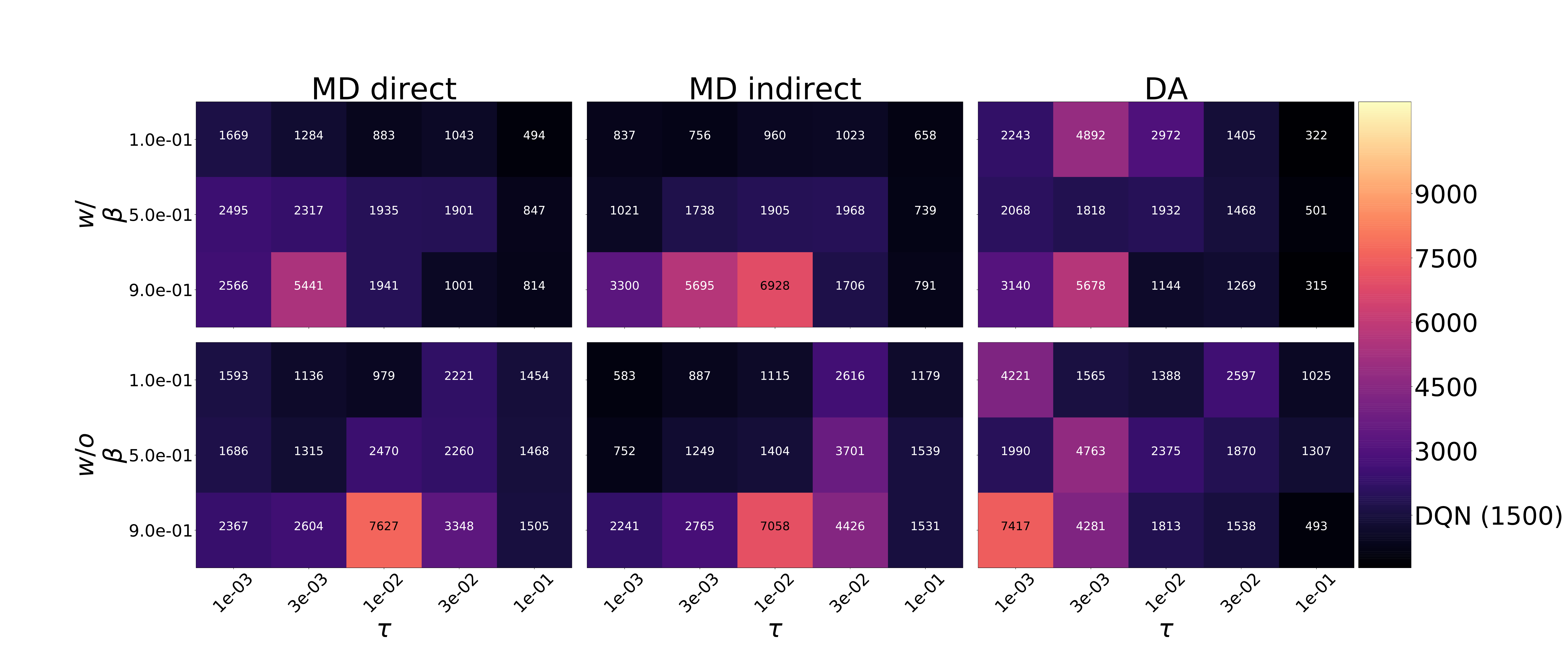}  
    \includegraphics[width=0.8\linewidth,trim={2cm .5cm 4cm 4cm},clip]{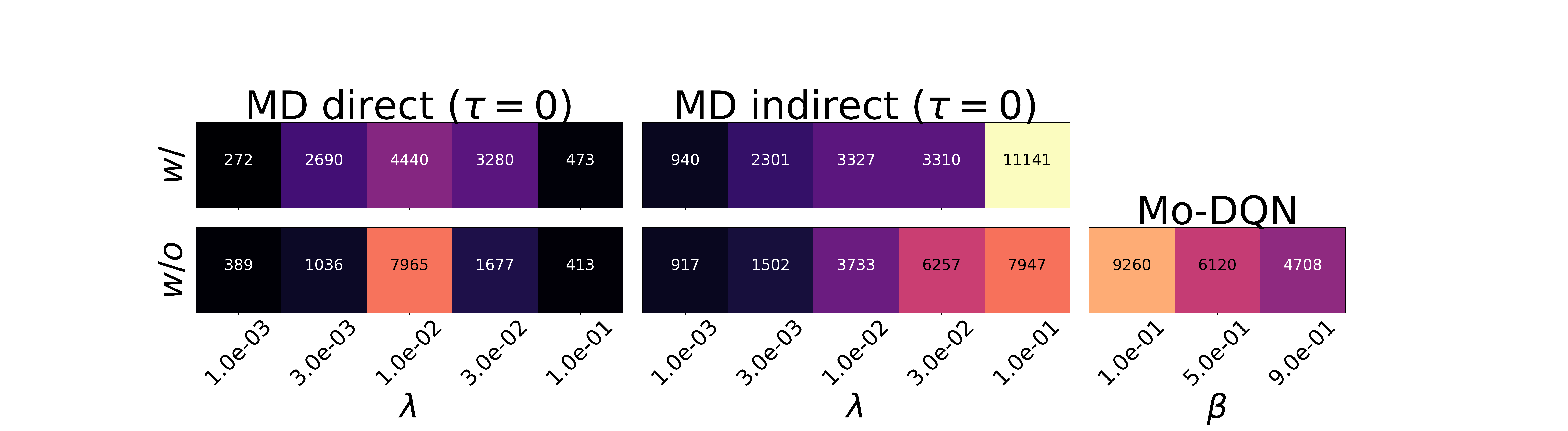}
    \vspace{-.2cm}
    \caption{Seaquest with complete values.}
    
    \label{fig:seaquest_values}
\end{figure}

\paragraph{Full tables.} We also provide the full results of the experiments (those from Section~\ref{sec:experiments} and the new ones). The same plots are reported, expect that we add the exact value of each grid cell for completeness. Results for Carpole and Lunarlander are provided in Figs.~\ref{fig:cartpole_values} and~\ref{fig:lunar_lander_values}, while results for the considered Atari games (Asterix, Breakout and Seaquest) are reported in Figs.~\ref{fig:asterix_values}, \ref{fig:breakout_values} and~\ref{fig:seaquest_values}.

\begin{figure}
\begin{center}
\begin{tabular}{c c c}
     \includegraphics[width=0.3\linewidth]{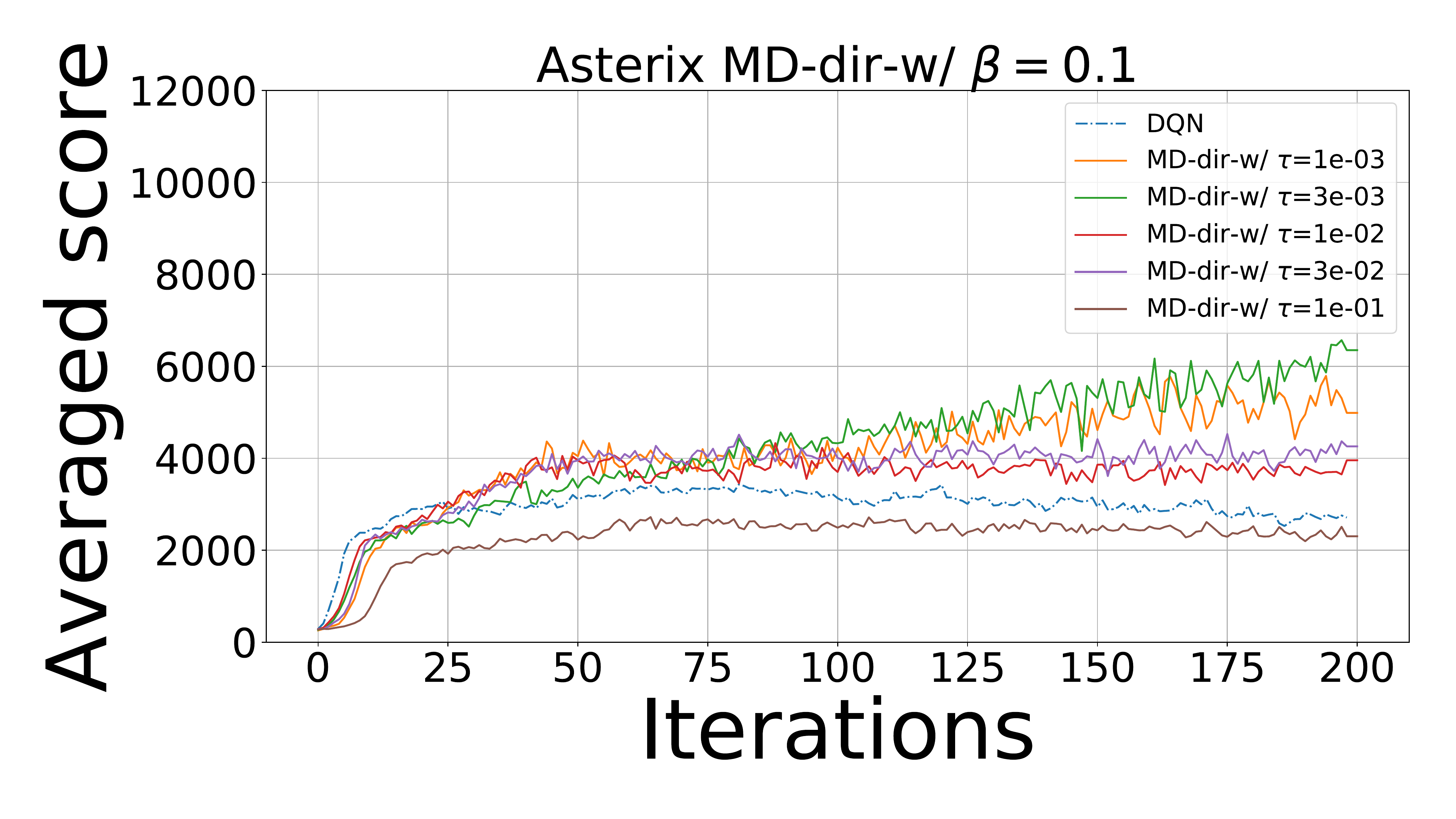} &
     \includegraphics[width=0.3\linewidth]{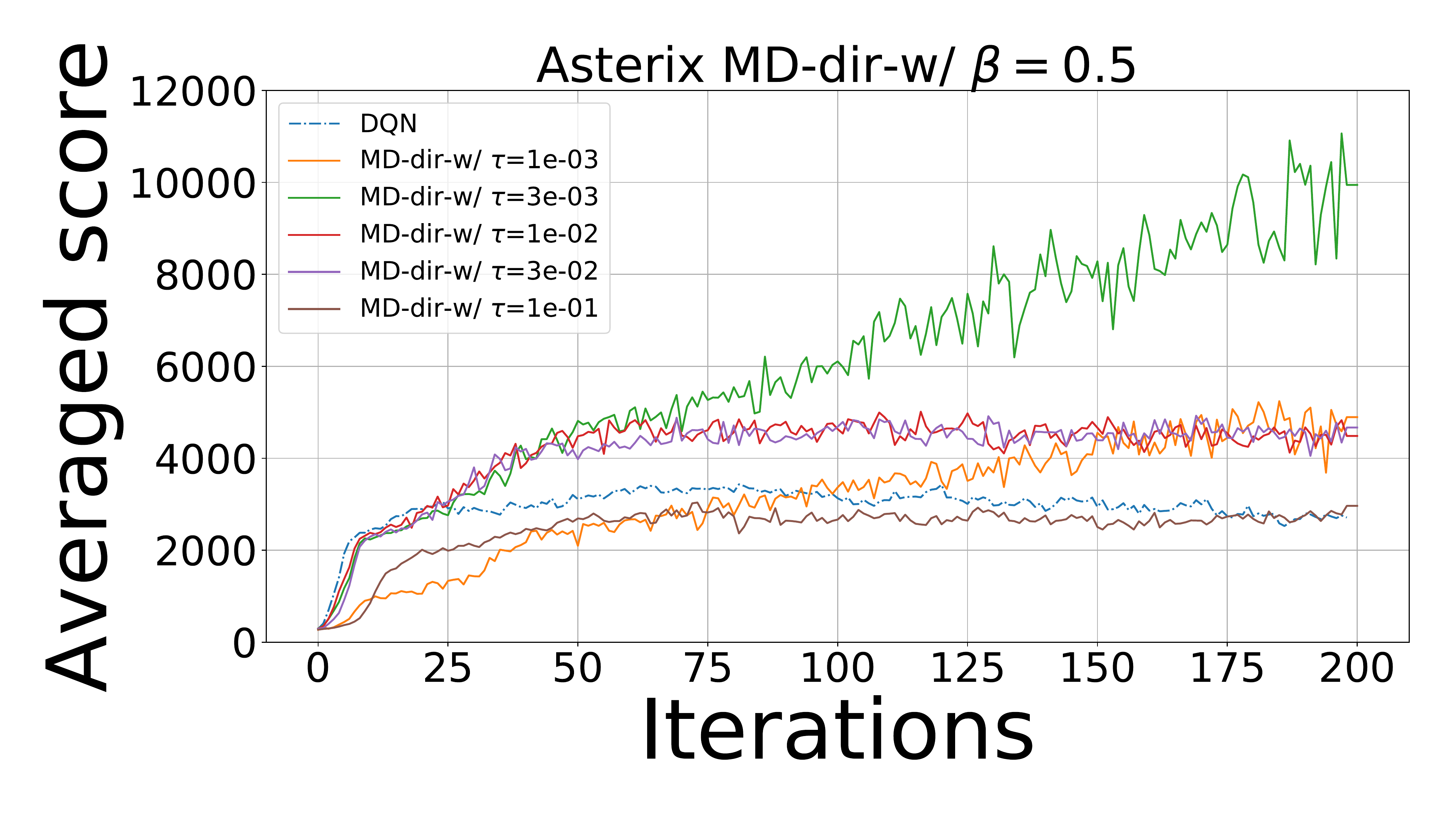} &
     \includegraphics[width=0.3\linewidth]{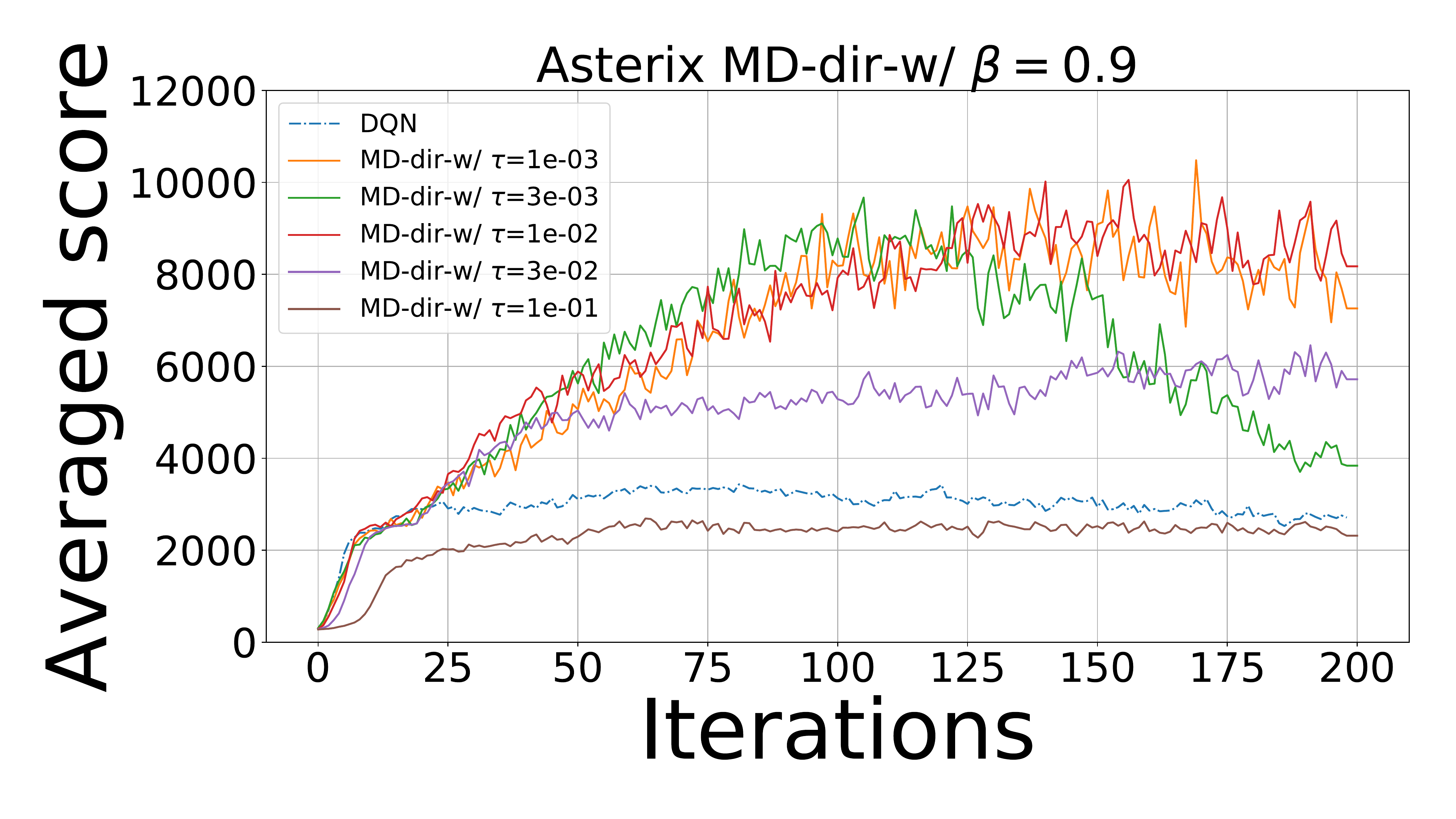} \\
     \includegraphics[width=0.3\linewidth]{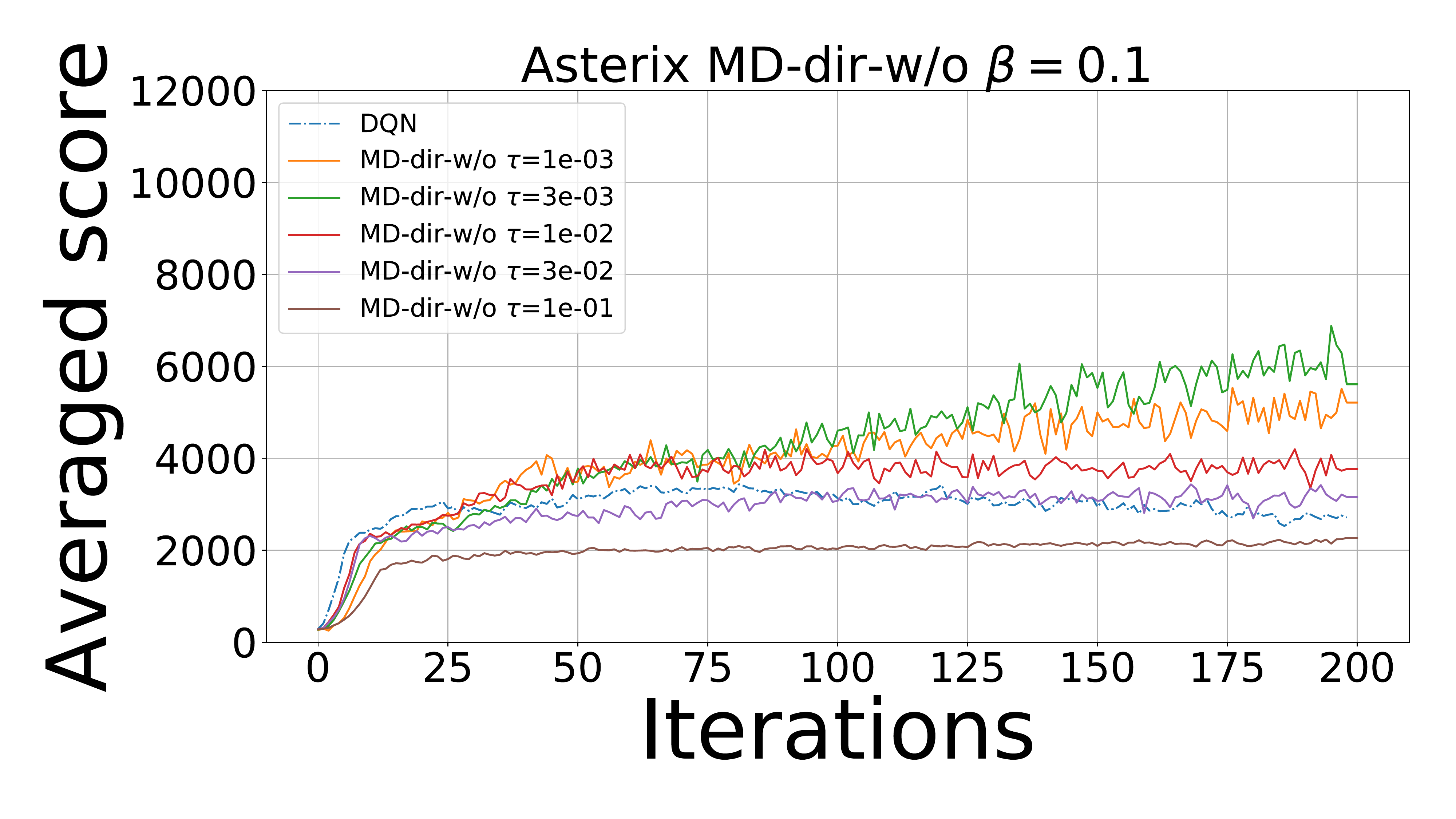} &
     \includegraphics[width=0.3\linewidth]{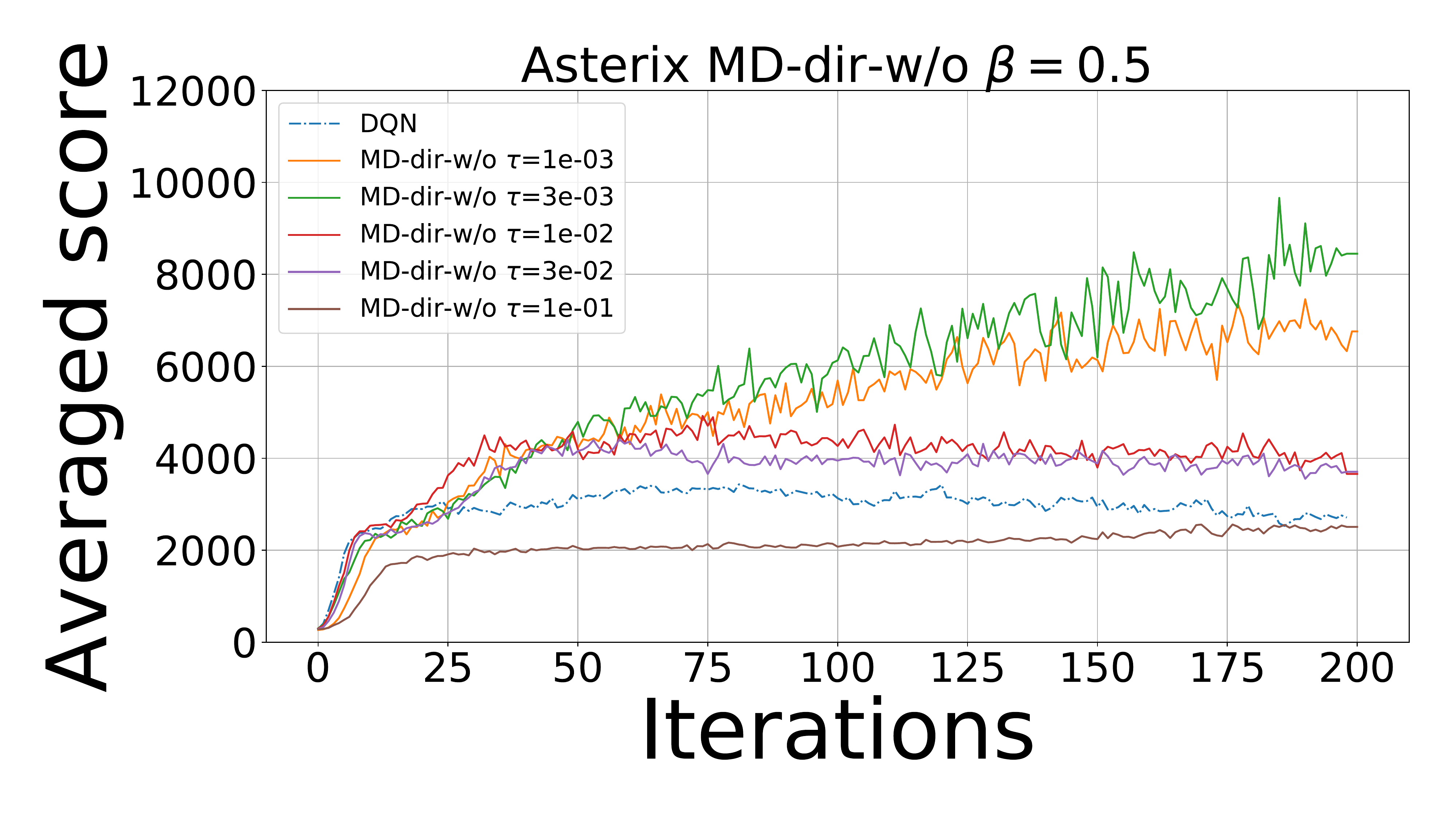} & 
     \includegraphics[width=0.3\linewidth]{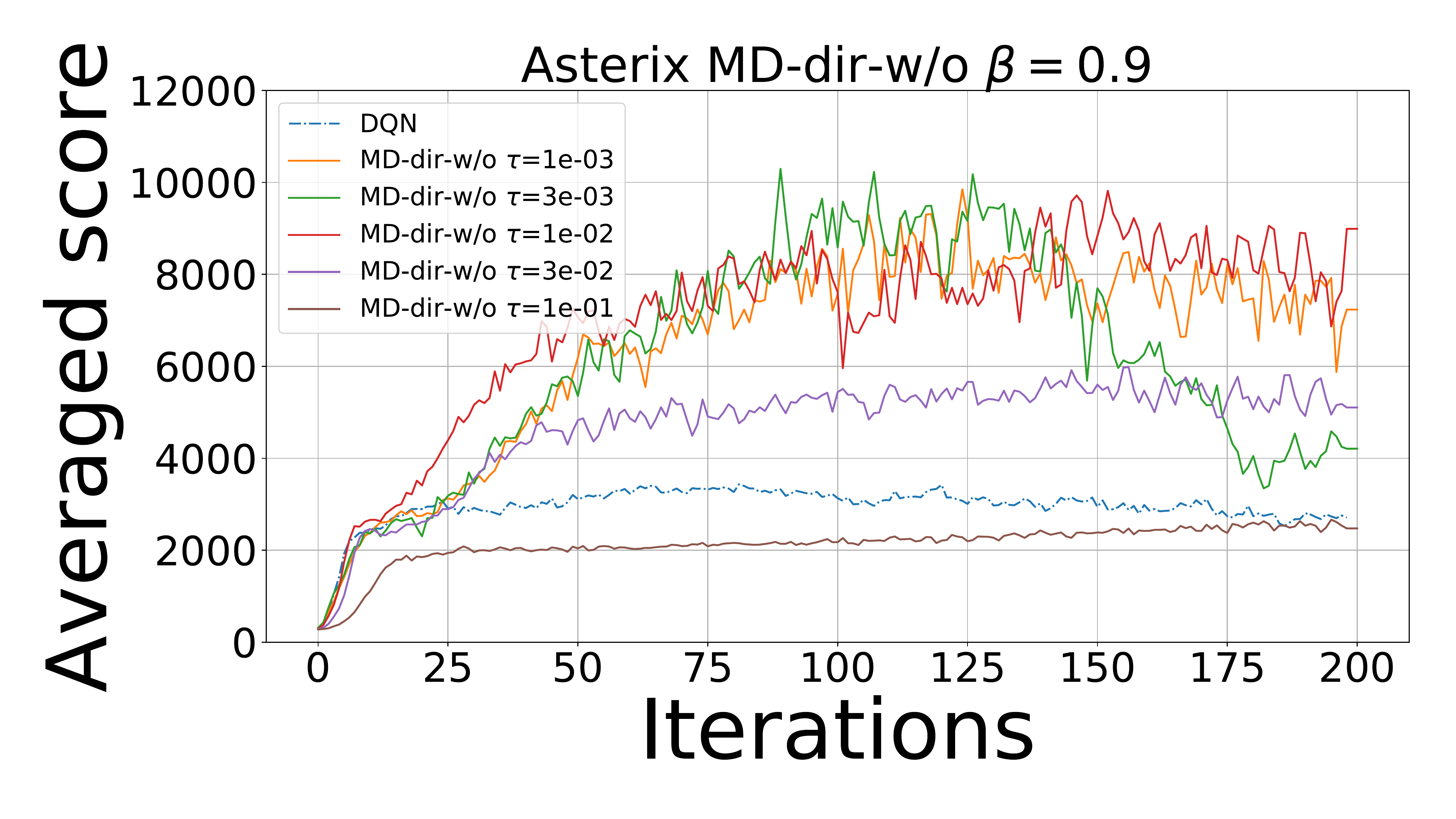}\\
\midrule
     \includegraphics[width=0.3\linewidth]{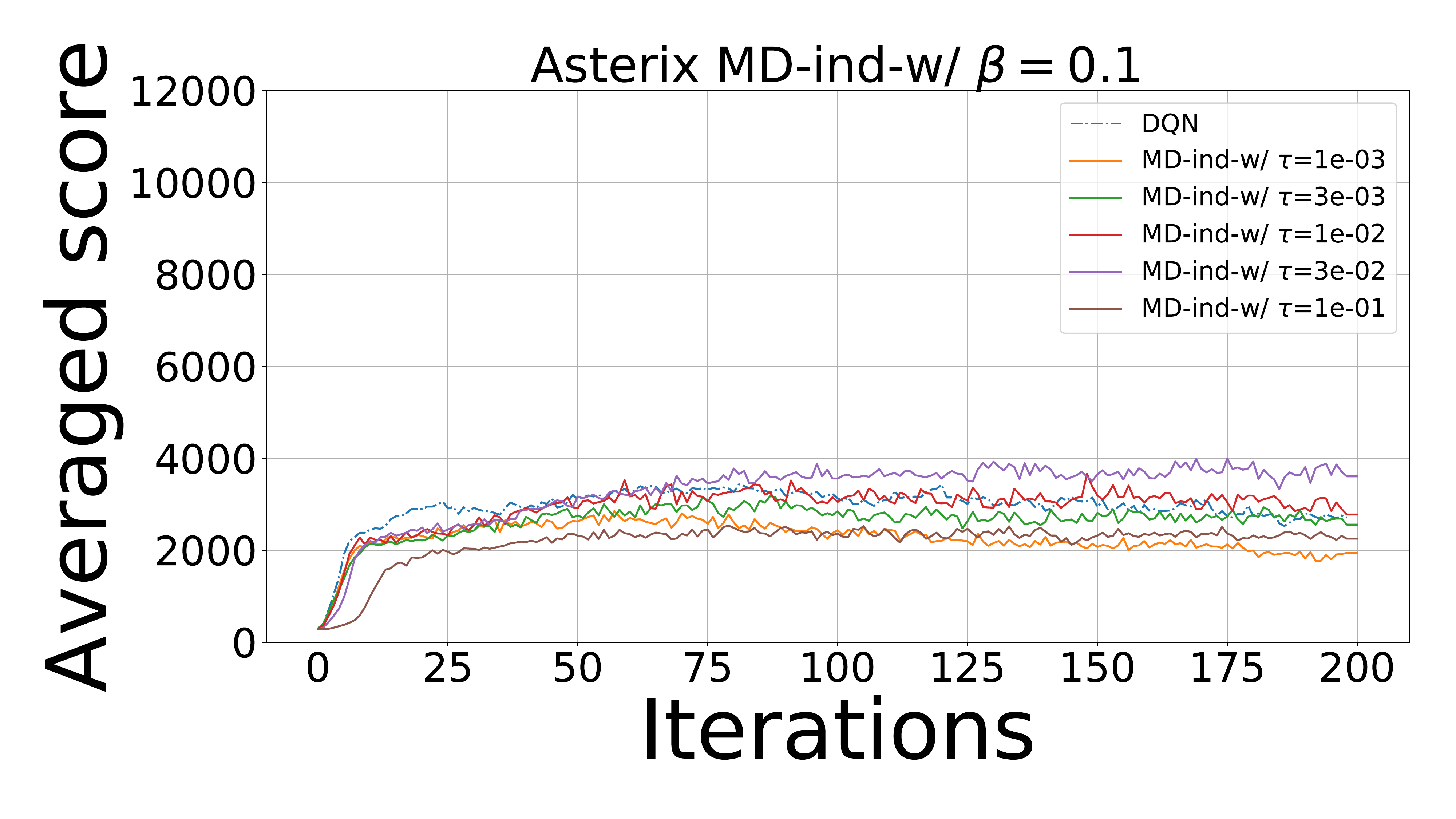} &
     \includegraphics[width=0.3\linewidth]{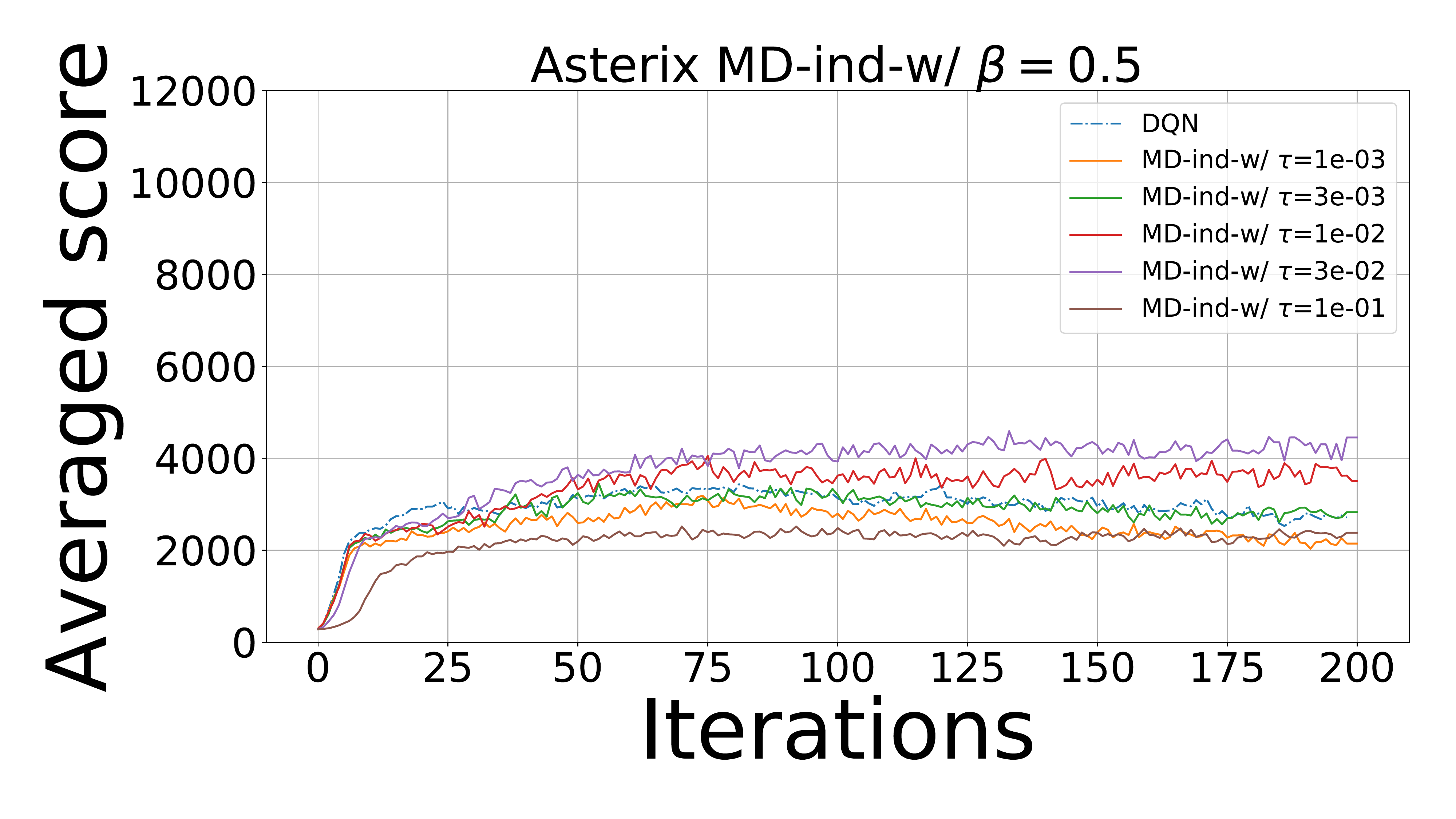} &
     \includegraphics[width=0.3\linewidth]{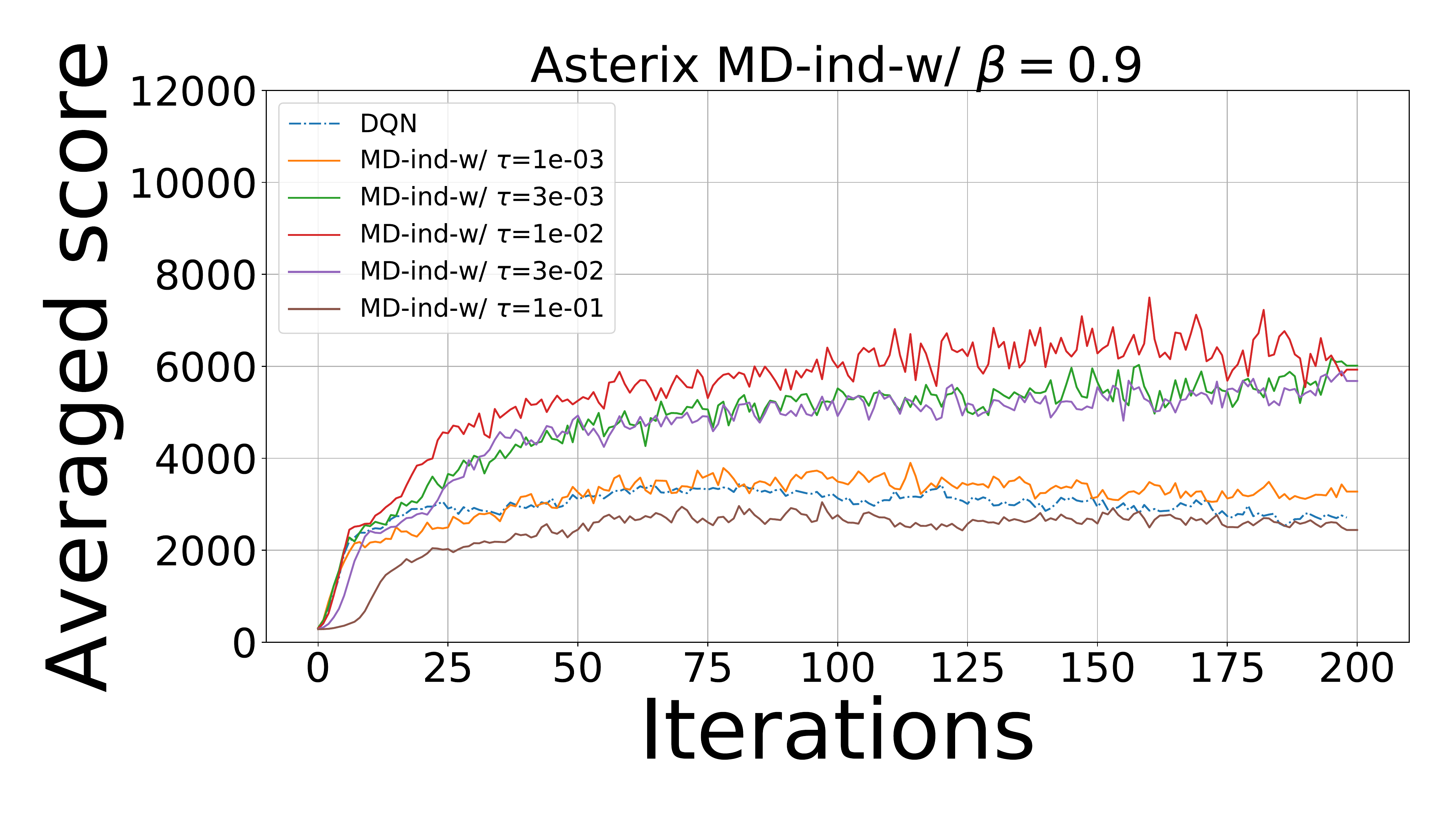}\\ 
     \includegraphics[width=0.3\linewidth]{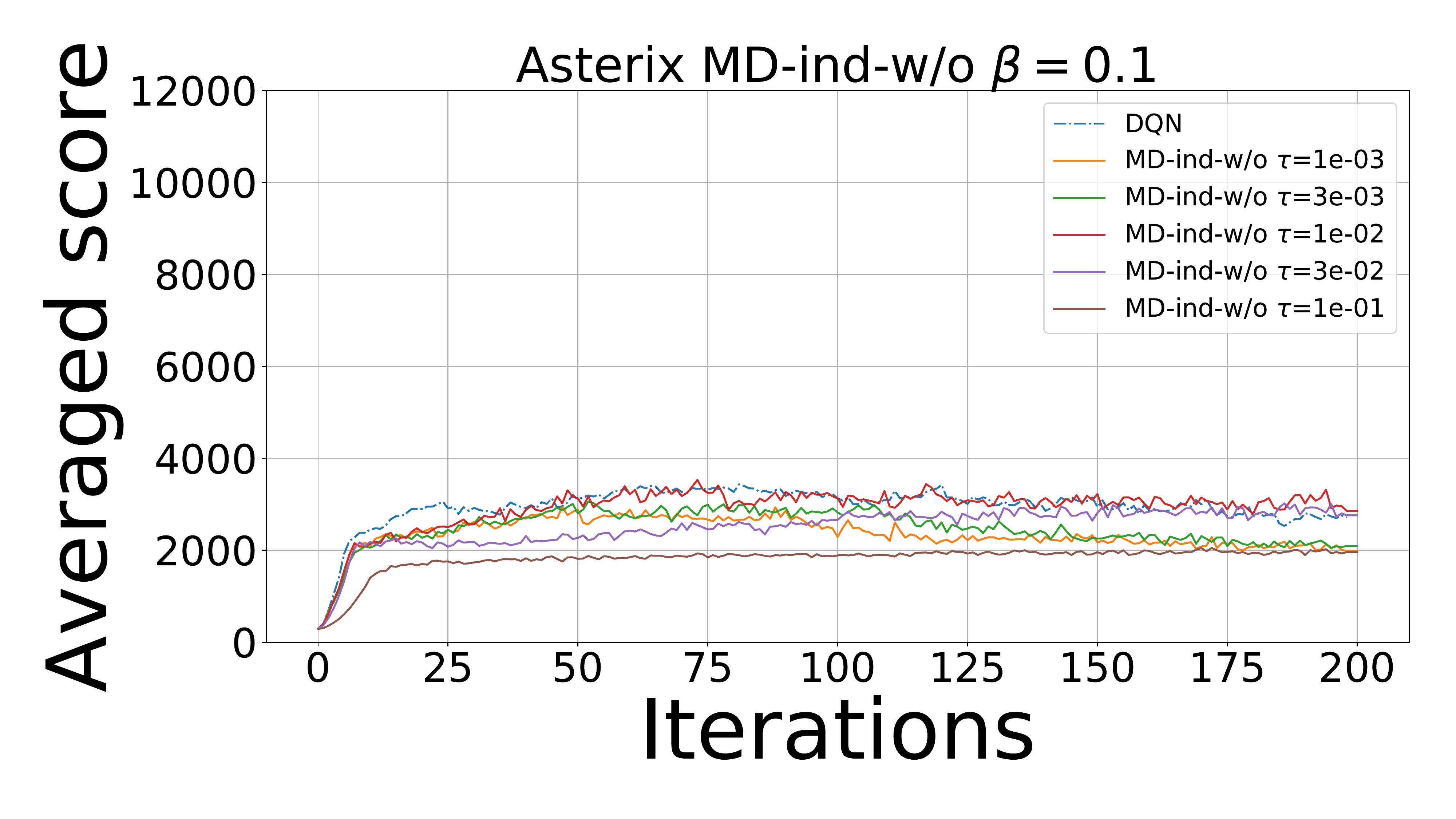} &
     \includegraphics[width=0.3\linewidth]{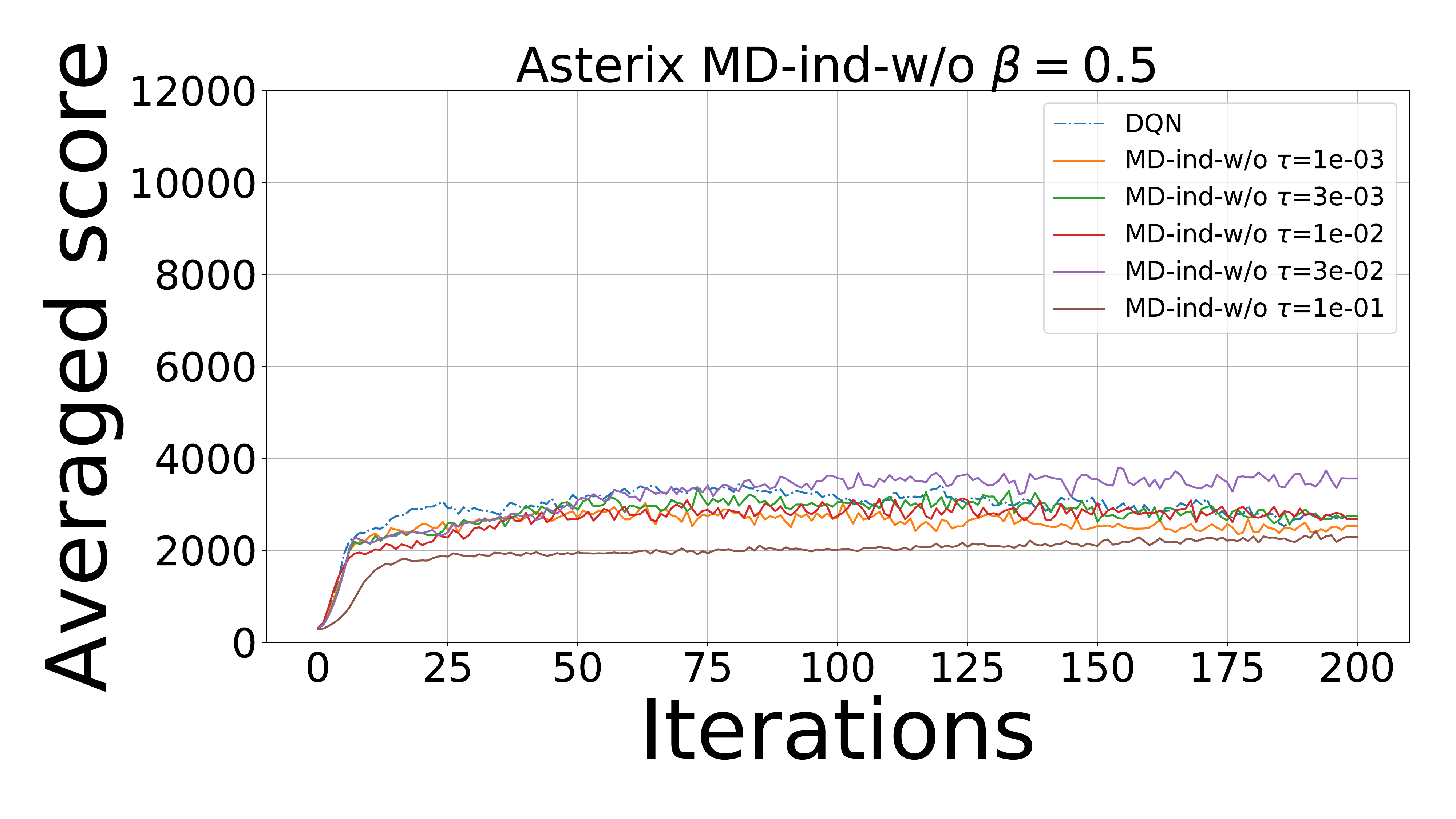} &
     \includegraphics[width=0.3\linewidth]{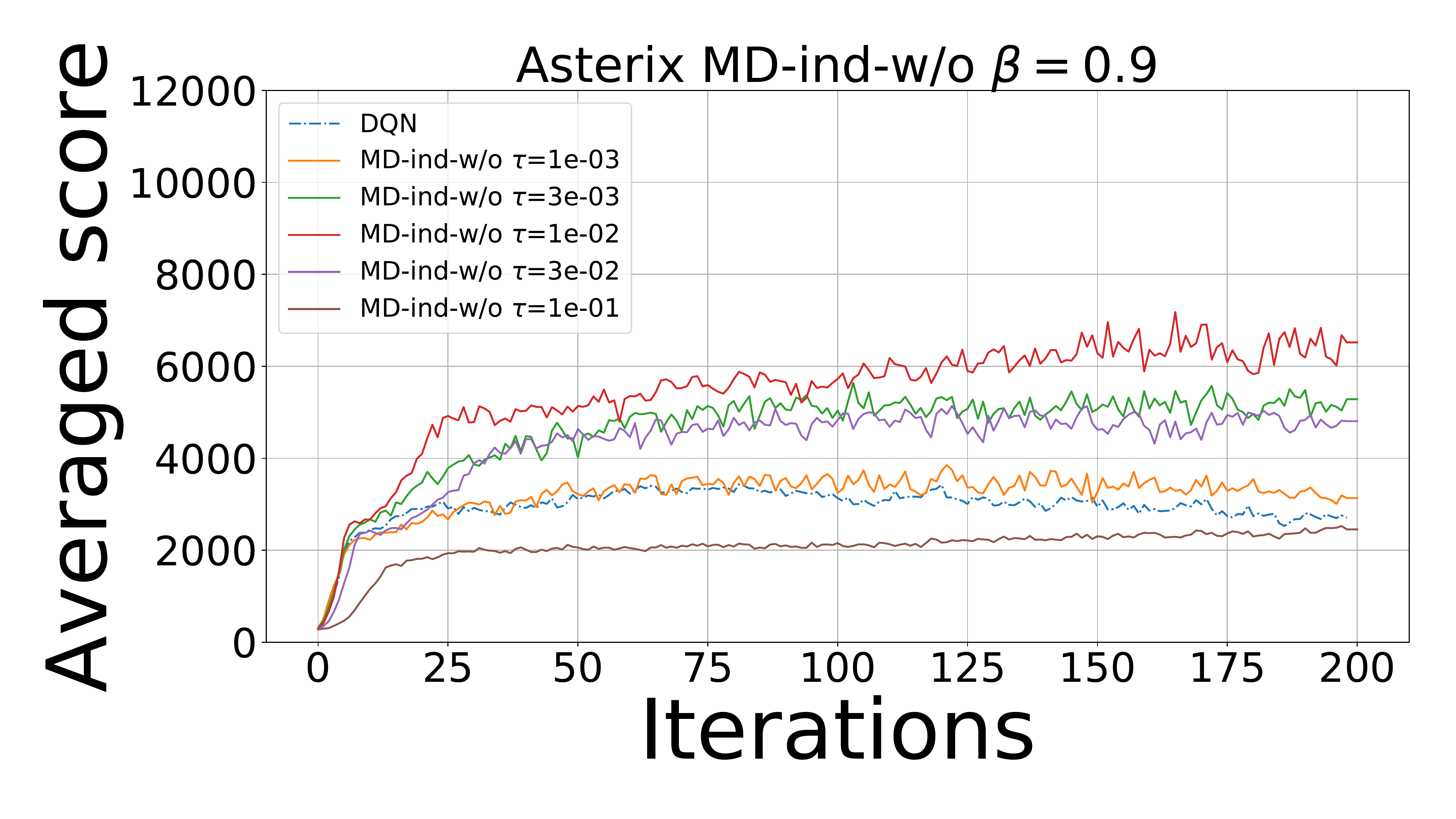}\\
\midrule
     \includegraphics[width=0.3\linewidth]{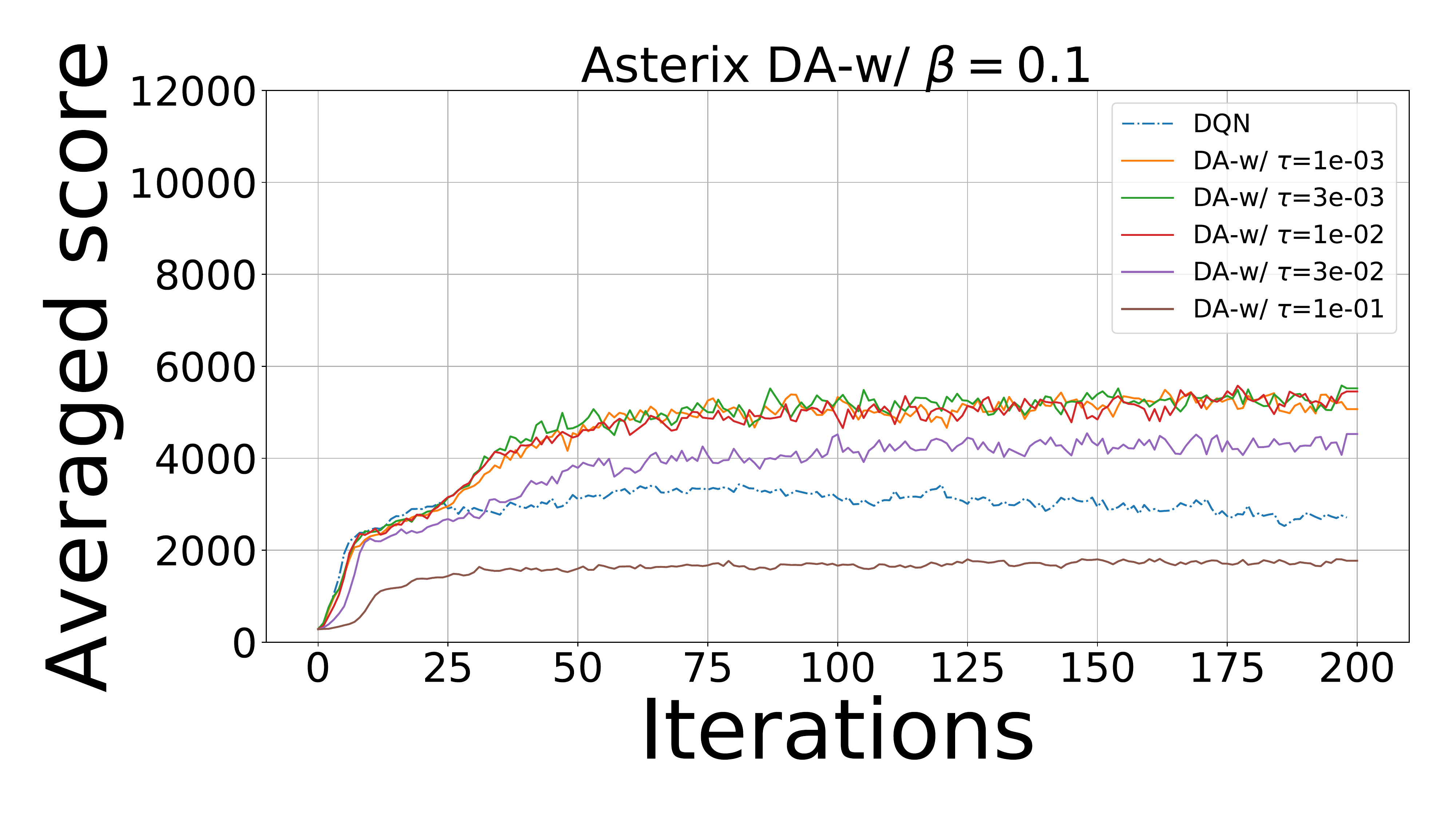} &
     \includegraphics[width=0.3\linewidth]{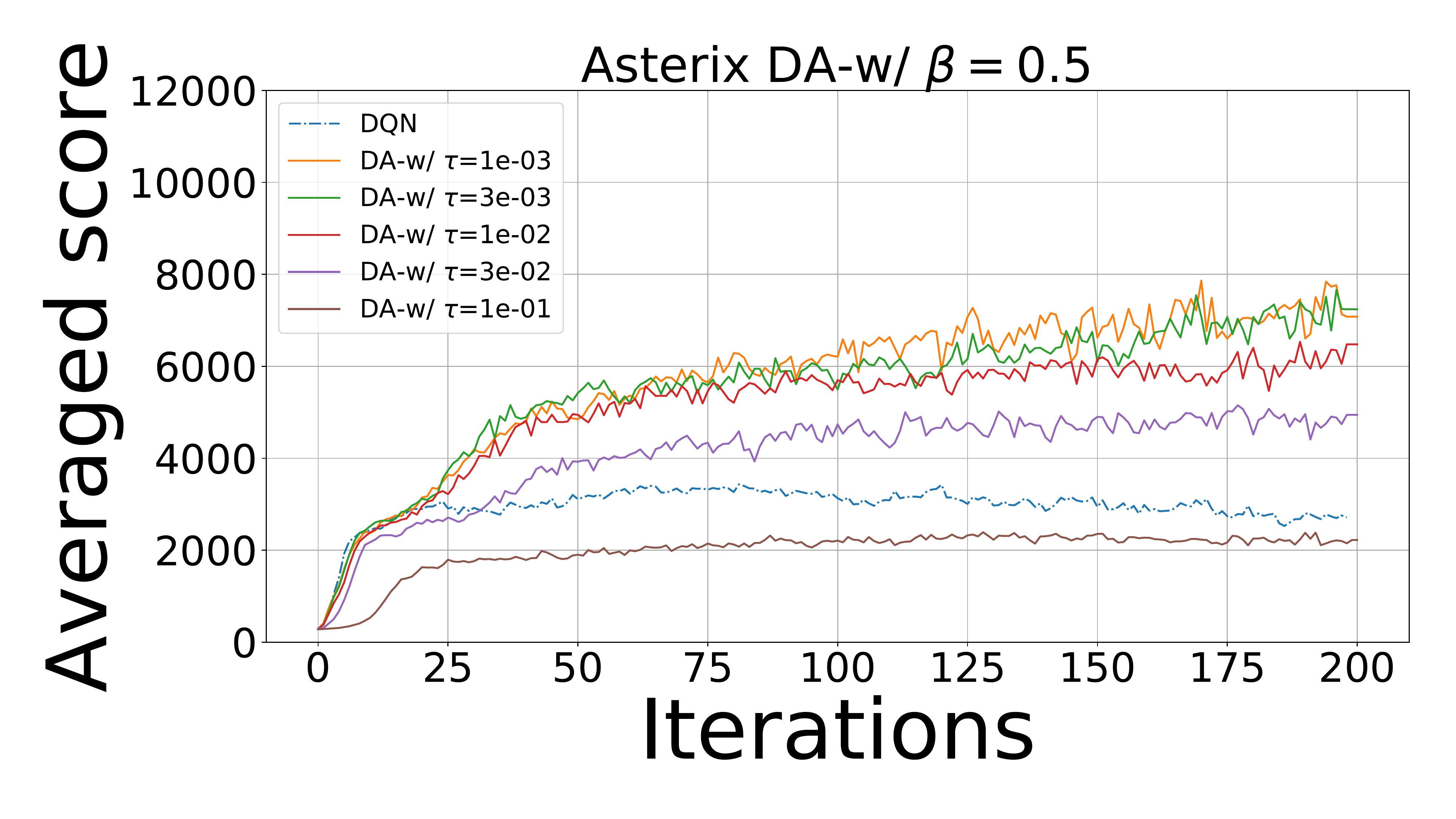} &
     \includegraphics[width=0.3\linewidth]{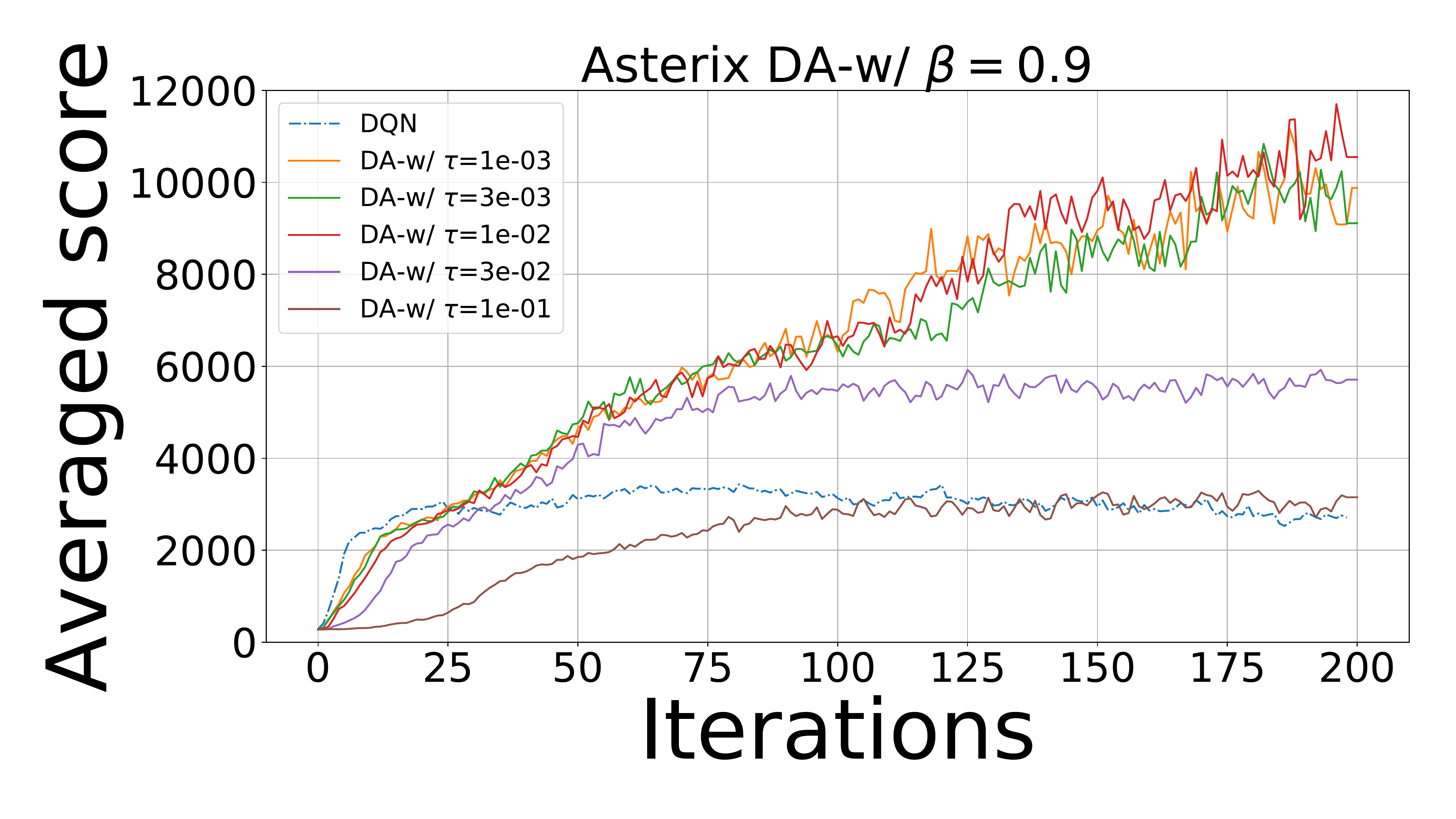} \\
     \includegraphics[width=0.3\linewidth]{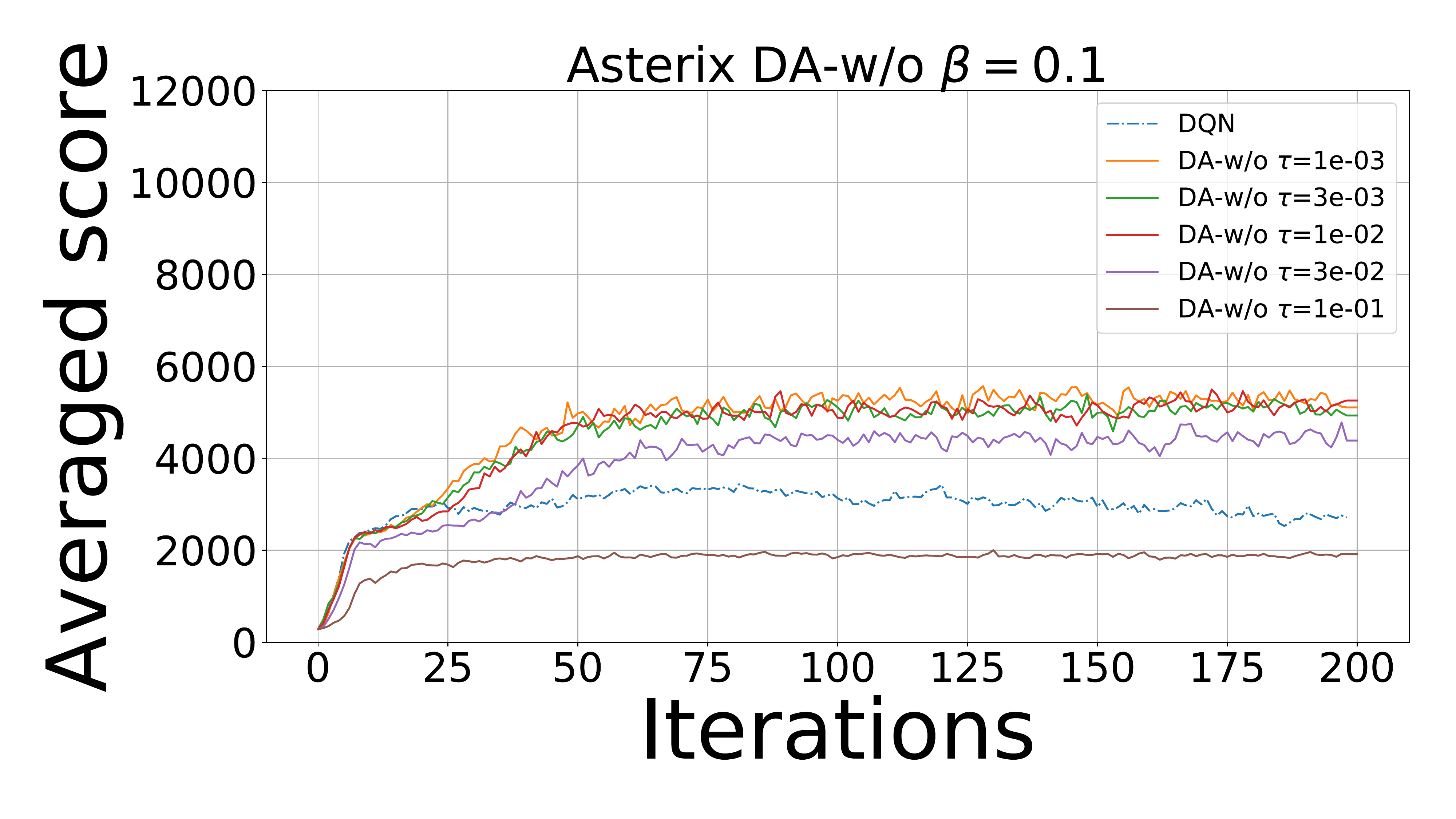} &
     \includegraphics[width=0.3\linewidth]{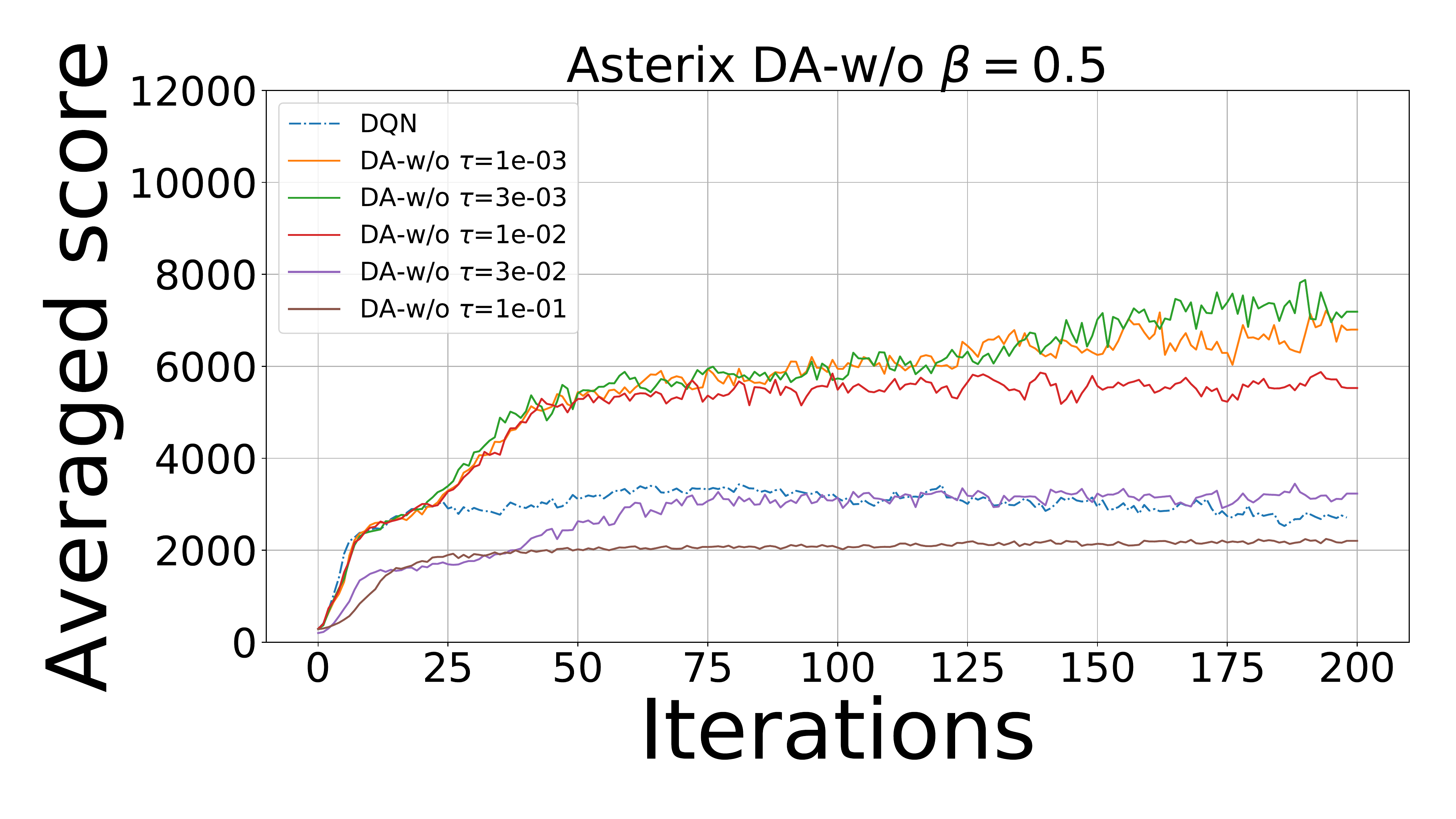} &
     \includegraphics[width=0.3\linewidth]{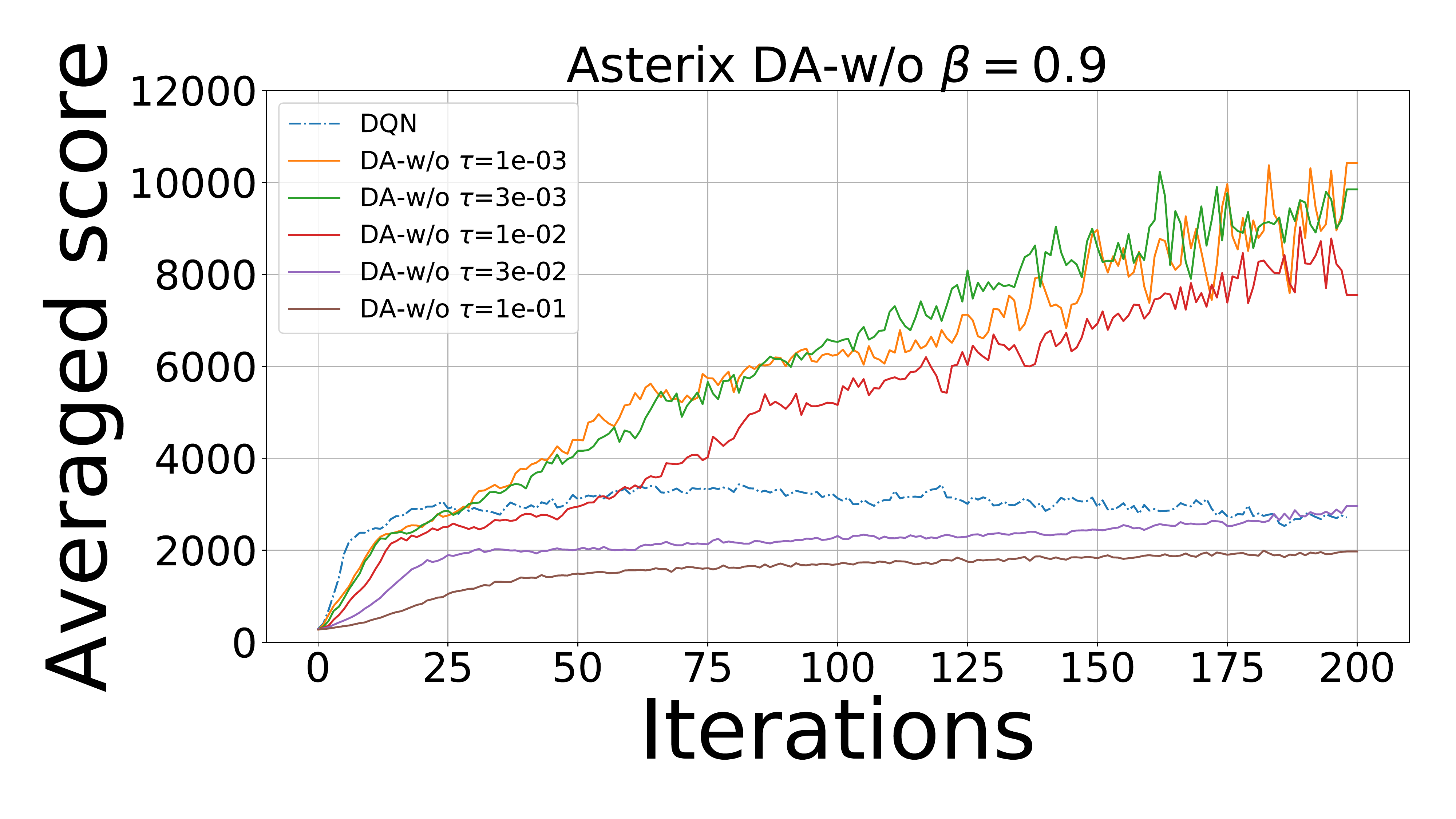} \\
\end{tabular}
\caption{All averaged training scores of MD-dir (top), MD-ind (middle) and DA (bottom), \wir{} and \wor{}, on Asterix, for several values of $\beta$ and $\tau$. Each plot corresponds to one value of $\beta$ (in the titles). In each plot, a curve corresponds to a value of $\tau$: $1e-3$ (orange), $3e-3$ (green), $1e-02$ (red), $3e-2$ (blue), $1e-1$ (brown). The blue dotted line is DQN.\label{fig:asterix_curves}}
\end{center}
\end{figure}

\begin{figure}
\begin{center}
\begin{tabular}{c c c}
     \includegraphics[width=0.3\linewidth]{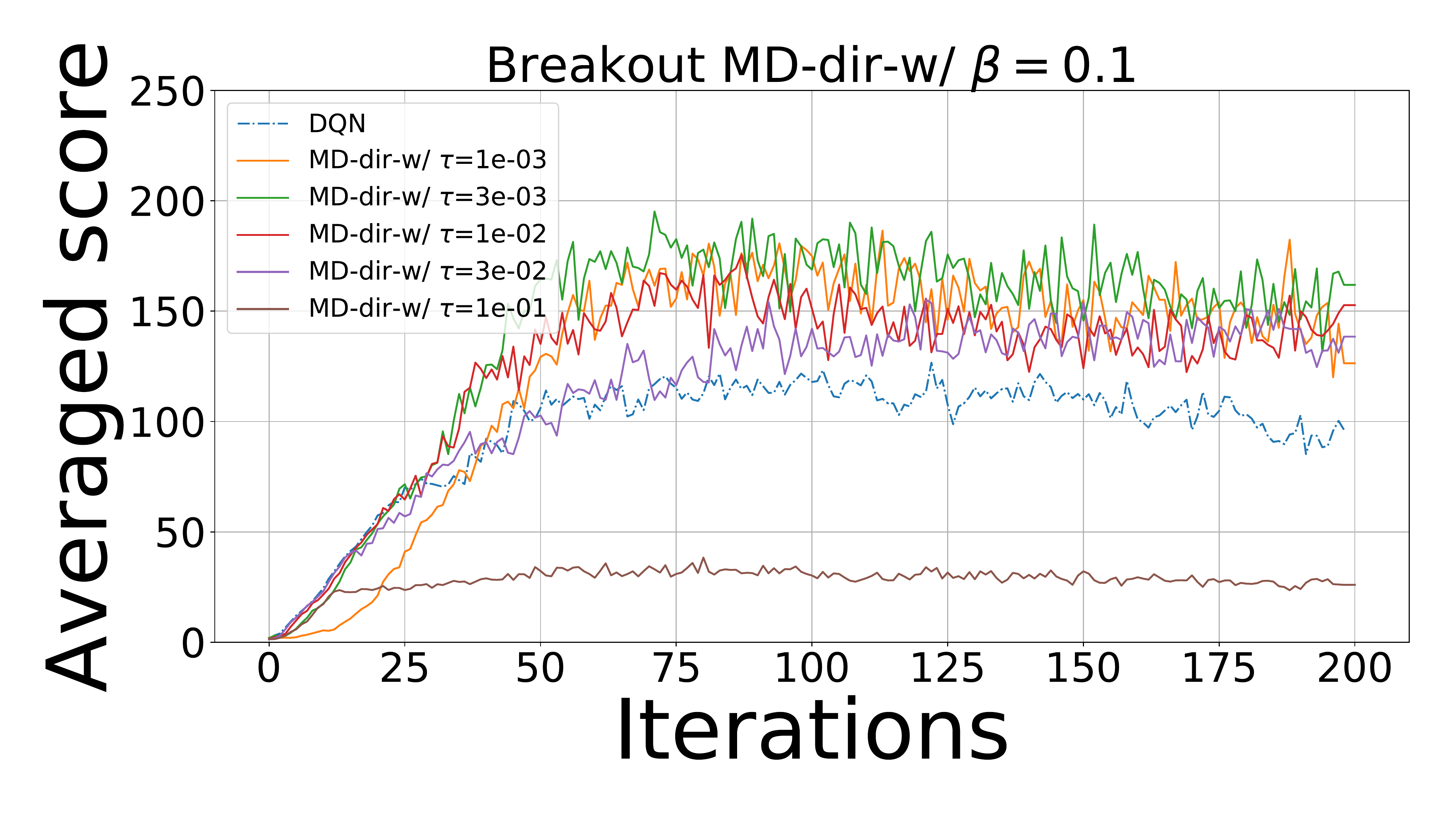} &
     \includegraphics[width=0.3\linewidth]{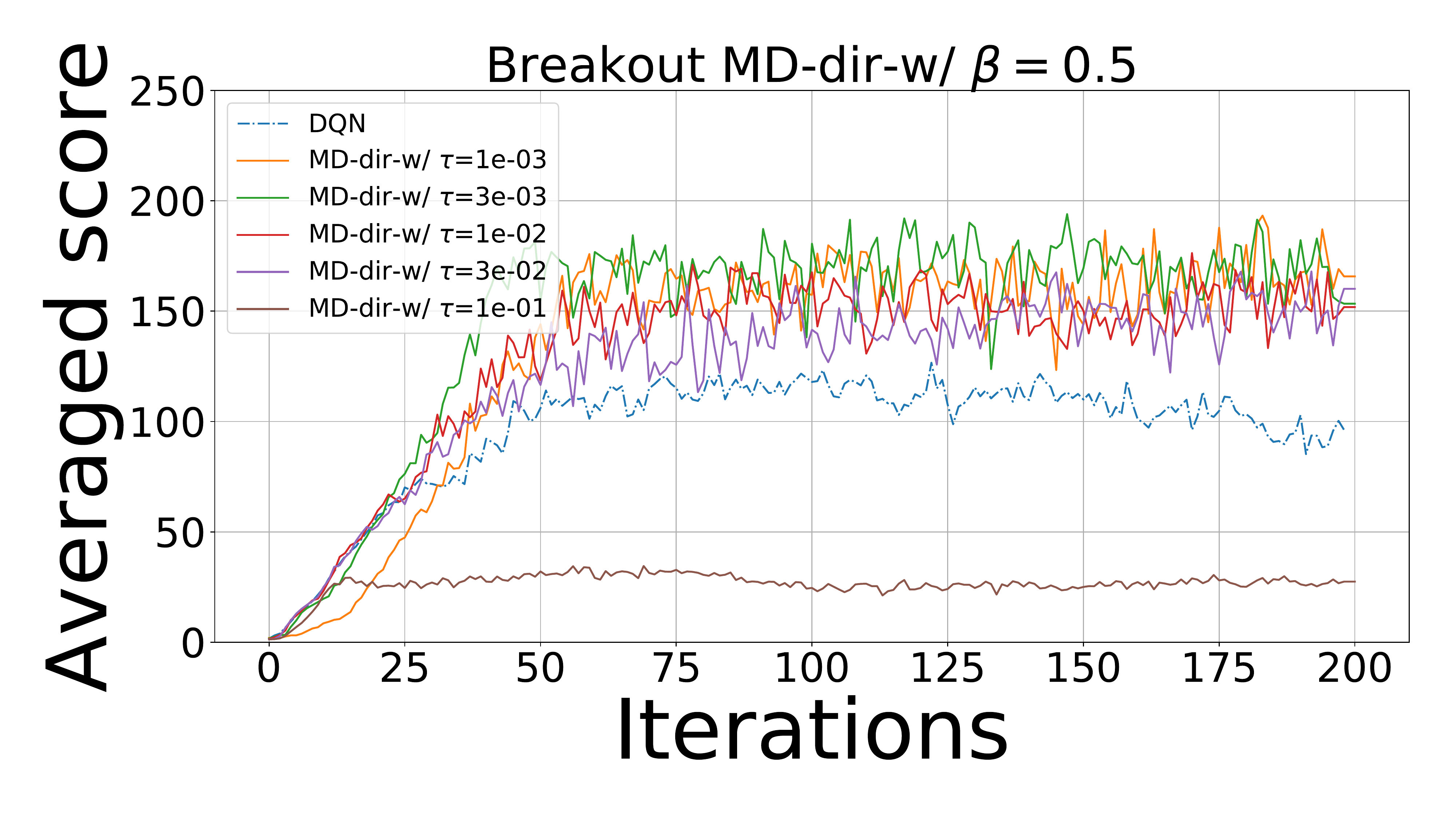} &
     \includegraphics[width=0.3\linewidth]{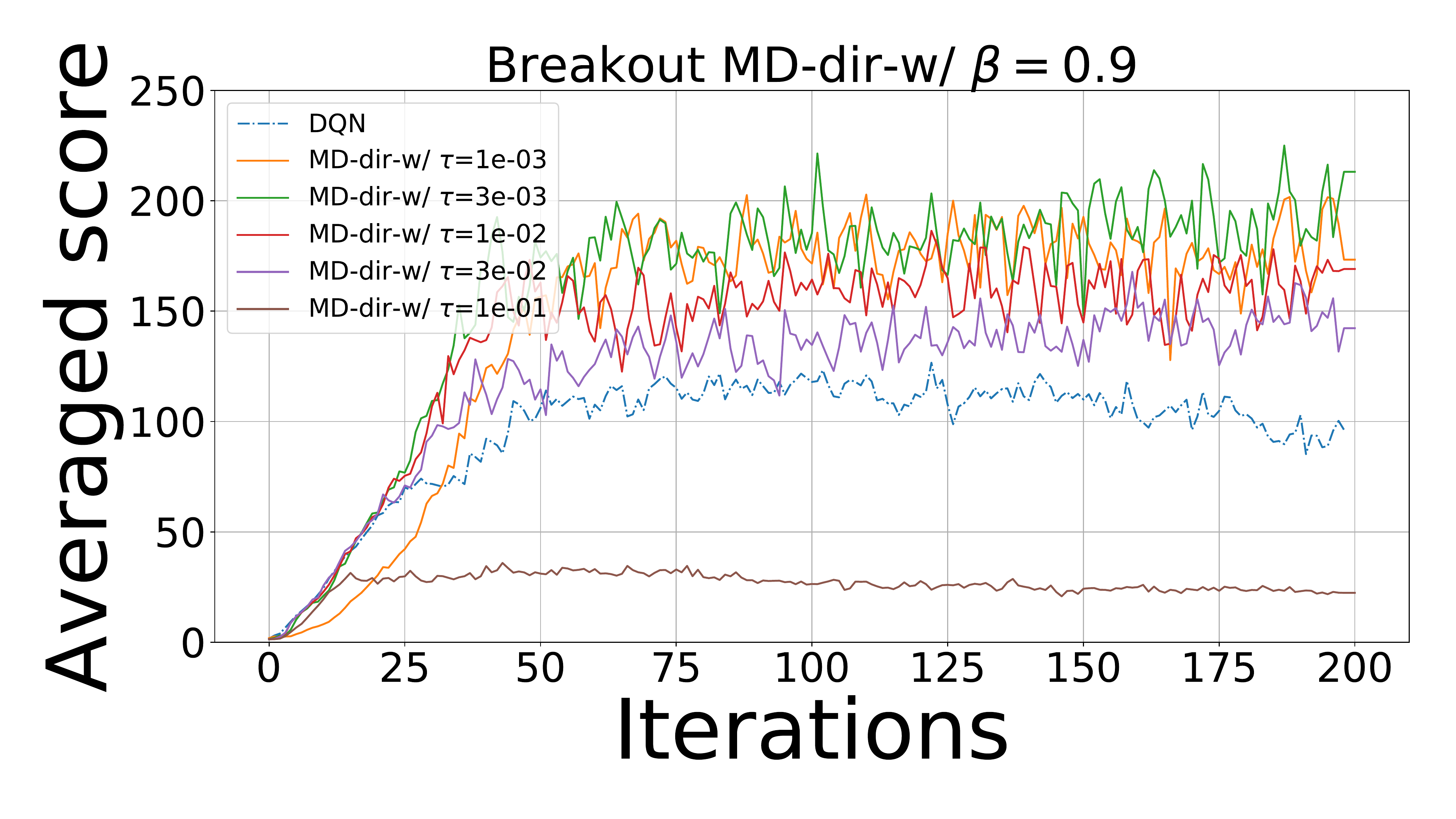}\\ 
     \includegraphics[width=0.3\linewidth]{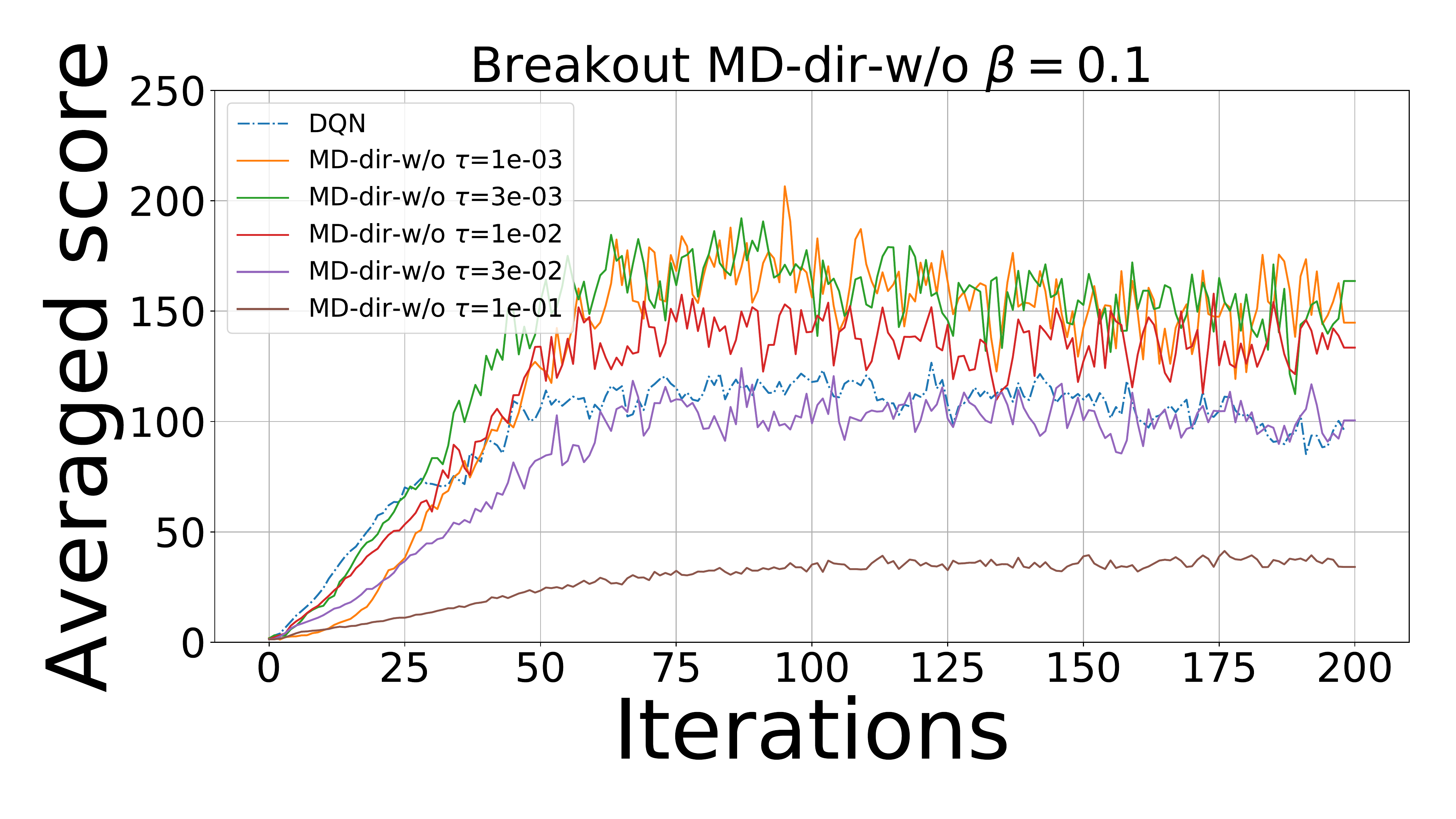}&
     \includegraphics[width=0.3\linewidth]{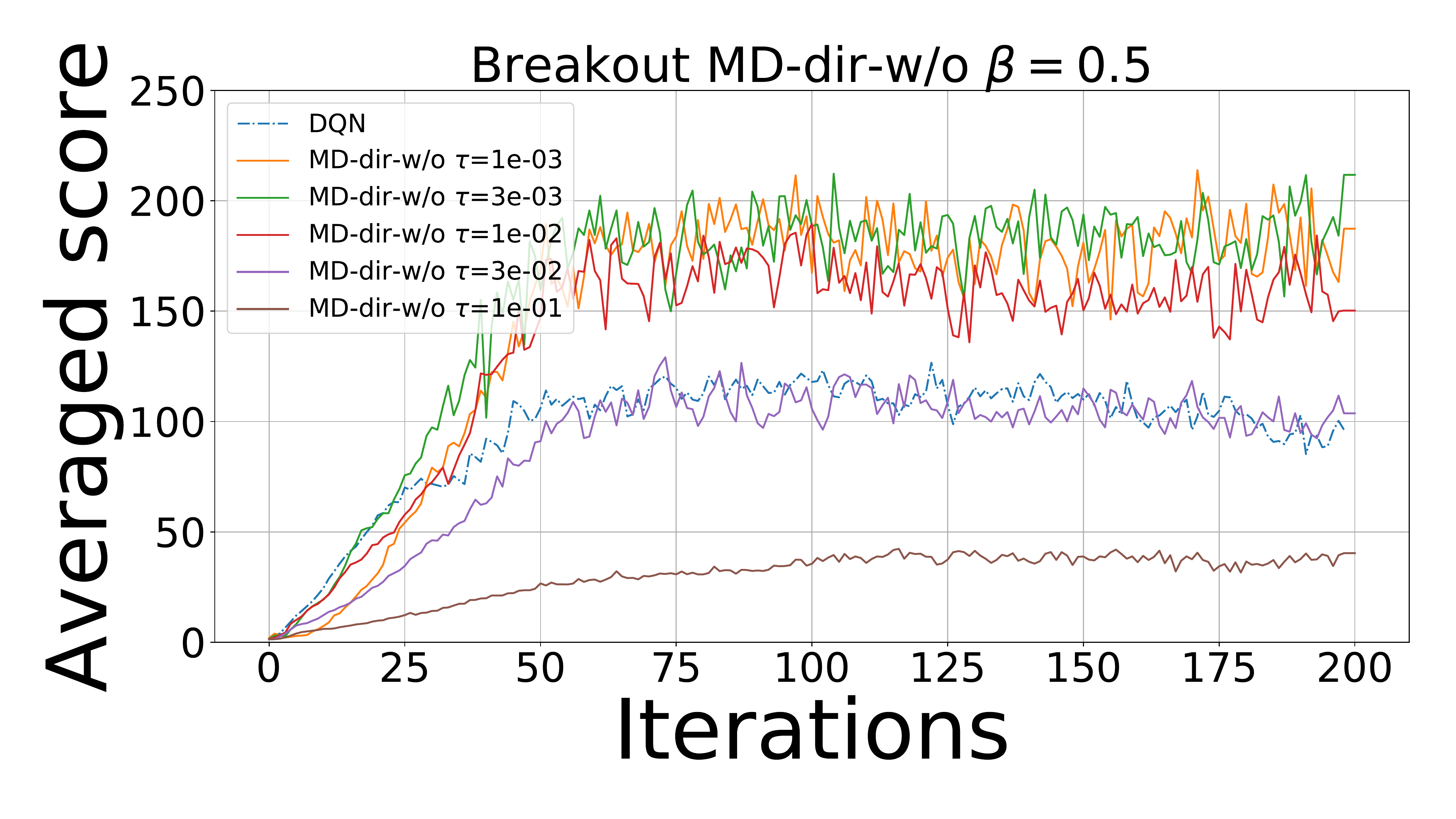} &
     \includegraphics[width=0.3\linewidth]{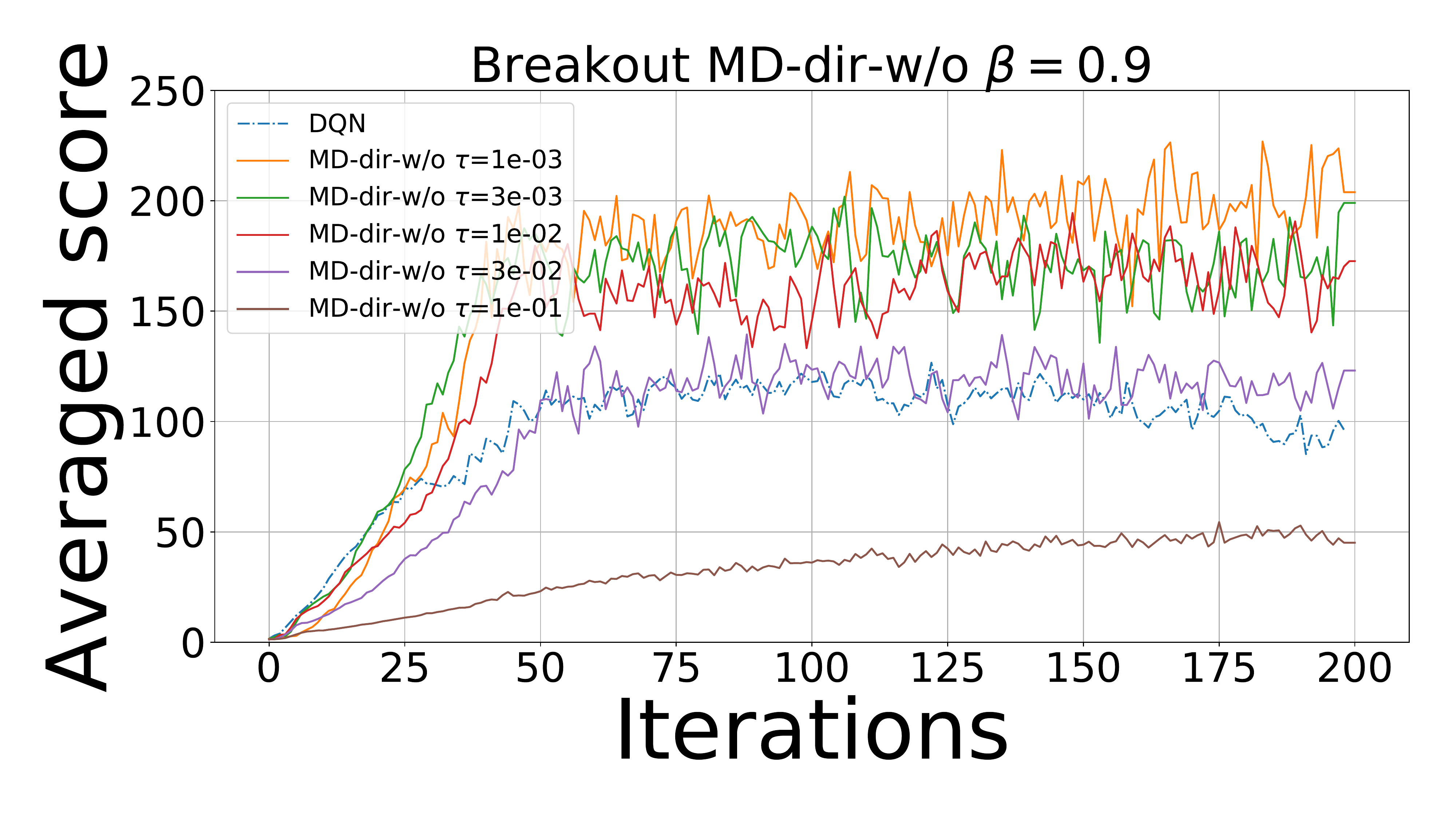}\\
\midrule
     \includegraphics[width=0.3\linewidth]{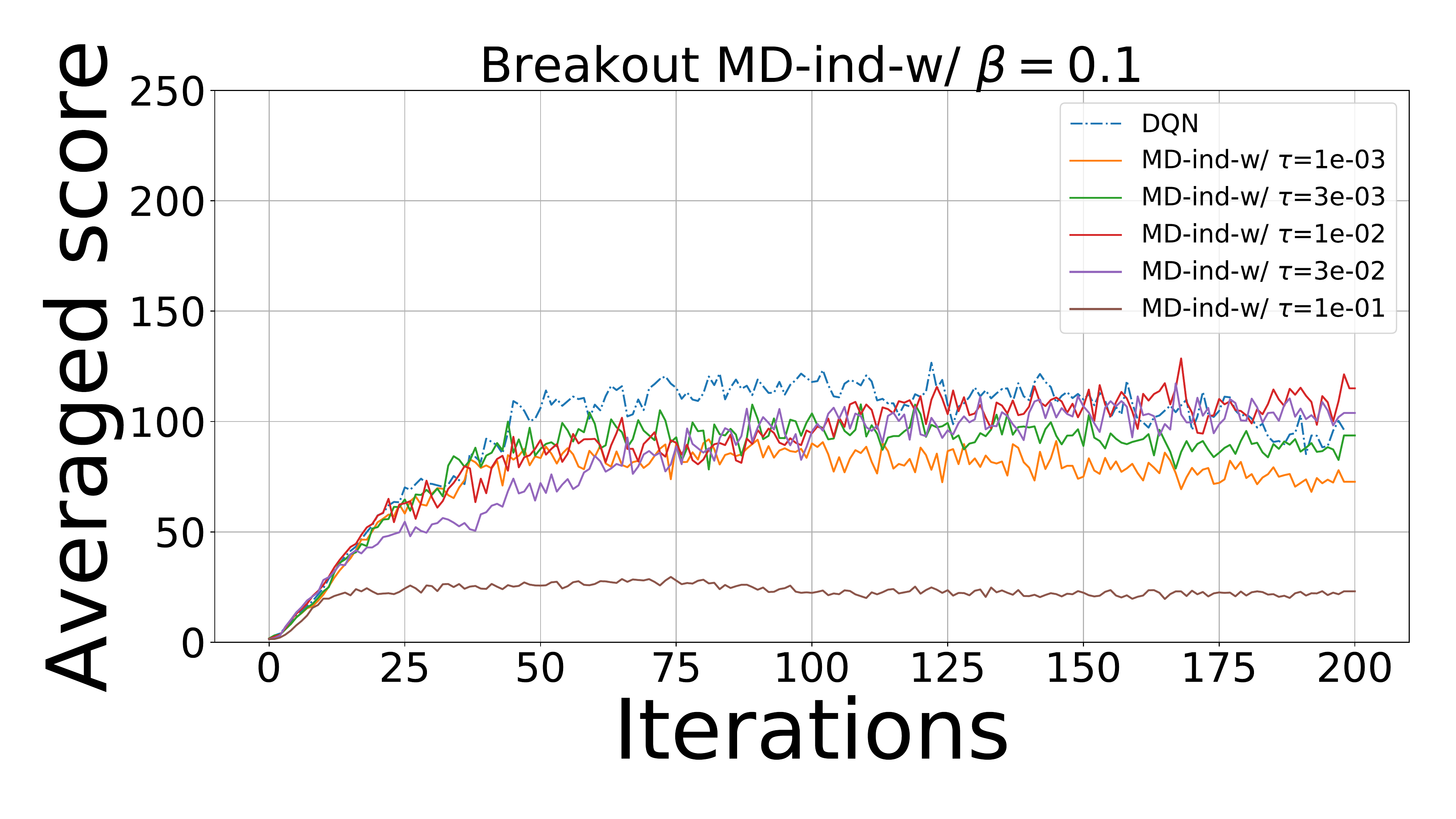} &
     \includegraphics[width=0.3\linewidth]{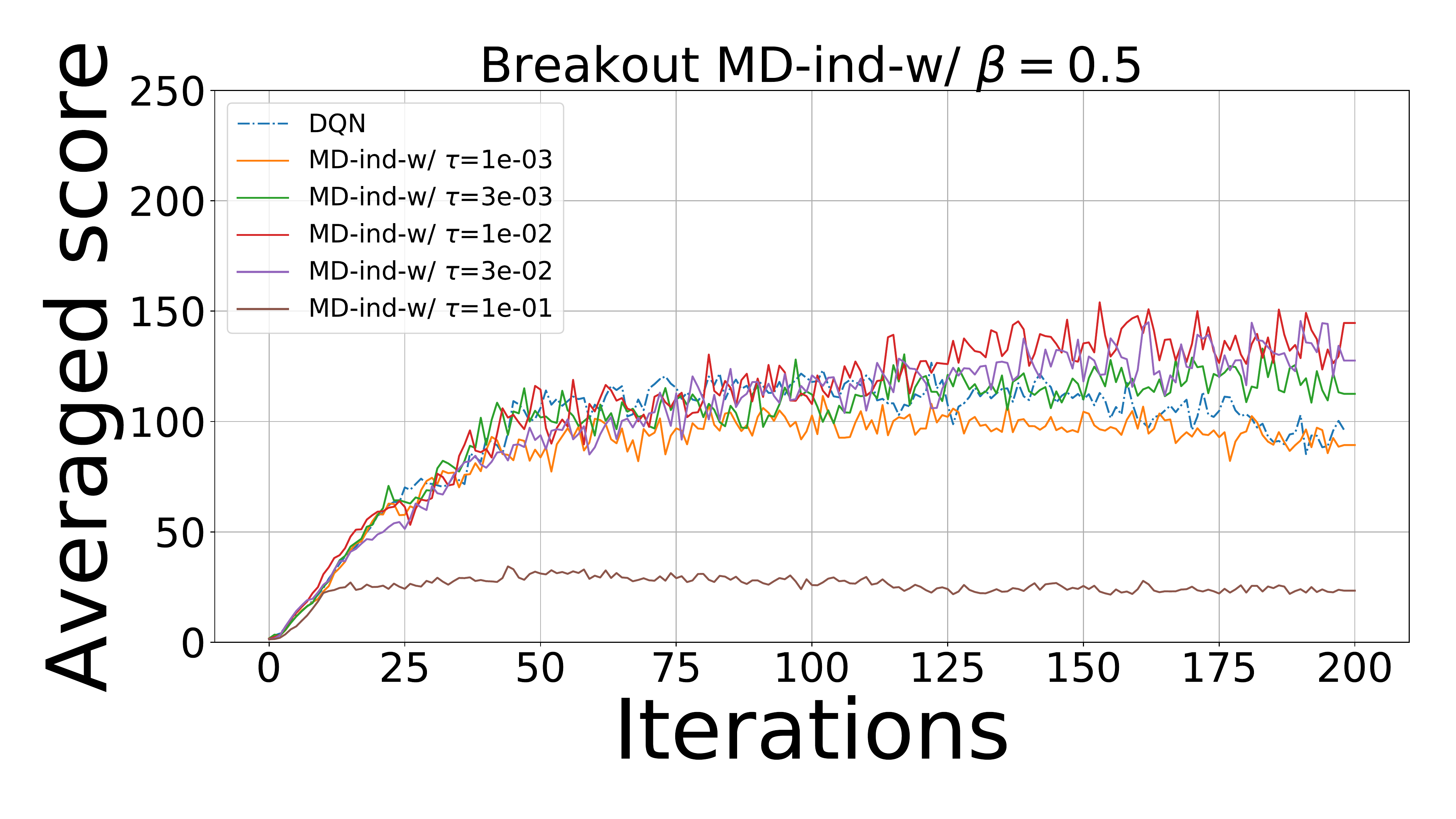} &
     \includegraphics[width=0.3\linewidth]{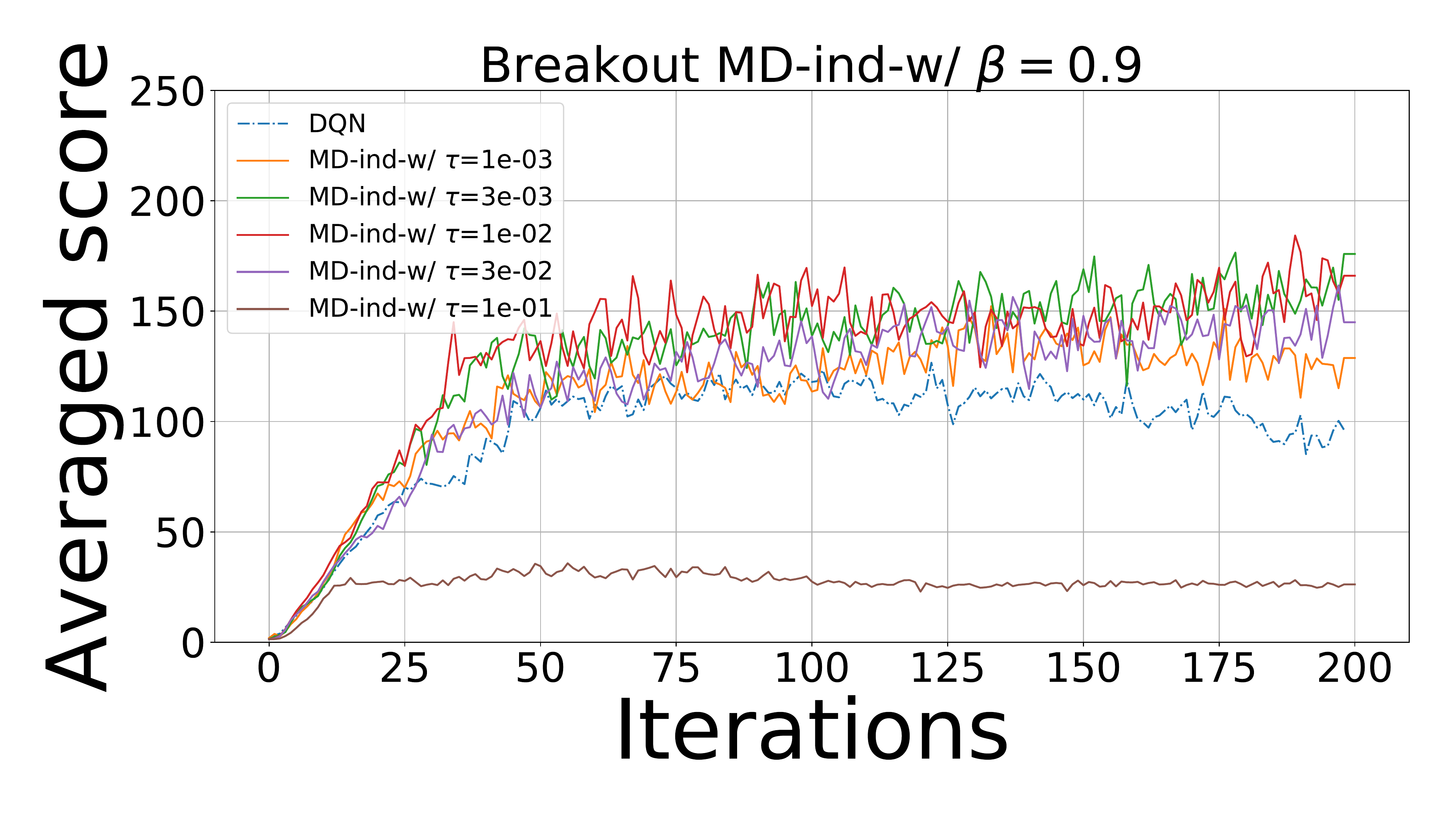} \\
     \includegraphics[width=0.3\linewidth]{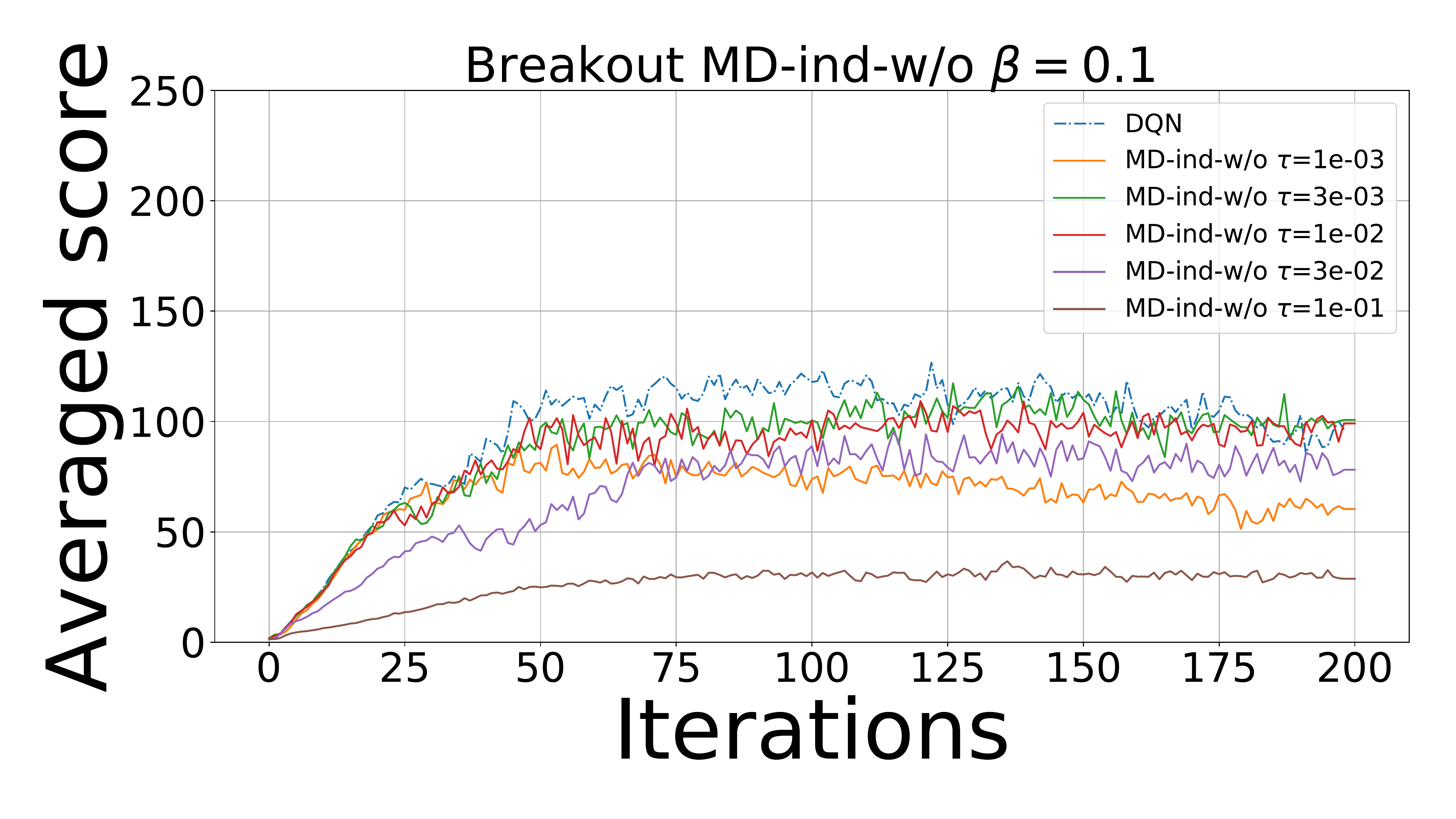} &
     \includegraphics[width=0.3\linewidth]{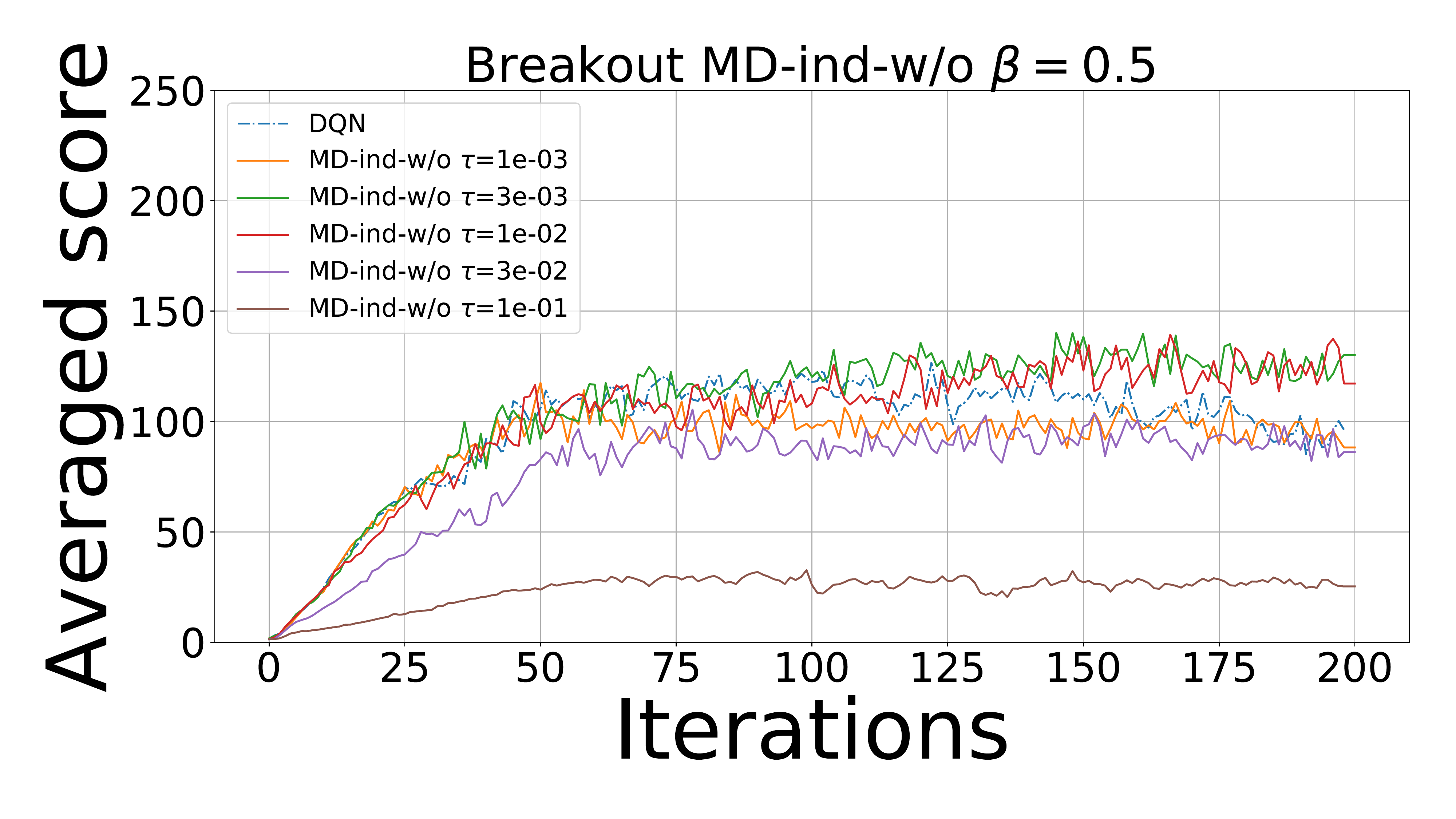} &
     \includegraphics[width=0.3\linewidth]{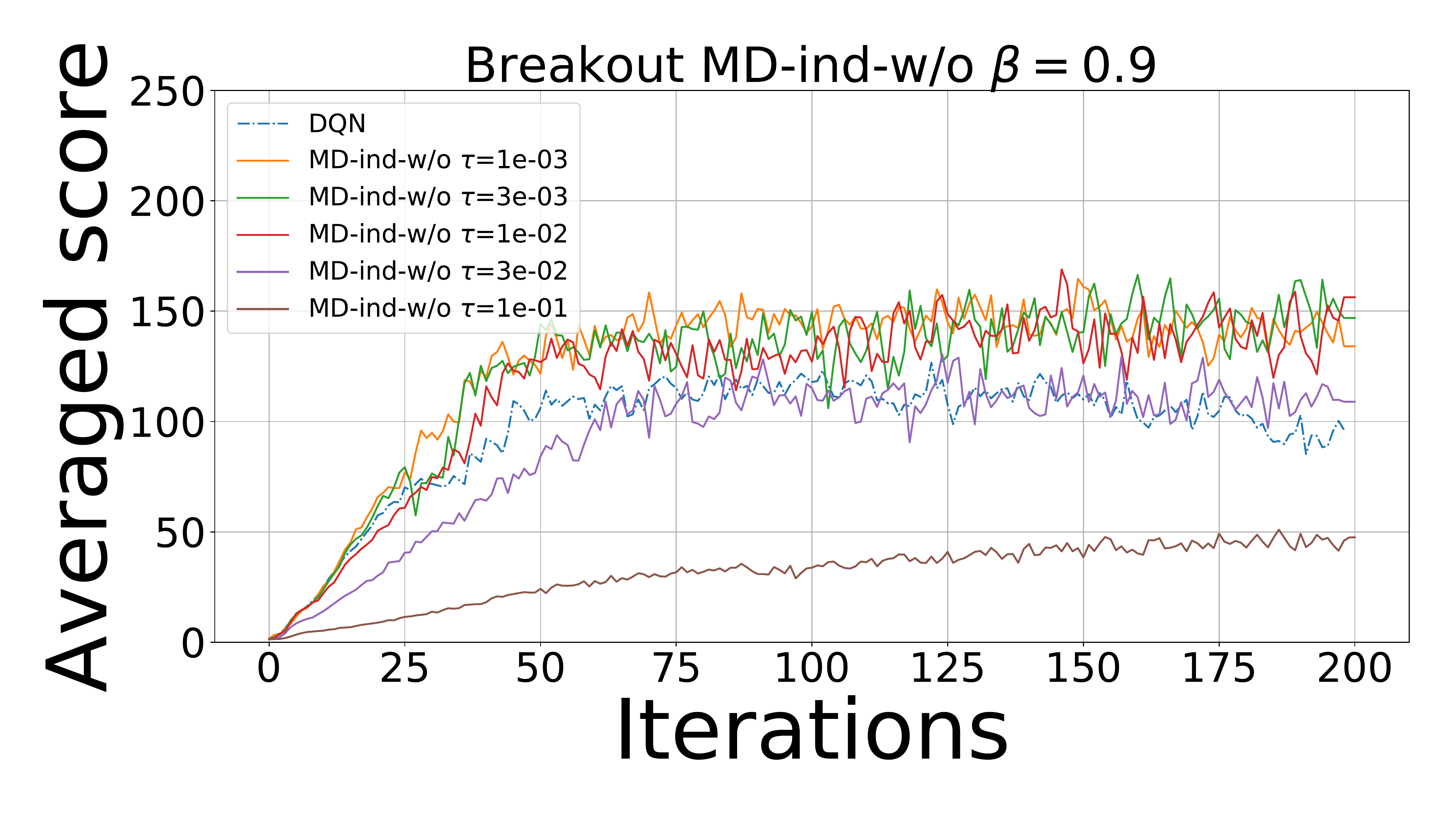}\\
\midrule
     \includegraphics[width=0.3\linewidth]{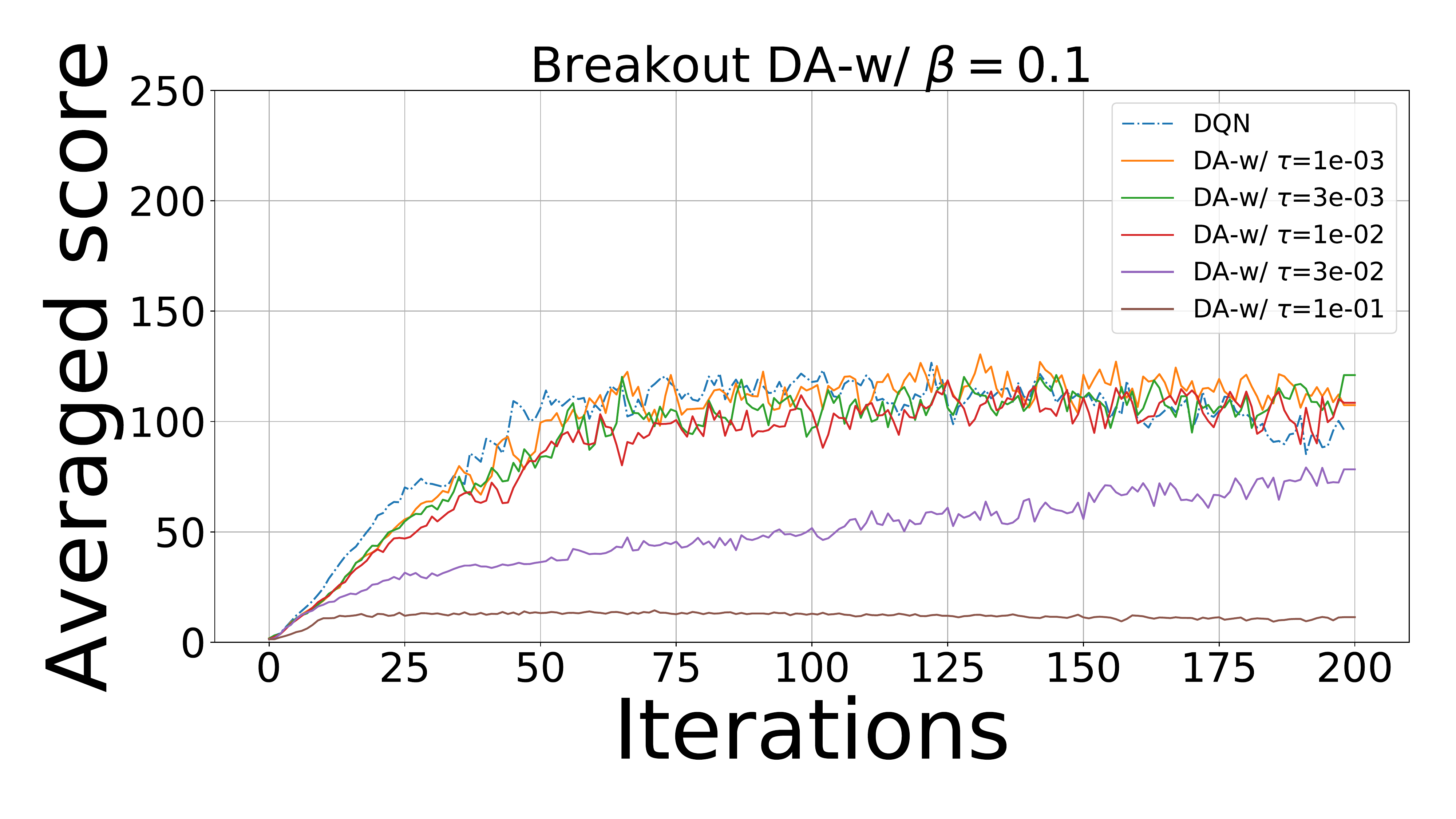} &
     \includegraphics[width=0.3\linewidth]{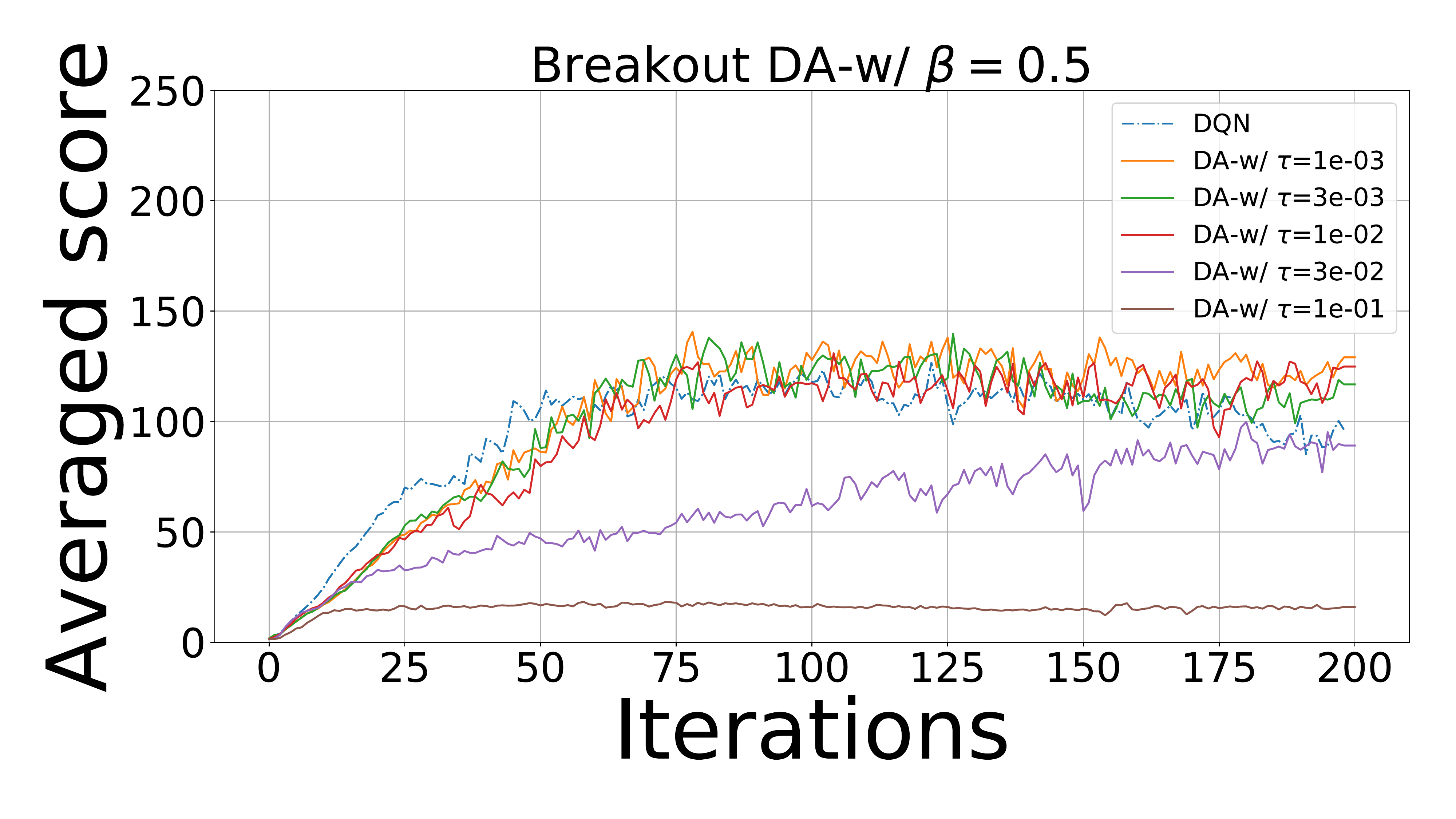} &
     \includegraphics[width=0.3\linewidth]{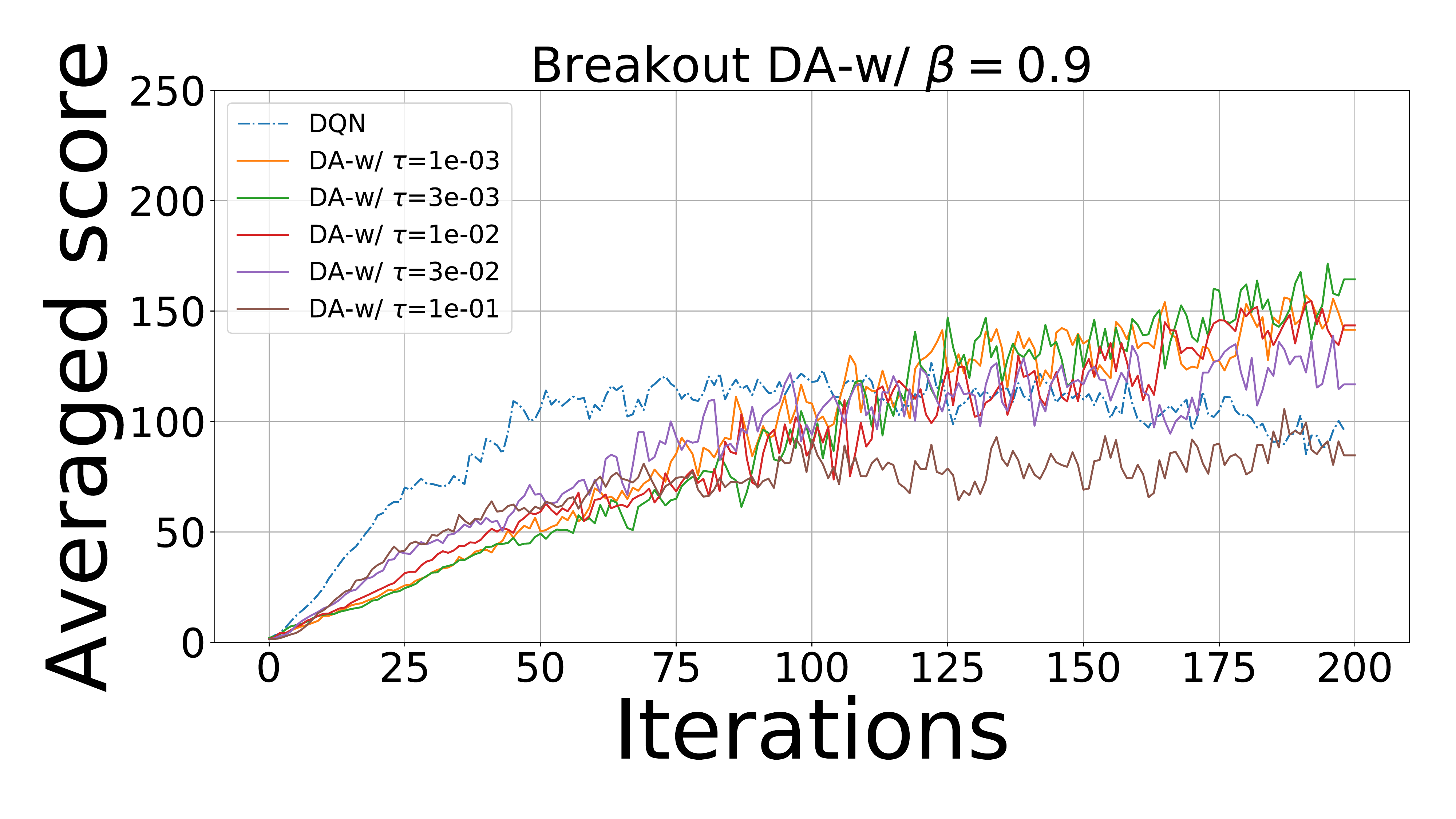} \\
     \includegraphics[width=0.3\linewidth]{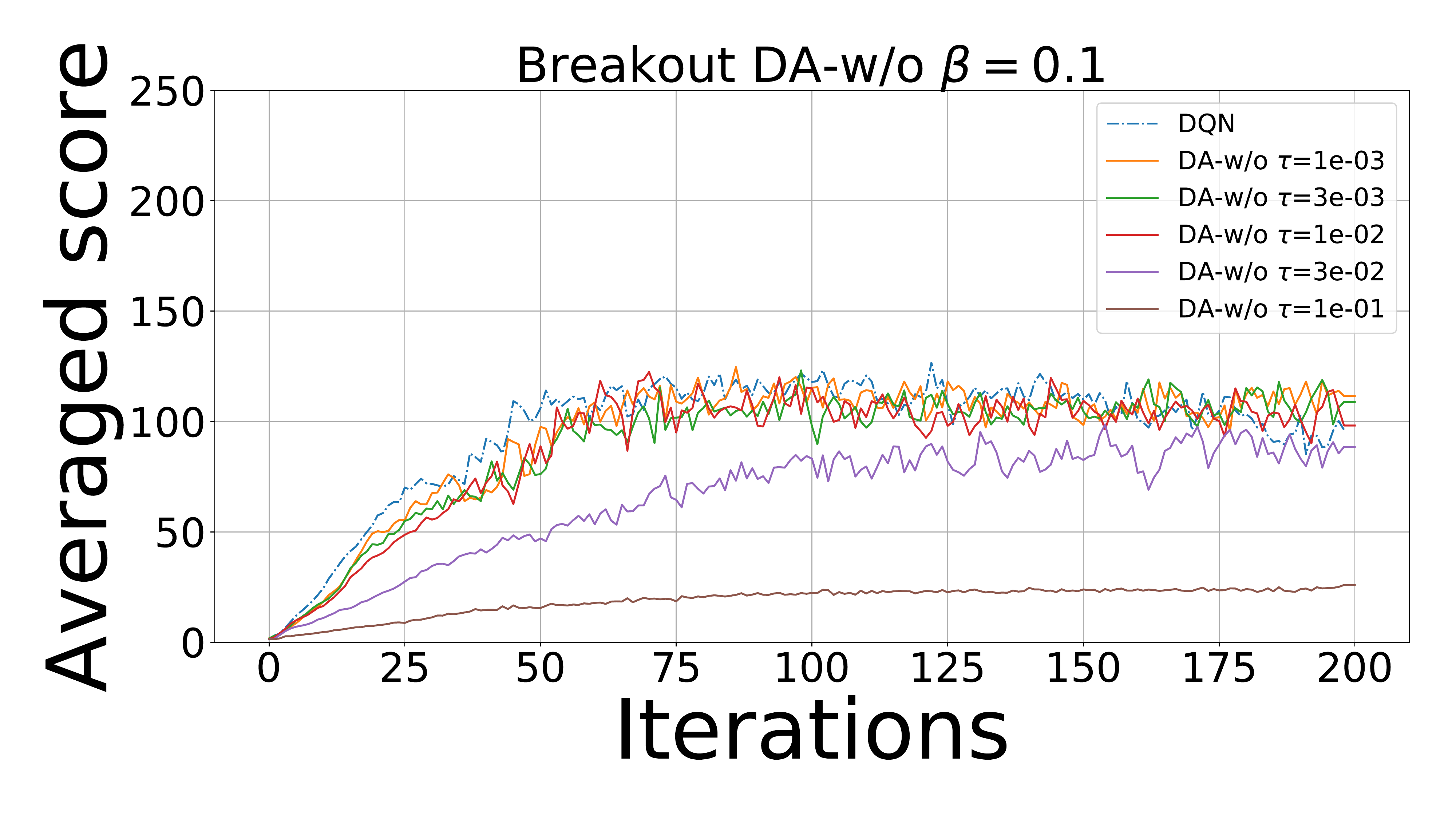} &
     \includegraphics[width=0.3\linewidth]{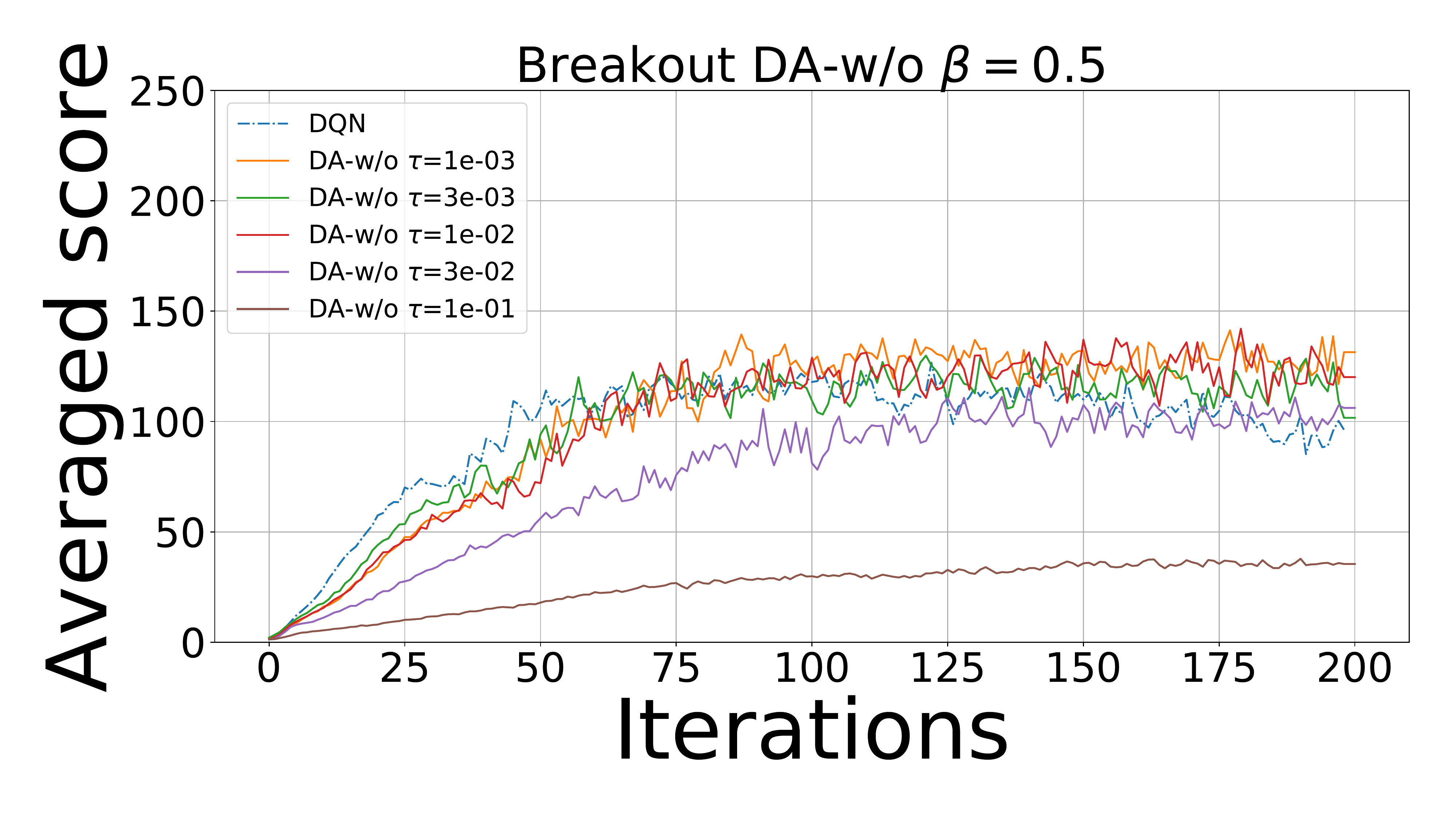} &
     \includegraphics[width=0.3\linewidth]{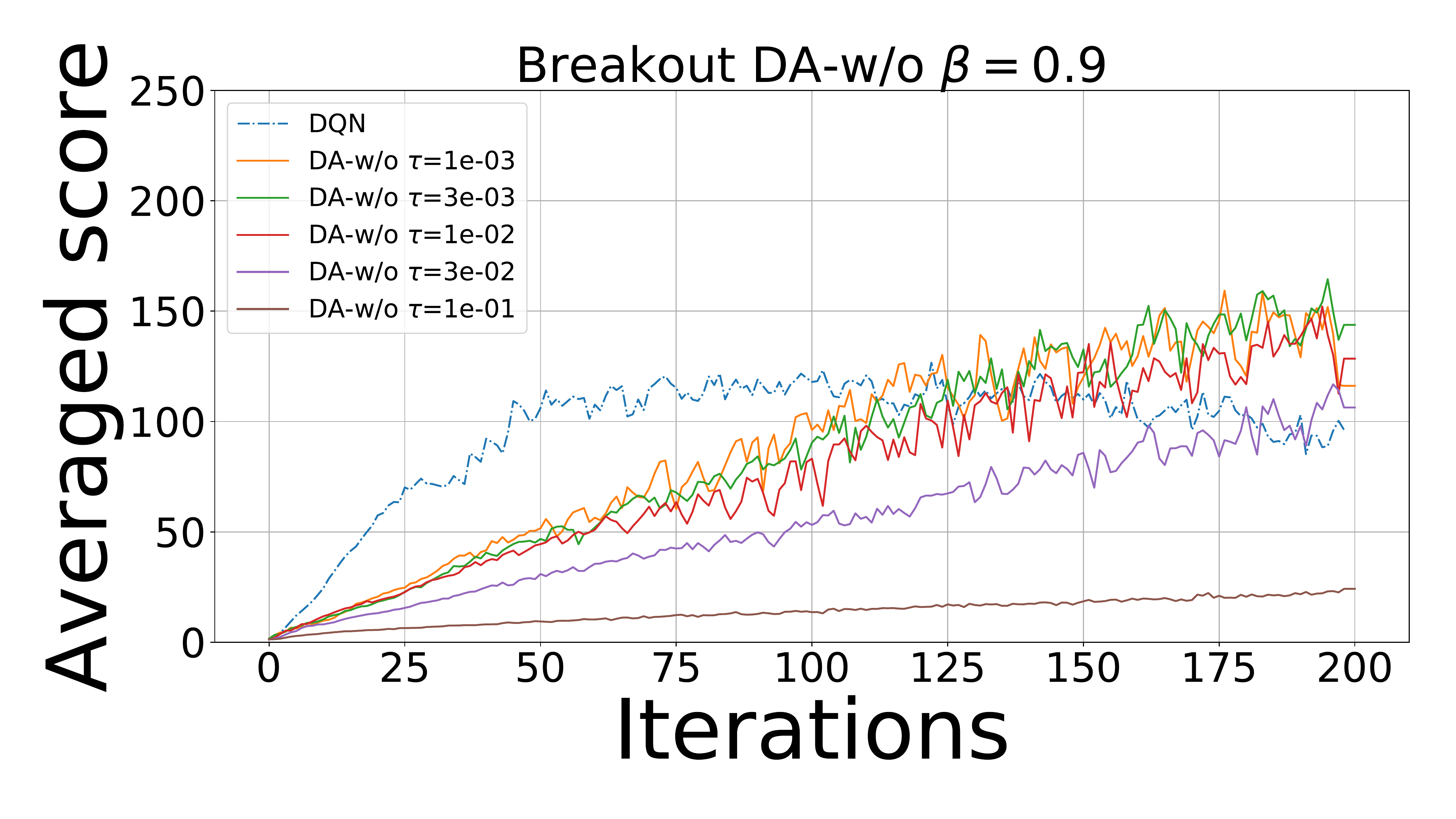}\\
\end{tabular}
\caption{All averaged training scores of MD-dir (top), MD-ind (middle) and DA (bottom), \wir{} and \wor{}, on Breakout, for several values of $\beta$ and $\tau$. Each plot corresponds to one value of $\beta$ (in the titles). In each plot, a curve corresponds to a value of $\tau$: $1e-3$ (orange), $3e-3$ (green), $1e-02$ (red), $3e-2$ (blue), $1e-1$ (brown). The blue dotted line is DQN.\label{fig:breakout_curves}}
\end{center}
\end{figure}

\begin{figure}
\begin{center}
\begin{tabular}{c c c}
     \includegraphics[width=0.3\linewidth]{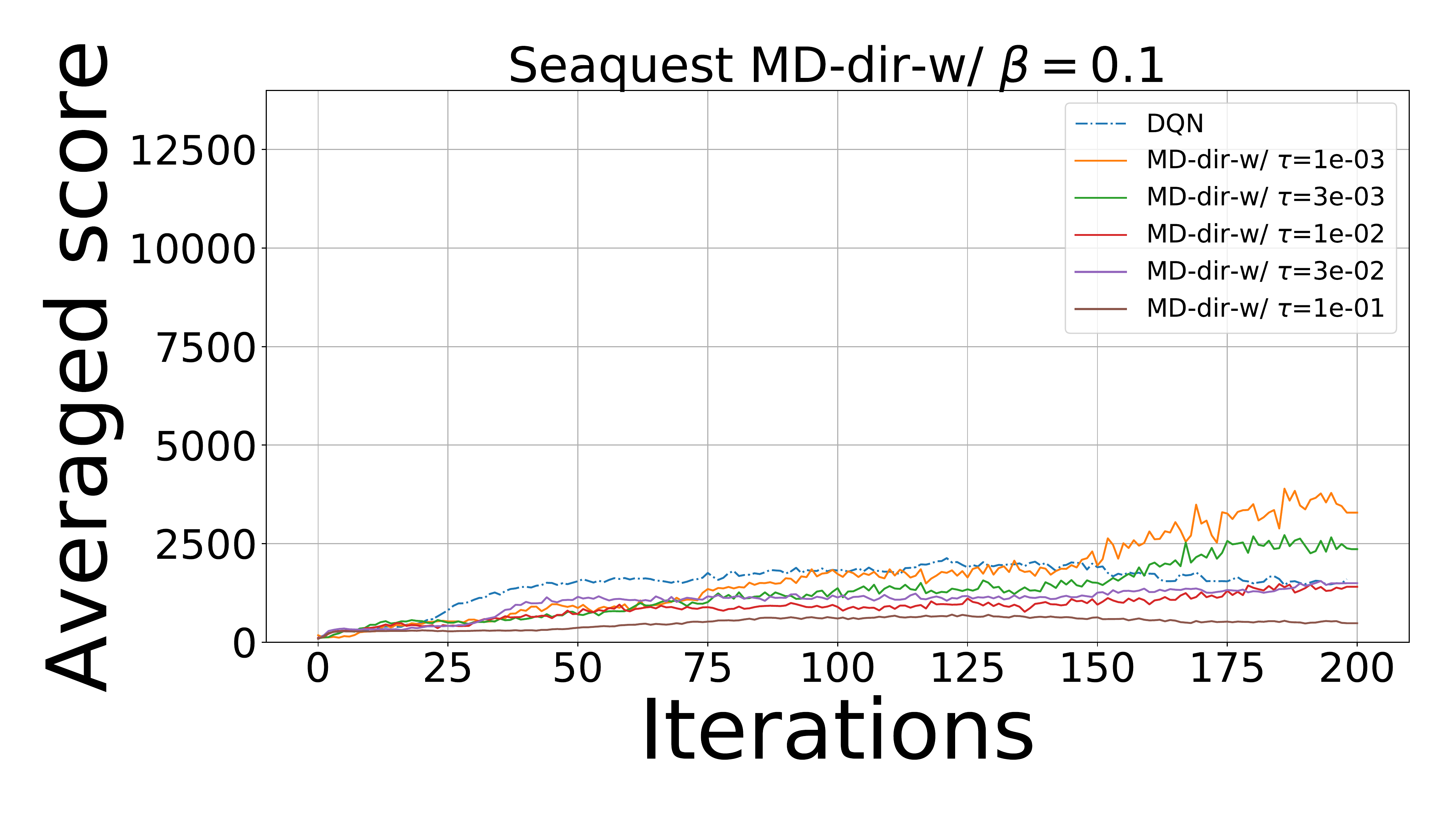} &
     \includegraphics[width=0.3\linewidth]{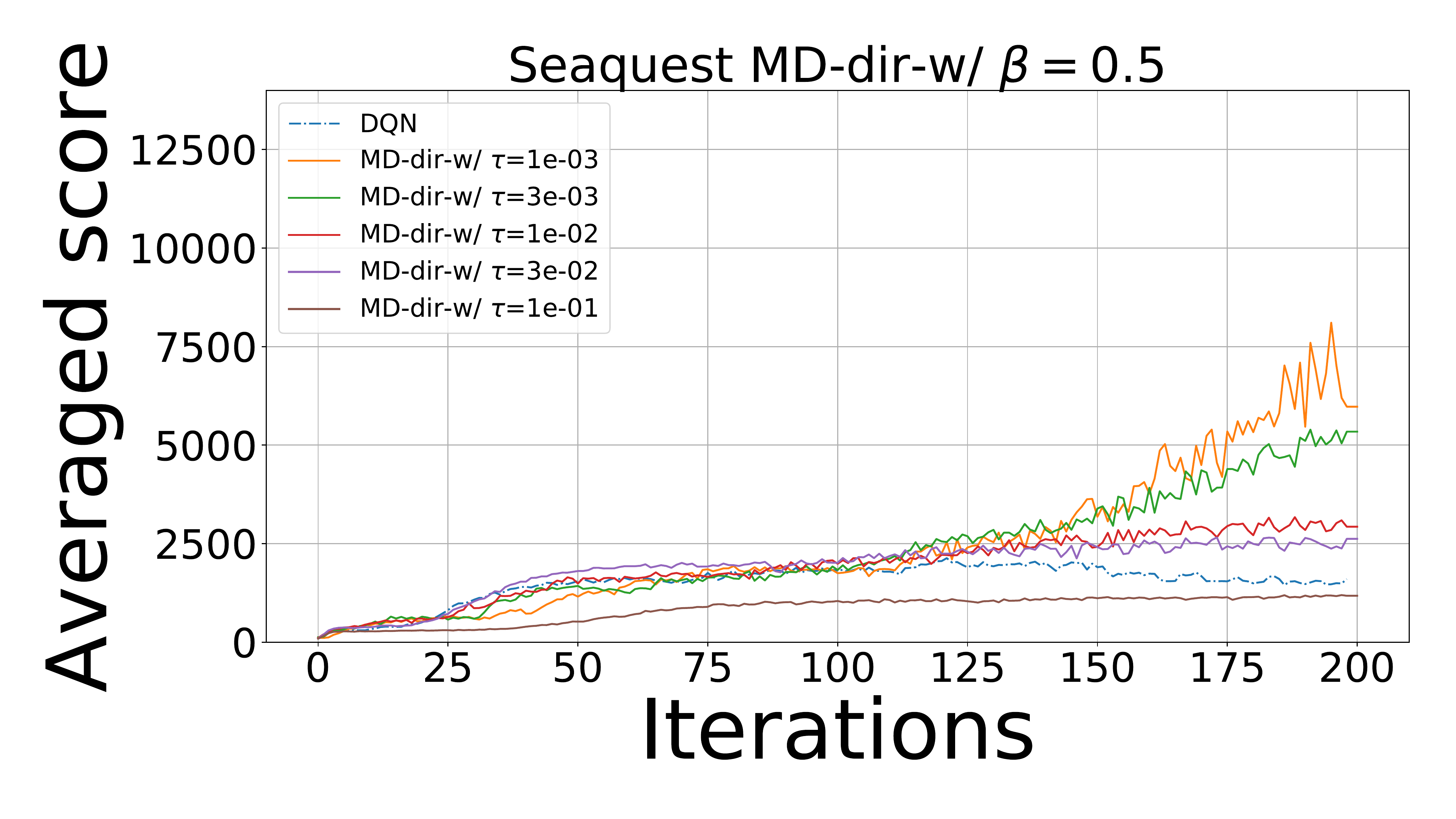} &
     \includegraphics[width=0.3\linewidth]{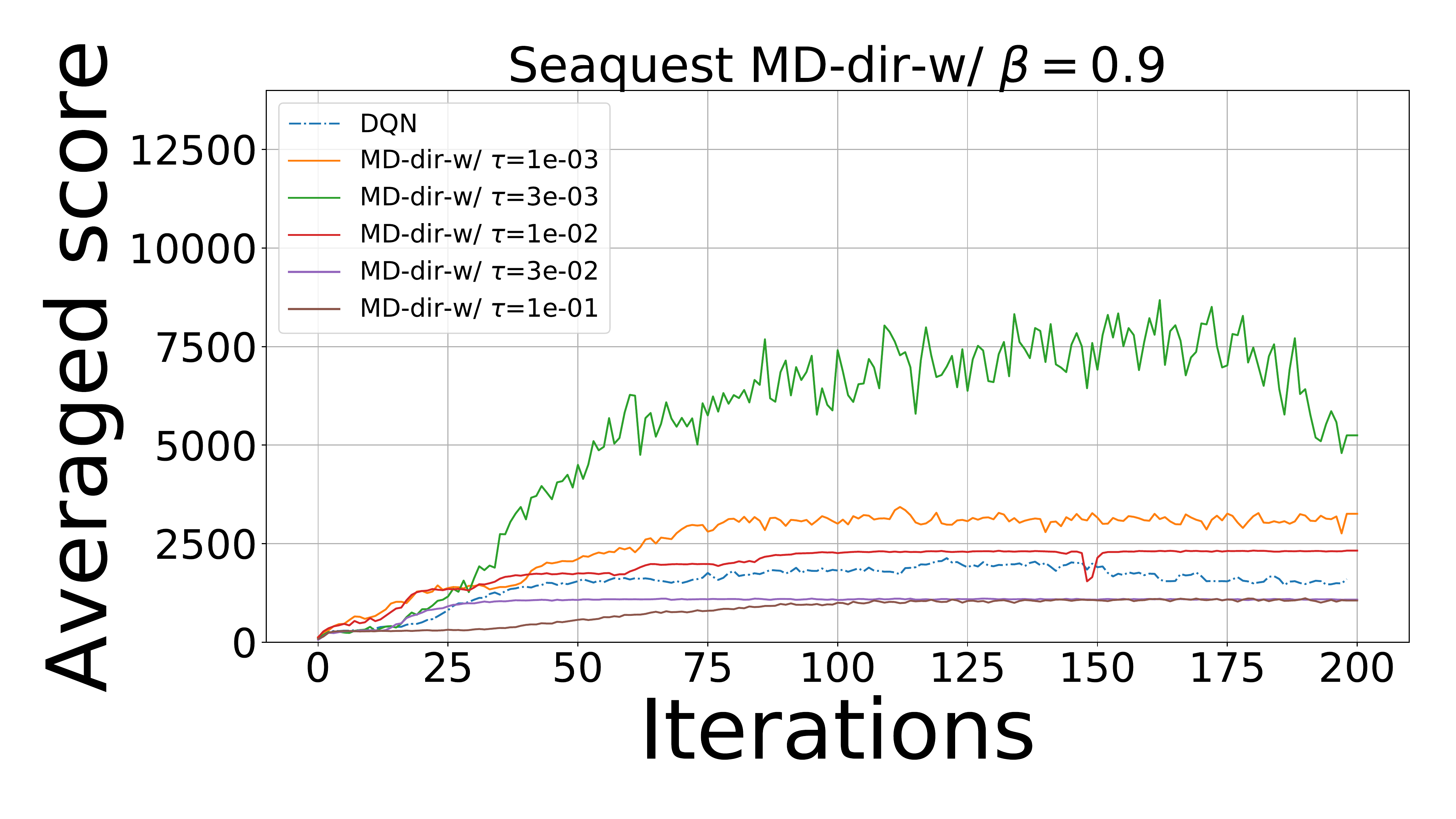} \\
     \includegraphics[width=0.3\linewidth]{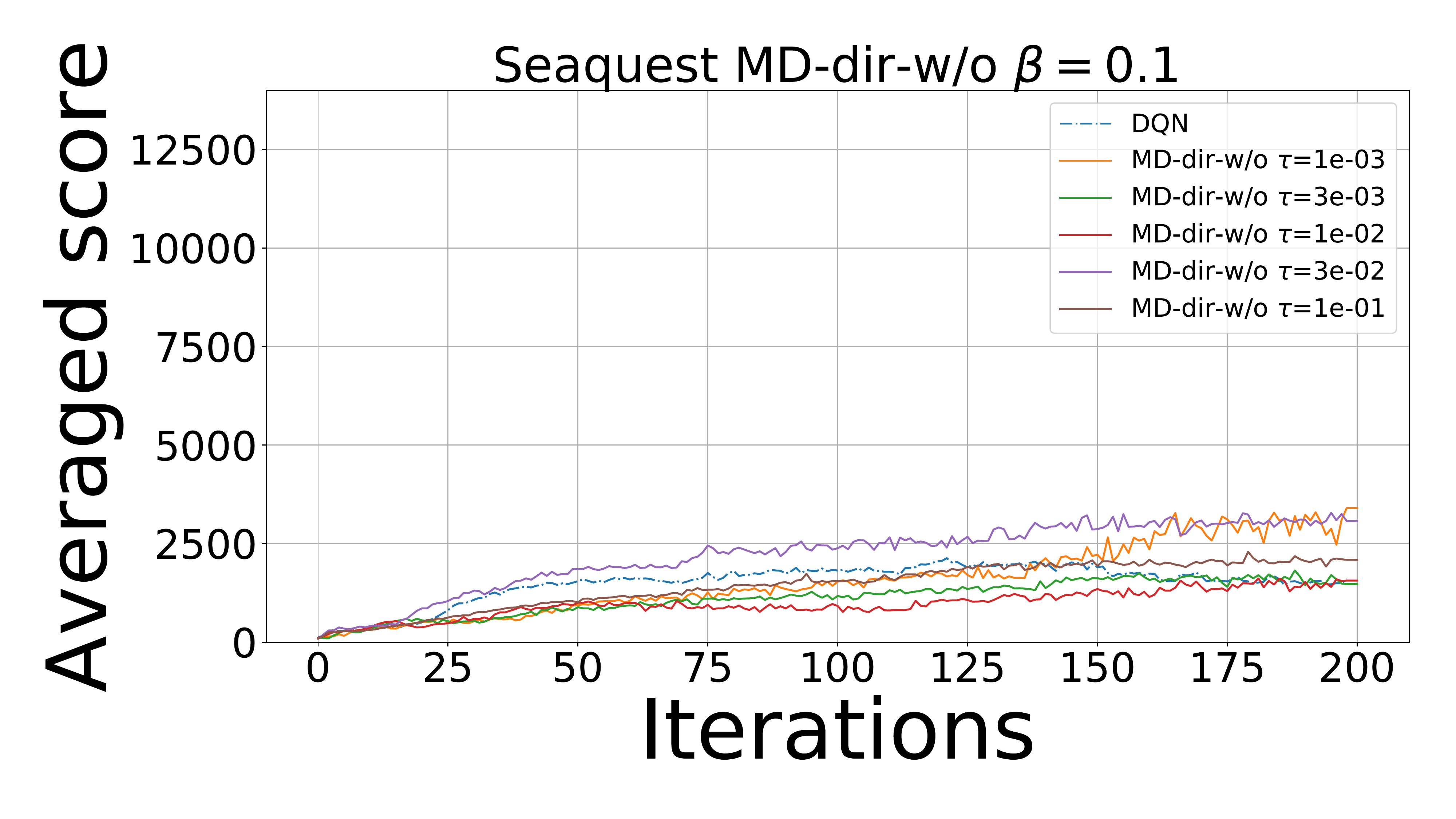} &
     \includegraphics[width=0.3\linewidth]{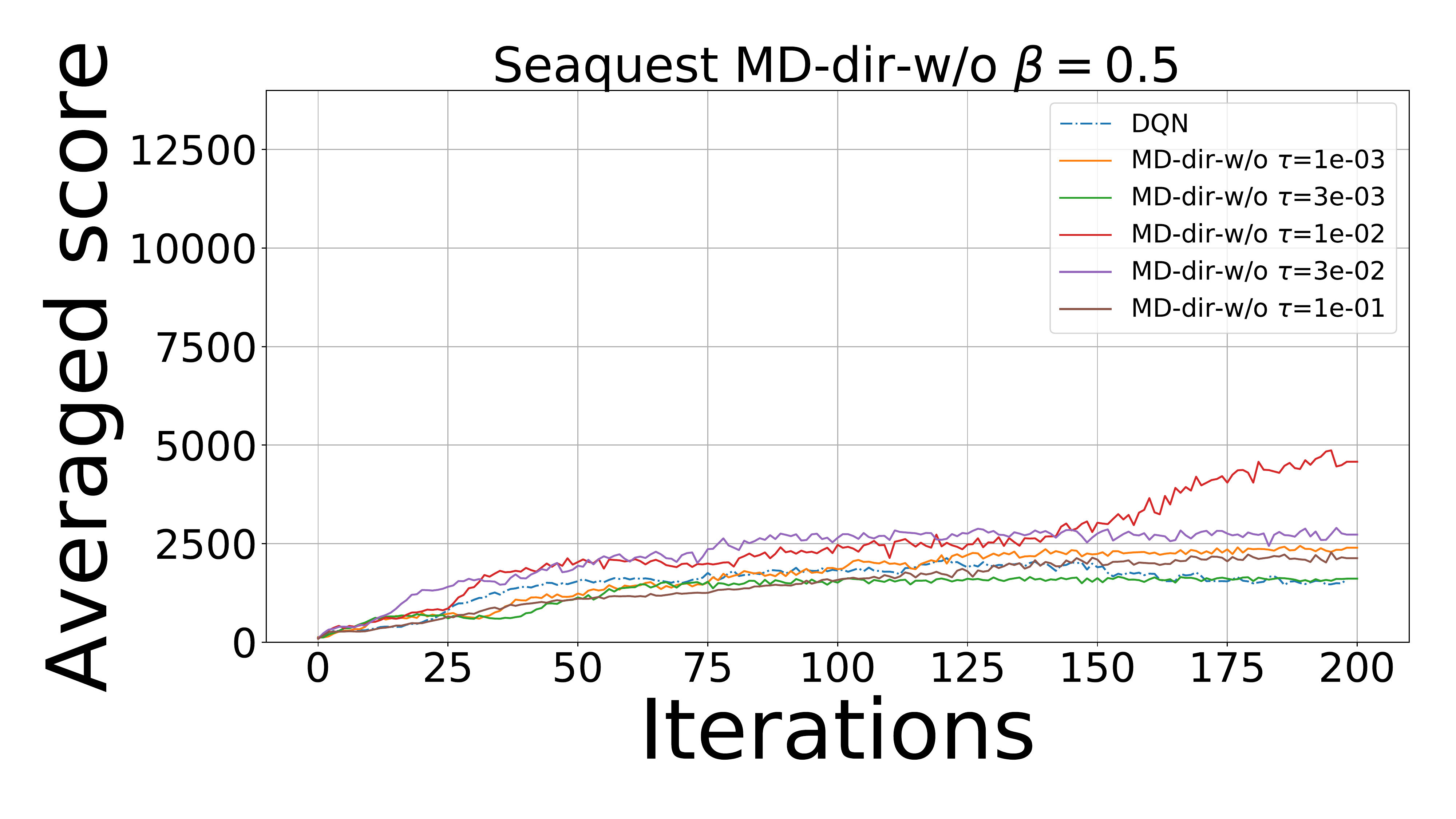} &
     \includegraphics[width=0.3\linewidth]{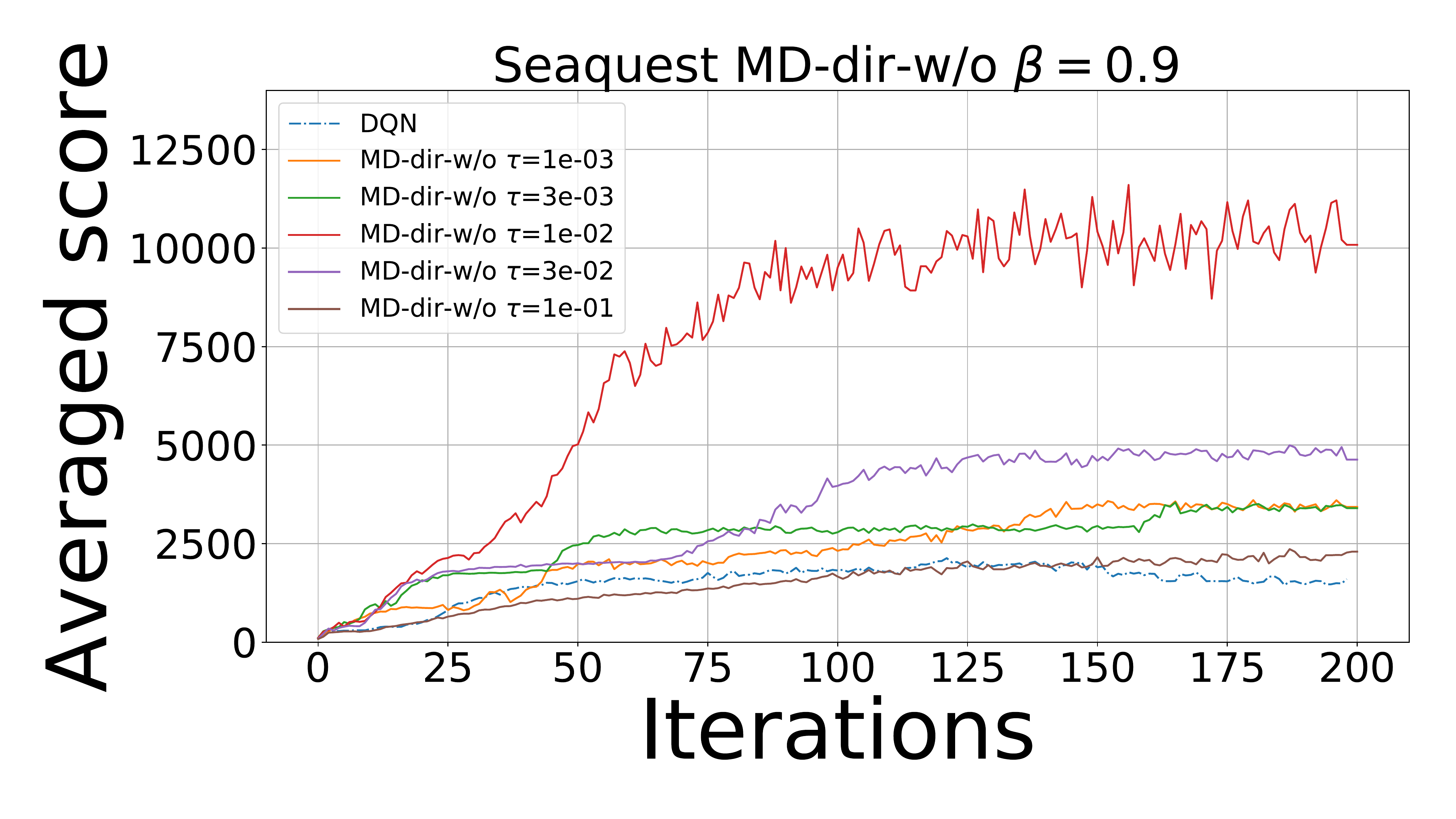}\\
\midrule
     \includegraphics[width=0.3\linewidth]{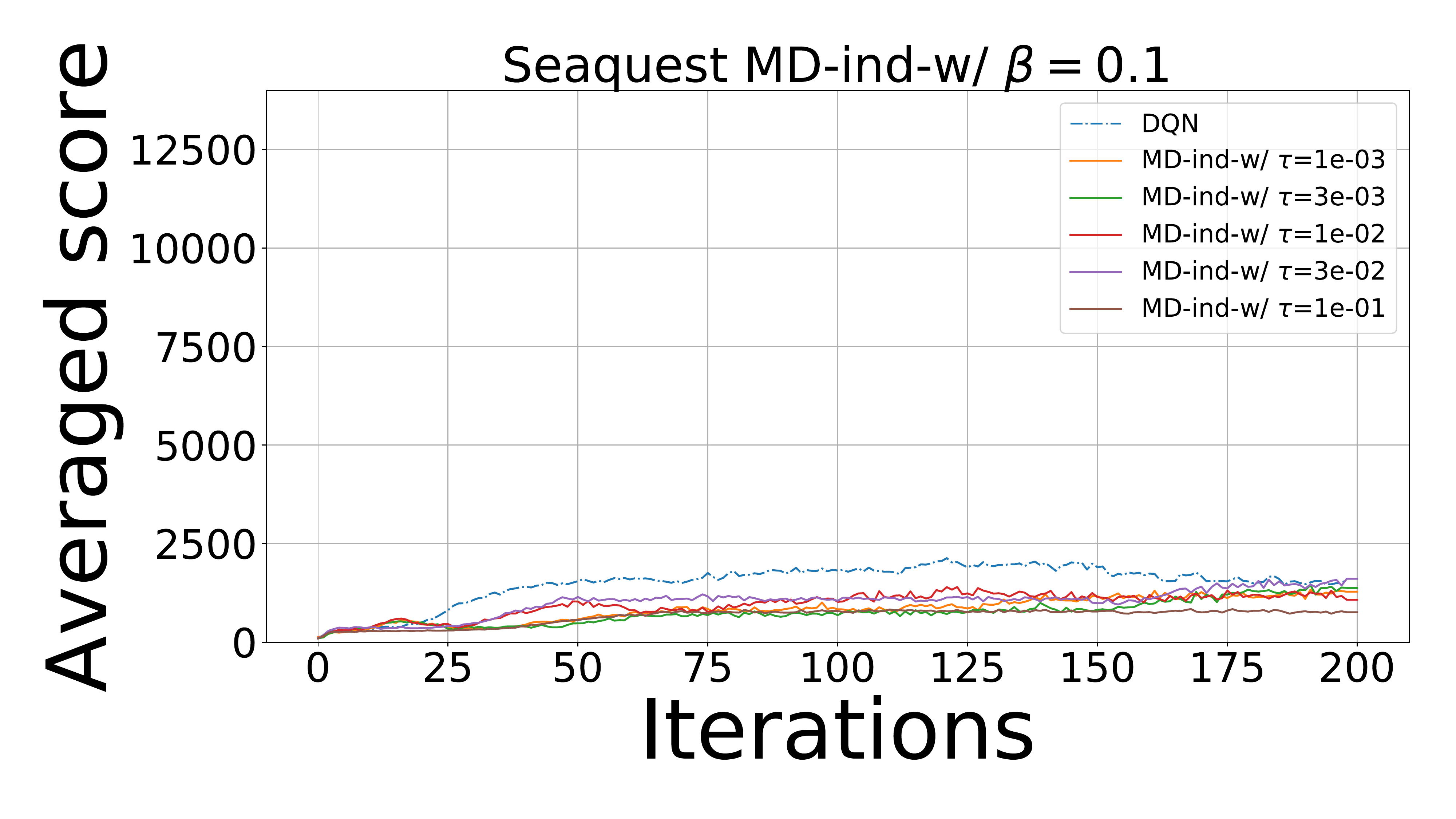} &
     \includegraphics[width=0.3\linewidth]{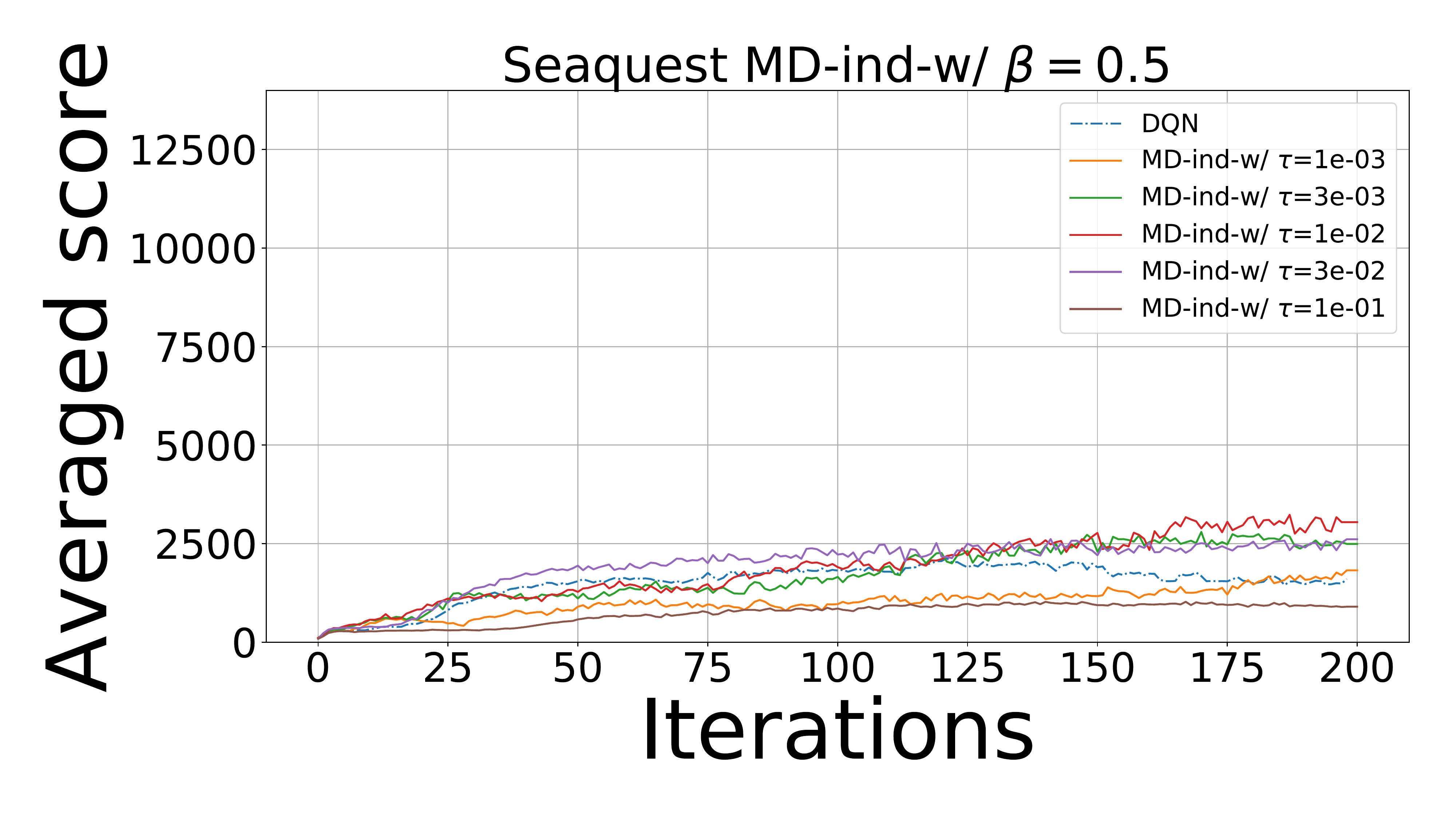} &
     \includegraphics[width=0.3\linewidth]{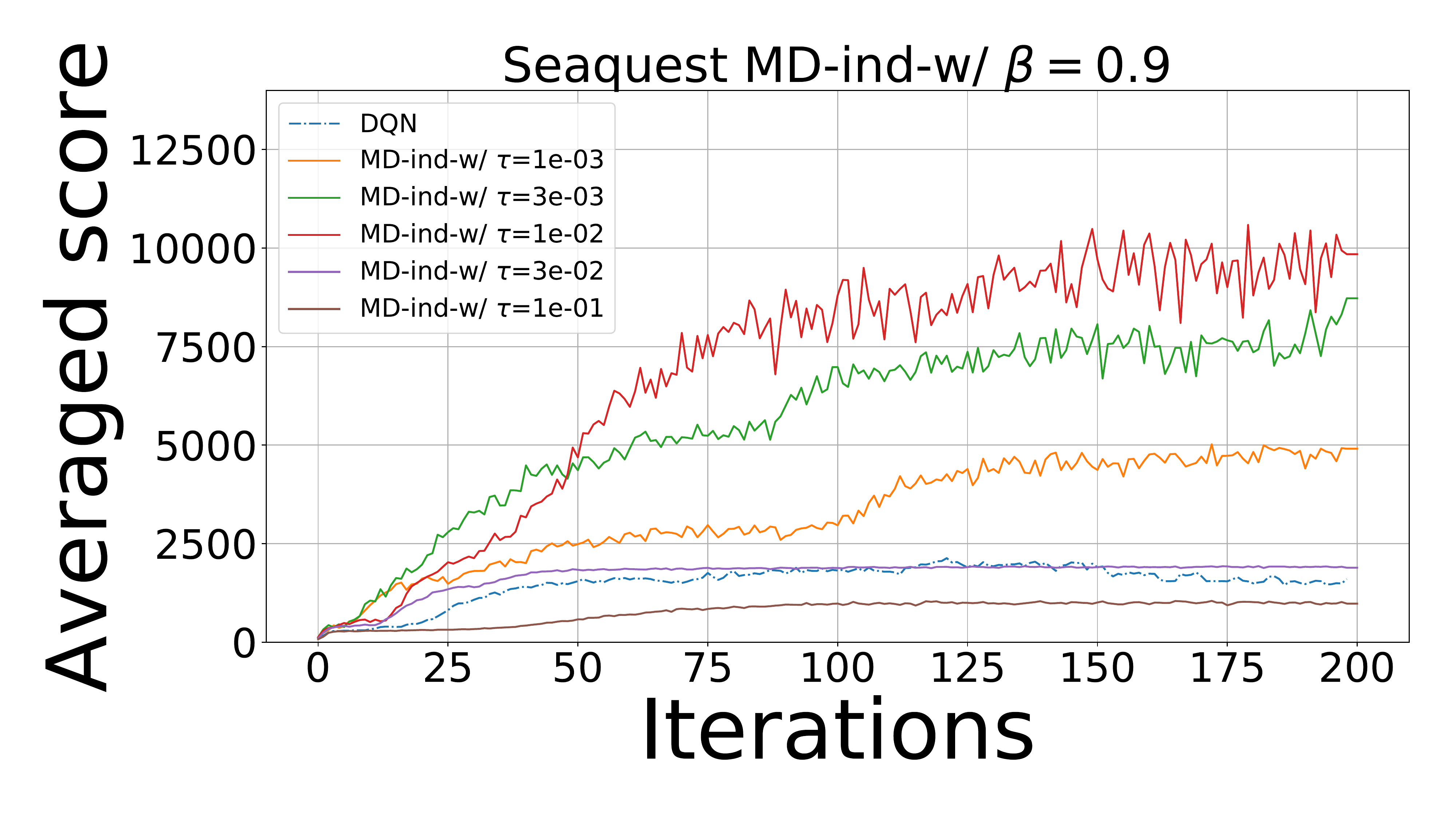} \\
     \includegraphics[width=0.3\linewidth]{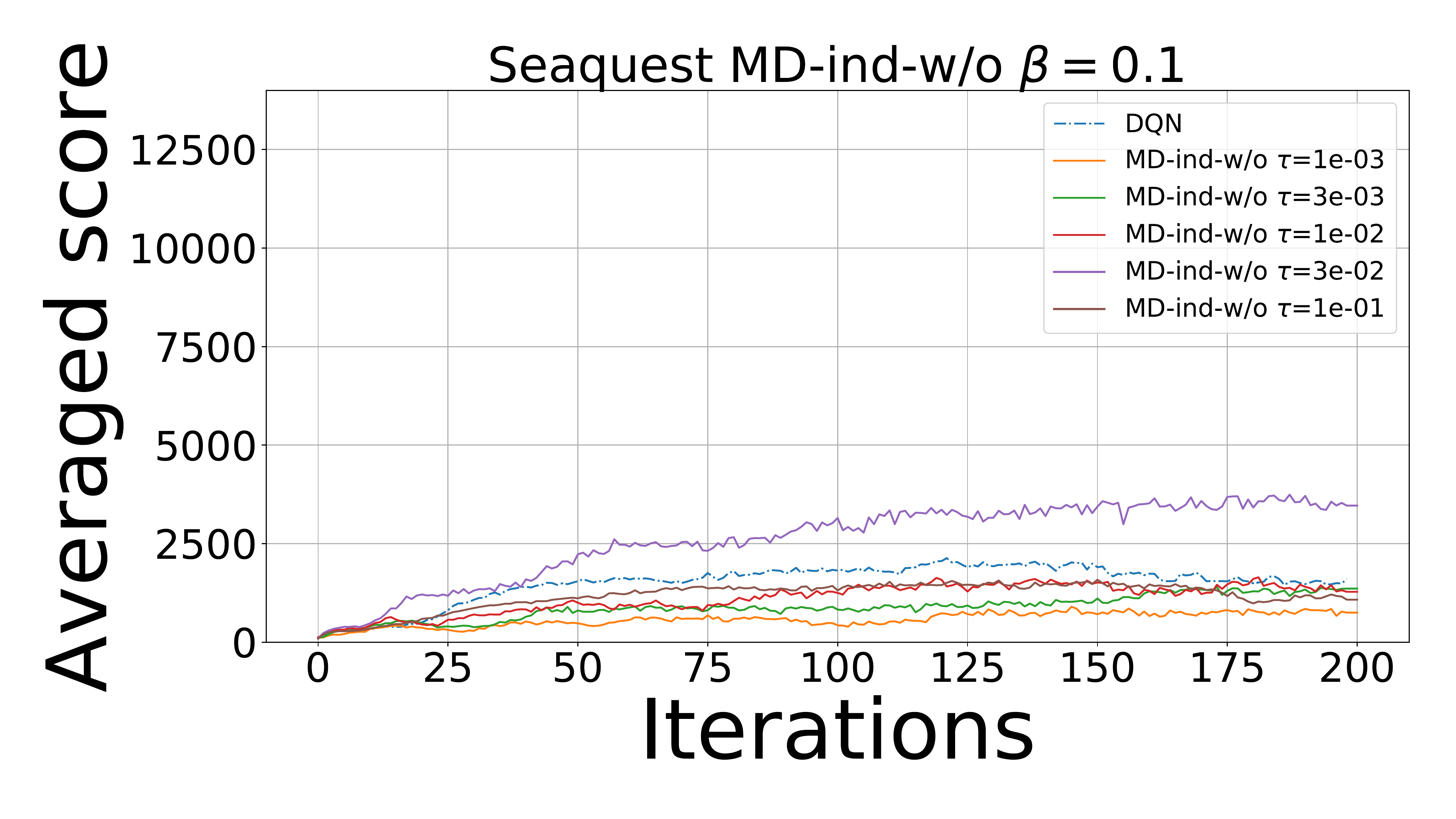} &
     \includegraphics[width=0.3\linewidth]{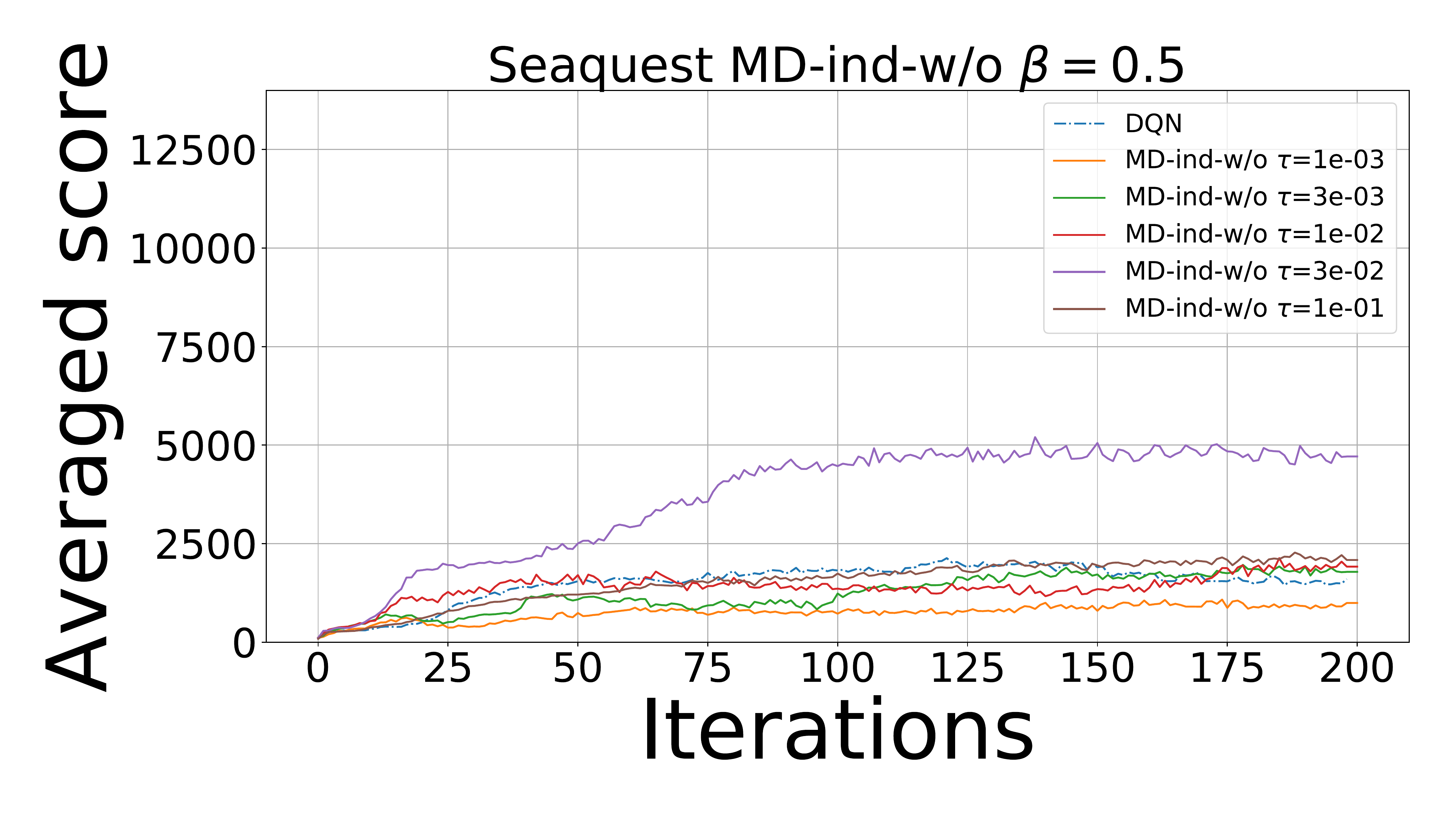} &
     \includegraphics[width=0.3\linewidth]{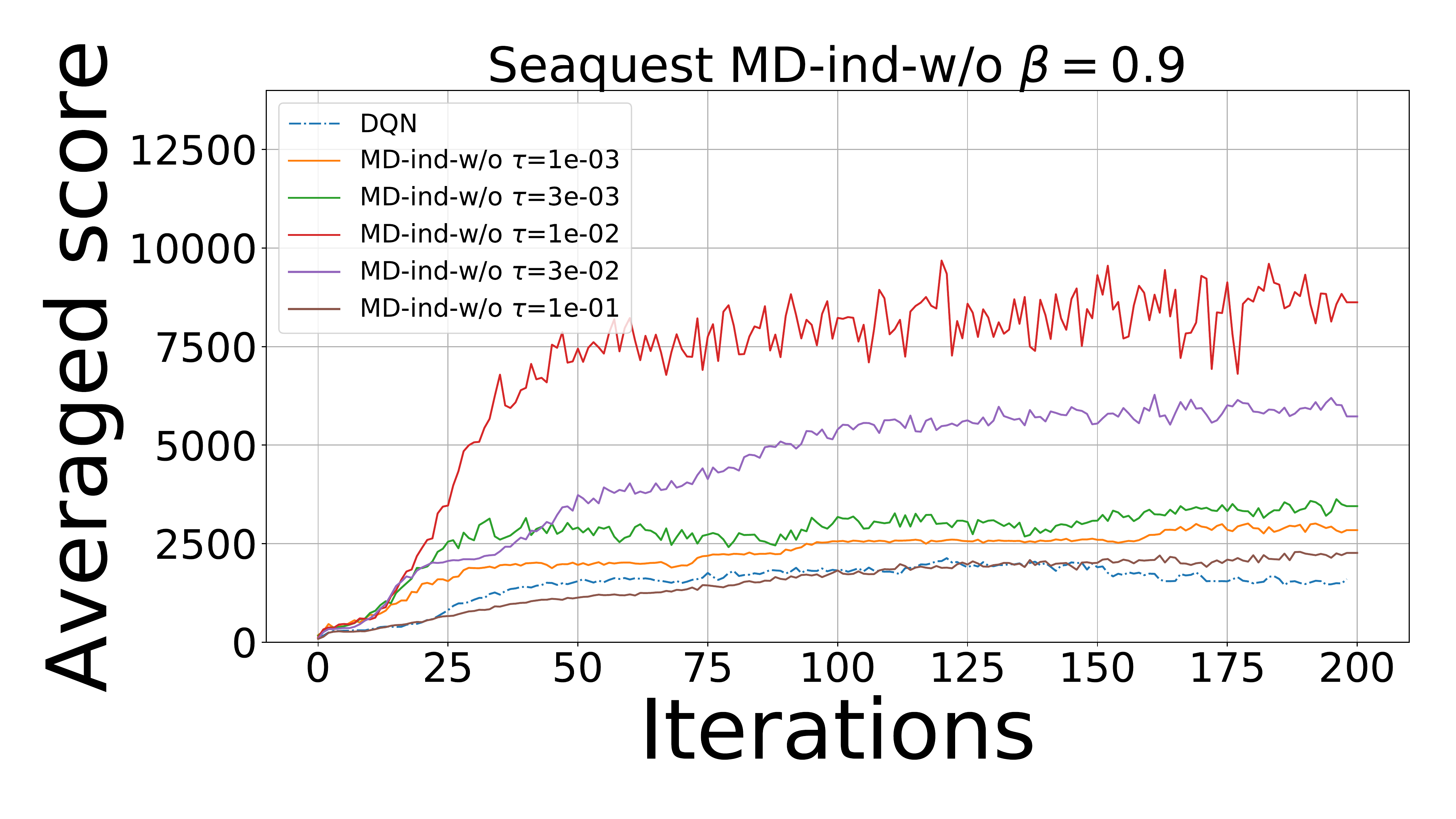}\\
\midrule
     \includegraphics[width=0.3\linewidth]{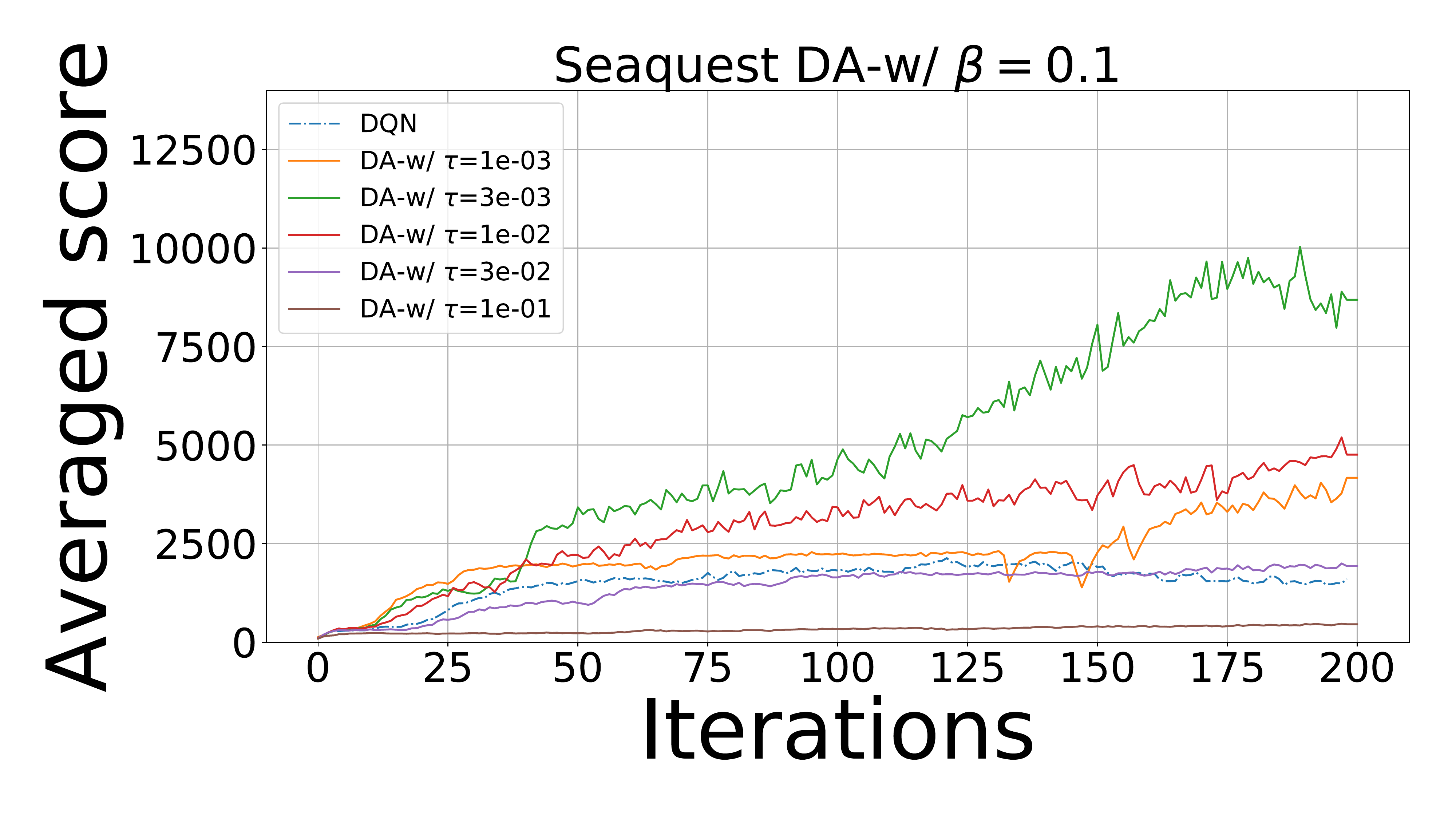} &
     \includegraphics[width=0.3\linewidth]{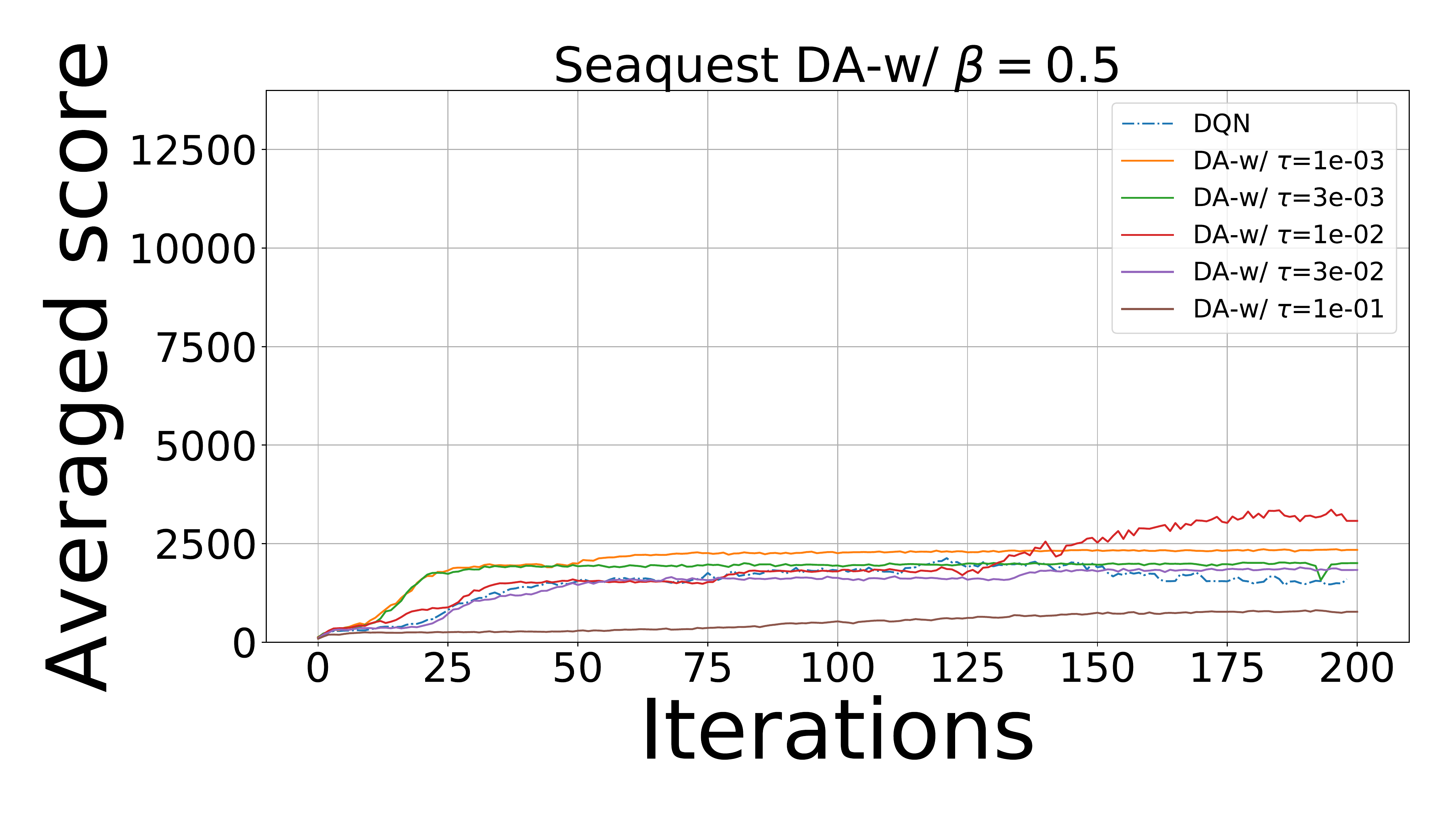} &
     \includegraphics[width=0.3\linewidth]{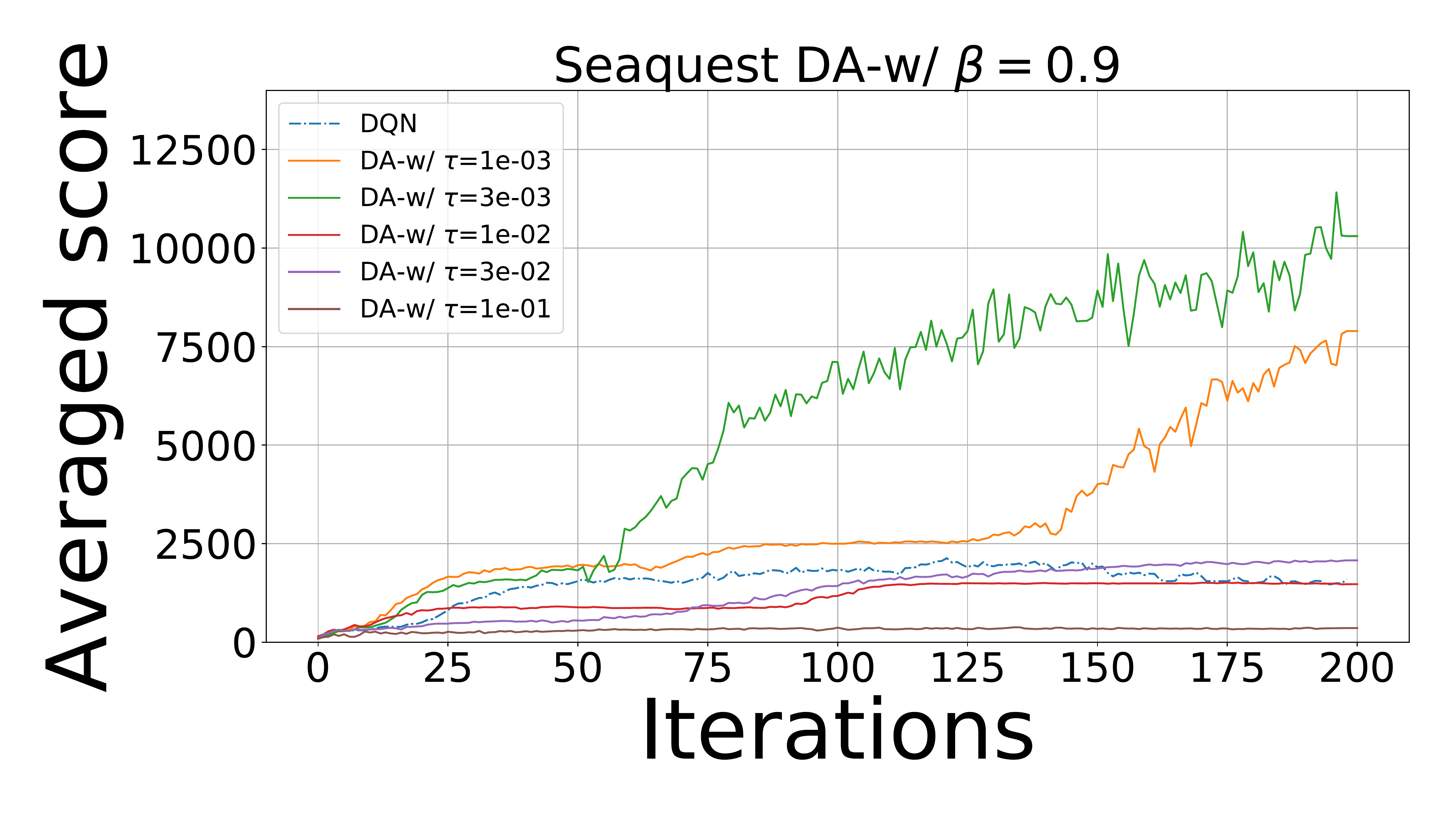} \\
     \includegraphics[width=0.3\linewidth]{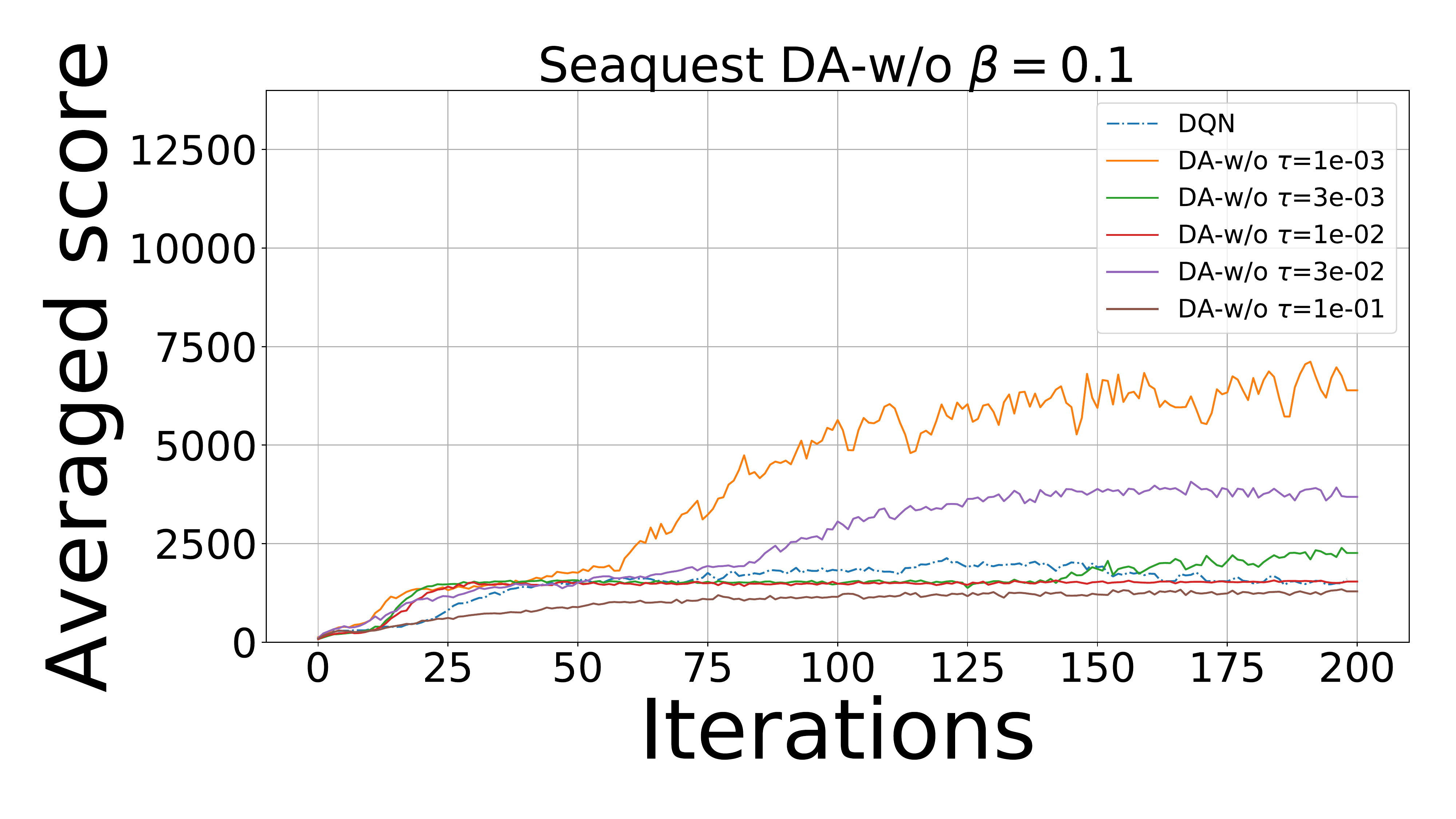} &
     \includegraphics[width=0.3\linewidth]{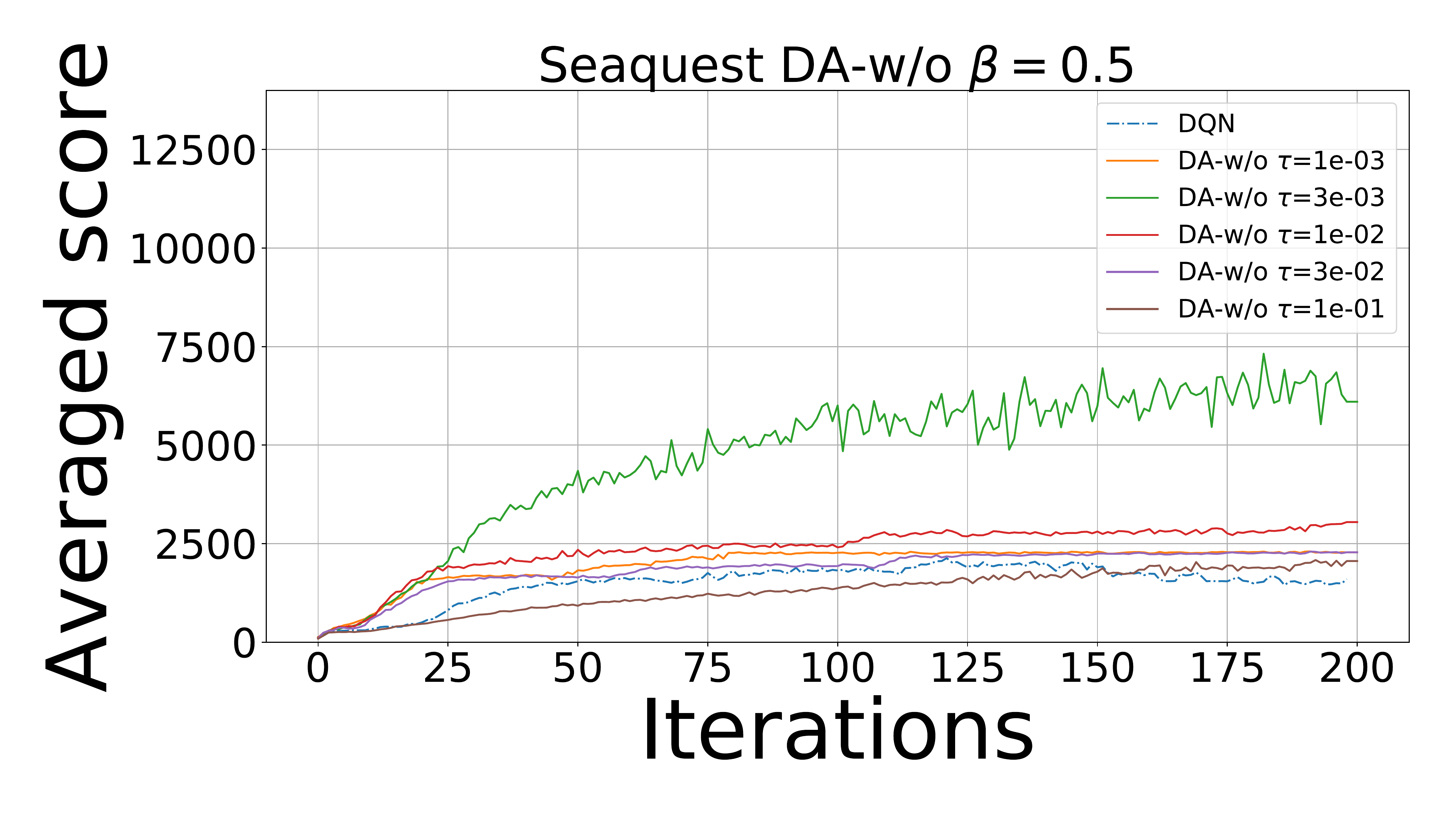} &
     \includegraphics[width=0.3\linewidth]{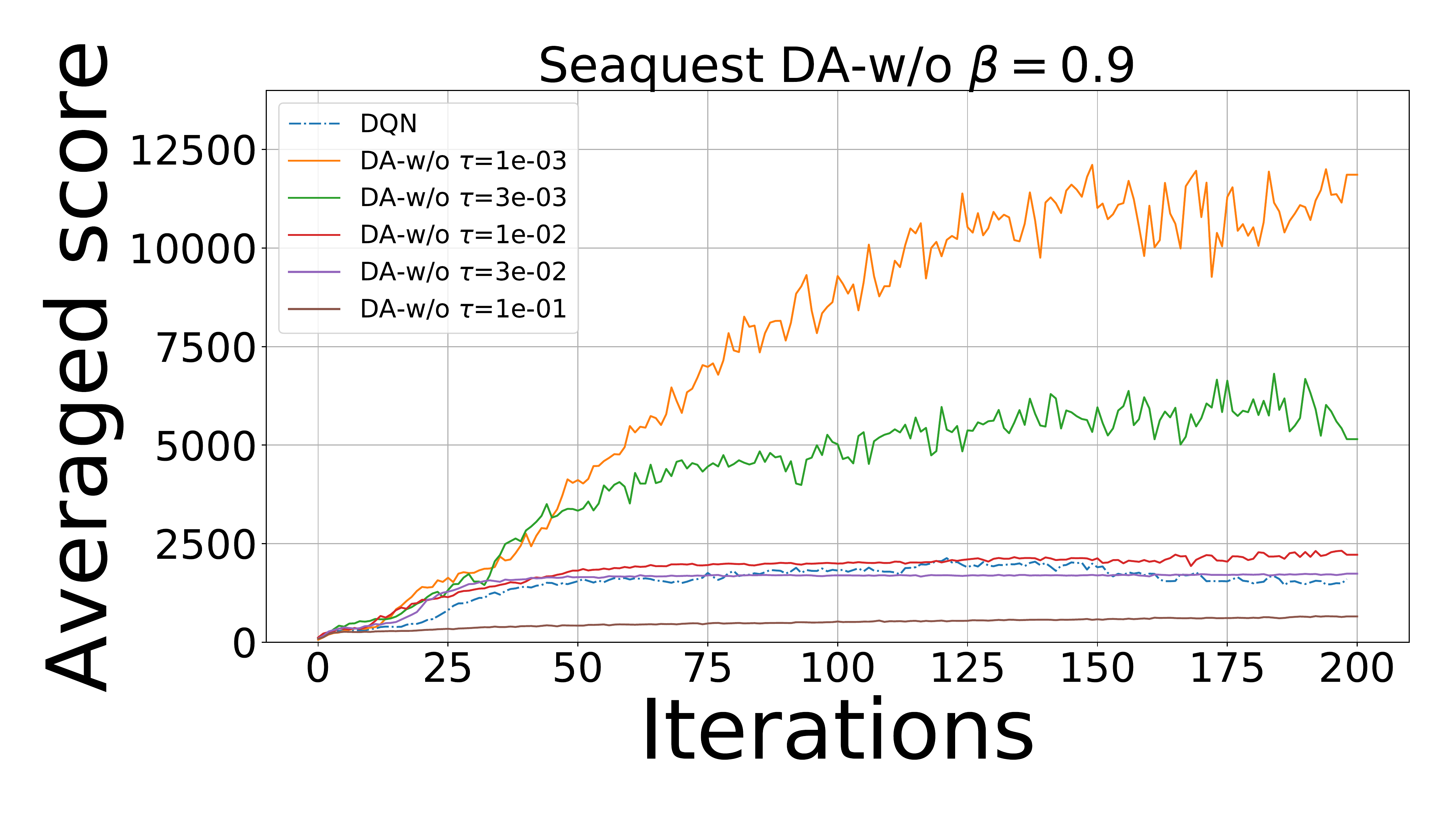}\\
\end{tabular}
\caption{All averaged training scores of MD-dir (top), MD-ind (middle) and DA (bottom), \wir{} and \wor{}, on Seaquest, for several values of $\beta$ and $\tau$. Each plot corresponds to one value of $\beta$ (in the titles). In each plot, a curve corresponds to a value of $\tau$: $1e-3$ (orange), $3e-3$ (green), $1e-02$ (red), $3e-2$ (blue), $1e-1$ (brown). The blue dotted line is DQN.\label{fig:seaquest_curves}}
\end{center}
\end{figure}

\begin{figure}
\begin{center}
\begin{tabular}{c c c}
     \includegraphics[width=0.3\linewidth]{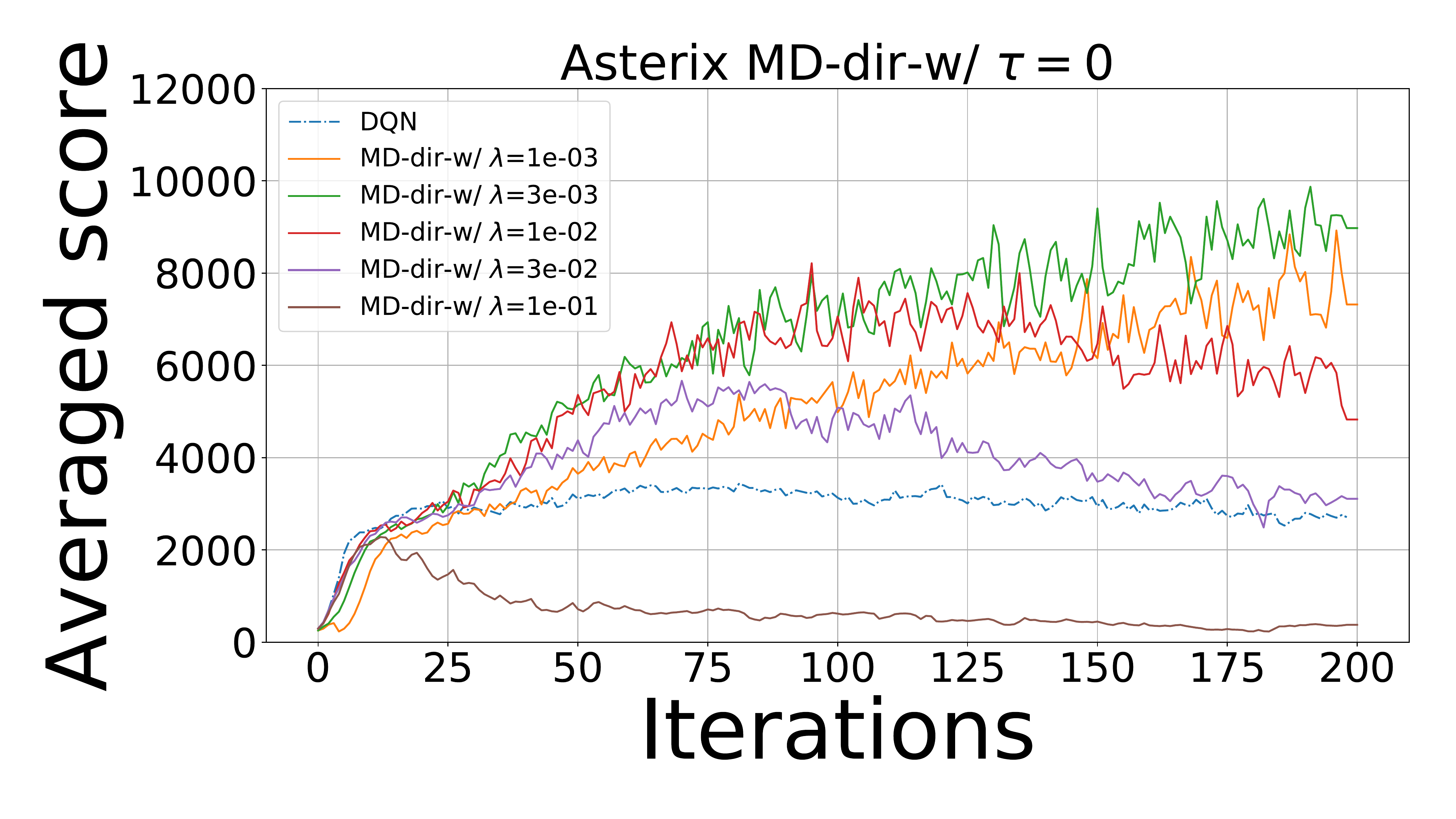}& 
     \includegraphics[width=0.3\linewidth]{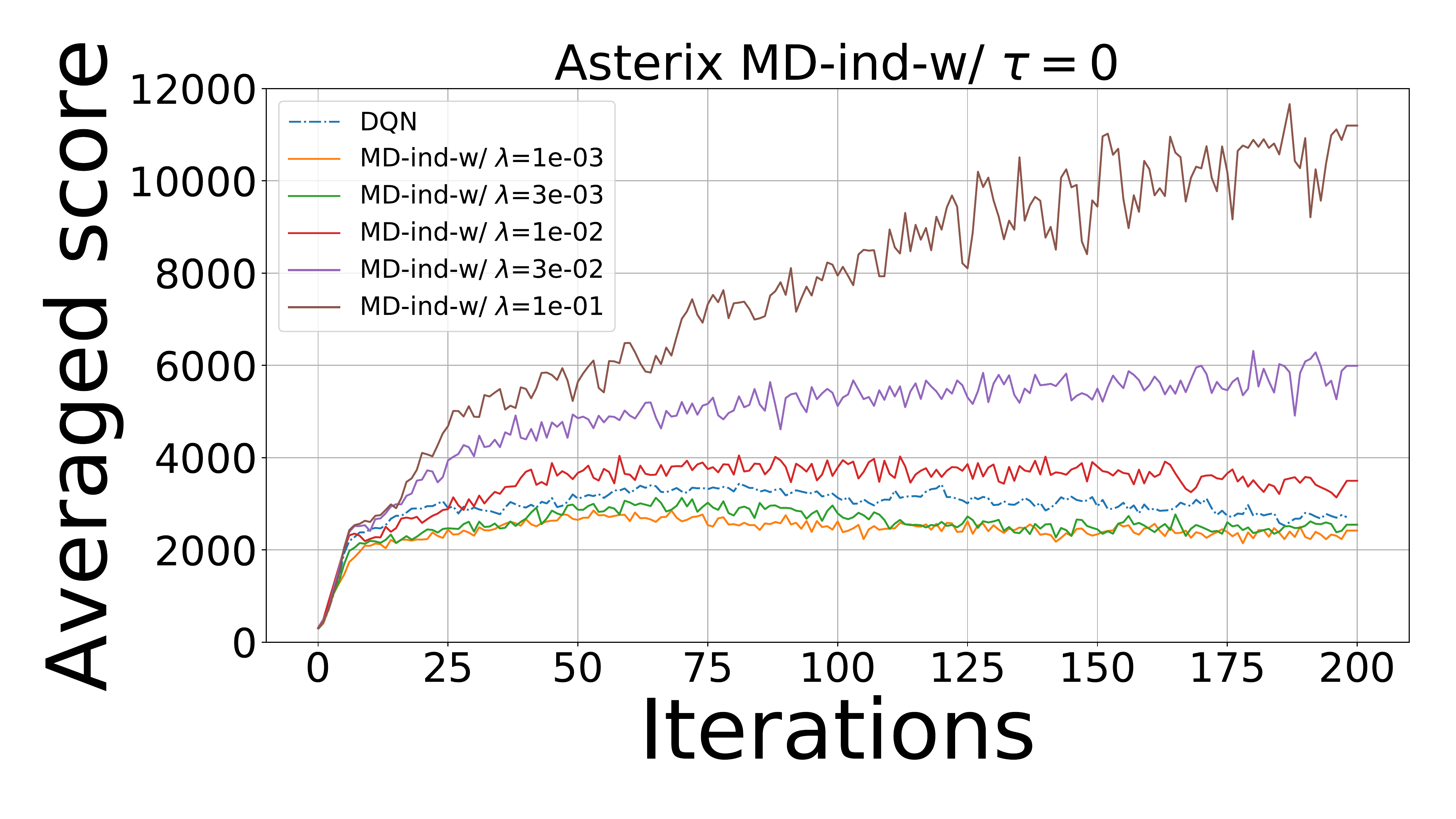}& 
     \multirow[b]{2}{*}[1cm]{\includegraphics[width=0.3\linewidth]{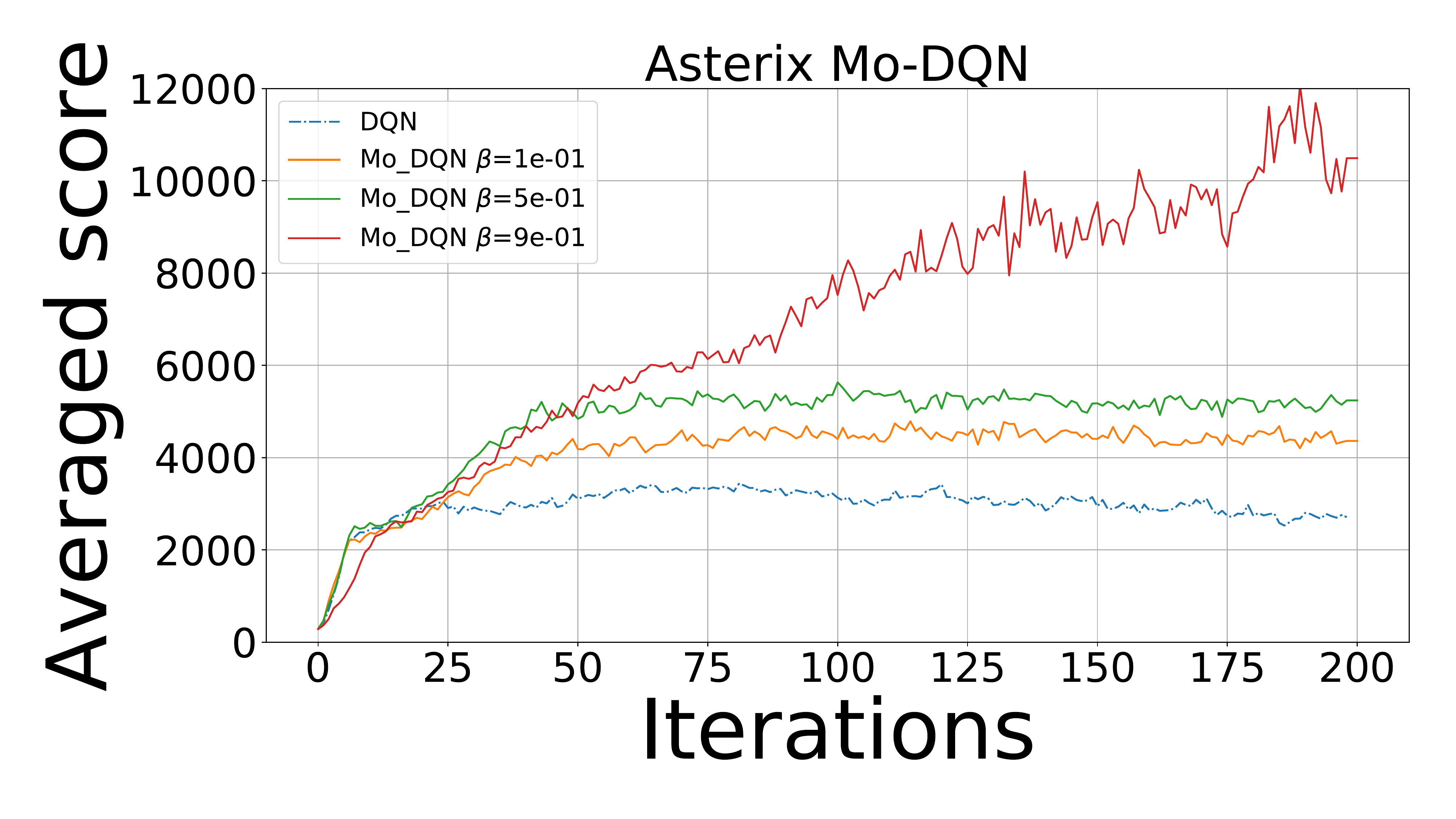}} \\
     \includegraphics[width=0.3\linewidth]{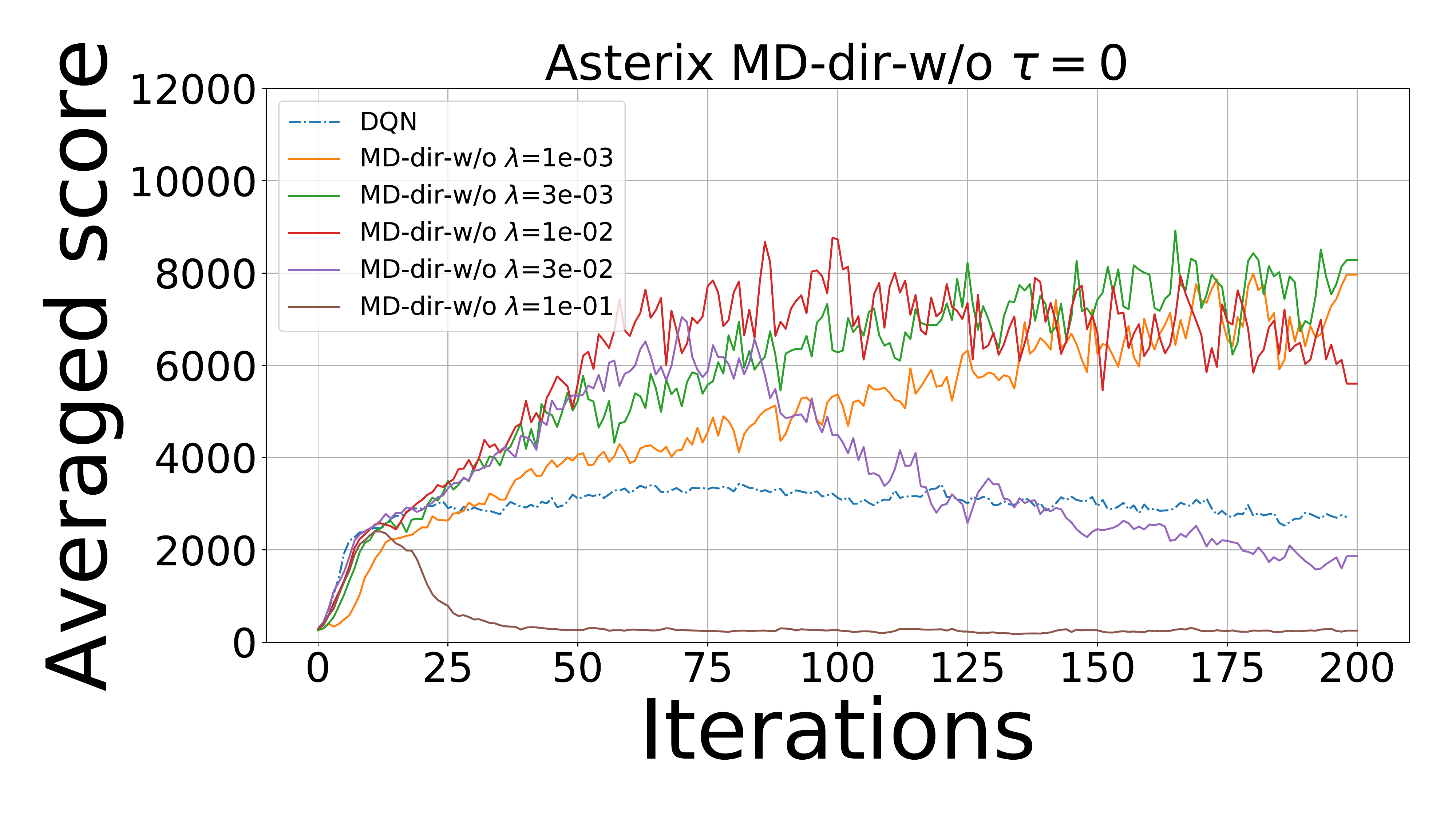}& 
     \includegraphics[width=0.3\linewidth]{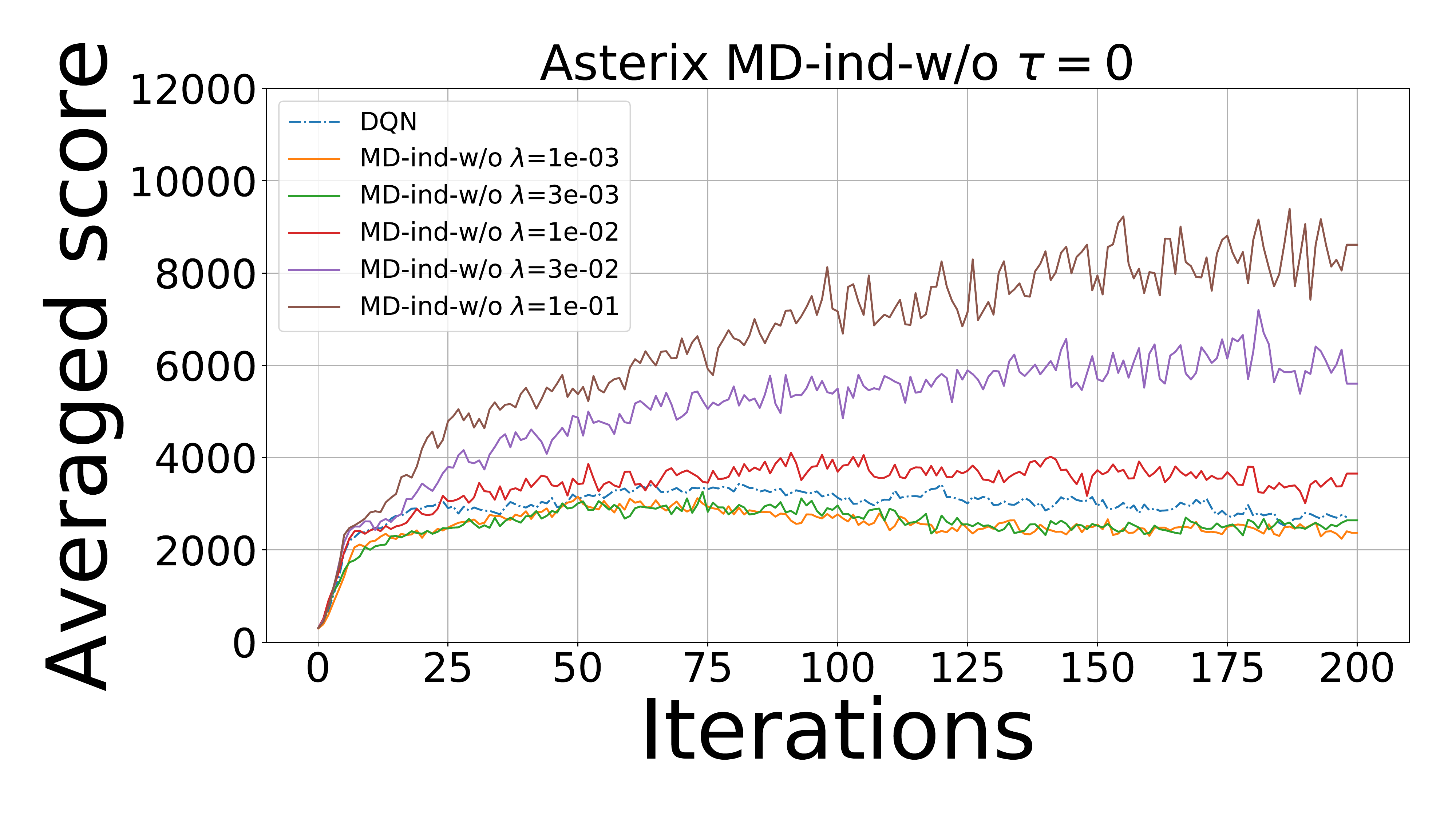}&
     \\
\end{tabular}
\caption{All averaged training scores of limit cases on Asterix, for several values of $\beta$ and $\lambda$. In each plot, a curve corresponds to a value of $\lambda$ for MD-ind and MD-dir, and to a value of $\beta$ for Mo-DQN. The blue dotted line is DQN.\label{fig:asterix_curves_lim}}
\end{center}
\end{figure}

\begin{figure}
\begin{center}
\begin{tabular}{c c c}
     \includegraphics[width=0.3\linewidth]{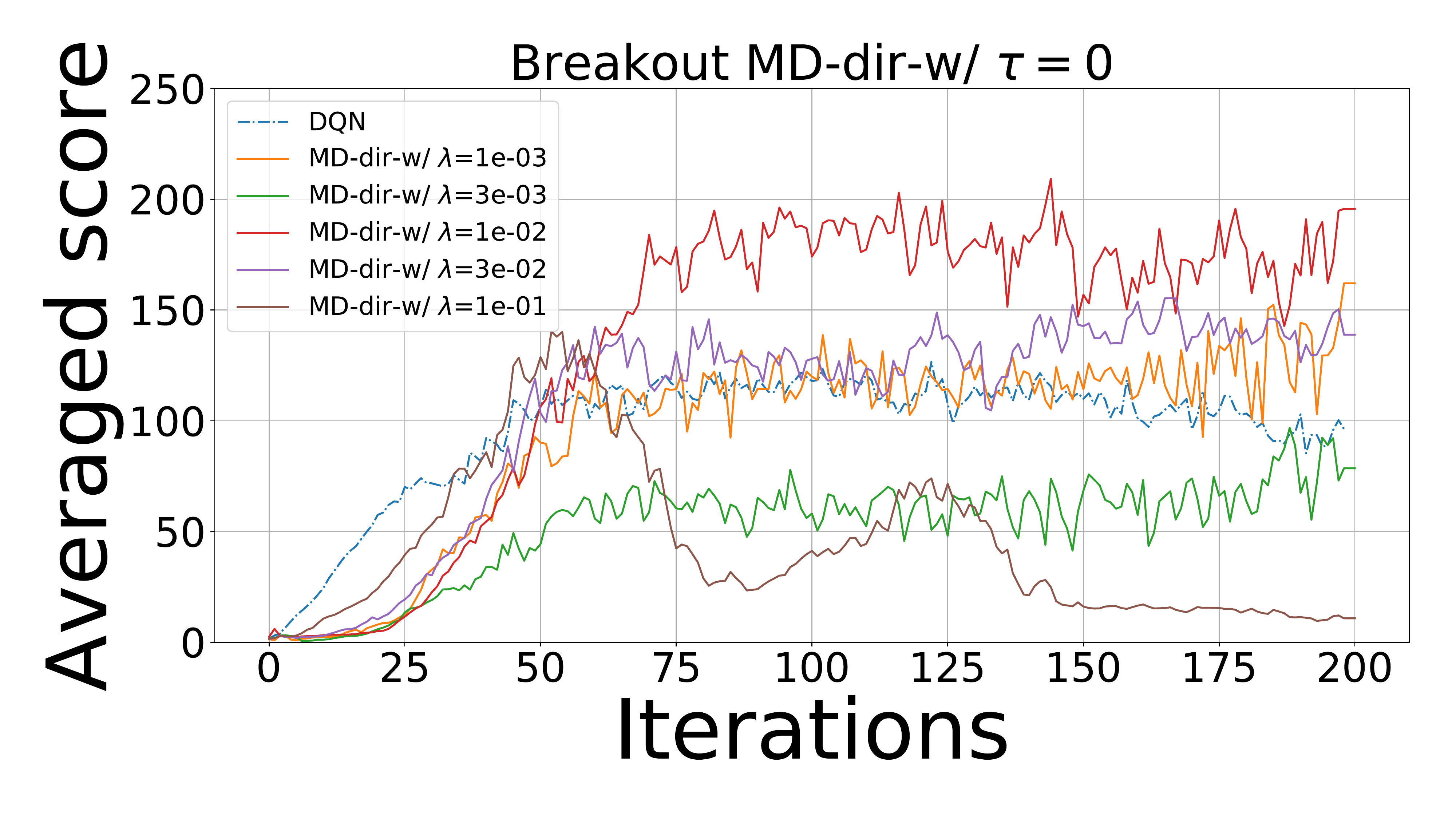}& 
     \includegraphics[width=0.3\linewidth]{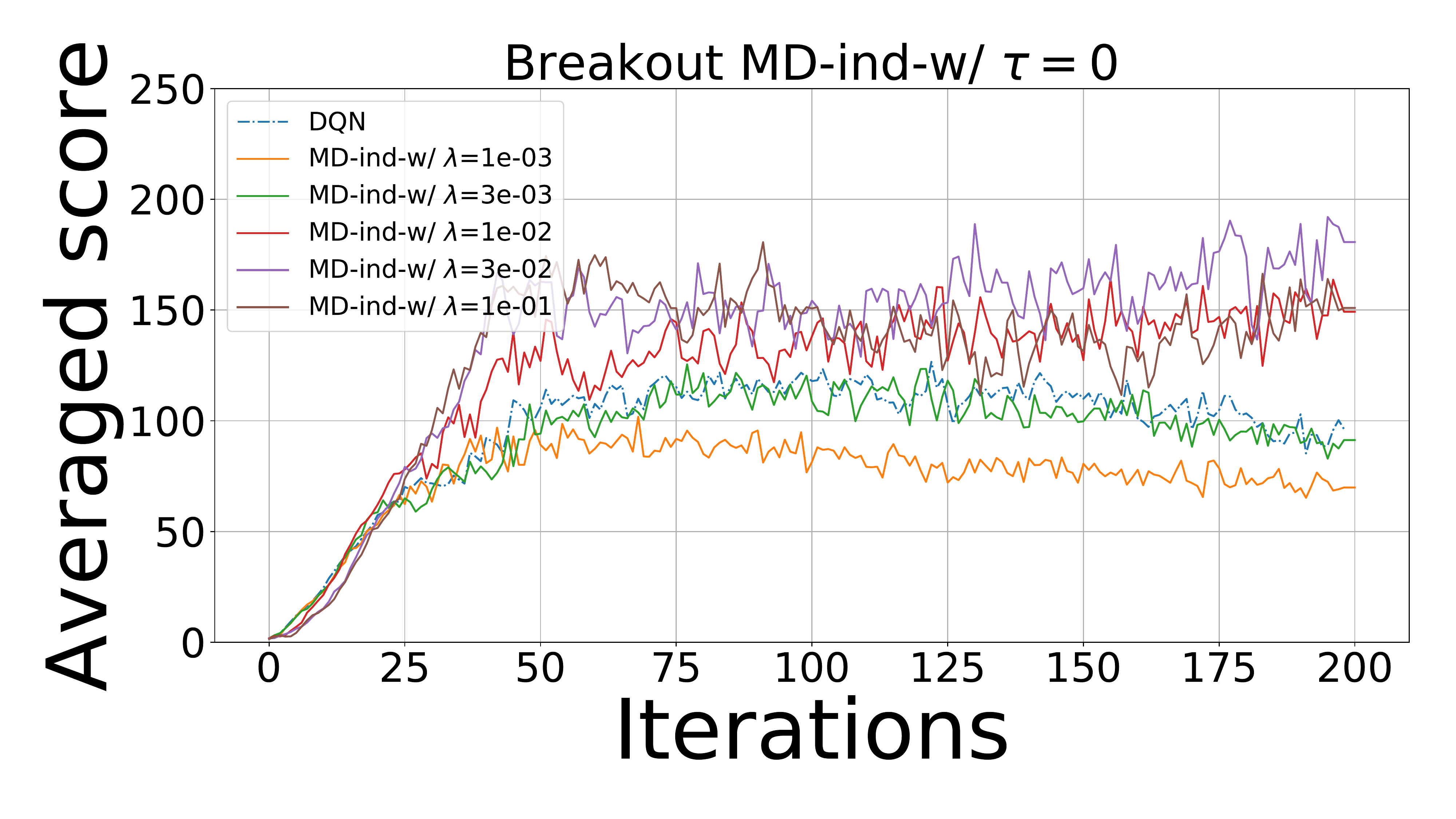}& 
     \multirow[b]{2}{*}[1cm]{\includegraphics[width=0.3\linewidth]{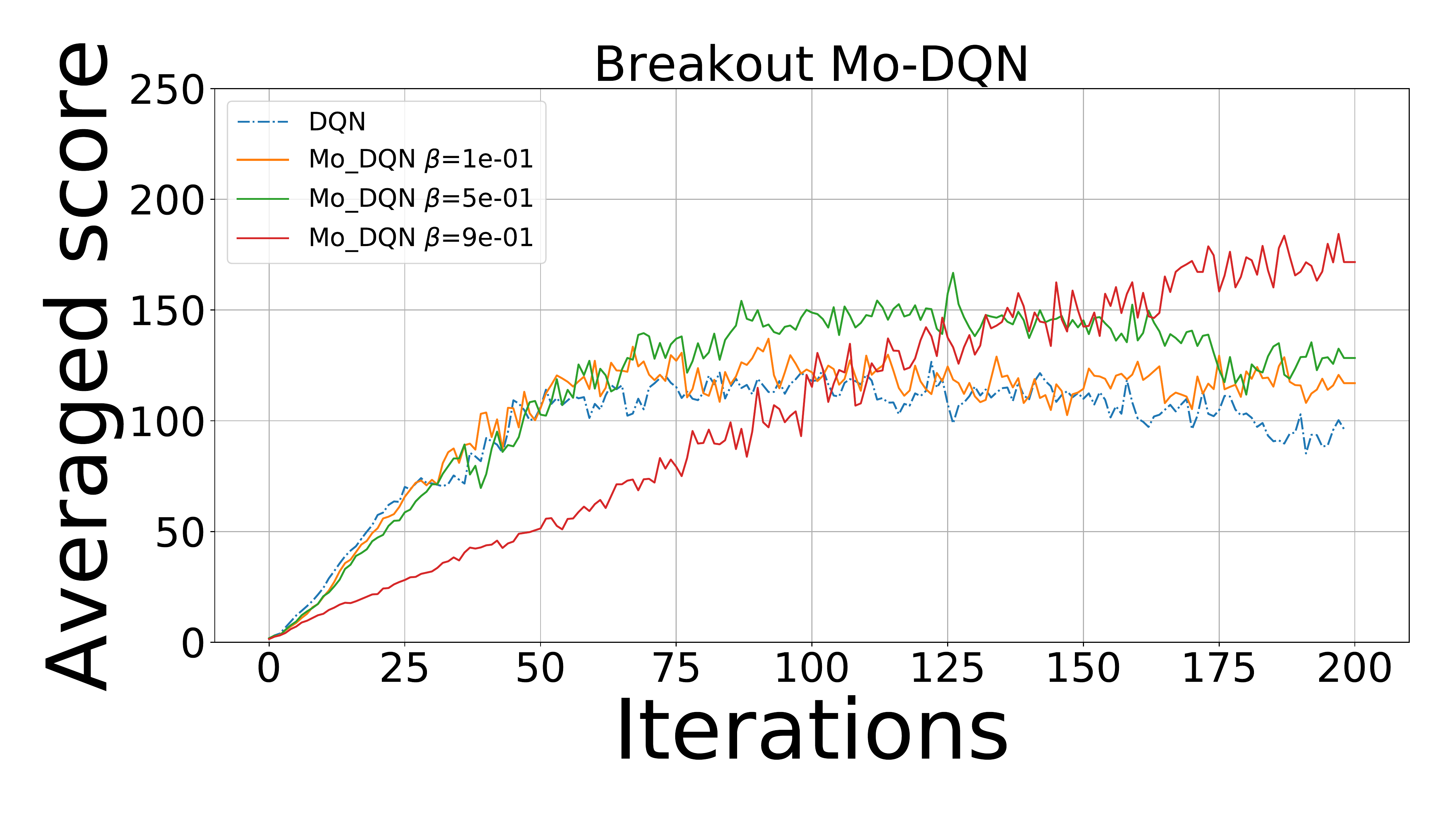}} \\
     \includegraphics[width=0.3\linewidth]{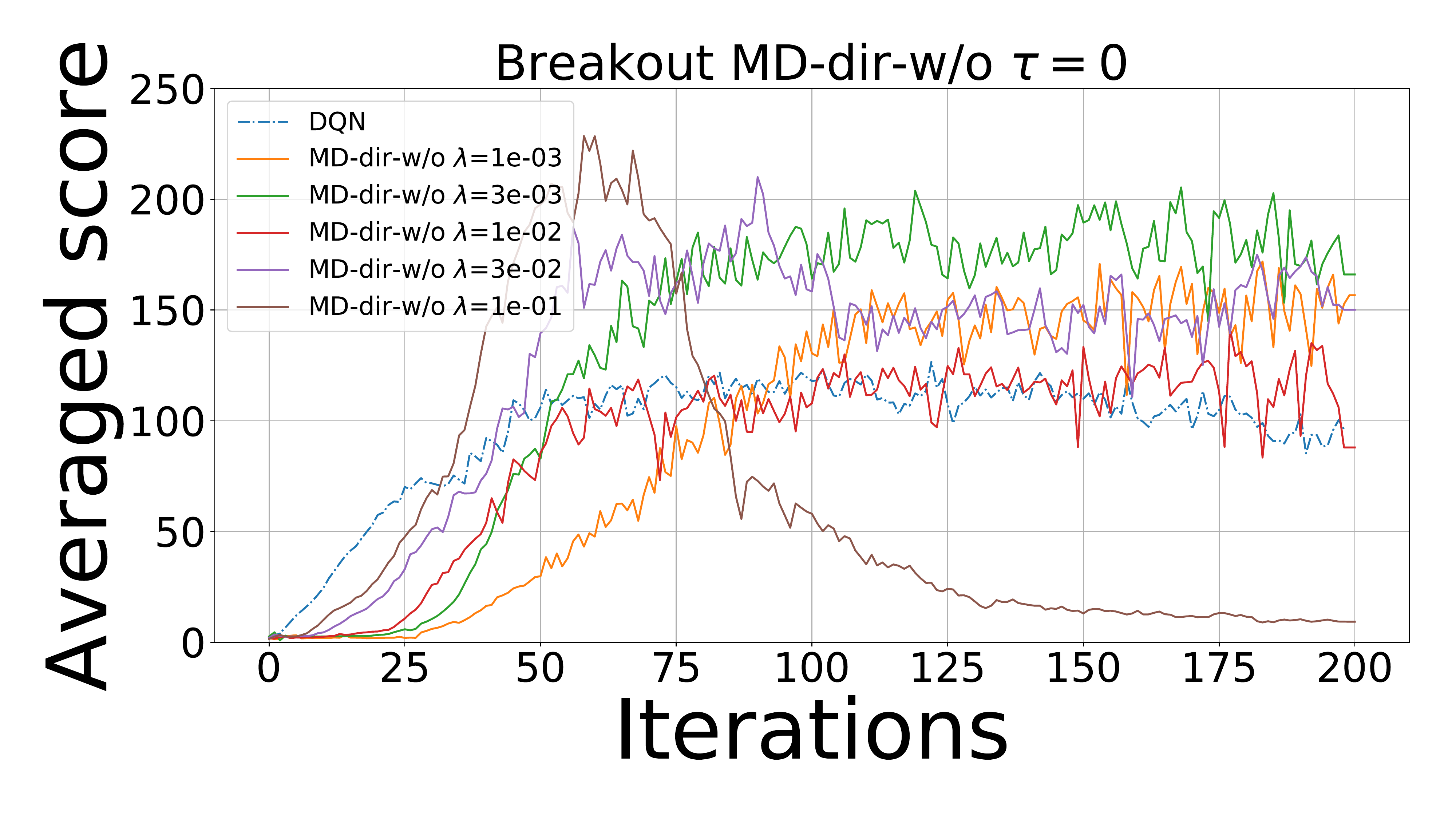}& 
     \includegraphics[width=0.3\linewidth]{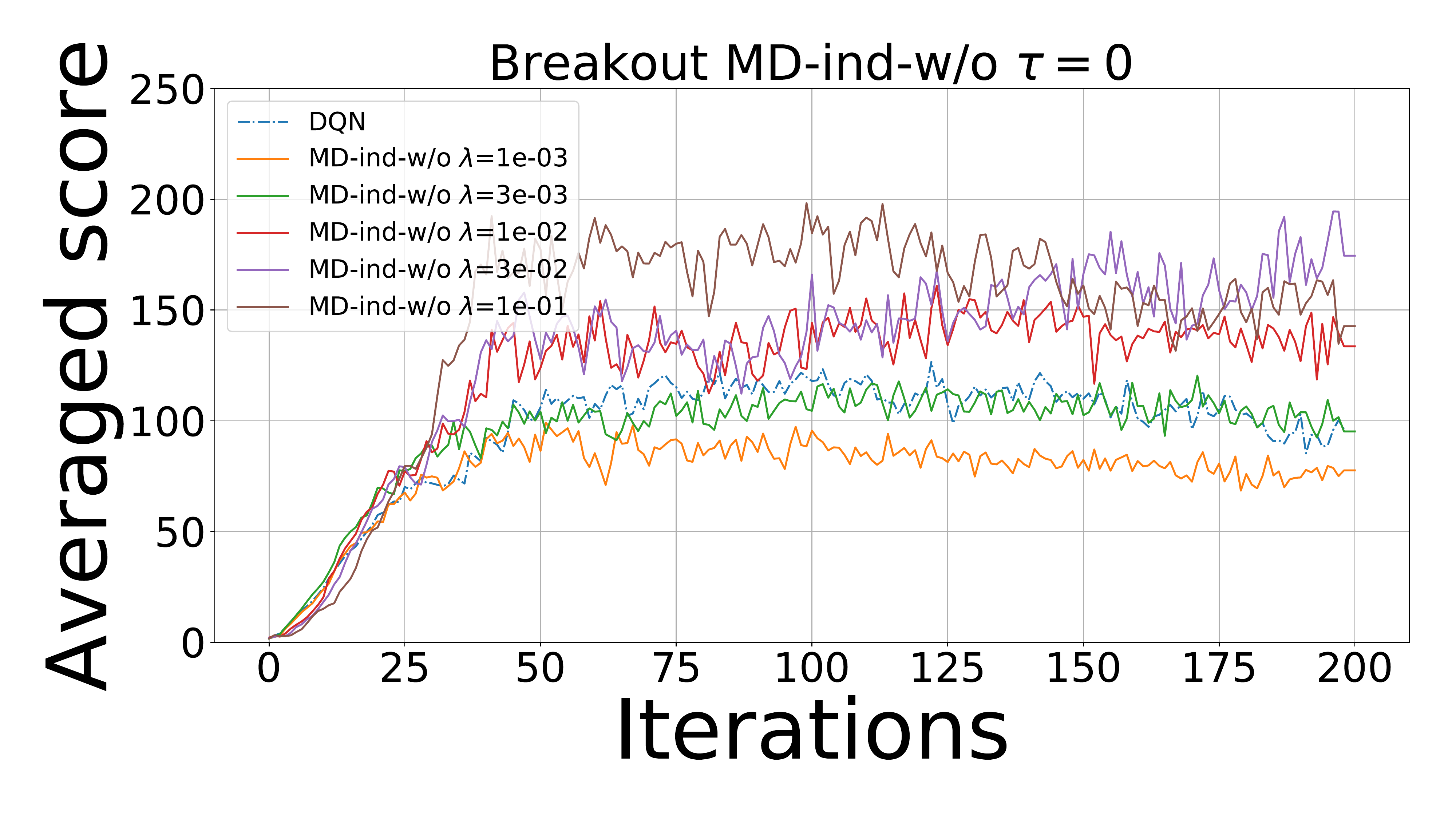}&
     \\
\end{tabular}
\caption{All averaged training scores of limit cases on Breakout, for several values of $\beta$ and $\lambda$. In each plot, a curve corresponds to a value of $\lambda$ for MD-ind and MD-dir, and to a value of $\beta$ for Mo-DQN. The blue dotted line is DQN. \label{fig:breakout_curves_lim}}
\end{center}
\end{figure}

\begin{figure}
\begin{center}
\begin{tabular}{c c c}
     \includegraphics[width=0.3\linewidth]{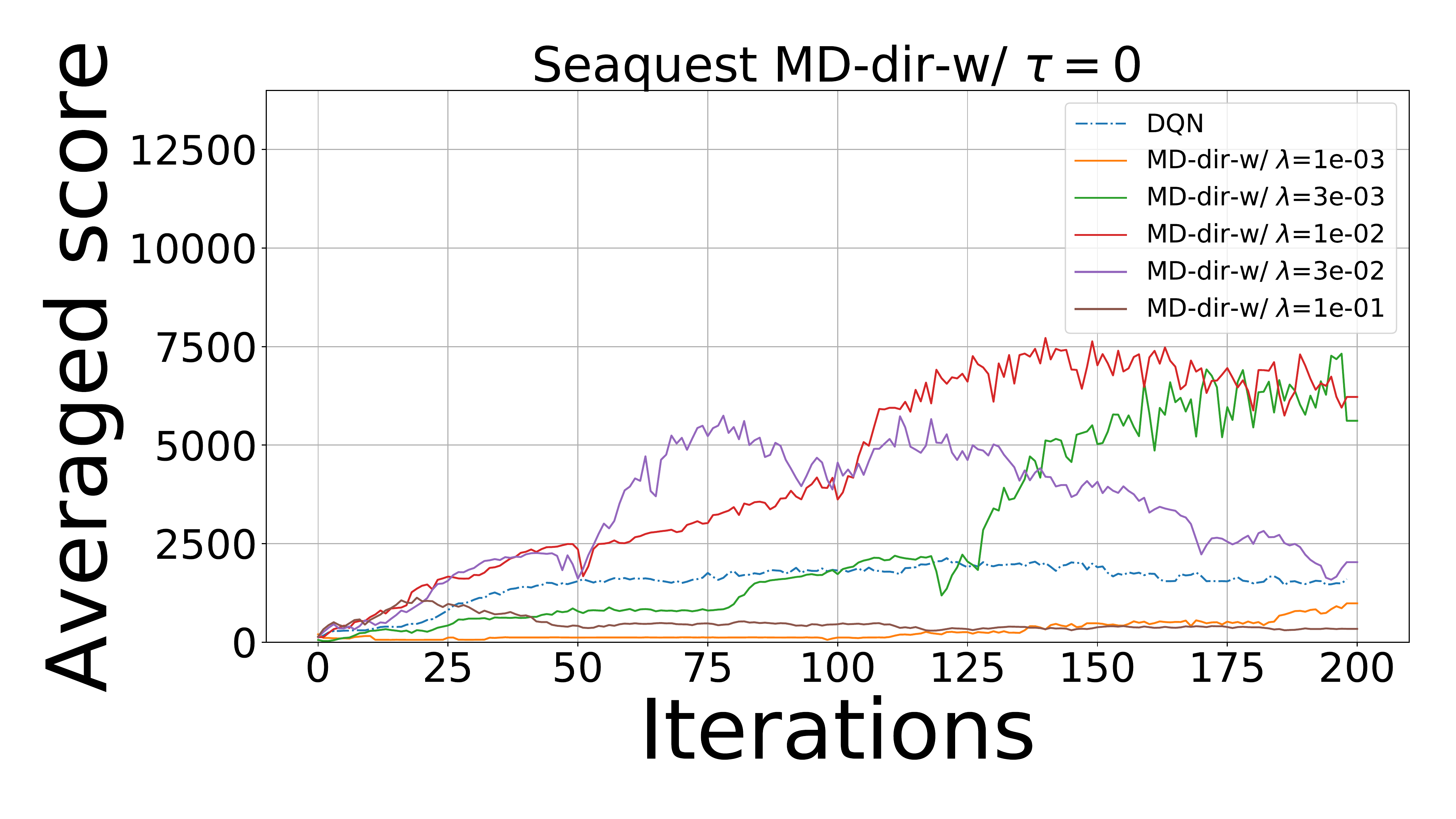}& 
     \includegraphics[width=0.3\linewidth]{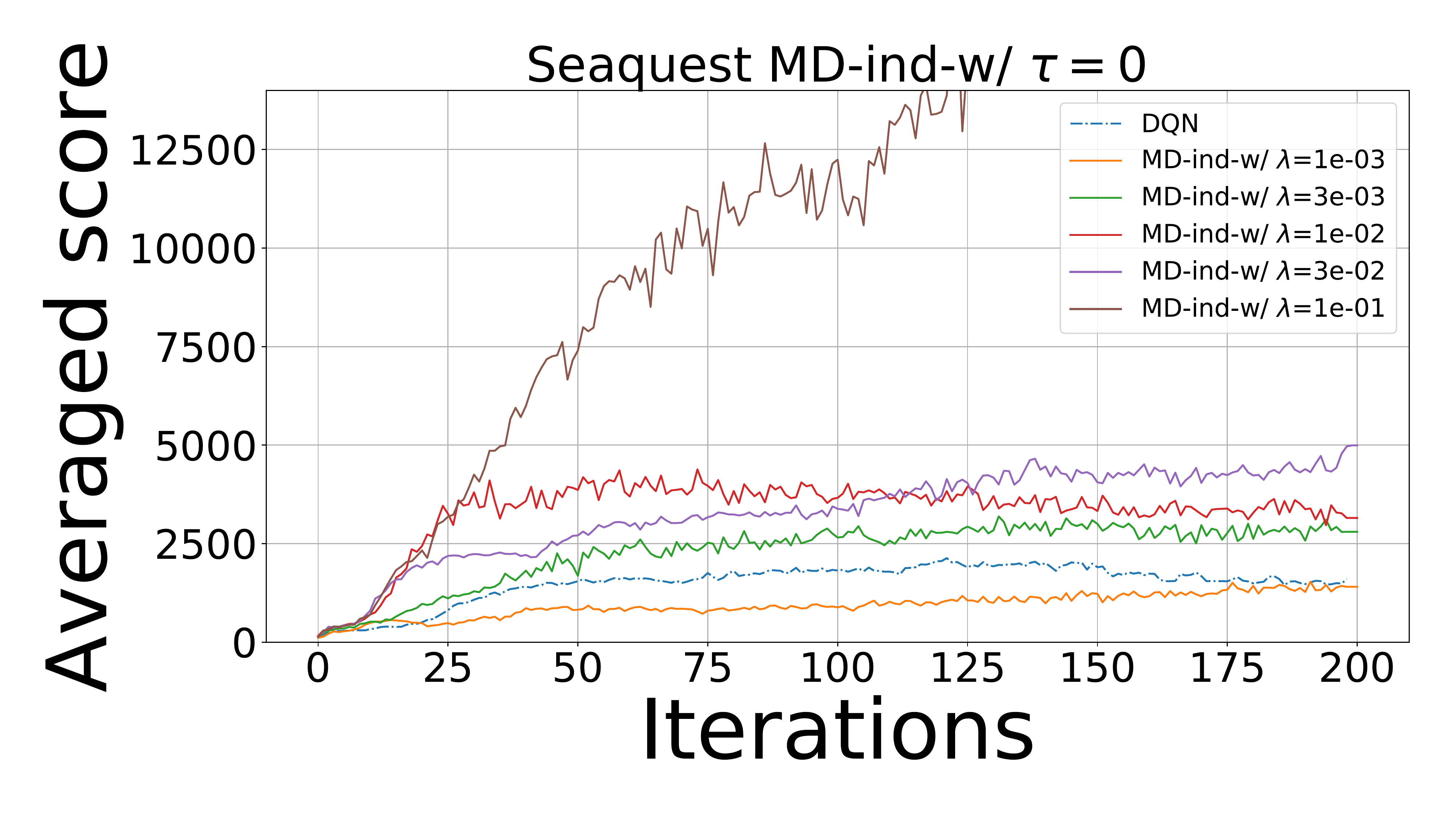}& 
     \multirow[b]{2}{*}[1cm]{\includegraphics[width=0.3\linewidth]{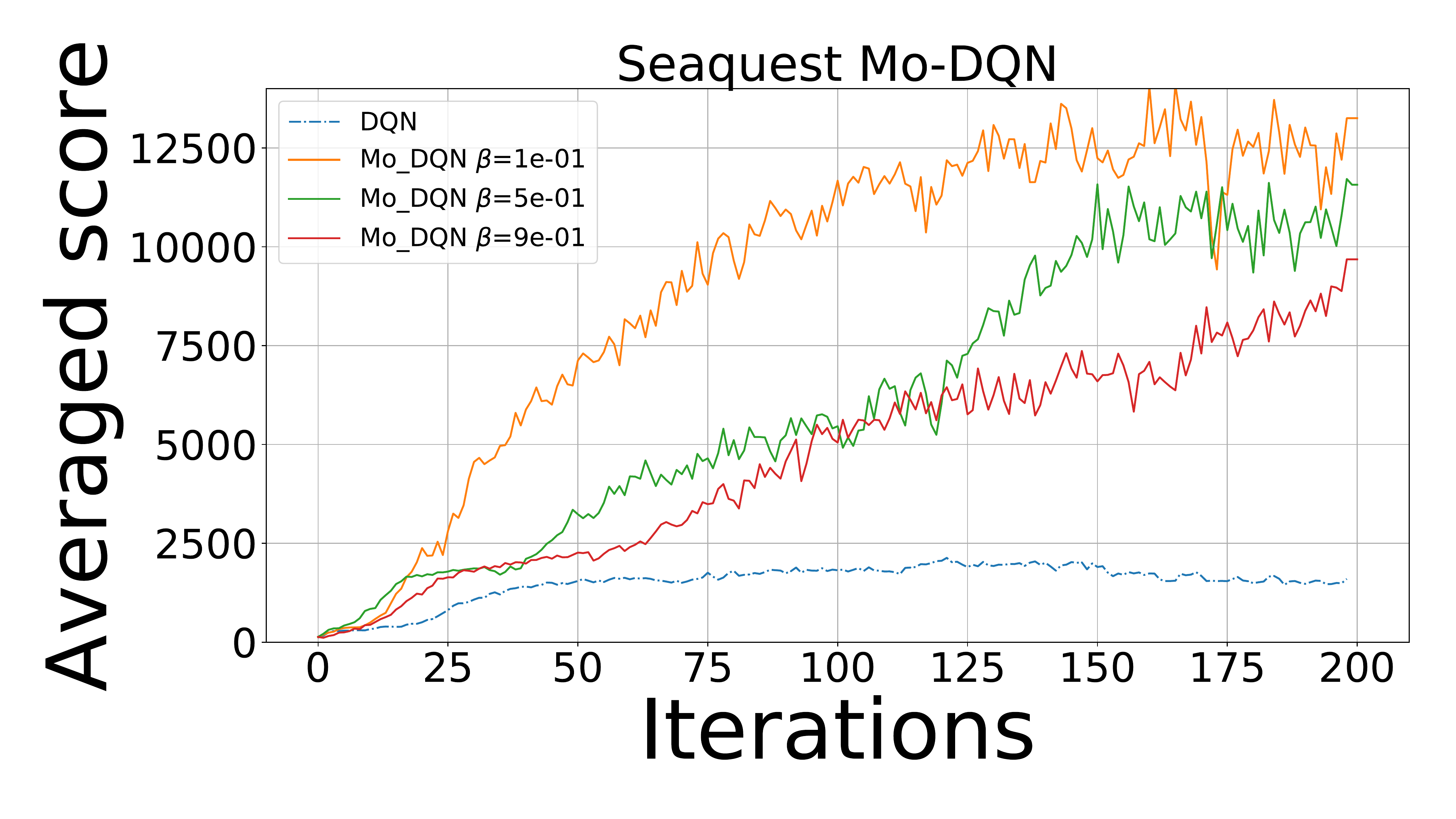}} \\
     \includegraphics[width=0.3\linewidth]{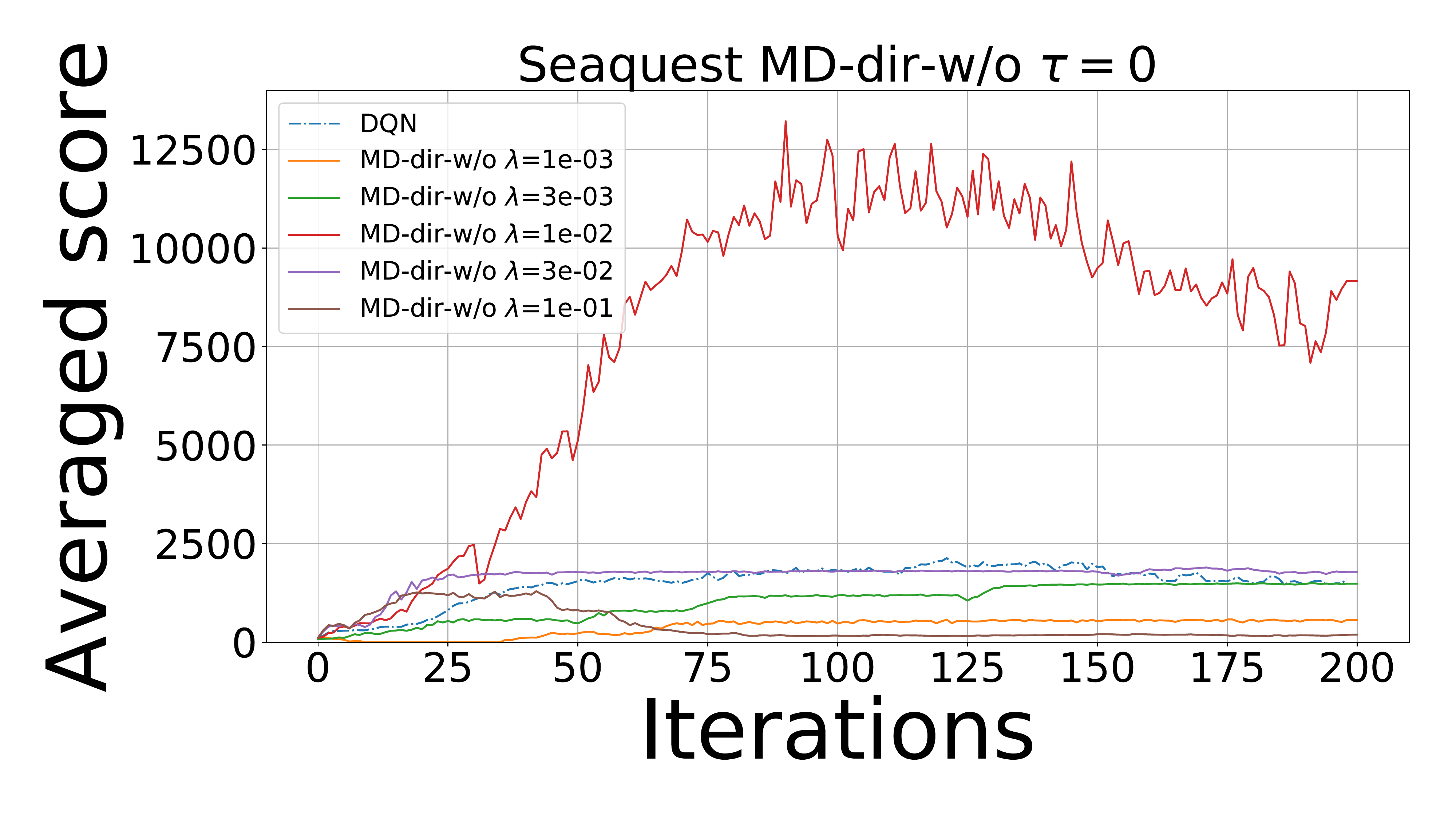}& 
     \includegraphics[width=0.3\linewidth]{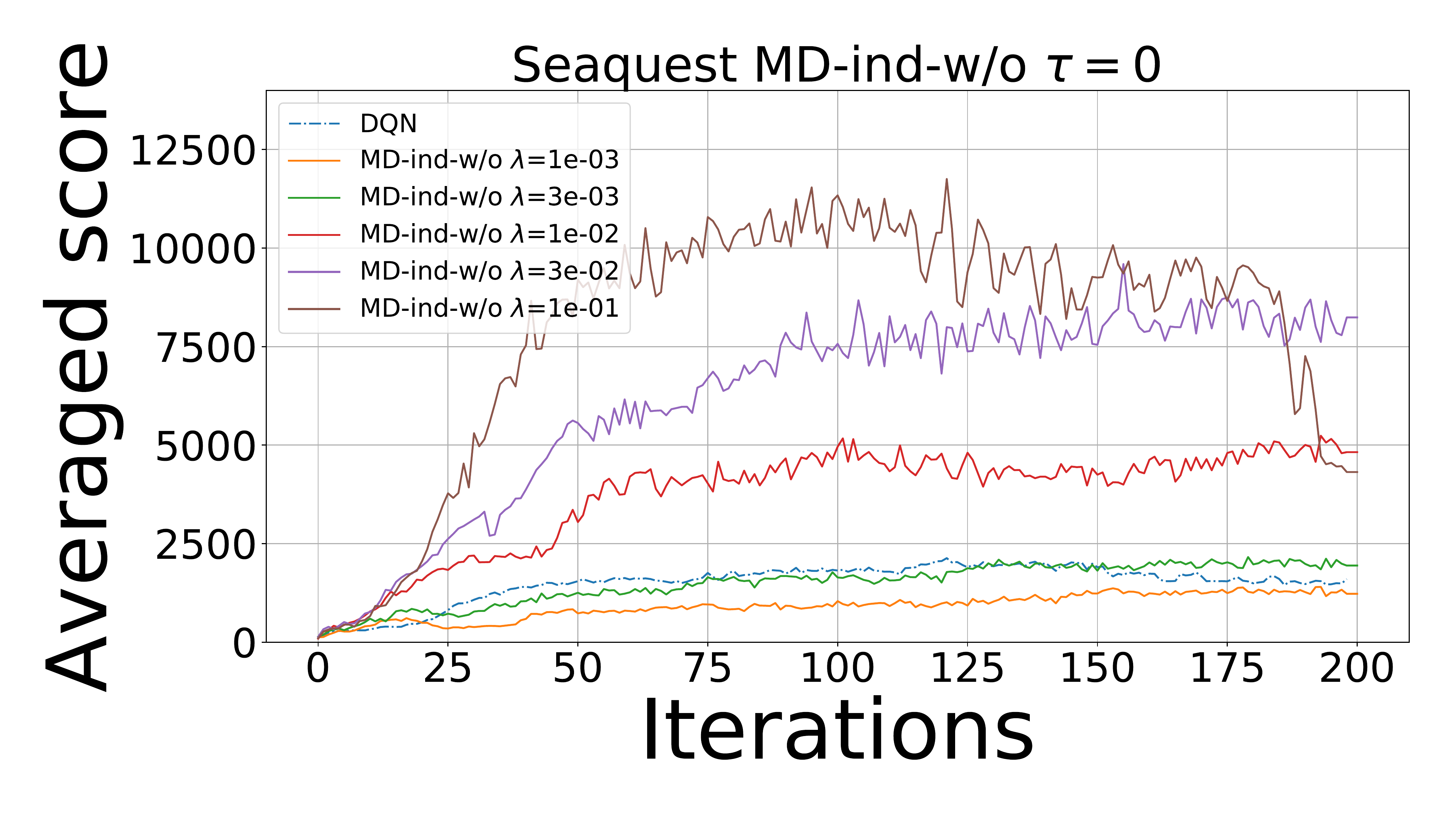}&
     \\
\end{tabular}
\caption{All averaged training scores of limit cases on Seaquest, for several values of $\beta$ and $\lambda$. In each plot, a curve corresponds to a value of $\lambda$ for MD-ind and MD-dir, and to a value of $\beta$ for Mo-DQN. The blue dotted line is DQN. \label{fig:seaquest_curves_lim}}
\end{center}
\end{figure}

\paragraph{Training curves.} We also report training curves on Atari. We report training curves of DA, MD-dir and MD-ind in Fig.~\ref{fig:asterix_curves} for Asterix, on Fig.~\ref{fig:breakout_curves} for Breakout, and on Fig.~\ref{fig:seaquest_curves} for Seaquest. We report the training curves of the limit cases on these three games on Figs.~\ref{fig:asterix_curves_lim}, \ref{fig:breakout_curves_lim} and~\ref{fig:seaquest_curves_lim}. In these figures, an \emph{iteration} corresponds to $250000$ training steps, and we report every iteration the undiscounted reward averaged over the last $100$ episodes (the \emph{averaged score}). The training curves are averaged over $3$ random seeds. 

The training curves give more hindsights on the performance of the algorithms. Indeed, the metric we used in the tables (the averaged score over all iteration) is partly flawed, because it could give a high score to an algorithm with a performance drop at the end of training. For example, the MD-dir method on Atari seems to benefit from regularizing the evaluation step (as unregularized evaluation suffers from a performance drop), which is less visible from the score tables. In almost all the cases, we do not observe such behaviour, which validates the use of our metric.

5
\end{document}